\pdfoutput=1
\documentclass{article}%
\usepackage{amssymb}
\usepackage{amsfonts}
\usepackage{amsmath}
\usepackage{graphicx}
\usepackage{natbib}
\usepackage{hyperref}%
\providecommand{\U}[1]{\protect\rule{.1in}{.1in}}
\newtheorem{theorem}{Theorem}

\newtheorem{claim}[theorem]{Claim}
\newtheorem{conclusion}[theorem]{Conclusion}

\begin{document}

\title{Resolving the Geometric Locus Dilemma for Support Vector Learning Machines}
\author{Denise M. Reeves}
\date{}
\maketitle

\begin{abstract}
Capacity control, the bias/variance dilemma, and learning unknown functions
from data, are all concerned with identifying effective and consistent fits of
unknown geometric loci to random data points. A geometric locus is a curve or
surface formed by points, all of which possess some uniform property. A
geometric locus of an algebraic equation is the set of points whose
coordinates are solutions of the equation. Any given curve or surface must
pass through each point on a specified locus. This paper argues that it is
impossible to fit random data points to algebraic equations of partially
configured geometric loci that reference arbitrary Cartesian coordinate
systems. It also argues that the fundamental curve of a linear decision
boundary is actually a principal eigenaxis. It is shown that learning
principal eigenaxes of linear decision boundaries involves finding a point of
statistical equilibrium for which eigenenergies of principal eigenaxis
components are symmetrically balanced with each other. It is demonstrated that
learning linear decision boundaries involves strong duality relationships
between a statistical eigenlocus of principal eigenaxis components and its
algebraic forms, in primal and dual, correlated Hilbert spaces. Locus
equations are introduced and developed that describe principal
eigen-coordinate systems for lines, planes, and hyperplanes. These equations
are used to introduce and develop primal and dual statistical eigenlocus
equations of principal eigenaxes of linear decision boundaries. Important
generalizations for linear decision boundaries are shown to be encoded within
a dual statistical eigenlocus of principal eigenaxis components. Principal
eigenaxes of linear decision boundaries are shown to encode Bayes' likelihood
ratio for common covariance data and a robust likelihood ratio for all other data.


\end{abstract}
\tableofcontents

\section{Introduction and Motivation}

The design and development of learning machine architectures has primarily
been based on curve and surface fitting methods of interpolation or
regression, alongside statistical methods of reducing data to minimum numbers
of relevant parameters. For example, multilayer artificial neural networks
(ANNs) estimate nonlinear regressions with optimally pruned architectures.
Good generalization performance for ANNs is considered an effect of a good
nonlinear interpolation of the training data
\citet{Geman1992}%
,
\citet{Haykin2009}%
. Alternatively, support vector machines (SVMs) fit linear curves or surfaces
to minimum numbers of training points. Good generalization performance for
SVMs is largely attributed to maximally separated linear decision borders
\citet{Boser1992}%
,
\citet{Cortes1995}%
,
\citet{Bennett2000}%
,
\citet{Cristianini2000}%
,
\citet{Scholkopf2002}%
.

Most significant machine learning problems are considered extrapolation
problems for unknown functions, e.g., nontrivial black box estimates. Black
boxes are defined in terms of inputs, subsequent outputs, and the mathematical
functions that relate them. Because training points will never cover a space
of possible inputs, practical learning machines must extrapolate in manners
that provide effective generalization performance
\citet{Geman1992}%
,
\citet{Gershenfeld1999}%
,
\citet{Haykin2009}%
.

Identifying effective methods that provide consistent fits of unknown
functions to random data points remains a difficult and open problem. This
paper assumes that classical numerical methods for curve fitting and function
approximation provide little insight into effective designs for learning
machine architectures.

\subsection{Paradigm Shift in Machine Learning: Targeting the Right Curves}

This paper intends to introduce a paradigm shift in machine learning of curves
or surfaces from data. The paradigm concerns the \emph{strong dual principal
eigenlocus method}. Principal eigenlocus methods involve learning principal
eigenaxes of second-order decision boundaries that take the form of
$d$-dimensional circles, ellipses, hyperbolae, parabolas, or lines. This paper
will argue that learning second-order decision boundaries from training data
requires learning the geometric locus of a principal eigenaxis. To wit, the
primary curve of interest for second-order decision boundaries is the locus of
a principal eigenaxis.

The fundamental idea behind the paradigm shift is the simple\ general concept
of a geometric locus of points. A geometric locus is a curve or surface formed
by a set of points, all of which possess some uniform property. Any given
geometric locus is described by an algebraic equation, where the geometric
locus of an algebraic equation is the locus (location) of all those points
whose coordinates are solutions of the equation. Any point whose coordinate
locations do not satisfy the algebraic equation of a specified locus is not on
the given curve or surface. Likewise, any given curve or surface must pass
through each point on a specified locus. Classic examples of a geometric locus
include circles, ellipses, hyperbolae, parabolas, lines, and points. Material
on geometric locus methods can be found in
\citet{Nichols1893}%
,
\citet{Tanner1898}%
,
\citet{Whitehead1911}%
, and
\citet{Eisenhart1939}%
.

For any given set of points which possess a uniform property, there exists a
geometric locus, formed by those points which exhibit the uniform property,
that is described by an algebraic equation, the solutions of which determine
the locations of the points on the given locus.

This paper will examine effective designs for learning machine architectures
that involve identifying and exploiting algebraic systems of geometric loci
which provide \emph{regularized geometric architectures of statistical
decision systems}. Accordingly, this paper will consider effective designs for
learning machine architectures that involve \emph{targeting the right curves
and learning the right set of geometric loci}.

\subsection{The Paradigm Shift}

Generally speaking, learning an unknown function from data involves finding
the function that generated a set of points. \emph{This paper will show that
the locus of a principal eigenaxis is the place of origin for all forms of
separating lines, planes, and hyperplanes}. This paper will develop the
algebraic equations of the geometric locus of the principal eigenaxis of a
linear locus of points. These equations will be used to identify the uniform
geometric properties which are exhibited by all of the points on a linear
locus, which include the principal eigenaxis of the locus. All of these
results will be used to motivate and develop an algebraic system of locus
equations that describes principal eigenaxes of linear decision boundaries.
Moreover, the algebraic system of locus equations determines principal
eigenaxes of separating lines, planes, or hyperplanes for all forms of data
distributions, including overlapping and homogeneous distributions. Thereby,
it will be shown that learning linear decision boundaries from training data
essentially involves learning the locus of a principal eigenaxis.

\subsection{Learning the Locus of a Principal Eigenaxis}

The locus of a principal eigenaxis is an inherent part of any linear curve or
surface. Moreover, the geometric locus of a linear curve or surface is encoded
within the locus of its principal eigenaxis. It will be shown that the
eigen-coordinate locations of a principal eigenaxis determine the uniform
properties exhibited by the points on a linear curve or surface. It will be
demonstrated that a principal eigenaxis satisfies a linear locus in terms of
its eigenenergy. It will also be shown that principal eigenaxes of linear loci
provide exclusive and distinctive reference axes.

A principal eigenaxis is considered to be a \emph{characteristic locus} for a
linear locus of points. Accordingly, the term \emph{eigenlocus} will be used
to refer to the principal eigenaxis of a linear locus. The term
\emph{statistical eigenlocus} will be used to refer to a statistical estimate
of an eigenlocus of a linear decision boundary. The term \emph{strong dual
principal eigenlocus} will be used to refer to joint, statistical eigenlocus
estimates in dual, correlated Hilbert spaces.

This paper will motivate and develop a strong dual principal eigenlocus of
principal eigenaxis components which encodes the eigen-coordinate locations of
an unknown principal eigenaxis of a linear decision boundary. The paper will
examine how important generalizations for linear decision boundaries are
encoded within statistical representations of principal eigenaxis locations,
whereby principal eigenaxis locations describe statistical properties of
training data. The paper will demonstrate how statistical representations of
principal eigenaxes take the form of a strong dual principal eigenlocus of
principal eigenaxis components, where each component encodes an
eigen-transformed principal location of large covariance.

This paper will examine how encoding relevant statistical aspects of principal
eigenaxis locations within learning machine architectures requires an
algebraic system of primal and dual statistical eigenlocus equations that
properly specify the loci of principal eigenaxis components for a given set of
training data. The paper will consider how this algebraic system of eigenlocus
equations involves strong duality relationships between a statistical
eigenlocus of principal eigenaxis components and its algebraic forms, in
primal and dual, correlated Hilbert spaces. It will be shown that learning a
strong dual principal eigenlocus of principal eigenaxis components involves
finding a point of statistical equilibrium for which the eigenenergies of
eigenlocus components are symmetrically balanced with each other in relation
to a centrally located statistical fulcrum. It will be demonstrated that a
strong dual principal eigenlocus satisfies a linear decision boundary in terms
of a critical minimum, i.e., a total allowed, eigenenergy. It will also be
demonstrated that a strong dual principal eigenlocus encodes Bayes' likelihood
ratio for similar covariance data distributions and a robust likelihood ratio
for all other data distributions.

The findings presented in this paper will define substantial geometric
architectures and statistical models for linear kernel SVMs. Findings for
polynomial kernel SVMs will be presented in another paper. New and surprising
information, discovered from the creation and analysis of statistical
eigen-coordinate systems for linear and nonlinear second-order decision
boundaries, is expected to provide significant insights into the
identification and exploitation of effective kernel widths for Gaussian kernel SVMs.

This paper will show that linear kernel SVMs are a powerful and robust class
of statistical learning machines which are useful for the design, development,
and implementation of probabilistic, multiclass linear pattern recognition systems.

The work presented in this paper is motivated by the seminal works of Thomas
Cover (1965)
\citet{Cover1965}%
, Geman, Bienenstock, and Doursat (1992)
\citet{Geman1992}%
, Boser, Guyon, and Vapnik (1992)
\citet{Boser1992}%
, and Cortes and Vapnik (1995)
\citet{Cortes1995}%
.

\subsection{Organization of the Paper}

The paper is organized as follows. The issues of fitting learning machine
architectures to unknown functions of training data are laid out in Section
$2$. In addition, Section $2$ considers the system representation problem for
learning machine architectures, methods for solving locus problems, and a high
level overview of a statistical model for linear kernel SVMs. Section $3$
formulates the geometric locus dilemma for statistical learning machines and
SVMs. Section $3$ also motivates resolving the geometric locus dilemma for
linear and polynomial kernel SVMs. Section $4$ develops a principal
eigen-coordinate system that describes all forms of linear loci. Principal
eigenaxes of linear loci are given the name of normal eigenaxes. Section $5$
motivates the design of learning machine architectures in Hilbert spaces,
develops the notion of functional glue for learning machine architectures, and
motivates hardwiring the eigenlocus of a principal, i.e., normal, eigenaxis
into linear kernel SVM architectures. Section $6$ defines the primal and the
Wolfe dual normal eigenlocus equations of a probabilistic, linear binary
classification system. Section $7$ defines the Lagrangian of the primal normal
eigenlocus. Section $8$ begins the process of defining the primal normal
eigenlocus within the Wolfe dual eigenspace. Section $9$ examines width
regulation of large covariance linear decision regions, develops a normal
eigenlocus test statistic for classifying unknown data, considers strong dual
normal eigenlocus transforms for homogeneous data distributions, and examines
equilibrium conditions for strong dual normal eigenlocus transforms. Section
$10$ examines the strong dual normal eigenlocus problem within the context of
the eigenlocus equation of a Wolfe dual normal eigenlocus. Section $11$
considers how geometric and statistical properties of strong dual normal
eigenlocus transforms are sensitive to eigenspectrums of Gram matrices;
Section $11$ also consider how eigenspectrums of Gram matrices determine
shapes of quadratic surfaces. Section $12$ motivates the examination of point
and coordinate relationships between the constrained primal and the Wolfe dual
normal eigenaxis components. Pointwise covariance statistics are defined for
individual training points, and are used to find extreme data points which
possess large pointwise covariances. Section $12$ also considers the total
allowed eigenenergies of a strong dual normal eigenlocus. Section $13$
develops a general expression for a principal eigen-decomposition that is used
to examine point and coordinate relationships between the constrained primal
and the Wolfe dual normal eigenaxis components. Sections $14$ and $15$ develop
algebraic expressions for the eigenloci (the geometric locations) of the Wolfe
dual normal eigenaxis components. These expressions are used to define uniform
geometric and statistical properties which are jointly exhibited by Wolfe dual
and constrained primal normal eigenaxis components. Section $16$ outlines the
fundamental issue that must be resolved to ensure that eigenenergies of normal
eigenlocus components are symmetrically balanced with each other. Section $17$
examines the algebraic, geometric, and statistical nature of the remarkable
statistical balancing feat that is routinely accomplished by strong dual
normal eigenlocus transforms. Algebraic and statistical expressions are
developed for the total allowed eigenenergies of a strong dual normal
eigenlocus. These expressions are used to define the statistical machinery
behind the statistical balancing feat, and are used to derive the strong dual
normal eigenlocus identity. Section $18$ defines probabilistic properties
which are exhibited by strong dual normal eigenlocus discriminant functions.
An expression is obtained for the normal eigenlocus decision rule which
encodes likelihood ratios, and is used to show that strong dual normal
eigenlocus discriminant functions encode Bayes' likelihood ratio for common
covariance data, and a robust likelihood ratio for all other data
distributions. Section $19$ outlines dual-use of strong dual normal eigenlocus
discriminant functions which include probabilistic, multiclass linear pattern
recognition systems, a statistical multimeter for measuring class separability
and Bayes' error rate, and a robust indicator of homogeneous data
distributions. Section $20$ summarizes the geometric underpinnings and
statistical machinery of linear kernel SVMs. Section $21$ summarizes the major
findings and conclusions of the paper.

\section{Fitting Learning Machine Architectures to Unknown Functions of Data}

Fitting unknown functions to collections of random or arbitrary data points
has long been described as fraught with difficulties. Functions tend to be
overly sensitive to point coordinate locations and small perturbations of the
data. Minor changes in data point locations can produce largely different
functions that vary widely in performance. Function values and performance are
also affected by data quantities
\citet{Synge1957}%
,
\citet{Daniel1979}%
,
\citet{Lancaster1986}%
,
\citet{Wahba1987}%
,
\citet{Linz2003}%
.

Most function approximation methods hinge on the guarantee that a straight
line can be passed through two points, a parabola through three, a cubic
through four, and so on
\citet{Davis1963}%
. A large class of function approximation problems involve fitting a set of
curves or surfaces to some representative data set
\citet{Lancaster1986}%
,
\citet{Rao2002}%
,
\citet{Linz2003}%
,
\citet{Rahman2004}%
. For example, the Fourier series fits sinusoids or complex exponentials to
periodic signals
\citet{Lathi1998}%
, whereas wavelets are used to decompose signals into low frequency and high
frequency components
\citet{Mertins1999}%
. Generally speaking, curve fitting involves selecting a function that
generated a set of points.

Classical approximation methods were initially developed to replace more
complicated functions with simpler ones, such as polynomial functions and
piecewise polynomials. The primary concerns for classical approximation
methods are approximation errors and speed of convergence
\citet{Davis1963}%
,
\citet{Linz1979}%
,
\citet{Keener2000}%
,
\citet{Linz2003}%
.

Fitting learning machine architectures to unknown functions of training data
involves numerous and difficult problems. Learning machine architectures are
extremely sensitive to algebraic and topological structures that include
functionals, reproducing kernels, parameter sets, constraint sets,
regularization parameters, and eigenspectrums of data matrices
\citet{Geman1992}%
,
\citet{Byun2002}%
,
\citet{Haykin2009}%
,
\citet{Reeves2009}%
,
\citet{Reeves2011}%
. For example, SVM architectures based on polynomial and Gaussian kernel
matrices vary widely in terms of generalization performance
\citet{Byun2002}%
,
\citet{Eitrich2006}%
. It has also been shown that regularization parameters of linear kernel SVMs
determine SVM architectures that exhibit extreme variations of generalization
performance
\citet{Reeves2011}%
. The generalization performance of multilayer ANNs also varies substantially
with data samples and functional (hidden node) configurations
\citet{Haykin2009}%
. In addition to these difficulties, parameter and eigenspectrum estimates may
be ill-defined
\citet{Kay1993}%
,
\citet{Moon2000}%
,
\citet{Reeves2009}%
,
\citet{Reeves2011}%
. Identifying the correct form of an equation for a statistical model is also
a large concern
\citet{Daniel1979}%
,
\citet{Breiman1991}%
,
\citet{Geman1992}%
.

In general, learning algorithms that estimate decision boundaries for
classification problems introduce four sources of error into the final
classification system: $(1)$ Bayes' error, $(2)$ model error or bias, $(3)$
estimation error or variance, and $(4)$ computational error
\citet{VanTrees1968}%
,
\citet{Tikhovov1977}%
,
\citet{Goldberg1979}%
,
\citet{Linz1979}%
,
\citet{Wahba1987}%
,
\citet{Fukunaga1990}%
,
\citet{Geman1992}%
,
\citet{Mitchell1997}%
,
\citet{Engl2000}%
,
\citet{Duda2001}%
,
\citet{Zhdanov2002}%
,
\citet{Linz2003}%
,
\citet{Haykin2009}%
. Bayes' error defines an optimal error rate for any decision making task
\citet{VanTrees1968}%
,
\citet{Fukunaga1990}%
,
\citet{Duda2001}%
.

\subsection{Regulation of Learning Machine Capacities}

Learning machine architectures with $N$ free parameters have a learning
capacity to fit $N$ data points, so that curves or surfaces can be made to
pass through every data point. However, fitting all of the training data is
generally considered bad statistical practice
\citet{Wahba1987}%
,
\citet{Breiman1991}%
,
\citet{Geman1992}%
,
\citet{Barron1998}%
,
\citet{Cherkassky1998}%
,
\citet{Gershenfeld1999}%
,
\citet{Duda2001}%
,
\citet{Hastie2001}%
,
\citet{Haykin2009}%
. Random fluctuations (noise) in signals or images obscures information
contained in training data. Learning machine architectures that correctly
interpolate collections of noisy training points, fit the idiosyncrasies of
the noise and are not expected to exhibit good generalization performance.
Likewise, highly flexible architectures with indefinite parameter sets are
said to overfit the training data
\citet{Wahba1987}%
,
\citet{Geman1992}%
,
\citet{Cherkassky1998}%
,
\citet{Gershenfeld1999}%
,
\citet{Duda2001}%
,
\citet{Hastie2001}%
,
\citet{Scholkopf2002}%
,
\citet{Haykin2009}%
.

Yet again, learning machine architectures that interpolate insufficient
numbers of data points exhibit under-fitting difficulties. Architectures with
too few parameters ignore both the noise and the meaningful behavior of the
data. Parameterized architectures that cannot adequately describe a set of
data samples are said to underfit the training data
\citet{Wahba1987}%
,
\citet{Cherkassky1998}%
,
\citet{Gershenfeld1999}%
,
\citet{Scholkopf2002}%
,
\citet{Haykin2009}%
. Figure $1$ depicts a cartoon of an underfitting, overfitting, and balanced
fitting of a set of data points.

\begin{center}%
\begin{center}
\includegraphics[
natheight=7.499600in,
natwidth=9.999800in,
height=3.2897in,
width=4.3777in
]%
{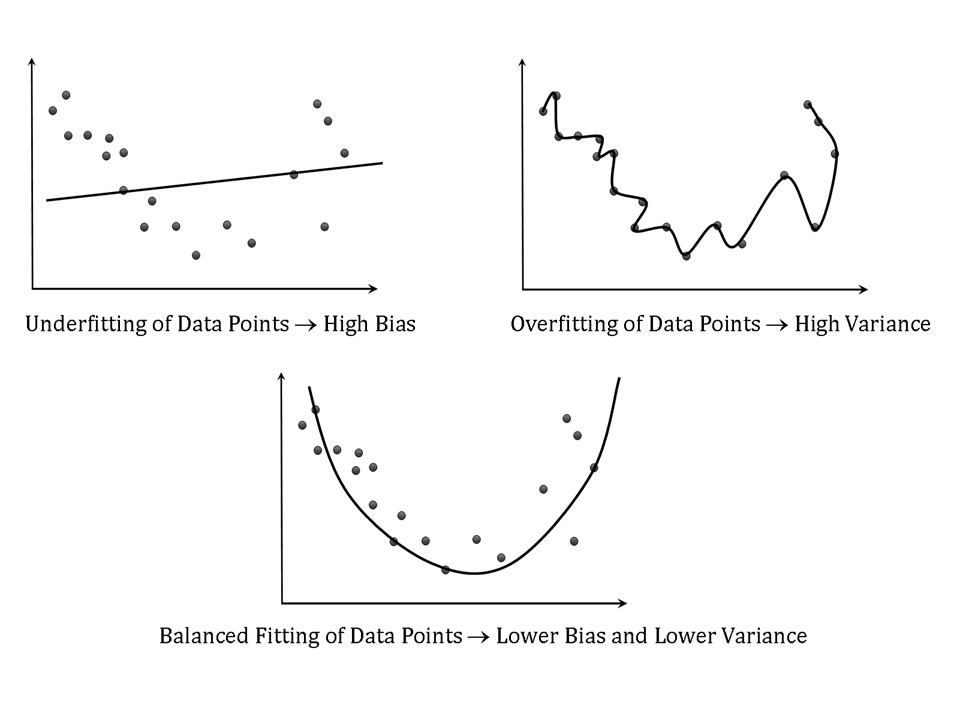}%
\end{center}

\end{center}

\begin{flushleft}
Figure $1$: Illustration of the difficulties associated with fitting an
unknown function to a collection of random data points.
\end{flushleft}

Discriminant functions that possess too much learning capacity are more likely
to overfit the training data
\citet{Boser1992}%
,
\citet{Scholkopf2002}%
. SVMs mitigate overfitting difficulties by a process termed capacity control
which is outlined next.

\subsection{SVM\ Capacity Control}

SVMs estimate linear decision boundaries by solving a quadratic programming
problem. The capacity or complexity of SVM decision boundaries is regulated by
means of a geometric margin of separation between two given sets of data. The
SVM method minimizes the capacity of a separating hyperplane by maximizing the
distance between a pair of margin hyperplanes or linear borders. Large
distances between margin hyperplanes $(1)$ allow for considerably fewer
hyperplane orientations, and $(2)$ enforce a limited capacity to separate
training data. Thus, maximizing the distance between margin hyperplanes
regulates the complexity of separating hyperplane estimates
\citet{Boser1992}%
,
\citet{Cortes1995}%
,
\citet{Burges1998}%
,
\citet{Bennett2000}%
,
\citet{Cristianini2000}%
,
\citet{Scholkopf2002}%
. Figure $2$ illustrates a cartoon of three geometric margins for a collection
of training data, where each linear border interpolates a data point from one
of the pattern classes. The SVM method chooses the green linear borders which
exhibit the largest geometric margin of separation.

\begin{center}%
\begin{center}
\includegraphics[
natheight=7.499600in,
natwidth=9.999800in,
height=3.2897in,
width=4.3777in
]%
{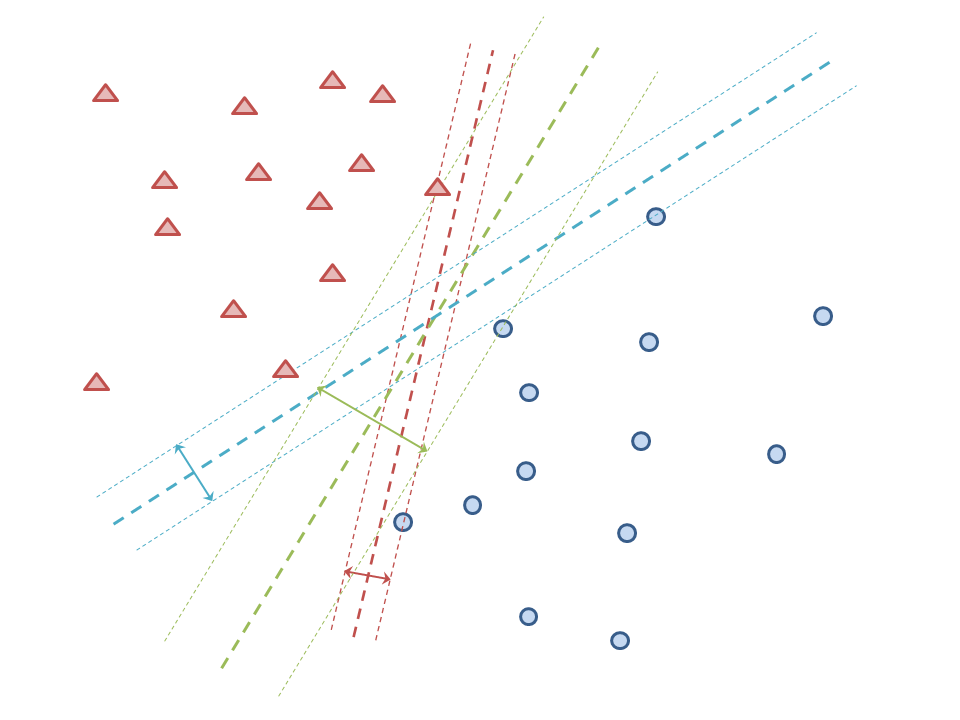}%
\end{center}

\end{center}

\begin{flushleft}
Figure $2$: Illustration of three geometric margins of separation, where data
points that belong to the same category are depicted by red triangles or blue
circles. The green linear borders exhibit the largest geometric margin,
whereas the red linear borders exhibit the smallest geometric margin.
\end{flushleft}

All of the above difficulties imply that learning unknown functions from
training data involves trade-offs between underfittings and overfittings of
data points. The bias-variance dilemma describes statistical facets of these
trade-offs
\citet{Geman1992}%
,
\citet{Gershenfeld1999}%
,
\citet{Duda2001}%
,
\citet{Hastie2001}%
,
\citet{Haykin2009}%
.

\subsection{The Bias \& Variance Dilemma}

All learning machine architectures are composed of training data. Moreover,
the estimation error between a learning machine and its target function
depends on the training data in a twofold manner. Geman, Bienenstock, and
Doursat examined these dual sources of estimation error in their seminal
article titled\textit{\ Neural Networks and the Bias/Variance Dilemma}
\citet{Geman1992}%
. The crux of the dilemma is that estimation error is composed of two distinct
components termed a bias and a variance. Large numbers of parameter estimates
raise the variance, whereas incorrect statistical models increase the bias
\citet{Geman1992}%
,
\citet{Gershenfeld1999}%
,
\citet{Duda2001}%
,
\citet{Hastie2001}%
,
\citet{Haykin2009}%
.

Model-free approaches to learning, which are also known as nonparametric
inference methods, assume no particular types of geometric structures, use
large numbers of parameters, and exhibit high estimation variance. For
example, learning machine architectures of multilayer ANNs, which are formed
by arbitrary sets and types of geometric surfaces, vary substantially with
data samples and hidden node configurations. Such variance can only be reduced
with sufficient amounts of data samples
\citet{Geman1992}%
,
\citet{Haykin2009}%
. On the other side of the dilemma, parametric or statistical models that use
improper representations exhibit large modeling biases. For instance, learning
Bayes' decision boundaries from training data drawn from Gaussian
distributions involves the estimation of conic sections or quadratic surfaces.
Discriminant functions that are based on incorrect statistical models do not
achieve the Bayes' error rate
\citet{VanTrees1968}%
,
\citet{Fukunaga1990}%
,
\citet{Duda2001}%
. Most statistical models have few parameters compared to model-free or
nonparametric inference methods.

Given sufficient numbers of parameters and data samples, model-free methods
are expected to achieve the best possible performance for any learning task
given to them. For instance, given enough training data, optimal decision
rules for discriminant functions can be arbitrarily well approximated by
consistent nonparametric estimators such as parzen windows, nearest neighbor
rules, projection pursuit methods, multilayer ANNs, and classification and
regression trees. But, model-free methods are slow to converge, due to
excessively large training sets necessary to reduce estimation variances.
Model-free architectures may also involve infinitely many parameters, which is
an impossible estimation task
\citet{Geman1992}%
.

Because convergence speeds for model-free methods are limited by training set
size, convergence rates cannot be increased by parallel architectures and fast
hardware. The only way to improve convergence speeds is to control the
estimation variance. But, controlling the estimation variance requires the use
of model-based architectures, which may increase the modeling bias
\citet{Geman1992}%
.

\subsubsection{Essence of the Bias/Variance Dilemma}

The essence of the bias/variance dilemma can be summarized as follows.
Model-free architectures based on insufficient data samples are unreliable and
have slow convergence speeds. However, model-based architectures based on
insufficient representations are also unreliable. Except, model-based
architectures with adequate representations are both reliable and have
reasonable convergence speeds. Even so, model-based architectures with
adequate representations are difficult to identify.

All of the above problems indicate that learning unknown functions from data
involves obfuscated problems, which remain to be identified.

\subsection{Prewiring of Important Generalizations}

The limitations imposed by the bias and variance dilemma led
\citet{Geman1992}
to argue that learning complex tasks from training data is essentially
impossible without some a priori introduction of carefully designed biases
into a learning machine's architecture. They also argue that the
identification and exploitation of proper biases or generalizations are the
more fundamental and difficult research issues in neural network modeling applications.%

\citet{Geman1992}
also suggest that the essential challenges in neural modeling applications are
about representation, rather than learning per se. Because most interesting
problems tend to be problems of extrapolation, the only way to avoid having to
densely cover an input space with training points is to prewire\ the important
generalizations, some of which can be achieved through proper data
representations. However, the identification of proper models for complex,
statistical inference tasks is a difficult problem. Consequently, any
model-based scheme for a complex inference problem may be incorrect, that is,
highly biased.

This paper will demonstrate how the identification and exploitation of proper
biases or generalizations enables effective designs of learning machine
architectures. The paper will address the matter of effective statistical
representations for probabilistic, binary, linear classification systems. For
the problem of learning decision boundaries, an important form of proper data
representations involves the identification and exploitation of pattern
vectors which exhibit sufficient class separability, i.e., a negligible
overlap exists between data distributions. In general, \emph{the design of
effective feature vectors which exhibit sufficient class separability is the
most fundamental and difficult problem in the overall design of a statistical
pattern recognition system}
\citet{Fukunaga1990}%
,
\citet{Duda2001}%
.

The essential problem of prewiring the significant statistical representations
into a learning machine's architecture remains ill-defined. What does it
really mean to introduce a carefully designed bias into a learning machine's
architecture? How do we identify the important generalizations for a given
problem? How should these generalizations be pre-wired? How does the
introduction of a proper bias involve the use of model-based estimation?\ This
paper will consider all of these problems in terms of the fundamental modeling
question posed next.

\subsection{The System Representation Problem}

Effective designs of learning machine architectures involve an underlying
system modeling problem
\citet{Naylor1971}%
,
\citet{Gershenfeld1999}%
. In particular, effective designs of model-based statistical architectures
involve the formulation of a mathematical system which simulates essential
stochastic behavior and models key aspects of a real statistical system.
Accordingly, \emph{statistical model formulation for learning machine
architectures involves the development of a mathematically tractable
statistical model that provides a useful representation of a statistical
decision system}.

In general terms, a system is an interconnected set of elements which are
coherently organized in a manner that achieves a useful function or purpose
\citet{Meadows2008}%
. \emph{This implies that encoding relevant aspects of statistical decision
systems within learning machine architectures involves effective
interconnections between suitable sets of coherently organized data points}.
The learning machine architecture examined in this paper will provide
substantial examples of effective interconnections between suitable sets of
coherently organized training points.

\subsection{Suitable Representations for Learning Machine Architectures}

The matter of identifying suitable representations for learning machine
architectures is extremely important. Indeed, for many scientific problems,
there is a natural and elegant way to represent the solution. For example,
each of the well-known special functions, e.g., Legendre polynomials, Bessel
functions, Fourier series, Fourier integrals, etc., have the common motivation
of being most appropriate for certain problems, and quite unsuitable for
others, where each special function represents the relevant aspects of a
physical system
\citet{Keener2000}%
.

Yet, most machine learning methods attempt to approximate unknown functions
with methods that assume no sort of representation, e.g., nonparametric
inference methods
\citet{Geman1992}%
,
\citet{Cherkassky1998}%
,
\citet{Duda2001}%
,
\citet{Hastie2001}%
,
\citet{Haykin2009}%
, or assume representations that are tentative and ill-defined, e.g.,
indefinite interpolations of SVM margin hyperplanes in unknown, high
dimensional spaces
\citet{Boser1992}%
,
\citet{Cortes1995}%
. Likewise, consider the notion of the asymptotic convergence of a learning
machine architecture to some unknown function. Can we picture what this
actually means?

\subsection{Tractable Statistical Models for SVM Architectures}

Tangible representations provide objects and forms which can be seen and
imagined, along with a perspective for seeing and imagining them
\citet{Hillman2012}%
. This paper will develop substantial geometric architectures for linear SVMs,
which are based on a mathematically tractable statistical model that can be
depicted and understood in two and three-dimensional vector spaces, and fully
comprehended in higher dimensions. The statistical model represents the
relevant aspects of a probabilistic, binary linear classification system. The
corresponding geometric architecture is determined by correlated, primal and
dual, algebraic systems of locus equations of interconnected sets of primal
and dual principal eigenaxis components, all of which jointly delineate a
linear decision boundary that is bounded by bilaterally symmetrical borders.

The locations of the primal and dual principal eigenaxis components are
determined by the geometric and statistical properties of a training data
collection. Because a geometric locus of points represents a mathematically
tractable geometric curve or surface, the learning machine architecture is
completely illustrated in two and three-dimensional vector spaces, and is
readily envisioned in higher dimensional vector spaces. The statistical model
involves a dual statistical eigenlocus of principal eigenaxis components
formed by eigen-scaled extreme data points, all of which encode an
eigen-transformed principal location of large covariance. Extreme data points
are located in regions of large covariance between two overlapping or
non-overlapping data distributions. The term dual statistical eigenlocus is
used to describe an eigenlocus, i.e., a characteristic curve, of principal
eigenaxis components which encodes essential geometric underpinnings and
statistical machinery for a statistical decision system. A\ dual statistical
eigenlocus is also referred to as a strong dual principal eigenlocus.

\subsection{A Tractable Dual Locus of Eigen-scaled Data Points}

The sections that follow will motivate and develop a learning machine
architecture that is based on a natural representation of second-order
statistical decision systems. The geometric architecture is based on a dual
statistical eigenlocus of principal eigenaxis components which encodes
likelihoods of extreme data points. The statistical model provides an elegant
solution to difficult interrelated problems that include the bias/variance
dilemma, capacity control, and overfitting or underfitting training data. The
analysis begins with normal vector directions for linear decision boundaries.

\subsubsection{Partially Specified Principal Eigenaxes}

When an optimum decision boundary is a linear curve or surface, the vector
direction deemed most significant is perpendicular to some separating line,
plane, or hyperplane. If this direction can be specified in some manner, then
no other vector directions contribute useful information for linear
discrimination
\citet{Cooper1962}%
. This paper will demonstrate that normal vector directions determine
\emph{partially specified principal eigenaxes}, which provide necessary, but
insufficient, information for linear decision boundary estimates.

\subsubsection{Properly Specified Normal Eigenaxes}

This paper will show that the most significant vector direction for linear
discrimination is specified by the geometric locus of a principal or major
eigenaxis, which will be referred to as a normal eigenaxis. It will be shown
that the magnitude of a normal eigenaxis contains essential information for
effective linear partitioning of pattern vector spaces. Thereby, it will be
shown that direction alone is insufficient for describing separating lines,
planes, or hyperplanes.

This paper will develop a class of mathematically tractable learning machine
architectures that is based on a dual statistical eigenlocus of normal
eigenaxis components, all of which encode principal magnitudes and principal
directions for linear decision boundary estimates. The paper will extend the
fundamental ideas behind the general notion of a geometric locus to develop
symmetrical algebraic systems of primal and dual normal eigenlocus equations
that jointly specify a normal eigenlocus of eigen-scaled extreme data points.
The term\ strong dual normal eigenlocus refers to a dual statistical
eigenlocus of normal eigenaxis components. Extreme data points are\ innermost
data points of large covariance which are located between overlapping or
non-overlapping data distributions. The analyses presented in this paper will
demonstrate how correlated algebraic systems of primal and dual normal
eigenlocus equations provide an estimate of an unknown normal eigenaxis of an
unknown linear decision boundary. Because a strong dual normal eigenlocus is
based on graphs or geometric loci of equations, a strong dual normal
eigenlocus is mathematically tractable.

To motivate the development of a strong dual normal eigenlocus of eigen-scaled
extreme data points, Section $3$ will consider the fundamental limitations of
classical geometric locus methods applied to collections of training data.
These limitations will be described in terms of the geometric locus dilemma
for statistical learning machines, a terminology inspired by
\citet{Geman1992}%
. Existing locus methods are outlined next.

\subsection{The Graph or Locus of an Equation}

The graph or locus of an equation is the locus (place) of all points whose
coordinates are solutions of the equation. Any point whose coordinates are
solutions of a locus equation is on the geometric locus of the equation. Any
given point on a geometric locus possesses a geometric property which is
common to all points on the locus, and \textit{no other points}
\citet{Nichols1893}%
,
\citet{Tanner1898}%
,
\citet{Whitehead1911}%
,
\citet{Eisenhart1939}%
. For example, consider the geometric locus of a circle. A\ circle is a locus
of points $\left(  x,y\right)  $, all of which are at the same distance, the
radius $r$, from a fixed point $\left(  x_{0},y_{0}\right)  $, the center. The
algebraic equation for the geometric locus of a circle in Cartesian
coordinates is:%
\begin{equation}
\left(  x-x_{0}\right)  ^{2}+\left(  y-y_{0}\right)  ^{2}=r^{2}\text{.}
\label{Coordinate Equation of Circle}%
\end{equation}
For any given center $\left(  x_{0},y_{0}\right)  $ and radius $r$, only those
coordinates $\left(  x,y\right)  $ that satisfy Eq.
(\ref{Coordinate Equation of Circle}) contribute to the geometric locus of the
specified circle
\citet{Eisenhart1939}%
.

The identification of the geometric property of a locus of points is a central
problem in coordinate geometry. The inverse problem finds the algebraic form
of an equation, whose solution gives the coordinates of all of the points on a
locus which has been defined geometrically. Geometric figures are defined in
two ways: $(1)$ as a figure with certain known properties, and $(2)$ as the
path of a point which moves under known conditions
\citet{Nichols1893}%
,
\citet{Tanner1898}%
.

\subsection{Methods for Solving Locus Problems}

Methods for solving geometric locus problems hinge on the identification of
algebraic and geometric correlations for a given locus of points. Geometric
correlations between a set of points which lie on a definite curve or surface,
correspond to geometric and algebraic constraints that are satisfied by the
coordinates of any point on a given locus
\citet{Nichols1893}%
,
\citet{Tanner1898}%
,
\citet{Whitehead1911}%
,
\citet{Eisenhart1939}%
.

Finding the algebraic form of an equation for a given geometric figure or
locus is often a difficult problem. However, because some type of
relationships exist between a geometric locus and its algebraic form, careful
examination into the point and coordinate relationships specified by the
algebraic form of a locus equation may yield additional insight into the
uniform property exhibited by a geometric locus of points
\citet{Tanner1898}%
.

\subsection{Changing the Loci of Reference Axes}

The algebraic form of a locus equation hinges on both the geometric property
and the frame of reference (coordinate system) of the locus. Thereby, changing
the position of the coordinate axes changes both $(1)$ the algebraic form of
the locus that references the new axes and $(2)$ the coordinates of any point
on the locus. It follows that the equation of a locus and the identification
of the geometric property of the locus can be greatly simplified by changing
the position of the axes to which the locus of points is referenced
\citet{Nichols1893}%
,
\citet{Tanner1898}%
,
\citet{Eisenhart1939}%
.

\subsection{Geometric Locus of a Straight Line}

The geometric locus of every equation of the first degree is a straight line.
Only two geometric conditions are deemed necessary to determine the equation
of a particular line. Either a line should pass through two given points, or
should pass through a given point and have a given slope. Standard equations
of a straight line include the point-slope, slope-intercept, two-point,
intercept, and normal forms.

The general equation of the first degree in two coordinate variables $x$ and
$y$ has the form:%
\[
Ax+By+C=0\text{,}%
\]
where $A$, $B$, $C$ are constants which may have any real values, subject to
the restriction that $A$ and $B$ cannot both be zero
\citet{Nichols1893}%
,
\citet{Tanner1898}%
,
\citet{Eisenhart1939}%
.

Excluding the point, a straight line appears to be the simplest type of
geometric locus. Nonetheless, the uniform geometric property of a straight
line remains undefined. This paper will soon identify several, correlated
uniform properties exhibited by all of the points on a linear locus.

The next section will outline the algebraic, geometric, and statistical
essence of the geometric locus dilemma for statistical learning machines. The
essence of the dilemma will be defined for support vector learning machines,
and a method will be outlined that resolves the dilemma for linear and
polynomial kernel SVMs. By way of motivation, simulation studies will be
presented which demonstrate that linear kernel SVMs learn Bayes' decision
boundaries for training data drawn from overlapping Gaussian distributions.

\section{The Geometric Locus Dilemma}

An insoluble dilemma in the design and development of learning machine
architectures will now be identified. The dilemma underlies interrelated
problems that include the bias/variance dilemma, capacity control, and
overfitting or underfitting training data. The underlying issue involves
determining an effective fit of an $N$-dimensional set of $d$-dimensional
random data points to geometric loci in $d$-dimensional Cartesian space. More
specifically, the problem involves defining suitable fits for given
collections of $N\times d$ random vector coordinates to algebraic equations of
prespecified (explicit) geometric loci that reference arbitrary Cartesian
coordinate systems.

A\ classical locus of points is an explicit, and thus fixed, geometric
configuration of vectors, whose Cartesian coordinate locations are determined
by, and therefore satisfy, an algebraic equation. Curves or surfaces of
classical locus equations are determined by properties of geometric loci with
respect to coordinate axes of arbitrary Cartesian coordinate systems. Thereby,
an algebraic equation of a classical locus of points generates an explicit
point, curve, or surface, in an arbitrarily specified Cartesian space. It
follows that any point on a classical geometric locus naturally exhibits the
uniform property of the locus. Indeed, any point on a classical geometric
locus satisfies the uniform property of the given locus in an innate manner.
It will now be argued that fitting collections of training data to classical
geometric locus equations involves an impossible estimation task.

\subsection{An Impossible Estimation Task}

Consider fitting a collection of training data to some classical or standard
locus equation(s). It follows that any given second-order curve or surface
must pass through any training points on the specified locus. Therefore, any
training point whose coordinate locations do not satisfy the given locus
equation and correlated geometric property of the specified locus is simply
not on the given curve or surface. Likewise, training points that are not on a
given curve or surface do not contribute to the locus of a specified curve or surface.

Given the correlated algebraic and geometric constraints on a classical locus
of points, it follows that any attempt to \emph{fit}\textit{\ }an
$N$-dimensional set of $d$-dimensional random data points to the equation(s)
of a classical geometric locus, involves the unfeasible problem of determining
an \emph{effective constellation} of an $\left(  N-M\right)  \times d$ subset
of $N\times d$ random vector coordinates that $(1)$ inherently satisfy
prespecified, fixed magnitude (length) constraints on each of the respective
$d$ Cartesian coordinate axes, and thereby $(2)$ generate explicit, fixed
points, curves, or surfaces in $%
\mathbb{R}
^{d}$. Such an estimation process is clearly unfeasible. It follows that
fitting collections of random data points to classical locus equations is an
impossible estimation task. The essence of the geometric locus dilemma for
support vector learning machines is defined next.

\subsection{SVMs and the Geometric Locus Dilemma}

So far, it has been argued that the insoluble aspect of the geometric locus
dilemma concerns finding suitable fits for collections of random vector
coordinates to algebraic equations of partially configured geometric loci that
reference arbitrary Cartesian coordinate systems. It has also been argued that
such estimation processes are impracticable methods that involve impossible
estimation tasks. The next section will argue that SVM capacity control
involves an impossible estimation task.

\subsubsection{Linear Interpolation Using Random Slack Variables}

SVM methods are based on the idea of specifying a pair of maximally separated
linear curves or surfaces that interpolate two sets of data points
\citet{Cortes1995}%
,
\citet{Bennett2000}%
,
\citet{Cristianini2000}%
,
\citet{Hastie2001}%
,
\citet{Scholkopf2002}%
. Given two non-overlapping sets of data points, linear SVM finds a pair of
maximally separated linear decision borders, such that the decision borders
pass through data points called support vectors. For example, in Section $2$,
Fig. $2$ illustrates how SVM\ decision borders interpolate two sets of data
points, where the green decision borders exhibit the largest geometric margin.

Identifying interpolation methods that provide effective fits of separating
lines, planes, or hyperplanes involves the long standing problem of fitting
linear decision boundaries to overlapping sets of data points
\citet{Cover1965}%
. Soft margin linear SVM\ is said to resolve this problem by means of
non-negative random slack variables $\xi_{i}\geq0$, each of which allows a
correlated data point $\mathbf{x}_{i}$, that lies between or beyond a pair of
linear decision borders, to satisfy a linear border. Nonlinear kernel SVMs
also employ non-negative random slack variables, each of which allows a
transformed,\ correlated data point to satisfy a hyperplane\textit{\ }decision
border in some higher dimensional feature space
\citet{Cortes1995}%
,
\citet{Bennett2000}%
,
\citet{Cristianini2000}%
,
\citet{Hastie2001}%
,
\citet{Scholkopf2002}%
.

\emph{This implies that non-negative random slack variables, for both linear
and nonlinear kernel SVMs, encode effective distances of data points from
unknown linear curves or surfaces}. Clearly, this is an impossible estimation
task. It follows that computing effective values for $l$ non-negative random
slack variables $\left\{  \xi_{i}|\xi_{i}\geq0\right\}  _{i=1}^{l}$ is an
impossible estimation task. Figure $3$ depicts the insoluble aspect of the
geometric locus dilemma for linear kernel SVMs.

\begin{center}%
\begin{center}
\includegraphics[
natheight=7.499600in,
natwidth=9.999800in,
height=3.2897in,
width=4.3777in
]%
{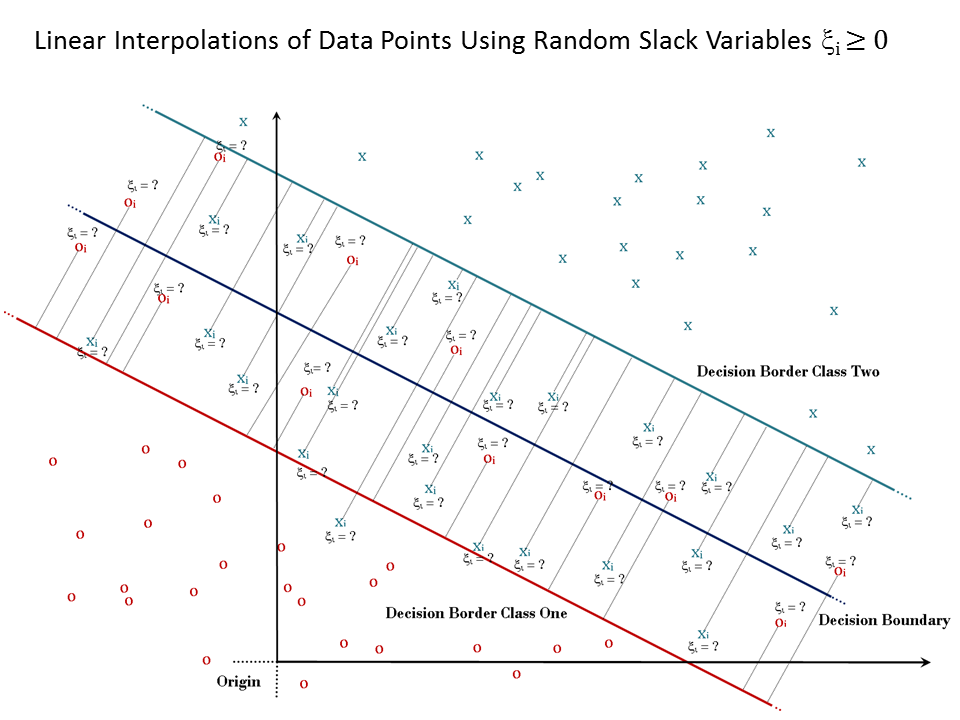}%
\end{center}

\end{center}

\begin{flushleft}
Figure $3$: Illustration of the geometric locus dilemma for linear kernel
SVMs: $l$ non-negative random slack variables $\left\{  \xi_{i}|\xi_{i}%
\geq0\right\}  _{i=1}^{l}$ must be estimated for effective linear
interpolations of the $l-k$ blue $\left\{  \mathbf{x}_{i}\right\}
_{i=1}^{l-k}$ and the $l-k+1$ red $\left\{  \mathbf{o}_{i}\right\}
_{i=l-k+1}^{l}$ overlapping data points. Computing effective values for these
$l$ random slack variables is an impossible estimation task.
\end{flushleft}

\subsection{Ill-defined SVM Architectures}

SVM architectures and regularization parameters are largely ill-defined. For
example, the polynomial degree, kernel width, and regularization parameters of
nonlinear kernel SVMs, and the regularization parameters of soft margin linear
kernel SVMs, are mostly determined by trial and error
\citet{Byun2002}%
,
\citet{Eitrich2006}%
,
\citet{Liang2011}%
. Likewise, the total citation count for the $1998$ article titled
\textit{A\ Tutorial on Support Vector Machines for Pattern Recognition }%
\citet{Burges1998}
currently exceeds $15,000$.

This paper will demonstrate that resolving the geometric locus dilemma for
linear and polynomial kernel SVMs involves two correlated and fundamental
problems that involve graphs or loci of properly specified locus equations.
This paper will show that effective design of linear kernel SVM architectures
are based on $(1)$ a suitable set of geometric loci that provide the basis of
a statistical decision system, and $(2)$ a properly specified algebraic system
of locus equations for a given statistical decision system. This paper will
also establish that the fundamental geometric locus of interest for linear
kernel SVM architectures is the \emph{locus of a principal eigenaxis}.
Polynomial kernel SVMs will be extensively examined in an upcoming paper. At
this time, the paper will motivate taking an extensive look under the hood of
linear kernel SVMs.

\subsection{How Does Linear Kernel SVM Learn from Data?}

Linear kernel SVMs have been applied to training data drawn from overlapping
Gaussian distributions
\citet{Reeves2007}%
,
\citet{Reeves2009}%
. The results obtained from these simulation studies indicate that linear
kernel SVM learns Bayes' decision boundaries for highly overlapping Gaussian
data sets. Bayes' discriminant functions are the gold standard for linear
discrimination tasks. Bayes' classification error rate is the best error rate
that can be achieved by any classifier
\citet{Fukunaga1990}%
,
\citet{Duda2001}%
. Results from two of the simulation studies are outlined below.

\subsubsection{Classification Example One}

Gaussian data set one has the covariance matrix:%

\[
\mathbf{\Sigma}_{1}=\mathbf{\Sigma}_{2}=%
\begin{pmatrix}
0.5 & 0\\
0 & 2
\end{pmatrix}
\text{,}%
\]
and the mean vectors $\mathbf{\mu}_{1}=%
\begin{pmatrix}
3, & 0.5
\end{pmatrix}
^{T}$ and $\mathbf{\mu}_{2}=%
\begin{pmatrix}
3, & -0.5
\end{pmatrix}
^{T}$. The Bayes' decision boundary%
\[
x_{2}=0\text{,}%
\]
which is depicted in Fig. $4$, enforces the Bayes' error rate of $36.5\%$.

\begin{center}%
\begin{center}
\includegraphics[
natheight=7.385500in,
natwidth=14.791700in,
height=2.2866in,
width=4.5515in
]%
{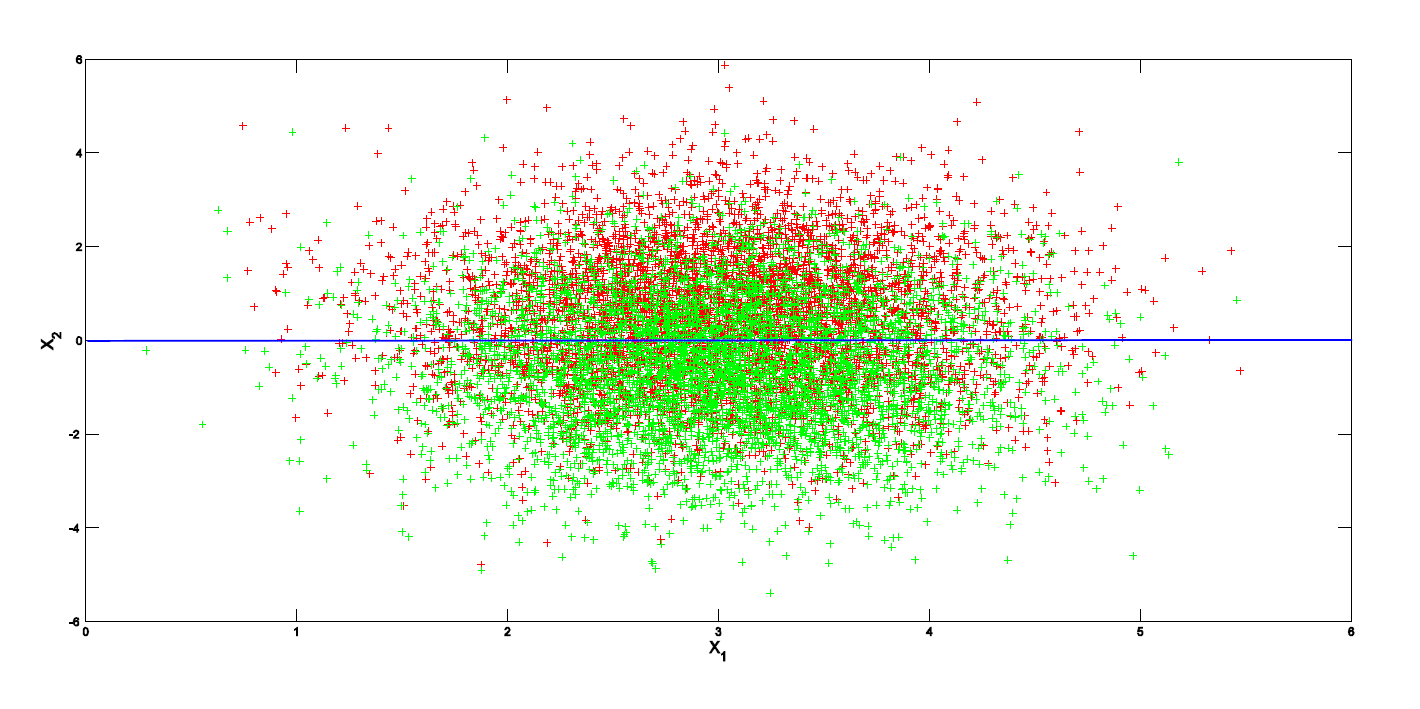}%
\end{center}

\end{center}

\begin{flushleft}
Figure $4$: The Bayes' decision boundary for the overlapping Gaussian data
sets of classification example one.
\end{flushleft}

Figure $5$ illustrates that linear kernel SVM learns the Bayes' decision
boundary for the overlapping Gaussian data in classification example one.
Linear kernel SVM uses $596$ support vectors ($99\%$ of the training data) to
learn the Bayes' decision boundary depicted in Fig. $4$.

\begin{center}%
\begin{center}
\includegraphics[
natheight=7.864600in,
natwidth=15.155800in,
height=2.2866in,
width=4.3803in
]%
{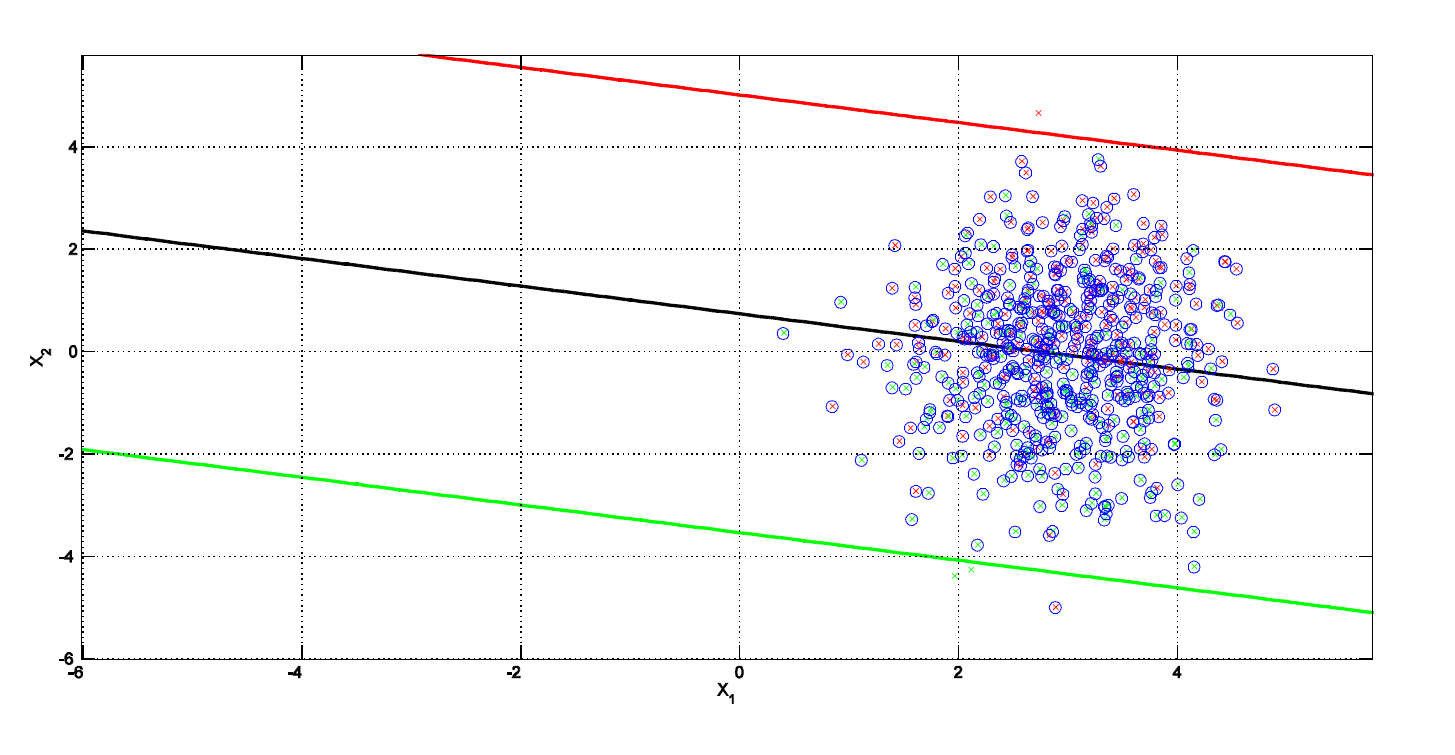}%
\end{center}

\end{center}

\begin{flushleft}
Figure $5$: Linear kernel SVM learns the Bayes' decision boundary for the
overlapping Gaussian data sets of classification example one. Each support
vector is enclosed in a blue circle.
\end{flushleft}

\subsubsection{Classification Example Two}

Gaussian data set two has the covariance matrix:%
\[
\mathbf{\Sigma}_{1}=\mathbf{\Sigma}_{2}=%
\begin{pmatrix}
0.95 & 0.45\\
0.45 & 0.35
\end{pmatrix}
\text{,}%
\]
and the mean vectors $\mathbf{\mu}_{1}=%
\begin{pmatrix}
3, & 0.25
\end{pmatrix}
^{T}$ and $\mathbf{\mu}_{2}=%
\begin{pmatrix}
3, & -0.25
\end{pmatrix}
^{T}$. The Bayes' decision boundary%
\[
x_{2}=0.47x_{1}-1.42\text{,}%
\]
which is depicted in Fig. $6$, enforces the Bayes' error rate of $25\%$.

\begin{center}%
\begin{center}
\includegraphics[
natheight=12.396200in,
natwidth=16.499800in,
height=3.039in,
width=4.0352in
]%
{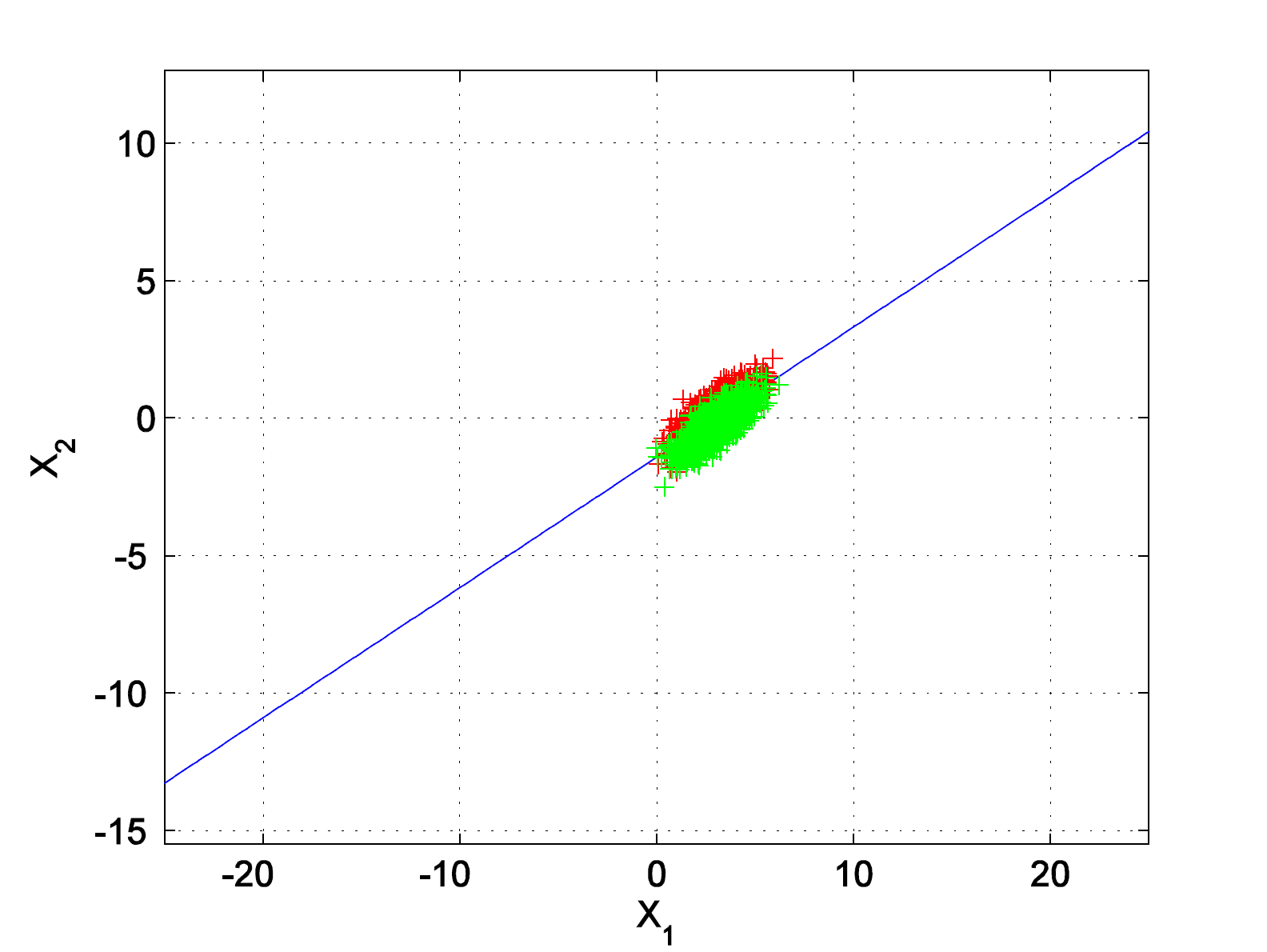}%
\end{center}

\end{center}

\begin{flushleft}
Figure $6$: The Bayes' decision boundary for the overlapping Gaussian data
sets of classification example two.
\end{flushleft}

Figure $7$ illustrates that linear kernel SVM also learns the Bayes' decision
boundary for the overlapping Gaussian data in classification example two.
Linear kernel SVM uses $547$ support vectors ($91\%$ of the training data) to
learn the Bayes' decision boundary depicted in Fig. $6$.

\begin{center}%
\begin{center}
\includegraphics[
natheight=7.604300in,
natwidth=14.937900in,
height=2.2866in,
width=4.4642in
]%
{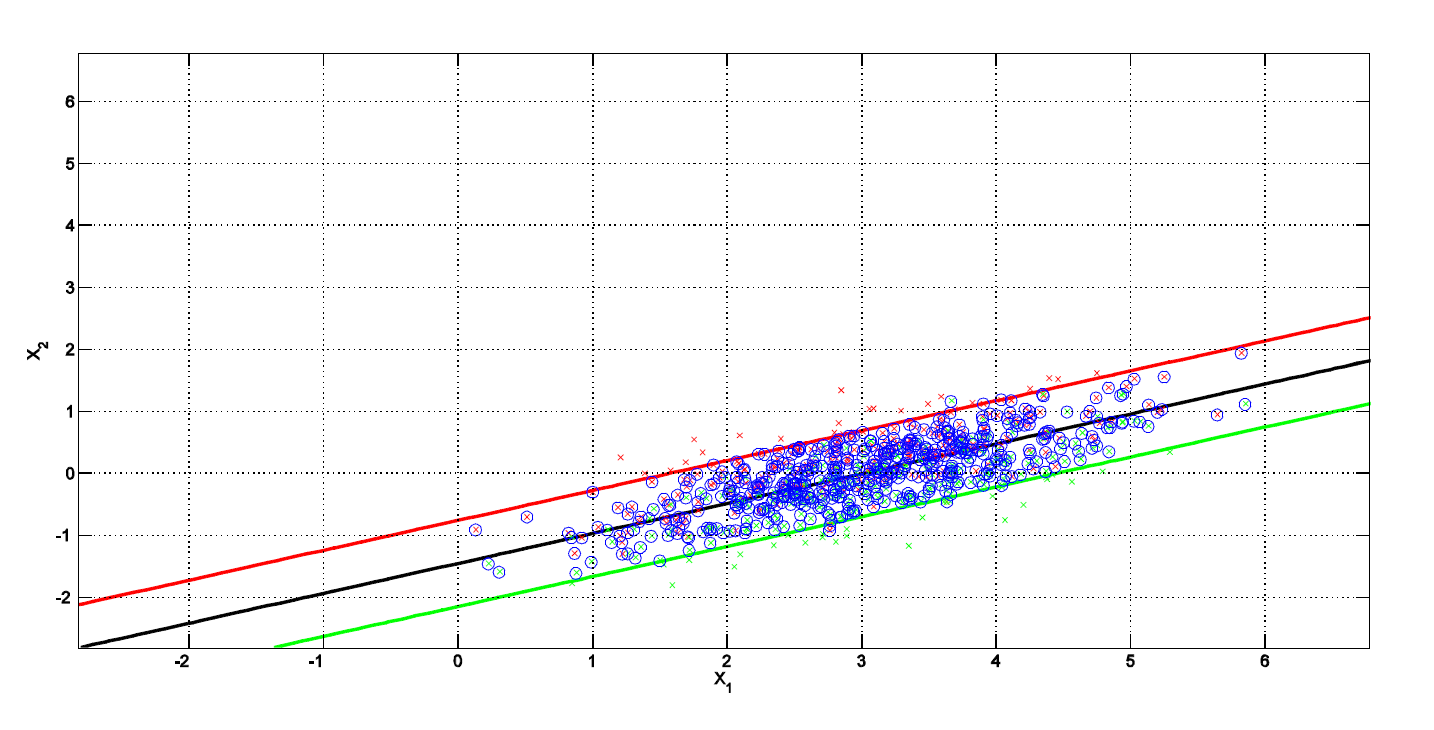}%
\end{center}

\end{center}

\begin{flushleft}
Figure $7$: Linear kernel SVM learns the Bayes' decision boundary for the
overlapping Gaussian data sets in classification example two. Each support
vector is enclosed in a blue circle.
\end{flushleft}

Consideration of the above simulation studies \emph{motivates an extensive
investigation into what is really happening under the hood of linear kernel
SVMs}. How does linear SVM\ actually learn Bayes' linear decision boundaries
for overlapping Gaussian data distributions? How does the geometric
architecture of linear kernel SVM encode Bayes' likelihood ratio? How do we
describe linear kernel SVM architectures? What types of geometric
underpinnings make up the statistical machinery of linear kernel SVM architectures?

All of these questions will be resolved in the sections that follow. The next
section of the paper will begin the process of taking a comprehensive look
under the hood of linear kernel SVMs. Section $4$ will develop an elegant
principal eigen-coordinate system that describes all forms of linear loci.

\section{An Elegant Principal Eigen-coordinate System for Linear Loci}

What types of geometric underpinnings and statistical machinery are encoded
within linear kernel SVM architectures? How does linear SVM learn optimal
linear decision boundaries for overlapping data distributions? All of these
questions will be answered by describing the linear SVM method in a geometric
locus framework. Moreover, answers to these question will identify an
estimation method that resolves the geometric locus dilemma for linear SVMs.
It will be shown that learning linear decision boundaries involves strong
duality relationships, between a statistical eigenlocus of principal eigenaxis
components and its algebraic forms, in primal and dual, correlated inner
product, i.e., Hilbert, spaces. Thereby, a computer-implemented method will be
formulated that generates regularized, data-driven geometric architectures
which encode Bayes' likelihood ratio for common covariance data and a robust
likelihood ratio for all other data distributions.

Section $4$ begins by introducing and developing a locus equation of the
principal eigenaxis of a straight line. Overviews of the equations of a
straight line can be found in
\citet{Nichols1893}%
,
\citet{Tanner1898}%
,
\citet{Eisenhart1939}%
, and
\citet{Davis1973}%
. The vector equation of a straight line, which is outlined in
\citet{Davis1973}%
, is the hinge point and central principle for the chain of arguments on
linear decision boundary estimates that follow.

\subsection{Locus Equations of a Normal Eigenaxis}

At this time, a\ few remarks on notation are necessary. Strictly speaking, a
vector $\mathbf{x}$ is a directed straight line segment that emanates from a
chosen point $P_{0}$, termed the origin, such that the endpoint of the
directed straight line segment, termed the tip, defines a real specific point
$P$. Thereby, a point is an inherent part of a vector, such that correlated
points $P_{\mathbf{x}}$ and vectors $\mathbf{x}$ both describe the same
ordered pair of real numbers in the real Euclidean plane or the same ordered
$d$-tuple of real numbers in real Euclidean space. Given that correlated
points $P_{\mathbf{x}}$ and vectors $\mathbf{x}$ specify equivalent ordered
pairs or $d$-tuples of real numbers, correlated points $P_{\mathbf{x}}$ and
vectors $\mathbf{x}$ have common geometric representations as points in $%
\mathbb{R}
^{2}$ or $%
\mathbb{R}
^{d}$. Depending on the geometric context, an ordered pair of real numbers or
an ordered $d$-tuple of real numbers will be referred to as either a point or
a vector. The analysis that follows will denote points and vectors by
$\mathbf{x}$.

A locus equation is now introduced that describes lines simply in terms of
algebraic correlations between the points on a line. Moreover, the\ locus
equation contains no constants or parameters. It will be argued that one of
the points on any given line determines the principal eigenaxis of the line.
The principal eigenaxis of a linear locus will be called a \emph{normal
eigenaxis}. Several locus equations of a normal eigenaxis will be developed,
all of which describe geometric loci of lines, planes, and hyperplanes. All of
these algebraic expressions will be used to identify the correlated uniform
properties exhibited by any point on a linear locus.

The development of normal eigenaxis locus equations and the identification of
the correlated uniform properties exhibited by any point on a linear locus,
will lead to far-reaching consequences for the matter of linear decision
boundary estimates. The locus equations of a normal eigenaxis and the
invariant geometric properties of a normal eigenaxis will be used to motivate
and develop symmetric primal and dual algebraic systems of strong dual normal
eigenlocus equations, which will be demonstrated to jointly specify robust and
optimal estimates of separating lines, planes, and hyperplanes for a large
number of data distributions, including homogeneous data distributions.

The paper will now derive the primary locus equation of a normal eigenaxis
that describes geometric loci of lines, planes, and hyperplanes. By way of
introduction, Fig. $8$ depicts the geometric locus of a line in the real
Euclidean plane $%
\mathbb{R}
^{2}$.

\begin{center}%
\begin{center}
\includegraphics[
natheight=7.499600in,
natwidth=9.999800in,
height=3.2897in,
width=4.3777in
]%
{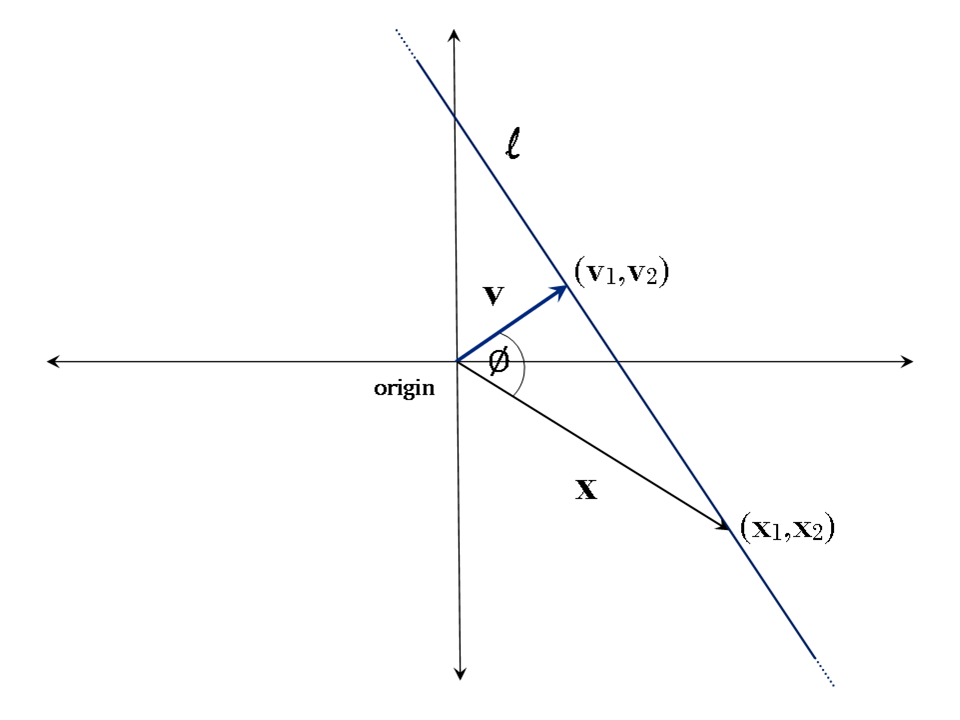}%
\end{center}

\end{center}

\begin{flushleft}
Figure $8$: Illustration of an elegant principal eigen-coordinate system for
lines that is readily generalized to planes and hyperplanes. Any vector
$\mathbf{x}_{i}=%
\begin{pmatrix}
x_{1i}, & x_{2i}%
\end{pmatrix}
^{T}$ whose tip $%
\begin{pmatrix}
x_{1i}, & x_{2i}%
\end{pmatrix}
$ is on the line $l$ explicitly and exclusively references the vector
$\mathbf{v}=%
\begin{pmatrix}
v_{1}, & v_{2}%
\end{pmatrix}
^{T}$, which is shown to be the principal, i.e., normal, eigenaxis of the line
$l$.
\end{flushleft}

\subsection{The Normal Eigenaxis of a Linear Locus}

Let $\mathbf{v}\triangleq%
\begin{pmatrix}
v_{1}, & v_{2}%
\end{pmatrix}
^{T}$ be a fixed vector in the real Euclidean plane and consider the line $l$
at the tip of $\mathbf{v}$ that is perpendicular to $\mathbf{v}$. To wit, the
vector $\mathbf{v}$ is a point on the line $l$. Therefore, the coordinates
$\left(  v_{1},v_{2}\right)  $ of $\mathbf{v}$ delineate and satisfy $l$. In
addition, consider an arbitrary vector $\mathbf{x}\triangleq%
\begin{pmatrix}
x_{1}, & x_{2}%
\end{pmatrix}
^{T}$ whose tip is also on the line $l$. Thereby, the coordinates $\left(
x_{1},x_{2}\right)  $ of the point $\mathbf{x}$ also delineate and satisfy
$l$. Finally, let $\phi$ be the acute angle between the vectors $\mathbf{v}$
and $\mathbf{x}$, satisfying $0\leq\phi\leq\pi/2$ and the algebraic
relationship $\cos\phi=\frac{\left\Vert \mathbf{v}\right\Vert }{\left\Vert
\mathbf{x}\right\Vert }$.

Using all of the above assumptions, it follows that the locus of points
$\left(  x_{1},x_{2}\right)  $ on the line $l$ is described by the functional
locus equation:%
\begin{equation}
\mathbf{x}^{T}\mathbf{v}=\left\Vert \mathbf{x}\right\Vert \left\Vert
\mathbf{v}\right\Vert \cos\phi\text{,} \label{Linear Locus Functional}%
\end{equation}
which is the vector equation of a line
\citet{Davis1973}%
. It will shortly be demonstrated that the vector $\mathbf{v}$ is the
principal eigenaxis of the line $l$.

\subsection{Fundamental Equation of a Linear Locus}

Take any fixed vector $\mathbf{v}$, and consider the line $l$ that is
described by Eq. (\ref{Linear Locus Functional}), where the axis of
$\mathbf{v}$ is perpendicular to the specified line $l$ and the tip of
$\mathbf{v}$ is on $l$. Given that any vector $\mathbf{x}_{i}$ with its tip on
the given line $l$ satisfies the algebraic relationship:%
\[
\left\Vert \mathbf{x}_{i}\right\Vert \cos\phi_{i}=\left\Vert \mathbf{v}%
\right\Vert \text{,}%
\]
with the fixed vector $\mathbf{v}$, it follows that the geometric locus of a
line $l$ is also described by the locus equation:%
\begin{equation}
\mathbf{x}^{T}\mathbf{v}=\left\Vert \mathbf{v}\right\Vert ^{2}\text{.}
\label{Normal Eigenaxis Functional}%
\end{equation}
Equations (\ref{Linear Locus Functional}) and
(\ref{Normal Eigenaxis Functional}) are readily generalized to planes $p$ and
hyperplanes $h$ in $%
\mathbb{R}
^{d}$ by letting $\mathbf{v}\triangleq%
\begin{pmatrix}
v_{1}, & v_{2}, & \cdots, & v_{d}%
\end{pmatrix}
^{T}$ and $\mathbf{x}=%
\begin{pmatrix}
x_{1}, & x_{2}, & \cdots, & x_{d}%
\end{pmatrix}
^{T}$. Given that Eq. (\ref{Normal Eigenaxis Functional}) contains no
constants or parameters, it follows that Eq.
(\ref{Normal Eigenaxis Functional}) is the fundamental equation of a linear locus.

Assuming that $\left\Vert \mathbf{v}\right\Vert \neq0$, Eq.
(\ref{Normal Eigenaxis Functional}) can also be written as:%
\begin{equation}
\frac{\mathbf{x}^{T}\mathbf{v}}{\left\Vert \mathbf{v}\right\Vert }=\left\Vert
\mathbf{v}\right\Vert \text{.} \label{Normal Form Normal Eigenaxis}%
\end{equation}
The axis $\mathbf{v}/\left\Vert \mathbf{v}\right\Vert $ has length $1$ and
points in the direction of the vector $\mathbf{v}$, such that $\left\Vert
\mathbf{v}\right\Vert $ is the distance of a specified line $l$, plane $p$, or
hyperplane $h$ to the origin. Using Eq. (\ref{Normal Form Normal Eigenaxis}),
it follows that the distance $\mathbf{\Delta}$ of a line, plane, or hyperplane
from the origin is specified by the magnitude $\left\Vert \mathbf{v}%
\right\Vert $ of the axis $\mathbf{v}$.

It is claimed that the vector $\mathbf{v}$ provides an exclusive, intrinsic
reference axis for a linear locus of points. An intrinsic axis which coincides
as an exclusive, fixed reference axis for coordinates, delineates curves or
surfaces that are mirror images of each other. Such intrinsic reference axes
are principal eigenaxes of conic sections and quadratic surfaces
\citet{Hewson2009}%
.

\subsection{Major Intrinsic Axes of Second-order Loci}

All of the second-order geometric loci are characterized by one or more
intrinsic axes which are represented by the eigenvectors of a real symmetric
matrix associated with a quadratic form. This paper claims that a major
intrinsic axis, which may coincide as an exclusive, fixed reference axis, is
an inherent part of any second-order geometric locus. Examples of major
intrinsic axes include the major axes, i.e., principal eigenaxes, of ellipses,
parabolas, and hyperbolas. In general, geometric shapes and orientations of
conic sections and quadratic surfaces are described by eigenvalues and
eigenaxes
\citet{Hewson2009}%
.

Recall that the algebraic form of a locus equation hinges on both the
geometric property and the frame of reference of the locus. Because coordinate
versions of geometric loci reference Cartesian coordinate systems, the
positions of axes of coordinates to which a given locus is referenced are
arbitrary
\citet{Nichols1893}%
,
\citet{Tanner1898}%
,
\citet{Eisenhart1939}%
. However, this paper claims that the positions of the major axes of the
second-order geometric loci are not arbitrary. It can be argued that the locus
of a major axis is characteristic of a particular second-order locus of
points. It can also be demonstrated that a principal eigenaxis offers an
elegant principal eigen-coordinate system for a conic section or quadratic
surface. It will shortly be demonstrated that the locus of the major intrinsic
axis\ of a linear locus offers an elegant principal eigen-coordinate system
that is characteristic of a specific locus of points.

\subsection{The Principal Eigenaxis of a Linear Locus}

Figure $8$ shows how the geometric configuration of a fixed vector
$\mathbf{v}$ determines the geometric configuration of a linear locus $l$. It
will now be demonstrated that the axis $\mathbf{v}$ denoted in Eqs
(\ref{Linear Locus Functional}), (\ref{Normal Eigenaxis Functional}), and
(\ref{Normal Form Normal Eigenaxis}) is the principal eigenaxis of linear loci.

All of the major axes of conic sections and quadratic surfaces are major
intrinsic axes which may also coincide as exclusive, fixed reference axes
\citet{Nichols1893}%
,
\citet{Tanner1898}%
,
\citet{Eisenhart1939}%
. In order to demonstrate that $\mathbf{v}$ is the principal eigenaxis of
linear loci, it must be shown that $\mathbf{v}$ is a major intrinsic axis
which is also a reference axis. It will first be argued that $\mathbf{v}$ is a
major intrinsic axis for a linear locus of points.

Using the definitions of Eqs (\ref{Linear Locus Functional}),
(\ref{Normal Eigenaxis Functional}), or (\ref{Normal Form Normal Eigenaxis}),
the axis $\mathbf{v}$ is a major intrinsic axis because all of the points
$\mathbf{x}$ on a linear locus satisfy identical algebraic and
geometric\ constraints related to the locus of $\mathbf{v}$ that are
inherently specified by Eqs (\ref{Linear Locus Functional}),
(\ref{Normal Eigenaxis Functional}), and (\ref{Normal Form Normal Eigenaxis}).
Therefore, $\mathbf{v}$ is a major intrinsic axis of a linear locus. The
uniform algebraic and geometric constraints satisfied by all of the points on
a linear locus determine the uniform properties exhibited by each point on the
linear locus.

Again using the definitions of Eqs (\ref{Linear Locus Functional}),
(\ref{Normal Eigenaxis Functional}), or (\ref{Normal Form Normal Eigenaxis}),
the axis $\mathbf{v}$ is an exclusive reference axis because the uniform
properties possessed by all of the points $\mathbf{x}$ on a linear locus are
defined with respect to the axis of $\mathbf{v}$. Therefore, all of the points
$\mathbf{x}$ on a given line, plane, or hyperplane, explicitly and exclusively
reference the major intrinsic axis $\mathbf{v}$ of the linear locus. It is
concluded that the vector $\mathbf{v}$ provides an exclusive, fixed reference
axis for a linear locus.

In conclusion, it has been demonstrated that the vector $\mathbf{v}$ is a
major intrinsic axis that coincides as an exclusive, fixed reference axis for
a linear locus. It follows that $\mathbf{v}$ is the major axis of linear loci.
It is concluded that the vector $\mathbf{v}$ denoted in Eqs
(\ref{Linear Locus Functional}), (\ref{Normal Eigenaxis Functional}), and
(\ref{Normal Form Normal Eigenaxis}) is the principal eigenaxis of linear
curves and plane or hyperplane surfaces in $%
\mathbb{R}
^{d}$.

It will now be argued that the locus of a principal eigenaxis $\mathbf{v}$ is
unique. Take any given line $l$, plane $p$, or hyperplane $h$. Using the
definitions of Eqs (\ref{Linear Locus Functional}),
(\ref{Normal Eigenaxis Functional}), or (\ref{Normal Form Normal Eigenaxis}),
it follows that the given line $l$, plane $p$, or hyperplane $h$ is
perpendicular to the principal eigenaxis $\mathbf{v}$ of the specified line
$l$, plane $p$, or hyperplane $h$, at the tip of the principal eigenaxis
$\mathbf{v}$. Moreover, the line $l$, plane $p$, or hyperplane $h$ is
perpendicular to only one\textit{\ }major axis $\mathbf{v}$, at the tip of
that major axis $\mathbf{v}$, which is uniquely specified by the locus of
$\mathbf{v}$. It is concluded that the locus of a principal eigenaxis is
unique for any given linear locus of points.

The vector $\mathbf{v}$ will be referred to as the normal eigenaxis of linear
curves and surfaces. Equation (\ref{Normal Form Normal Eigenaxis}) will now be
used to develop a coordinate form locus equation, which will be used to
identify a uniform property which is exhibited by any point on a linear curve
or surface.

\subsection{Coordinate Form Locus Equation of a Unit Normal Eigenaxis}

Given Eq. (\ref{Normal Form Normal Eigenaxis}), it follows that any line $l$
in the Euclidean plane $%
\mathbb{R}
^{2}$ and any plane $p$ or hyperplane $h$ in Euclidean space $%
\mathbb{R}
^{d}$ is described by the locus equation%
\begin{equation}
\mathbf{x}^{T}\mathbf{u}_{N_{e}}\mathbf{=\Delta}\text{\textbf{,}}
\label{normal form linear functional}%
\end{equation}
where $\mathbf{u}_{N_{e}}$ is a unit length normal eigenaxis that is
perpendicular to $l$, $p$, or $h$, and $\mathbf{\Delta}$ denotes the distance
of $l$, $p$, or $h$ to the origin. The unit eigenvector $\mathbf{u}_{N_{e}}$
specifies the direction of a normal eigenaxis of a linear curve or surface,
while the distance $\mathbf{\Delta}$ of a line, plane, or hyperplane from the
origin is specified by the magnitude $\left\Vert \mathbf{v}\right\Vert $ of
its normal eigenaxis $\mathbf{v}$.

Now express $\mathbf{u}_{N_{e}}$ in terms of standard orthonormal basis
vectors%
\[
\left\{  \mathbf{e}_{1}=\left(  1,0,\ldots,0\right)  ,\ldots,\mathbf{e}%
_{d}=\left(  0,0,\ldots,1\right)  \right\}
\]
so that%
\[
\mathbf{u}_{N_{e}}=\cos\alpha_{1}\mathbf{e}_{1}+\cos\alpha_{2}\mathbf{e}%
_{2}+\cdots+\cos\alpha_{d}\mathbf{e}_{d}\text{,}%
\]
where $\cos\alpha_{i}$ are the direction cosines between $\mathbf{u}_{N_{e}}$
and $\mathbf{e}_{i}$. Each $\cos\alpha_{i}$ is the $i^{\text{th}} $ component
of the unit normal eigenaxis $\mathbf{u}_{N_{e}}$ along the coordinate axis
$\mathbf{e}_{i}$, where each eigen-scale $\cos\alpha_{i}$ is said to be normalized.

Substitution of the expression for $\mathbf{u}_{N_{e}}$ into Eq.
(\ref{normal form linear functional}) produces a coordinate form locus
equation%
\begin{equation}
x_{1}\cos\alpha_{1}+x_{2}\cos\alpha_{2}+\cdots+x_{d}\cos\alpha_{d}%
=\mathbf{\Delta}\text{,} \label{Unit Normal Coordinate Form Equation}%
\end{equation}
which is satisfied by the eigen-transformed coordinates $\left(  \cos
\alpha_{1}x_{1},\ldots,\cos\alpha_{d}x_{d}\right)  $ of all of the points
$\mathbf{x}$ on the geometric locus of a line, plane, or hyperplane. Equation
(\ref{Unit Normal Coordinate Form Equation}) is the well known coordinate
equation version of a linear locus%
\[
\cos\alpha_{1}x_{1}+\cos\alpha_{2}x_{2}+\cdots+\cos\alpha_{d}x_{d}%
=\mathbf{\Delta}\text{,}%
\]
which is usually written as%
\[
Ax_{1}+Bx_{2}+\ldots+Nx_{N}=P\text{.}%
\]
Equation (\ref{Unit Normal Coordinate Form Equation}) is now used to define a
uniform property which is exhibited by any point on a linear locus.

\subsection{Uniform Property Exhibited by Points on a Linear Locus}

Given Eq. (\ref{Unit Normal Coordinate Form Equation}), it follows that
a\ line, plane, or hyperplane is a locus of points $\mathbf{x}$, all of which
possess a set of normalized, eigen-scaled coordinates:%
\[
\mathbf{x}=\left(  \cos\alpha_{1}x_{1},\cos\alpha_{2}x_{2},\ldots,\cos
\alpha_{d}x_{d}\right)  ^{T}\text{,}%
\]
such that the sum of those coordinates equals the distance $\mathbf{\Delta}$
that the line, plane, or hyperplane is from the origin $\left(  0,0,\ldots
,0\right)  $:%
\begin{equation}
\sum\nolimits_{i=1}^{d}\cos\alpha_{i}x_{i}=\mathbf{\Delta}\text{,}
\label{Geometric Property of Linear Loci}%
\end{equation}
where $x_{i}$ are point coordinates or vector components, and $\cos\alpha_{i}$
are the direction cosines between a unit length normal eigenaxis
$\mathbf{u}_{N_{e}}$ and the coordinate axes $\mathbf{e}_{i}:\left\{
\mathbf{e}_{1}=\left(  1,0,\ldots,0\right)  ,\ldots,\mathbf{e}_{d}=\left(
0,0,\ldots,1\right)  \right\}  $.

It follows that a point $\mathbf{x}$ is on the geometric locus of a line $l$,
plane $p$, or hyperplane $h$, if, and only if, the normalized, eigen-scaled
coordinates of $\mathbf{x}$ satisfy Eq.
(\ref{Geometric Property of Linear Loci}); otherwise, the point $\mathbf{x}$
is not on the locus of points described by Eqs (\ref{Linear Locus Functional}%
), (\ref{Normal Eigenaxis Functional}), and
(\ref{normal form linear functional}). Given Eq.
(\ref{Geometric Property of Linear Loci}), it follows that the sum of
normalized, eigen-scaled coordinates of any point on a linear locus, equals
the magnitude of the normal eigenaxis of the linear locus. It is concluded
that all of the points $\mathbf{x}$ on a linear locus possess a characteristic
set of eigen-scaled coordinates, such that the inner product of each vector
$\mathbf{x}$ with $\mathbf{u}_{N_{e}}$ satisfies the distance $\mathbf{\Delta
}$ of the linear locus from the origin.

So far, the principal eigenaxis of linear curves and surfaces has been
identified, along with a correlated uniform property which is exhibited by any
point on the lines, planes, or hyperplanes of Eqs
(\ref{Linear Locus Functional}), (\ref{Normal Eigenaxis Functional}), and
(\ref{Normal Form Normal Eigenaxis}). Moreover, Eqs
(\ref{Linear Locus Functional}) - (\ref{Geometric Property of Linear Loci})
all indicate that the eigen-coordinate locations of a normal eigenaxis provide
a distinctive set of eigenfeatures which effectively characterize all forms of
lines, planes, and hyperplanes. To wit, the locus of a normal eigenaxis
effectively determines the locus of points on a linear curve or surface. This
implies that the\ important generalizations for a linear locus are encoded
within the invariant geometric location of its normal eigenaxis.

The next section will examine how important generalizations and properties for
linear loci are encoded within the geometric locus of a normal eigenaxis. It
will be demonstrated that a normal eigenaxis is an exclusive, intrinsic
reference axis which has a distinctive geometric configuration, a
characteristic set of eigen-coordinate loci, and a characteristic eigenenergy,
all of which determine a characteristic eigen-signature for any given linear
locus of points. The uniform properties which are satisfied by all of the
points on a linear locus will be identified. It will also be demonstrated that
each uniform property is uniquely determined by the locus of a normal
eigenaxis. Later on, these arguments will be extended to include linear
decision boundary estimates. The properties of normal eigenaxes are examined next.

\subsection{Properties of Normal Eigenaxes}

Take any line, plane, or hyperplane in $%
\mathbb{R}
^{d}$. Given Eqs (\ref{Linear Locus Functional}) or
(\ref{Normal Eigenaxis Functional}) and a particular line, plane, or
hyperplane, it follows that a normal eigenaxis $\mathbf{v}$ exists, such that
the tip of $\mathbf{v}$ is on the specified line, plane, or hyperplane, and
the axis of $\mathbf{v}$ is perpendicular to the line, plane, or hyperplane.
Given Eq. (\ref{Normal Form Normal Eigenaxis}), it follows that the length
$\left\Vert \mathbf{v}\right\Vert $ of $\mathbf{v}$\ is determined by the
given line, plane, or hyperplane. Given Eq.
(\ref{Unit Normal Coordinate Form Equation}), it follows that the unit normal
eigenaxis $\mathbf{u}_{N_{e}}$ of the specified linear curve or surface is
characterized by a unique set of direction cosines $\cos\alpha_{i}$ between
$\mathbf{u}_{N_{e}}$ and each standard basis vector $\mathbf{e}_{i}$.

Next, take any normal eigenaxis $\mathbf{v}$ in $%
\mathbb{R}
^{d}$. Given Eqs (\ref{Linear Locus Functional}) or
(\ref{Normal Eigenaxis Functional}) and a particular normal eigenaxis
$\mathbf{v}$, it follows that a line, plane, or hyperplane exists that is
perpendicular to $\mathbf{v}$, such that the tip of the given normal eigenaxis
$\mathbf{v}$ is on the line, plane, or hyperplane. Given Eq.
(\ref{Normal Form Normal Eigenaxis}), it follows that the distance of the
line, plane, or hyperplane from the origin is specified by the magnitude
$\left\Vert \mathbf{v}\right\Vert $ of the given normal eigenaxis $\mathbf{v}%
$. It will now be shown that the normal eigenaxis of any given linear locus
satisfies the linear locus in terms of its eigenenergy.

\subsubsection{Characteristic Eigenenergy of a Normal Eigenaxis}

Take the normal eigenaxis $\mathbf{v}$ of any line, plane, or hyperplane in $%
\mathbb{R}
^{d}$. It follows that the normal eigenaxis $\mathbf{v}$ satisfies Eqs
(\ref{Linear Locus Functional}), (\ref{Normal Eigenaxis Functional}), and
(\ref{Normal Form Normal Eigenaxis}). Given Eqs (\ref{Linear Locus Functional}%
) or (\ref{Normal Eigenaxis Functional}) and a particular normal eigenaxis
$\mathbf{v}$, it follows that the normal eigenaxis $\mathbf{v}$ satisfies the
linear locus in terms of its eigenenergy:%
\[
\mathbf{v}^{T}\mathbf{v}=\left\Vert \mathbf{v}\right\Vert ^{2}\text{.}%
\]
Therefore, the normal eigenaxis $\mathbf{v}$ of any given line, plane, or
hyperplane exhibits a characteristic eigenenergy $\left\Vert \mathbf{v}%
\right\Vert ^{2}$. It follows that the line, plane, or hyperplane delineated
by a normal eigenaxis $\mathbf{v}$ exhibits a characteristic eigen-signature
in the form of the characteristic eigenenergy $\left\Vert \mathbf{v}%
\right\Vert ^{2}$ of its normal eigenaxis $\mathbf{v}$.

It is concluded that the normal eigenaxis of any given linear locus satisfies
the linear locus in terms of its eigenenergy. It is also concluded that the
fundamental property of any given normal eigenaxis $\mathbf{v}$ is its
characteristic eigenenergy $\left\Vert \mathbf{v}\right\Vert ^{2}$.

\subsection{Correlated Uniform Properties Exhibited by Points on a Linear
Locus}

Take any point $\mathbf{x}$ on any linear locus. Given Eq.
(\ref{Linear Locus Functional}) and a specific point $\mathbf{x}$ on a
particular locus, it follows that the length of the component $\left\Vert
\mathbf{x}\right\Vert \cos\phi$ of the given vector $\mathbf{x}$ along the
normal eigenaxis $\mathbf{v}$ of the given locus satisfies the length
$\left\Vert \mathbf{v}\right\Vert $ of $\mathbf{v}$:%
\[
\left\Vert \mathbf{x}\right\Vert \cos\phi=\left\Vert \mathbf{v}\right\Vert
\text{,}%
\]
where the length $\left\Vert \mathbf{v}\right\Vert $ of $\mathbf{v}$
determines the distance $\mathbf{\Delta}$ of the linear locus from the origin.

Given Eq. (\ref{normal form linear functional}) and the same point
$\mathbf{x}$ on the given locus, it follows that the inner product
$\mathbf{x}^{T}\mathbf{u}_{N_{e}}$ of the given vector $\mathbf{x}$ with the
unit normal eigenaxis $\mathbf{u}_{N_{e}}$:%
\[
\mathbf{x}^{T}\mathbf{u}_{N_{e}}\mathbf{=\Delta}\text{,}%
\]
of the given locus, satisfies the distance $\mathbf{\Delta}$ of the linear
locus from the origin. Likewise, given Eq.
(\ref{Geometric Property of Linear Loci}), it follows that the normalized,
eigen-scaled coordinates of the given point $\mathbf{x}$%
\[
\sum\nolimits_{i=1}^{d}\cos\alpha_{i}x_{i}=\mathbf{\Delta}\text{,}%
\]
also satisfy the distance $\mathbf{\Delta}$ of the given locus from the origin.

Finally, given Eq. (\ref{Normal Eigenaxis Functional}), it follows that the
inner product $\mathbf{x}^{T}\mathbf{v}$ of the given vector $\mathbf{x}$ with
the normal eigenaxis $\mathbf{v}$ of the given locus, satisfies the
characteristic eigenenergy $\left\Vert \mathbf{v}\right\Vert ^{2}$%
\[
\mathbf{x}^{T}\mathbf{v}=\left\Vert \mathbf{v}\right\Vert ^{2}\text{,}%
\]
of the normal eigenaxis $\mathbf{v}$ of the linear locus.

It is concluded that the uniform properties which are satisfied by all of the
points $\mathbf{x}$ on a linear locus are uniquely determined by the geometric
locus of the normal eigenaxis $\mathbf{v}$ of the linear locus.

Given that the uniform, correlated properties exhibited by the points
$\mathbf{x}$ on a linear locus are uniquely determined by the geometric locus
of a normal eigenaxis $\mathbf{v}$, it is concluded that all of the points
$\mathbf{x}$ on a line, plane, or hyperplane explicitly and exclusively
reference the normal eigenaxis $\mathbf{v}$ of Eqs
(\ref{Linear Locus Functional}), (\ref{Normal Eigenaxis Functional}), and
(\ref{Normal Form Normal Eigenaxis}).

In summary, it has been demonstrated that the uniform, correlated properties
exhibited by any point $\mathbf{x}$ on any linear locus are functions of the
eigen-coordinate locations and the corresponding magnitude and eigenenergy of
the normal eigenaxis $\mathbf{v}$ of the given locus. Thereby, it has been
demonstrated that the vector components of a normal eigenaxis provide a
characteristic set of eigenaxis locations that effectively specify all forms
of linear curves and surfaces. A characteristic set of normal eigenaxis
locations will be referred to as eigenloci. It is concluded that a normal
eigenaxis $\mathbf{v}$ coincides as an exclusive and distinctive coordinate
axis that effectively characterizes all of the points on a linear locus.
Thereby, a normal eigenaxis offers an elegant principal eigen-coordinate
system for a linear locus of points.

The next section of the paper will argue that the rich set of geometric
properties exhibited by all of the points on a linear locus, which include the
normal eigenaxis of a given locus, involve inner product correlations between
the \emph{geometric loci of vectors}. Section $5$ will examine the idea of the
geometric locus of a vector, which will be demonstrated to be the primary
building block of regularized, data-driven geometric architectures in Hilbert
spaces. It will also be argued that inner product statistics between the
geometric loci of training vectors offer a natural \emph{functional glue} for
generating learning machine architectures.

\section{Design of Learning Machine Architectures in Hilbert Spaces}

Geometric locus methods restricted to Cartesian coordinate spaces are
essentially restricted to static and fixed representations of geometric loci.
Indeed, Cartesian coordinate spaces only permit algebraic equations of
geometric loci in terms of Euclidean distances between point coordinates and
algebraic constraints on point coordinate values. Alternatively, Hilbert
spaces permit algebraic systems of geometric loci in terms of correlated
algebraic systems of inner product statistics between the geometric loci of
vectors, where the magnitude and direction of any given vector determines an
endpoint formed by a unique set of point coordinates.

In general, the geometric structures of Euclidean spaces equipped with inner
product structures, i.e., Hilbert spaces, are much richer than the geometric
structures of Cartesian coordinate spaces
\citet{Naylor1971}%
. It will be demonstrated that inner product structures in Hilbert spaces
actually involve the geometric loci of vectors. Accordingly, the notion of the
geometric locus of a point is ill-defined. Given that a geometric locus is a
\emph{curve or surface formed by a set of points which possess some uniform
property}, it will be shown that the locus of a point actually involves the
locus of a vector. To clearly distinguish a vector from a point in the
discussion that follows, a vector will be denoted by $\mathbf{x}$ or
$\mathbf{x}=%
\begin{pmatrix}
x_{1}, & x_{2}, & \cdots, & x_{d}%
\end{pmatrix}
^{T}$ and a point will be denoted by $P$ or $P%
\begin{pmatrix}
x_{1}, & x_{2}, & \cdots, & x_{d}%
\end{pmatrix}
$.

This section of the paper will demonstrate that a vector
$\widetilde{\mathbf{x}}\in%
\mathbb{R}
^{d}$ is a geometric locus of a directed straight line segment formed by two
points $P_{\mathbf{0}}%
\begin{pmatrix}
0, & 0, & \cdots, & 0
\end{pmatrix}
$ and $P_{\widetilde{\mathbf{x}}}%
\begin{pmatrix}
\widetilde{x}_{1}, & \widetilde{x}_{2}, & \cdots, & \widetilde{x}_{d}%
\end{pmatrix}
$, which are at a distance of%
\[
\left\Vert \widetilde{\mathbf{x}}\right\Vert =\left(  \widetilde{x}_{1}%
^{2}+\widetilde{x}_{2}^{2}+\cdots+\widetilde{x}_{d}^{2}\right)  ^{1/2}\text{,}%
\]
where each point coordinate $\widetilde{x}_{i}$ is at a distance of
$\left\Vert \widetilde{\mathbf{x}}\right\Vert \cos\mathbb{\alpha}_{ij}$ from
the origin $P_{0}$, along the direction of an orthonormal coordinate axis
$\mathbf{e}_{j}$, where $\mathbb{\alpha}_{ij}$ is the angle between the vector
$\widetilde{\mathbf{x}}$ and an orthonormal coordinate axis $\mathbf{e}_{j}%
$\textbf{. }This section will also demonstrate how algebraic and geometric
structures generated by an inner product statistic between two vectors
describe rich topological and algebraic relationships between the geometric
loci of two vectors.

\subsection{Geometric Properties of a Vector}

In geometric terms, a vector $\mathbf{x}$%
\[
\mathbf{x}=%
\begin{pmatrix}
x_{1}, & x_{2}, & \cdots, & x_{d}%
\end{pmatrix}
^{T}\text{,}%
\]
is a directed straight line segment that emanates from the point of
intersection of coordinate axes $P_{\mathbf{0}}$%
\[
P_{\mathbf{0}}%
\begin{pmatrix}
0, & 0, & \cdots, & 0
\end{pmatrix}
\text{,}%
\]
where the values of the coordinates are all zero, commonly known as the
origin, such that the endpoint $P_{e}$ of the directed straight line segment
defines a real specific point%
\[
P_{e}%
\begin{pmatrix}
x_{1}, & x_{2}, & \cdots, & x_{d}%
\end{pmatrix}
\text{.}%
\]
Thereby, the point coordinates%
\[
P%
\begin{pmatrix}
x_{1}, & x_{2}, & \cdots, & x_{d}%
\end{pmatrix}
\text{,}%
\]
of any given point $P$ coincide with the components%
\[
\mathbf{x}=%
\begin{pmatrix}
x_{1}, & x_{2}, & \cdots, & x_{d}%
\end{pmatrix}
^{T}\text{,}%
\]
of a vector $\mathbf{x}$. Given orthogonal coordinate axes, each point
coordinate $x_{i}$ of $P$ specifies a scaling for an orthogonal unit
coordinate axis $\mathbf{e}_{i}$ in $%
\mathbb{R}
^{d}$, where each variable length coordinate axis $x_{i}\mathbf{e}_{i}$
determines a component of a vector $\mathbf{x}=%
\begin{pmatrix}
x_{1}, & x_{2}, & \cdots, & x_{d}%
\end{pmatrix}
^{T}$. For example, Fig. $8$ depicts the vectors $\mathbf{v}$ and $\mathbf{x}
$ in the Euclidean plane $%
\mathbb{R}
^{2}$, where the endpoint of the vector $\mathbf{v}$ is on the locus of the
point $P_{\mathbf{v}}$, with point coordinates and vector components $%
\begin{pmatrix}
x_{v_{1}}, & x_{v_{2}}%
\end{pmatrix}
$, and the endpoint of the vector $\mathbf{x}$ is on the locus of the point
$P_{\mathbf{x}}$, with point coordinates and vector components $%
\begin{pmatrix}
x_{1}, & x_{2}%
\end{pmatrix}
$.

So, take any directed straight line segment $\widetilde{\mathbf{x}}$ formed by
the points%
\[
P_{\mathbf{0}}%
\begin{pmatrix}
0, & 0, & \cdots, & 0
\end{pmatrix}
\text{,}%
\]
and%
\[
P_{\widetilde{\mathbf{x}}}%
\begin{pmatrix}
\widetilde{x}_{1}, & \widetilde{x}_{2}, & \cdots, & \widetilde{x}_{d}%
\end{pmatrix}
\text{,}%
\]
where the point $P_{\widetilde{\mathbf{x}}}$ is the endpoint of the vector
$\widetilde{\mathbf{x}}$.

The coordinates of the point $P_{\widetilde{\mathbf{x}}}$ and the components
of the vector $\widetilde{\mathbf{x}}$ are both described by the unique,
ordered $d$-tuple of real numbers:%
\[%
\begin{pmatrix}
\widetilde{x}_{1}, & \widetilde{x}_{2}, & \cdots, & \widetilde{x}_{d}%
\end{pmatrix}
\text{,}%
\]
where the Euclidean distance $D_{E}$ between $P_{\mathbf{0}}$ and
$P_{\widetilde{\mathbf{x}}}$ is%
\begin{align*}
D_{E}\left(  P_{\mathbf{0}},P_{\widetilde{\mathbf{x}}}\right)   &  =\left(
\left\vert 0-\widetilde{x}_{1}\right\vert ^{2}+\ldots+\left\vert
0-\widetilde{x}_{d}\right\vert ^{2}\right)  ^{1/2}\text{,}\\
&  =\left(  \widetilde{x}_{1}^{2}+\widetilde{x}_{2}^{2}+\cdots+\widetilde{x}%
_{d}^{2}\right)  ^{1/2}\text{.}%
\end{align*}
Now, it makes no sense to speak of the length of a point or to consider the
angle between two points. However, any given point $P_{j}\in%
\mathbb{R}
^{d}$ has an enhanced representation as the endpoint of a vector
$\mathbf{x}_{j}\in%
\mathbb{R}
^{d}$. The geometric locus of a vector is defined next.

\subsection{Geometric Locus of a Vector}

Take any given point $P_{\widetilde{\mathbf{x}}}$ which is also the endpoint
of a vector $\widetilde{\mathbf{x}}$ in $%
\mathbb{R}
^{d}$. Next, take the standard set of orthonormal basis vectors in $%
\mathbb{R}
^{d}$:%
\[
\left\{  \mathbf{e}_{1}=\left(  1,0,\ldots,0\right)  ,\ldots,\mathbf{e}%
_{d}=\left(  0,0,\ldots,1\right)  \right\}  \text{,}%
\]
and consider the scalar projection of the vector $\widetilde{\mathbf{x}}$ onto
the above set of standard basis vectors. The component of the vector
$\widetilde{\mathbf{x}}$ along each basis vector $\mathbf{e}_{j}$%
\[
\operatorname{comp}_{\overrightarrow{\mathbf{e}_{j}}}\left(
\overrightarrow{\widetilde{\mathbf{x}}}\right)  =\left\Vert
\widetilde{\mathbf{x}}\right\Vert \cos\mathbb{\alpha}_{j}\text{,}%
\]
where $\mathbb{\alpha}_{j}$ is the angle between $\widetilde{\mathbf{x}}$ and
$\mathbf{e}_{j}$, determines a set of signed magnitudes along the axes of the
basis vectors
\citet{Stewart2009}%
\[%
\begin{pmatrix}
\left\Vert \widetilde{\mathbf{x}}\right\Vert \cos\mathbb{\alpha}_{1}, &
\left\Vert \widetilde{\mathbf{x}}\right\Vert \cos\mathbb{\alpha}_{2}, &
\cdots, & \left\Vert \widetilde{\mathbf{x}}\right\Vert \cos\mathbb{\alpha}_{d}%
\end{pmatrix}
\text{,}%
\]
all of which describe a unique, ordered $d$-tuple of geometric loci, where the
distance of each point coordinate or vector component $\widetilde{x}_{i}$ from
the origin $P_{o}$, along the axis of the basis vector $\mathbf{e}_{j}$, is
$\left\Vert \widetilde{\mathbf{x}}\right\Vert \cos\mathbb{\alpha}_{j}$.

\subsection{Uniform Property Exhibited by Vector Components}

Generally speaking, the geometric locus of any vector $\mathbf{x}_{k}$ and
correlated point $P_{\mathbf{x}_{k}}$ is characterized by a unique, ordered
$d$-tuple of geometric loci:%
\begin{equation}%
\begin{pmatrix}
\left\Vert \mathbf{x}_{k}\right\Vert \cos\mathbb{\alpha}_{\mathbf{x}_{k1}1}, &
\left\Vert \mathbf{x}_{k}\right\Vert \cos\mathbb{\alpha}_{\mathbf{x}_{k2}2}, &
\cdots, & \left\Vert \mathbf{x}_{k}\right\Vert \cos\mathbb{\alpha}%
_{\mathbf{x}_{kd}d}%
\end{pmatrix}
\text{,} \label{Geometric Locus of Vector}%
\end{equation}
where $\left(  \cos\alpha_{\mathbf{x}_{k1}1},\cdots,\cos\alpha_{\mathbf{x}%
_{kd}d}\right)  $ are the direction cosines of the components $%
\begin{pmatrix}
x_{k1}, & \cdots, & x_{kd}%
\end{pmatrix}
$ of the vector $\mathbf{x}_{k}$ relative to the standard set of orthonormal
coordinate axes $\left\{  \mathbf{e}_{j}\right\}  _{j=1}^{d}$.

Each of the $d$ point coordinates or vector components $\left\{
x_{\mathbf{x}_{ki}}\right\}  _{i=1}^{d}$ are at a distance of $\left\Vert
\mathbf{x}_{k}\right\Vert \cos\mathbb{\alpha}_{\mathbf{x}_{ki}j}$ from the
origin $P_{\mathbf{0}}$, along the direction of an orthonormal coordinate axis
$\mathbf{e}_{j}$. Thus, each point coordinate or vector component
$x_{\mathbf{x}_{ki}}$ exhibits a characteristic magnitude of $\left\Vert
\mathbf{x}_{k}\right\Vert \cos\mathbb{\alpha}_{\mathbf{x}_{ki}j}$ along an
orthonormal coordinate axis $\mathbf{e}_{j}$.

It is concluded that a vector $\widetilde{\mathbf{x}}\in%
\mathbb{R}
^{d}$ is a geometric locus of a directed straight line segment formed by two
points $P_{\mathbf{0}}$ and $P_{\widetilde{\mathbf{x}}}$, which are at a
distance of $\left\Vert \widetilde{\mathbf{x}}\right\Vert $ from each other,
where each point coordinate $\widetilde{x}_{i}$ is at a distance of
$\left\Vert \widetilde{\mathbf{x}}\right\Vert \cos\mathbb{\alpha}_{ij}$ from
the origin $P_{\mathbf{0}}$, along the direction of an orthonormal coordinate
axis $\mathbf{e}_{j}$.

It has been demonstrated how a\ vector provides an enhanced representation of
a point. Thereby, it has been demonstrated that the geometric locus of a point
is determined by the geometric locus of a vector. It will now be demonstrated
how inner product statistics encode a rich set of algebraic and topological
relationships between the geometric loci of two vectors.

\subsection{Inner Product Statistics}

The inner product expression $\mathbf{x}^{T}\mathbf{x}$ defined by%
\[
\mathbf{x}^{T}\mathbf{x}=\ x_{1}x_{1}+x_{2}x_{2}+\cdots+x_{d}x_{d}\text{,}%
\]
generates the norm $\left\Vert \mathbf{x}\right\Vert $ of the vector
$\mathbf{x}$%
\[
\left\Vert \mathbf{x}\right\Vert =\left(  x_{1}^{2}+x_{2}^{2}+\cdots+x_{d}%
^{2}\right)  ^{1/2}\text{,}%
\]
which determines the Euclidean distance between the endpoint of $\mathbf{x}$
and the origin, whereby the norm $\left\Vert \mathbf{x}\right\Vert $ measures
the length of the vector $\mathbf{x}$, which indicates the magnitude of
$\mathbf{x}$. The Euclidean space $%
\mathbb{R}
^{d}$ equipped with a norm $\left\Vert \mathbf{x}\right\Vert $ that is
generated by an inner product $\mathbf{x}^{T}\mathbf{x}$ is a Hilbert space
\citet{Naylor1971}%
.

The inner product function $\mathbf{x}^{T}\mathbf{y}$ also determines the
angle between two vectors $\mathbf{x}$ and $\mathbf{y}$ in $%
\mathbb{R}
^{d}$. Given any two vectors $\mathbf{x}$ and $\mathbf{y}$, the inner product
expression%
\begin{equation}
\mathbf{x}^{T}\mathbf{y}=\ x_{1}y_{1}+x_{2}y_{2}+\cdots+x_{d}y_{d}\text{,}
\label{Inner Product Expression1}%
\end{equation}
is also given by the expression%
\begin{equation}
\mathbf{x}^{T}\mathbf{y}=\left\Vert \mathbf{x}\right\Vert \left\Vert
\mathbf{y}\right\Vert \cos\theta\text{,} \label{Inner Product Expression2}%
\end{equation}
where $\theta$ is the angle between the vectors $\mathbf{x}$ and $\mathbf{y}%
$\textbf{.} If $\theta=90^{\circ}$, $\mathbf{x}$ and $\mathbf{y}$ are said to
be perpendicular to each other. Accordingly, the inner product expression in
Eq. (\ref{Inner Product Expression2}) allows us to describe vectors which are
orthogonal or perpendicular to each other
\citet{Naylor1971}%
. Two vectors $\mathbf{x}$ and $\mathbf{y}$ are said to be orthogonal to each
other if%
\[
\mathbf{x}^{T}\mathbf{y}=0\text{,}%
\]
which is denoted by $\mathbf{x}\perp\mathbf{y}$.

It will now be demonstrated how the algebraic relationships in Eqs
(\ref{Inner Product Expression1}) and (\ref{Inner Product Expression2}) are
derived from second-order distance statistics between the geometric loci of
two vectors. Second-order distance statistics will be shown to encode rich
algebraic and topological relationships between the geometric loci of vectors.

\subsection{Second-order Distance Statistics Between Loci of Vectors}

The algebraic relationship%
\[
\mathbf{\upsilon}^{T}\mathbf{\nu}=\left\Vert \mathbf{\upsilon}\right\Vert
\left\Vert \mathbf{\nu}\right\Vert \cos\varphi\text{,}%
\]
between two vectors $\mathbf{\upsilon}$ and $\mathbf{\nu}$ in Hilbert space
can be derived by using the law of cosines
\citet{Lay2006}%
:%
\begin{equation}
\left\Vert \mathbf{\upsilon}-\mathbf{\nu}\right\Vert ^{2}=\left\Vert
\mathbf{\upsilon}\right\Vert ^{2}+\left\Vert \mathbf{\nu}\right\Vert
^{2}-2\left\Vert \mathbf{\upsilon}\right\Vert \left\Vert \mathbf{\nu
}\right\Vert \cos\varphi\text{,} \label{Inner Product Statistic}%
\end{equation}
which reduces to%
\begin{align*}
\left\Vert \mathbf{\upsilon}\right\Vert \left\Vert \mathbf{\nu}\right\Vert
\cos\varphi &  =\upsilon_{1}\nu_{1}+\upsilon_{2}\nu_{2}+\cdots+\upsilon_{d}%
\nu_{d}\text{,}\\
&  =\mathbf{\upsilon}^{T}\mathbf{\nu}\text{,}\\
&  =\mathbf{\nu}^{T}\mathbf{\upsilon}\text{.}%
\end{align*}
This indicates that the inner product statistic $\mathbf{\upsilon}%
^{T}\mathbf{\nu}$ or $\left\Vert \mathbf{\upsilon}\right\Vert \left\Vert
\mathbf{\nu}\right\Vert \cos\varphi$ determines the length $\left\Vert
\mathbf{\upsilon}-\mathbf{\nu}\right\Vert $ of the vector from $\mathbf{\nu}$
to $\mathbf{\upsilon}$, which is the distance between the endpoints of
$\mathbf{\upsilon}$ and $\mathbf{\nu}$.

It follows that the inner product statistic between any two vectors
$\mathbf{\upsilon}$ and $\mathbf{\nu}$ in Hilbert space
\begin{align*}
\mathbf{\upsilon}^{T}\mathbf{\nu}  &  =\upsilon_{1}\nu_{1}+\upsilon_{2}\nu
_{2}+\cdots+\upsilon_{d}\nu_{d}\text{,}\\
&  =\left\Vert \mathbf{\upsilon}\right\Vert \left\Vert \mathbf{\nu}\right\Vert
\cos\varphi\text{,}%
\end{align*}
determines the distance between the geometric loci of $\mathbf{\upsilon}$ and
$\mathbf{\nu}$.

It is concluded that the algebraic relationships defined within Eq.
(\ref{Inner Product Statistic}) describe a rich set of topological
relationships between the geometric loci of two vectors. Figure $9$ depicts
the rich set of correlated algebraic and topological structures encoded within
an inner product statistic of the geometric loci of two vectors.

\begin{center}%
\begin{center}
\includegraphics[
natheight=7.499600in,
natwidth=9.999800in,
height=3.2897in,
width=4.3777in
]%
{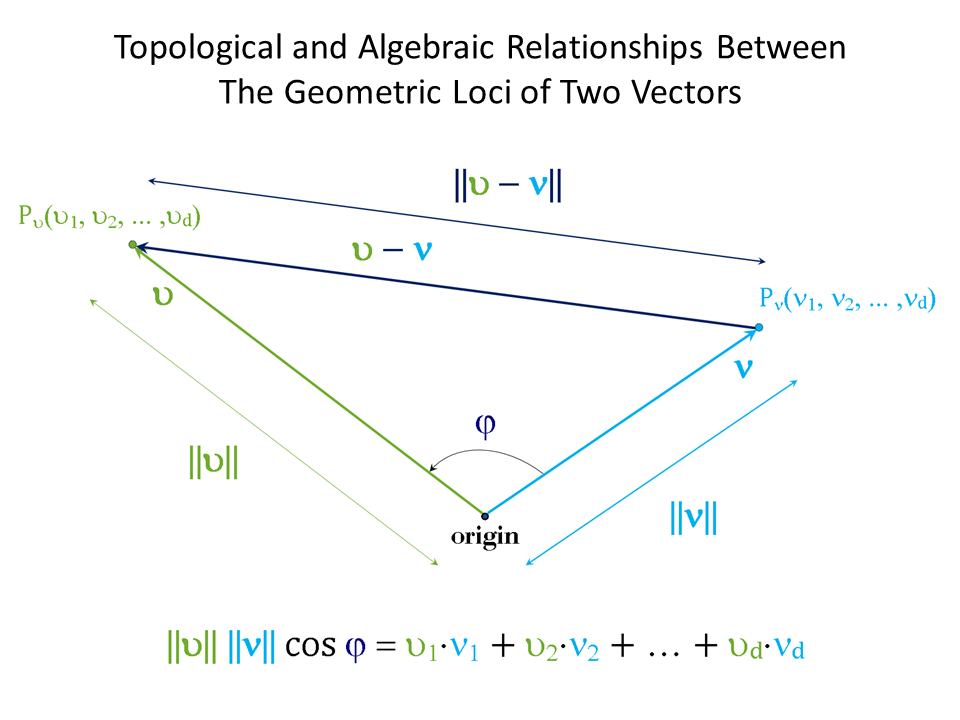}%
\end{center}

\end{center}

\begin{flushleft}
Figure $9$: Illustration of the rich set of algebraic and topological
relationships encoded within the inner product statistic $\mathbf{\upsilon
}^{T}\mathbf{\nu}$ of the geometric loci of the vectors $\mathbf{\upsilon}$
and $\mathbf{\nu}$.
\end{flushleft}

Equation (\ref{Inner Product Statistic}) also determines the component of a
vector along another vector, which is also known as a scalar projection. The
algebraic and geometric nature of scalar projections are examined next.

\subsection{Signed Magnitudes of Vector Projections}

The inner product statistic%
\[
\mathbf{x}^{T}\mathbf{y}=\left\Vert \mathbf{x}\right\Vert \left\Vert
\mathbf{y}\right\Vert \cos\theta\text{,}%
\]
can be interpreted as the length $\left\Vert \mathbf{x}\right\Vert $ of
$\mathbf{x}$ times the scalar projection of $\mathbf{y}$ onto $\mathbf{x}$%
\begin{equation}
\mathbf{x}^{T}\mathbf{y}=\left\Vert \mathbf{x}\right\Vert \times\left[
\left\Vert \mathbf{y}\right\Vert \cos\theta\right]  \text{,}
\label{Scalar Projection}%
\end{equation}
where the scalar projection of $\mathbf{y}$ onto $\mathbf{x}$, also known as
the component of $\mathbf{y}$ along $\mathbf{x}$, is defined to be the signed
magnitude of the vector projection%
\[
\left\Vert \mathbf{y}\right\Vert \cos\theta\text{,}%
\]
where $\theta$ is the angle between $\mathbf{x}$ and $\mathbf{y}$
\citet{Stewart2009}%
. Scalar projections are denoted by $\operatorname{comp}%
_{\overrightarrow{\mathbf{x}}}\left(  \overrightarrow{\mathbf{y}}\right)  $,
where $\operatorname{comp}_{\overrightarrow{\mathbf{x}}}\left(
\overrightarrow{\mathbf{y}}\right)  <0$ if $\pi/2<\theta\leq\pi$.

The scalar projection statistic also satisfies the inner product relationship%
\begin{align*}
\left\Vert \mathbf{y}\right\Vert \cos\theta &  =\frac{\mathbf{x}^{T}%
\mathbf{y}}{\left\Vert \mathbf{x}\right\Vert }\\
&  =\left(  \frac{\mathbf{x}}{\left\Vert \mathbf{x}\right\Vert }\right)
^{T}\mathbf{y}\text{,}%
\end{align*}
between the unit vector $\frac{\mathbf{x}}{\left\Vert \mathbf{x}\right\Vert }$
and $\mathbf{y}$.

Figure $10$ depicts the geometric nature of scalar projections for acute and
obtuse angles between vectors. Scalar projection statistics determine signed
magnitudes along the axes of given vectors.

\begin{center}%
\begin{center}
\includegraphics[
natheight=7.499600in,
natwidth=9.999800in,
height=3.2897in,
width=4.3777in
]%
{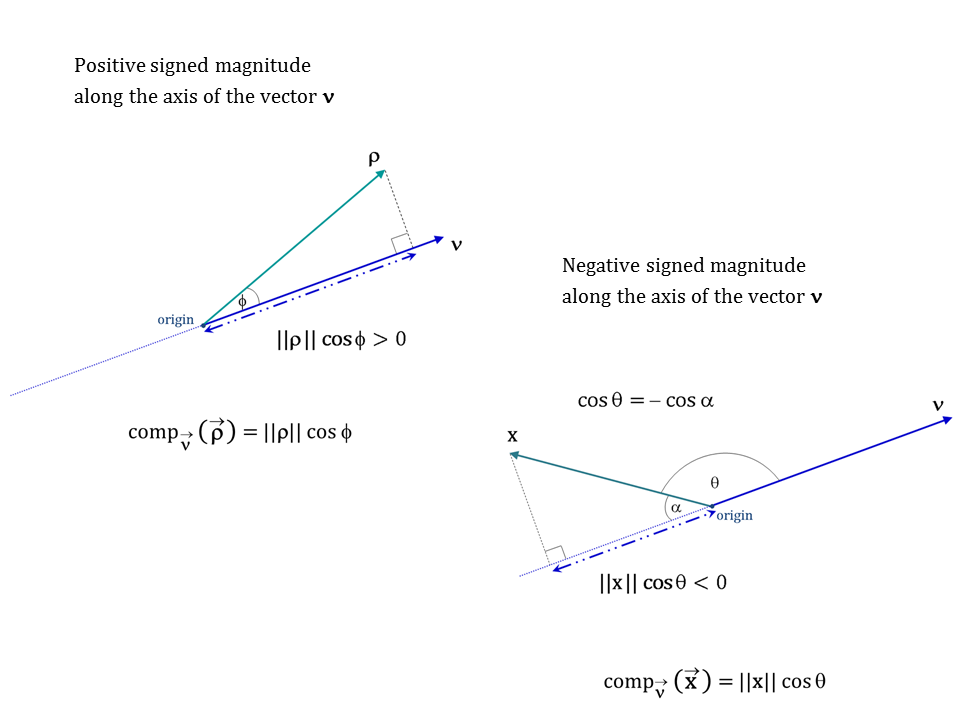}%
\end{center}

\end{center}

\begin{flushleft}
Figure $10$: Illustration of how scalar projection statistics determine
components (signed magnitudes) of vectors along the axis of a given vector.
\end{flushleft}

The findings presented in this paper will demonstrate how the algebraic and
topological relationships encoded within inner product statistics offer a
natural \emph{functional glue} for learning machine architectures.

\subsection{Designing Functional Glue for Learning Machine Architectures}

This paper will demonstrate how algebraic systems of properly specified,
correlated inner product statistics between training data give rise to
regularized, data-driven geometric architectures which encode robust decision
statistics for complex discrimination tasks. The remaining sections of this
paper will develop a regularized, data-driven geometric architecture, which
describes linear decision boundaries for overlapping and non-overlapping data
distributions, that is determined by correlated sets of dual, i.e., primal and
dual, principal (normal) eigenaxis components, all of which are jointly and
symmetrically located in primal and dual, correlated Hilbert spaces. It will
be shown that data-driven sets of primal and dual normal eigenaxis components
encode robust likelihood ratios for complex discrimination tasks. It will also
be demonstrated that eigenenergies of data-driven sets of primal and dual
normal eigenaxis components satisfy the law of cosines in a surprisingly
elegant and symmetric manner.

The analyses that follow will make extensive use of inner product and scalar
projection statistics. Inner product and scalar projection statistics will be
shown to provide a natural functional glue for adaptable geometric
architectures. Regularized, adaptable geometric architectures which encode
relevant aspects of statistical decision systems will be demonstrated to be
the principal foundation of learning machine architectures.

It has previously been argued that Cartesian coordinate spaces only permit
static and fixed descriptions of a geometric locus of points. The analyses
that follow will demonstrate how algebraic systems of correlated inner product
statistics between training vectors in dual, correlated Hilbert spaces
generate robust, data-driven, symmetrical geometric architectures that
represent statistical decision systems. Thereby, it will be demonstrated how
collections of training data are transformed into regularized geometric
architectures which encode relevant geometric and statistical aspects of
statistical decision systems.

Naylor and Sell note that a truly amazing number of problems in engineering
and science can be fruitfully treated with geometric methods in Hilbert space
\citet{Naylor1971}%
. \emph{This paper will use geometric and statistical methods in dual,
correlated Hilbert spaces to solve the long-standing problem of learning
robust or optimal linear decision boundaries for overlapping sets of data.}

The set of analyses which follow involve the examination of dual,
interconnected, symmetrical geometric architectures which are generated by
algebraic systems of correlated inner product statistics between training
vectors in dual, correlated Hilbert spaces in $%
\mathbb{R}
^{d}$ and $%
\mathbb{R}
^{N}$. The analyses will demonstrate how symmetrical, interconnected geometric
architectures in dual, correlated Hilbert spaces provide the geometric basis
of the statistical representation of a primal normal eigenaxis in a Wolfe dual
eigenspace. The analyses will use all of the algebraic and geometric
properties of the normal eigen-coordinate system outlined earlier, to examine
how robust estimates of constrained normal eigen-coordinate locations provide
robust, stable, and optimal statistical representations of linear decision
boundaries. The analyses will demonstrate that robust estimates of constrained
normal eigenaxis components provide optimal statistical descriptions of linear
decision boundaries for normally distributed training data with common
covariance matrices. The analyses will also demonstrate that robust estimates
of constrained normal eigen-coordinate locations provide robust or optimal
estimates of linear decision boundaries for training data drawn from various
distributions, including completely overlapping data distributions.

More generally, it can be demonstrated that the eigen-coordinate locations of
the principal eigenaxis of any given second-order curve or surface offer a
characteristic set of eigenloci that specify the given curve or surface. An
upcoming paper will consider how principal eigenaxes provide exclusive,
intrinsic coordinate axes for the geometric loci of $d$-dimensional circles,
ellipses, hyperbolae, and parabolas. The paper will examine how robust
statistical representations of constrained principal eigen-coordinate
locations provide the primary statistical basis for second-order decision
boundary estimates. Indeed, robust estimates of constrained principal
eigen-coordinate locations can be shown to describe optimal binary decision
boundaries for all forms of normally distributed training data.

A high level description of the linear SVM method in a geometric locus
framework is outlined next. The outline is intended to motivate the
development of a computer-implemented method that effectively hardwires the
geometric locus of a normal eigenaxis into linear kernel SVM architectures.
The term eigenlocus is used to refer to the locus of a normal eigenaxis.

\subsection{Hardwiring the Eigenlocus of a Normal Eigenaxis into Linear Kernel
SVM Architectures}

So far, the principal eigenaxis of linear curves and surfaces has been
identified and given the name \emph{normal eigenaxis}. It has been established
that normal eigenaxes of linear curves and surfaces are major intrinsic axes
that coincide as exclusive reference axes. It has been demonstrated that all
of the points on a linear locus are specified by the eigen-coordinate
locations and corresponding magnitude and eigenenergy of its normal eigenaxis.
Therefore, given the normal eigenaxis of any linear locus of points, it
follows that the geometric locus of any given normal eigenaxis has a
distinctive geometric configuration that is specified by a characteristic set
of eigen-coordinates, all of which jointly determine the characteristic
location and eigenenergy of the normal eigenaxis. Because normal eigenaxes of
distinct linear curves or surfaces possess invariant and distinctive geometric
locations, it follows that the location of a normal eigenaxis is an\ invariant
and hardwired geometric property of lines, planes, and hyperplanes. Thus,
\emph{the important generalizations for lines, planes, and hyperplanes,
are\ hardwired into (encoded within) the geometric locus of a normal
eigenaxis}. Furthermore, the normal eigenaxis of any given linear locus
satisfies the linear locus in terms of its eigenenergy. Thereby, \emph{the
fundamental property of a normal eigenaxis is its eigenenergy}.

Clearly, then, the primary curve of interest for learning linear decision
boundaries is a normal eigenaxis. This implies that the important
generalizations for linear decision boundaries involve an estimation process
that encodes the geometric locus of a normal eigenaxis within learning machine
architectures. Accordingly, robust and optimal estimates of linear decision
boundaries must be based on effective statistical representations of
constrained normal eigen-coordinate locations of unknown linear decision
boundaries. The remaining portions of this paper will refer to a normal
eigenaxis as a normal eigenlocus. The paper will use the term statistical
eigenlocus to refer to a statistical estimate of a normal eigenlocus. The
paper will use the term strong dual normal eigenlocus to refer to joint,
statistical eigenlocus estimates in dual, correlated Hilbert spaces. Given all
of the above assumptions, the remaining sections of this paper will develop
algebraic and statistical expressions for a strong dual normal eigenlocus of
normal eigenaxis components that provides a robust statistical representation
of the constrained normal eigen-coordinate locations of an unknown linear
decision boundary.

\subsubsection*{Definition}

A\ \ strong dual normal eigenlocus of normal eigenaxis components is a normal
eigenlocus of a linear decision boundary which encodes the constrained
eigen-coordinate locations of an unknown normal eigenaxis. A\ strong dual
normal eigenlocus satisfies a linear decision boundary in terms of a critical
minimum, i.e., a total allowed, eigenenergy. The term eigenlocus will be used
to refer to the location of a normal eigenaxis component on a normal
eigenlocus, or to the location of a normal eigenlocus; the context will make
the meaning clear.

\subsection{A Strong Dual Normal Eigenlocus of Normal Eigenaxis Components}

Consider a normal eigenlocus of a linear decision boundary formed by a strong
dual normal eigenlocus of normal eigenaxis components, all of which are
eigen-scaled extreme data points of large covariance, all of which encode a
robust likelihood ratio, each of which determines an eigen-transformed
principal location of large covariance. A strong dual normal eigenlocus of a
separating line, plane, or hyperplane will be shown to satisfy a linear
decision boundary in terms of a critical minimum eigenenergy. The analyses
that follow will use Eqs (\ref{Linear Locus Functional}),
(\ref{Normal Eigenaxis Functional}), (\ref{Normal Form Normal Eigenaxis}), and
(\ref{Geometric Property of Linear Loci}), along with the correlated algebraic
and geometric properties of these equations, to demonstrate how the
constrained normal eigen-coordinate locations of an unknown linear decision
boundary can be estimated from training data by means of a properly specified
strong dual normal eigenlocus of normal eigenaxis components, each of which
encodes the probability of finding an extreme data point in a particular
region of $\mathbf{%
\mathbb{R}
}^{d}$. The analyses will demonstrate that the eigenlocus of each normal
eigenaxis component encodes an eigen-balanced first and second order
statistical moment about the locus of an extreme training point, which is
shown to determine the likelihood of finding the extreme data point in a
particular region of $\mathbf{%
\mathbb{R}
}^{d}$. First and second order statistical moments, which involve
unidirectional estimates of joint variations between a given vector and a
collection of training data, provide an estimate of how the components of the
given vector are distributed within the training data. The eigen-balanced
first and second order statistical moments encoded within a strong dual normal
eigenlocus of normal eigenaxis components will be shown to describe a large
number of data distributions. A strong dual normal eigenlocus of normal
eigenaxis components will be formally referred to as a strong dual normal eigenlocus.

\subsection{High Level Overview of a Strong Dual Normal Eigenlocus}

Let the term \emph{strong dual normal eigenlocus}\textit{\ }refer to a dual
statistical eigenlocus of normal eigenaxis components which delineates and
satisfies three, symmetrical linear partitioning curves or surfaces. A strong
dual normal eigenlocus\textit{\ }satisfies three, symmetrical linear
partitioning curves or surfaces in terms of a critical minimum eigenenergy.
The normal eigenaxis components on a strong dual normal eigenlocus are
analogous to the sinusoidal components of a Fourier series. All of the
sinusoidal components of a given Fourier series have such amplitudes and
phases that they sum to an approximation of a distinct periodic function or
signal
\citet{Lathi1998}%
. Likewise, all of the normal eigenaxis components on a strong dual normal
eigenlocus have such magnitudes and directions that they sum to an estimate of
a normal eigenaxis of three, characteristic and symmetrical linear
partitioning curves or surfaces. How such a statistical balancing feat can be
routinely accomplished is a central idea of this paper. This paper will
examine how achieving this type of statistical equilibrium involves
identifying and exploiting effective statistical representations for
constrained eigen-coordinate locations of unknown normal eigenaxes of unknown
linear decision boundaries.

The sections that follow will examine an estimation process that transforms
two sets of pattern vectors, generated by any two probability distributions
whose expected values and covariance structures do not vary over time, into a
dual statistical eigenlocus of normal eigenaxis components, all of which are
jointly and symmetrically located in dual and primal, correlated Hilbert
spaces, all of which encode a robust likelihood ratio, all of which jointly
describe symmetrical, linear subspaces of $%
\mathbb{R}
^{N}$\ and $%
\mathbb{R}
^{d}$, each of which determines an eigen-balanced, pointwise covariance
estimate of an extreme data point located between two data distributions in $%
\mathbb{R}
^{d}$. All of the normal eigenaxis components on a strong dual normal
eigenlocus jointly determine a statistical decision system, of three,
symmetrical linear partitioning curves or surfaces in $%
\mathbb{R}
^{d}$, that delineates bipartite, congruent geometric regions of large
covariance located between two data distributions in $%
\mathbb{R}
^{d}$,\ such that the congruent geometric regions of large covariance
delineate regions of data distribution overlap for overlapping distributions.
The resultant loci of points on all three linear curves or surfaces in $%
\mathbb{R}
^{d}$ exclusively reference the dual statistical eigenlocus of normal
eigenaxis components. Likelihoods encoded within the eigenloci of all of the
normal eigenaxis components specify the stochastic behavior of a statistical
decision system.

The next section will begin to develop the primal and the Wolfe dual normal
eigenlocus equations of a probabilistic, binary linear classification system.

\section{The Primal and the Wolfe Dual Normal Eigenlocus Equations of a
Probabilistic Binary Linear Classification System}

The eigenlocus equations of a strong dual normal eigenlocus are commonly
referred to as soft margin linear support vector machines. The analyses that
follow will show that the subset of weighted training points, commonly called
support vectors, form a dual statistical eigenlocus of eigen-transformed
extreme training points, all of which jointly determine a statistical decision
system of three, symmetrical linear partitioning curves or surfaces in $%
\mathbb{R}
^{d}$. It will be demonstrated that each support vector is an eigen-scaled
extreme training point that determines a well-proportioned (eigen-balanced)
and properly-positioned normal eigenaxis component on a strong dual normal
eigenlocus. To wit, it will be demonstrated that support vectors are major
eigenaxis components of an intrinsic reference axis of a linear decision
boundary and bilaterally symmetrical borders.

Finding a separating line, plane, or hyperplane requires estimating the normal
eigenlocus of a linear decision boundary and the bilaterally symmetrical
borders which bound it. The analyses that follow will define the complete
statistical system of a strong dual normal eigenlocus, and thereby will
identify a probabilistic linear discriminant function that is Bayes' optimal
for common covariance data.

The study begins with the eigenlocus equation of a primal normal eigenlocus.

\subsection{Eigenlocus Equation of a Primal Normal Eigenlocus}

The strong dual normal eigenlocus $\mathbf{\tau}$ of a separating line, plane,
or hyperplane is estimated by solving an inequality constrained optimization
problem:%
\begin{align}
\min\Psi\left(  \mathbf{\tau}\right)   &  =\left\Vert \mathbf{\tau}\right\Vert
^{2}/2+C/2\sum\nolimits_{i=1}^{N}\xi_{i}^{2}\label{Primal Normal Eigenlocus}\\
\text{s.t. }y_{i}\left(  \mathbf{x}_{i}^{T}\mathbf{\tau}+\tau_{0}\right)   &
\geq1-\xi_{i},\ \xi_{i}\geq0,\ i=1,...,N\text{,}\nonumber
\end{align}
where $\mathbf{\tau}$ is a $d\times1$ constrained primal normal eigenlocus of
three, symmetrical linear partitioning curves or surfaces, $C$ and $\xi_{i}$
are regularization parameters, $y_{i}$ are training set labels (if
$\mathbf{x}_{i}\in H_{1}$, assign $y_{i}=1$; otherwise, assign $y_{i}=-1$),
and $\tau_{0}$ is a function of $\mathbf{\tau}$, extreme training points
$\mathbf{x}_{i\ast}$, and training set labels $y_{i}$.

It will be demonstrated that Eq. (\ref{Primal Normal Eigenlocus}) provides the
primal (elemental) specification of a linear decision boundary that is bounded
by bilaterally symmetrical decision borders. Any given linear decision
boundary is centrally and symmetrically positioned between any two given data
distributions, such that the linear decision borders span symmetrical regions
of large covariance. The strong dual solution of Eq.
(\ref{Primal Normal Eigenlocus}) involves solving a complementary and
essential optimization problem that determines the fundamental unknowns of Eq.
(\ref{Primal Normal Eigenlocus}). It is claimed that the actual unknowns in
Eq. (\ref{Primal Normal Eigenlocus}) are the constrained eigen-coordinate
locations of a normal eigenaxis $\mathbf{v}$ that delineates and satisfies
three, symmetrical lines, planes, or hyperplanes, all of which jointly
delineate a symmetrical partitioning of a feature space in $%
\mathbb{R}
^{d\text{ }}$.

It will be shown that the locations of the normal eigenaxis components on
$\mathbf{\tau}$ provide estimates for the constrained eigen-coordinate
locations of $\mathbf{v}$. It will also be demonstrated that $\mathbf{\tau}$
provides an exclusive, intrinsic reference axis for any given linear decision
boundary and decision borders.

Moreover, it will be shown that the strong dual solution of Eq.
(\ref{Primal Normal Eigenlocus}) determines a statistical equilibrium point,
i.e., the eigenlocus of $\mathbf{\tau}$\textbf{,} such that a constrained
discriminant function $\mathbf{\tau}^{T}\mathbf{x}+\tau_{0}$ delineates a
centrally and symmetrically positioned, bipartite geometric region of
constrained, constant, and equal widths, that spans a region of high
variability (large covariance) between two data distributions, whereby the
bipartite, congruent regions of large covariance delineate symmetrical regions
of data distribution overlap for overlapping data distributions.

It will also be shown that a\ strong dual normal eigenlocus $\mathbf{\tau}$
satisfies three, symmetrical linear partitioning curves or surfaces in terms
of a critical minimum eigenenergy. Thereby, it will be shown that a\ strong
dual normal eigenlocus $\mathbf{\tau}$ possesses a critical minimum
eigenenergy which is the fundamental geometric and statistical property of
$\mathbf{\tau}$.

\subsection{The Critical Minimum Eigenenergy Constraint on $\mathbf{\tau}$}

Given Eq. (\ref{Primal Normal Eigenlocus}) and the assumptions outlined above,
it follows that $N$ primal normal eigenlocus equations must be satisfied:%
\[
y_{i}\left(  \mathbf{x}_{i}^{T}\mathbf{\tau}+\tau_{0}\right)  \geq1-\xi
_{i},\ \xi_{i}\geq0,\ i=1,...,N\text{,}%
\]
such that a constrained primal normal eigenlocus $\mathbf{\tau}$ satisfies a
critical minimum eigenenergy constraint:%
\begin{equation}
\gamma\left(  \mathbf{\tau}\right)  =\left\Vert \mathbf{\tau}\right\Vert
_{\min_{c}}^{2}\text{,}
\label{Minimum Total Eigenenergy Primal Normal Eigenlocus}%
\end{equation}
where the total allowed eigenenergy $\left\Vert \mathbf{\tau}\right\Vert
_{\min_{c}}^{2}$ of $\mathbf{\tau}$ is the fundamental geometric and
statistical property of $\mathbf{\tau}$. It will be shown that $\mathbf{\tau}$
possesses a critical minimum eigenenergy%
\begin{align*}
\left\Vert \mathbf{\tau}\right\Vert _{\min_{c}}^{2}  &  =\left\Vert
\mathbf{\tau}_{1}-\mathbf{\tau}_{2}\right\Vert _{\min_{c}}^{2}\text{,}\\
&  =\left\Vert \mathbf{\tau}_{1}\right\Vert _{\min_{c}}^{2}+\left\Vert
\mathbf{\tau}_{2}\right\Vert _{\min_{c}}^{2}-2\left\Vert \mathbf{\tau}%
_{1}\right\Vert \left\Vert \mathbf{\tau}_{2}\right\Vert \cos\theta
_{\mathbf{\tau}_{1}\mathbf{\tau}_{2}}\text{,}%
\end{align*}
where $\mathbf{\tau}_{1}$ and $\mathbf{\tau}_{2}$ are components of
$\mathbf{\tau}$%
\[
\mathbf{\tau=\tau}_{1}-\mathbf{\tau}_{2}\text{,}%
\]
such that the total allowed eigenenergies of $\mathbf{\tau}_{1}$ and
$\mathbf{\tau}_{2}$ are effectively balanced by means of a symmetric equalizer
statistic $\nabla_{eq}$%
\[
\left(  \left\Vert \mathbf{\tau}_{1}\right\Vert _{\min_{c}}^{2}-\left\Vert
\mathbf{\tau}_{1}\right\Vert \left\Vert \mathbf{\tau}_{2}\right\Vert
\cos\theta_{\mathbf{\tau}_{1}\mathbf{\tau}_{2}}\right)  +\nabla_{eq}%
\Leftrightarrow\left(  \left\Vert \mathbf{\tau}_{2}\right\Vert _{\min_{c}}%
^{2}-\left\Vert \mathbf{\tau}_{1}\right\Vert \left\Vert \mathbf{\tau}%
_{2}\right\Vert \cos\theta_{\mathbf{\tau}_{1}\mathbf{\tau}_{2}}\right)
-\nabla_{eq}\text{,}%
\]
in relation to a centrally located statistical fulcrum $f_{s}$. It will also
be demonstrated that the critical minimum eigenenergy $\left\Vert
\mathbf{\tau}\right\Vert _{\min_{c}}^{2}$ exhibited by $\mathbf{\tau}$
determines a statistical equilibrium point which encodes a robust likelihood
ratio for all data distributions.

Figure $11$ illustrates the algebraic, geometric, and statistical nature of
the remarkable statistical balancing feat that is routinely accomplished by
solving the inequality constrained optimization problem in Eq.
(\ref{Primal Normal Eigenlocus}).

\begin{center}%
\begin{center}
\includegraphics[
natheight=7.499600in,
natwidth=9.999800in,
height=3.2897in,
width=4.3777in
]%
{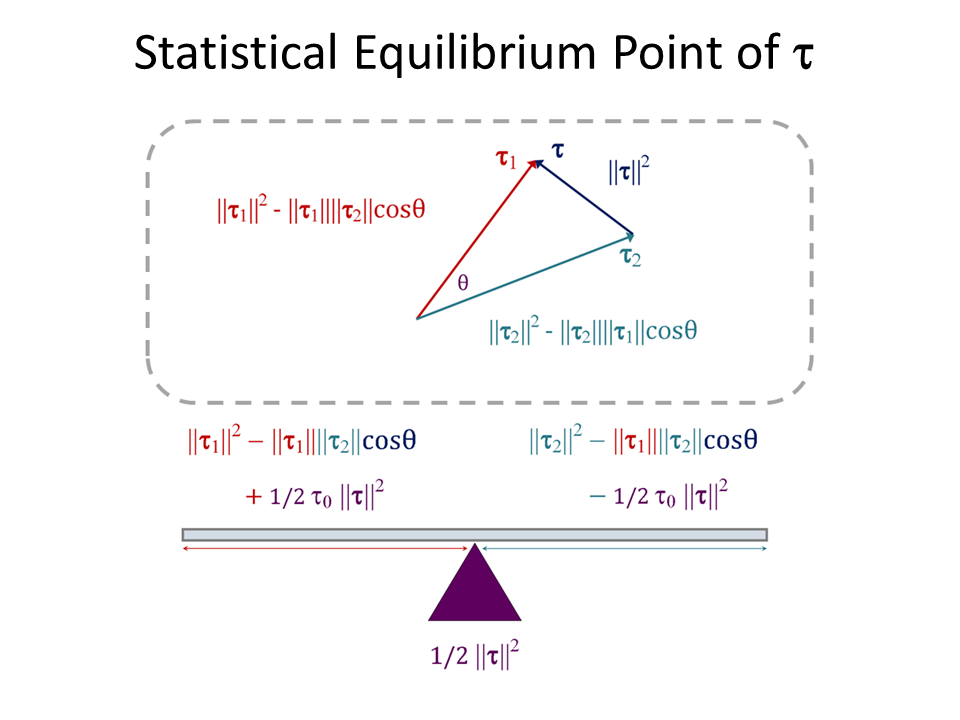}%
\end{center}

\end{center}

\begin{flushleft}
Figure $11$: Illustration of the algebraic and topological constraints which
determine a strong dual normal eigenlocus equilibrium point. The center of
eigenenergy of $\mathbf{\tau}$ is subjected to equal and opposite
eigenenergies, so that a strong dual normal eigenlocus $\mathbf{\tau=\tau}%
_{1}-\mathbf{\tau}_{2}$ achieves a state of statistical equilibrium.
\end{flushleft}

\subsection{The Strong Dual Normal Eigenlocus Equilibrium Point}

Section $17$ will show that the critical minimum eigenenergy constraint on
$\mathbf{\tau}$ determines a strong dual normal eigenlocus equilibrium point,
i.e., the location or eigenlocus of $\mathbf{\tau}$, whereby a constrained
discriminant function $\mathbf{\tau}^{T}\mathbf{x}+\tau_{0}$ delineates the
positions of three, symmetrical linear partitioning curves or surfaces.
Section $18$ will demonstrate that the total allowed eigenenergy and
statistical equilibrium point of $\mathbf{\tau}$ is specified by likelihood
statistics encoded within correlated normal eigenaxis components on a Wolfe
dual normal eigenlocus. Denote a Wolfe dual normal eigenlocus by
$\mathbf{\psi}$ and its eigenlocus equation by $\max\Xi\left(  \mathbf{\psi
}\right)  $.

Let $\mathbf{\psi}$ be a Wolfe dual of $\mathbf{\tau}$, such that proper and
effective strong duality relationships exist between the algebraic systems of
$\min\Psi\left(  \mathbf{\tau}\right)  $ and $\max\Xi\left(  \mathbf{\psi
}\right)  $. Thereby, let $\mathbf{\psi}$ be related with $\mathbf{\tau}$ in a
symmetrical manner that specifies the locations of the normal eigenaxis
components on $\mathbf{\tau}$. The Wolfe dual normal eigenlocus $\mathbf{\psi
}$ is important for the following reasons.

\subsection{Why the Wolfe Dual Normal Eigenlocus Matters}

Duality relationships for Lagrange multiplier problems are based on the
premise that it is the Lagrange multipliers which are the fundamental unknowns
associated with a constrained problem. Dual methods solve an alternate
problem, termed the dual problem, whose unknowns are the Lagrange multipliers
of the first problem, termed the primal problem. Once the Lagrange multipliers
are known, the solution to a primal problem can be determined
\citet{Luenberger2003}%
.

It is assumed that the real unknowns associated with the inequality
constrained optimization problem in Eq. (\ref{Primal Normal Eigenlocus}) are
the constrained eigen-coordinate locations of a normal eigenaxis $\mathbf{v}$
that delineates the geometric configuration of a linear decision boundary and
the widths of its decision borders. It is also assumed that a normal eigenaxis
$\mathbf{v}$ satisfies a linear decision boundary and its decision borders in
terms of a critical minimum, i.e., a total allowed, eigenenergy. The main
issue concerns how the constrained eigen-coordinate locations of a normal
eigenaxis $\mathbf{v}$ are determined.

It will be demonstrated that the constrained eigen-coordinate locations of
$\mathbf{v}$ are estimated by the locations of normal eigenaxis components on
a constrained primal normal eigenlocus $\mathbf{\tau}$, all of which are
effectively determined by the locations of normal eigenaxis components on a
Wolfe dual normal eigenlocus $\mathbf{\psi}$. To wit, it will be shown that
the constrained eigen-coordinate locations of $\mathbf{v}$ are essentially
determined by the eigenloci of normal eigenaxis components on a Wolfe dual
normal eigenlocus $\mathbf{\psi}$. It will be demonstrated that the eigenloci
of the Wolfe dual normal eigenaxis components on $\mathbf{\psi}$ determine
critical minimum eigenenergies for $\mathbf{\psi}$ and $\mathbf{\tau}$, both
of which jointly determine a statistical equilibrium point for $\mathbf{\tau}%
$. Section $17$ will define a strong dual normal eigenlocus equilibrium point
for which the critical minimum eigenenergies of $\mathbf{\tau=\tau}%
_{1}-\mathbf{\tau}_{2}$ satisfy a linear decision boundary and its decision borders.

\subsection{Fundamental Unknowns for Strong Dual Normal Eigenlocus Estimates}

For the problem of strong dual normal eigenlocus estimates, the Lagrange
multipliers method introduces a Wolfe dual normal eigenlocus $\mathbf{\psi}$
of normal eigenaxis components, for which the Lagrange multipliers $\left\{
\psi_{i}\right\}  _{i=1}^{N}$ are the magnitudes or lengths of a set of Wolfe
dual normal eigenaxis components $\left\{  \psi_{i}\overrightarrow{\mathbf{e}%
}_{i}\right\}  _{i=1}^{N}$, and finds extrema for the restriction of a primal
normal eigenlocus $\mathbf{\tau}$ to a Wolfe dual eigenspace. Accordingly, the
fundamental unknowns associated with Eq. (\ref{Primal Normal Eigenlocus}) are
the magnitudes or lengths of the Wolfe dual normal eigenaxis components on
$\mathbf{\psi}$. It will be shown that each Lagrange multiplier provides an
eigen-scale that determines the length of a correlated, constrained primal
normal eigenaxis component on $\mathbf{\tau}$.

Because Eq. (\ref{Primal Normal Eigenlocus}) is a convex programming problem,
the theorem for convex duality guarantees some type of equivalence and
corresponding symmetry between a constrained primal normal eigenlocus and its
Wolfe dual. Strong duality holds between the algebraic systems of $\min
\Psi\left(  \mathbf{\tau}\right)  $ and $\max\Xi\left(  \mathbf{\psi}\right)
$, so that the duality gap between the constrained primal and the Wolfe dual
normal eigenlocus solution is zero
\citet{Luenberger1969}%
,
\citet{Nash1996}%
,
\citet{Fletcher2000}%
,
\citet{Luenberger2003}%
.

This paper will demonstrate how strong duality relationships between the
algebraic systems of $\min\Psi\left(  \mathbf{\tau}\right)  $ and $\max
\Xi\left(  \mathbf{\psi}\right)  $ ensure that the total allowed eigenenergy
$\left\Vert \mathbf{\psi}\right\Vert _{\min_{c}}^{2}$ and the eigenlocus of
$\mathbf{\psi}$ are symmetrically related to the total allowed eigenenergy
$\left\Vert \mathbf{\tau}\right\Vert _{\min_{c}}^{2}$ and the statistical
equilibrium point of $\mathbf{\tau}$. These relationships will be defined in
Sections $12$, $14$, $15$, $16$, and $17$. Section $18$ will show that the
total allowed eigenenergy $\left\Vert \mathbf{\tau}\right\Vert _{\min_{c}}%
^{2}$ of $\mathbf{\tau}$ describes the likelihood of finding data points in
particular regions of $%
\mathbb{R}
^{d}$. Section $18$ will also demonstrate that the statistical equilibrium
point of $\mathbf{\tau}$ encodes Bayes' likelihood ratio for common covariance
data and a robust likelihood ratio for all other data.

This paper will also demonstrate how strong duality relationships between the
algebraic systems of $\min\Psi\left(  \mathbf{\tau}\right)  $ and $\max
\Xi\left(  \mathbf{\psi}\right)  $ ensure that the geometric architecture
described by $\max\Xi\left(  \mathbf{\psi}\right)  $ is symmetrically related
to the geometric architecture of the statistical decision system described by
$\min\Psi\left(  \mathbf{\tau}\right)  $. The strong duality relationships
between the algebraic systems of $\min\Psi\left(  \mathbf{\tau}\right)  $ and
$\max\Xi\left(  \mathbf{\psi}\right)  $ will be shown to determine symmetrical
linear partitioning systems in $%
\mathbb{R}
^{N}$ and $%
\mathbb{R}
^{d}$, which jointly determine a learning machine architecture in $%
\mathbb{R}
^{d}$ that exhibits a surprising amount of bilateral symmetry for arbitrary
data distributions.

The term \emph{strong dual} will frequently be used to emphasize the joint
geometric and statistical properties exhibited by a constrained primal and a
Wolfe dual normal eigenlocus. The matrix version of the Wolfe dual normal
eigenlocus equation is summarized below.

\subsection{The Wolfe Dual Normal Eigenlocus of a Separating Hyperplane}

The complementary and essential normal eigenlocus estimate, which is specified
by the Wolfe dual normal eigenlocus of Eqs (\ref{Wolfe Dual Normal Eigenlocus}%
) or (\ref{Vector Form Wolfe Dual}), involves finding normal eigenaxis
components that are determined by the minimization of a constrained quadratic
form%
\[
\max\Xi\left(  \mathbf{\psi}\right)  =\mathbf{1}^{T}\mathbf{\psi}%
-\frac{\mathbf{\psi}^{T}\mathbf{Q\psi}}{2}\text{,}%
\]
subject to the constraints $\mathbf{\psi}^{T}\mathbf{y}=0$ and $\psi_{i}\geq
0$, where $\mathbf{Q}\triangleq\epsilon\mathbf{I}+\widetilde{\mathbf{X}%
}\widetilde{\mathbf{X}}^{T}$, the matrix $\widetilde{\mathbf{X}}%
\triangleq\mathbf{D}_{y}\mathbf{X}$, $\mathbf{D}_{y}$ is an $N\times N$
diagonal matrix of training labels $y_{i}$ and the $N\times d$ data matrix is
$\mathbf{X}$ $=%
\begin{pmatrix}
\mathbf{x}_{1}, & \mathbf{x}_{2}, & \ldots, & \mathbf{x}_{N}%
\end{pmatrix}
^{T}$. The eigenlocus equation of $\max\Xi\left(  \mathbf{\psi}\right)  $ will
be derived in sections that follow.

The analyses that follow will examine how the strong duality relationships
between the algebraic systems of $\min\Psi\left(  \mathbf{\tau}\right)  $ and
$\max\Xi\left(  \mathbf{\psi}\right)  $ determine a strong dual normal
eigenlocus of symmetrical linear partitioning systems in $%
\mathbb{R}
^{N}$ and $%
\mathbb{R}
^{d}$. The critical minimum eigenenergy $\left\Vert \mathbf{\psi}\right\Vert
_{\min_{c}}^{2}$ of $\mathbf{\psi}$ will be shown to be symmetrically related
to the critical minimum eigenenergy $\left\Vert \mathbf{\tau}\right\Vert
_{\min_{c}}^{2}$ of $\mathbf{\tau}$. Thereby, this paper will demonstrate how
the geometric configuration of a Wolfe dual normal eigenlocus $\mathbf{\psi}$
determines the geometric configuration of a constrained primal normal
eigenlocus $\mathbf{\tau}$.

\subsection{Symmetrical Linear Partitioning Systems in $%
\mathbb{R}
^{N}$ and $%
\mathbb{R}
^{d}$}

Equation (\ref{Primal Normal Eigenlocus}) and the existence of strong duality
relationships between the algebraic systems of $\min\Psi\left(  \mathbf{\tau
}\right)  $ and $\max\Xi\left(  \mathbf{\psi}\right)  $ indicate that three,
symmetrical hyperplane partitioning surfaces are delineated by the constrained
quadratic form denoted by $\max\Xi\left(  \mathbf{\psi}\right)  $. Given these
assumptions and Eqs (\ref{Normal Form Normal Eigenaxis}) and
(\ref{Geometric Property of Linear Loci}), it follows that any point on a
hyperplane surface possesses a set of normalized, eigen-scaled coordinates
which satisfy the distance of the hyperplane surface from the origin, where
each distance is determined by a correlated constraint on the constrained
discriminant function of Eq. (\ref{Discriminant Function}). Section $11$ will
show that the geometric configurations of all three hyperplane surfaces are an
inherent function of the inner product elements of the Gram matrix
$\mathbf{Q}$ associated with the constrained quadratic form in the equation
denoted by $\max\Xi\left(  \mathbf{\psi}\right)  $ or Eq.
(\ref{Vector Form Wolfe Dual}). Sections $12$ - $16$ will examine how a Wolfe
dual normal eigenlocus $\mathbf{\psi}$ delineates and satisfies three,
symmetrical hyperplane partitioning surfaces in terms of a critical minimum
eigenenergy constraint.

Sections $14$ and $15$ will examine how the Lagrange multipliers of a primal
normal eigenlocus problem provide an estimate of constrained normal
eigen-coordinate locations that implicitly delineate a separating hyperplane
in $%
\mathbb{R}
^{N}$ which is effectively bounded by bilateral symmetrical hyperplane
borders. Sections $14$ and $15$ will show how each of the normal eigenaxis
components on $\mathbf{\psi}\in%
\mathbb{R}
^{N}$ encodes an eigen-scale that determines a critical length for a
symmetrical normal eigenaxis component on $\mathbf{\tau}\in%
\mathbb{R}
^{d}$, such that $\mathbf{\tau}$ delineates a statistical decision system of
three, symmetrical linear partitioning curves or surfaces in $%
\mathbb{R}
^{d}$.

Figure $12$\ depicts a high level overview of the symmetrical relationships
between a constrained primal normal eigenlocus $\mathbf{\tau}$ and its Wolfe
dual $\mathbf{\psi}$, where symmetry involves regularized correlations between
the critical minimum eigenenergies of $\mathbf{\tau}$ and $\mathbf{\psi}$,
which jointly determine the statistical equilibrium point $\mathbf{\tau}$ that
is satisfied by $\mathbf{\tau}$ and $\mathbf{\psi}$, all of which jointly
determine regularized correlations between dual, linear partitioning systems
in $%
\mathbb{R}
^{d}$ and $%
\mathbb{R}
^{N}$.

Denote the set of hyperplane partitioning surfaces in $%
\mathbb{R}
^{N}$ by $H_{0}$, $H_{+1}$, and $H_{-1}$, where the critical minimum
eigenenergy $\left\Vert \mathbf{\psi}\right\Vert _{\min_{c}}^{2}$ of
$\mathbf{\psi}$ satisfies $H_{0}$, $H_{+1}$, and $H_{-1}$. Let the set of
linear partitioning surfaces in $%
\mathbb{R}
^{d}$, which are determined by constraining the expression $\mathbf{\tau}%
^{T}\mathbf{x}+\tau_{0}$ to be equal to $0$, $+1$, and $-1$, be denoted by
$D_{0}\left(  \mathbf{x}\right)  $, $D_{+1}\left(  \mathbf{x}\right)  $, and
$D_{-1}\left(  \mathbf{x}\right)  $, where the critical minimum eigenenergy
$\left\Vert \mathbf{\tau}\right\Vert _{\min_{c}}^{2}$ of $\mathbf{\tau}$
satisfies $D_{0}\left(  \mathbf{x}\right)  $, $D_{+1}\left(  \mathbf{x}%
\right)  $, and $D_{-1}\left(  \mathbf{x}\right)  $. Figure $12$ illustrates
how strong duality relationships between the algebraic systems of $\min
\Psi\left(  \mathbf{\tau}\right)  $ and $\max\Xi\left(  \mathbf{\psi}\right)
$ ensure that the geometric configurations of the hyperplane partitioning
surfaces $H_{0}$, $H_{+1}$, and $H_{-1}$ in $%
\mathbb{R}
^{N}$ regulate the geometric configurations of the linear partitioning
surfaces $D_{0}\left(  \mathbf{x}\right)  $, $D_{+1}\left(  \mathbf{x}\right)
$, and $D_{-1}\left(  \mathbf{x}\right)  $ in $%
\mathbb{R}
^{d}$. Accordingly, the geometric configuration of a separating hyperplane
$H_{0}$ is symmetrically related to the geometric configuration of a linear
decision boundary $D_{0}\left(  \mathbf{x}\right)  $. Likewise, the geometric
configurations of the hyperplane decision borders $H_{+1}$ and $H_{-1}$ are
symmetrically related to the geometric configurations of the linear decision
borders $D_{+1}\left(  \mathbf{x}\right)  $, and $D_{-1}\left(  \mathbf{x}%
\right)  $.

\begin{center}
\textbf{%
\begin{center}
\includegraphics[
natheight=7.499600in,
natwidth=9.999800in,
height=3.2897in,
width=4.3777in
]%
{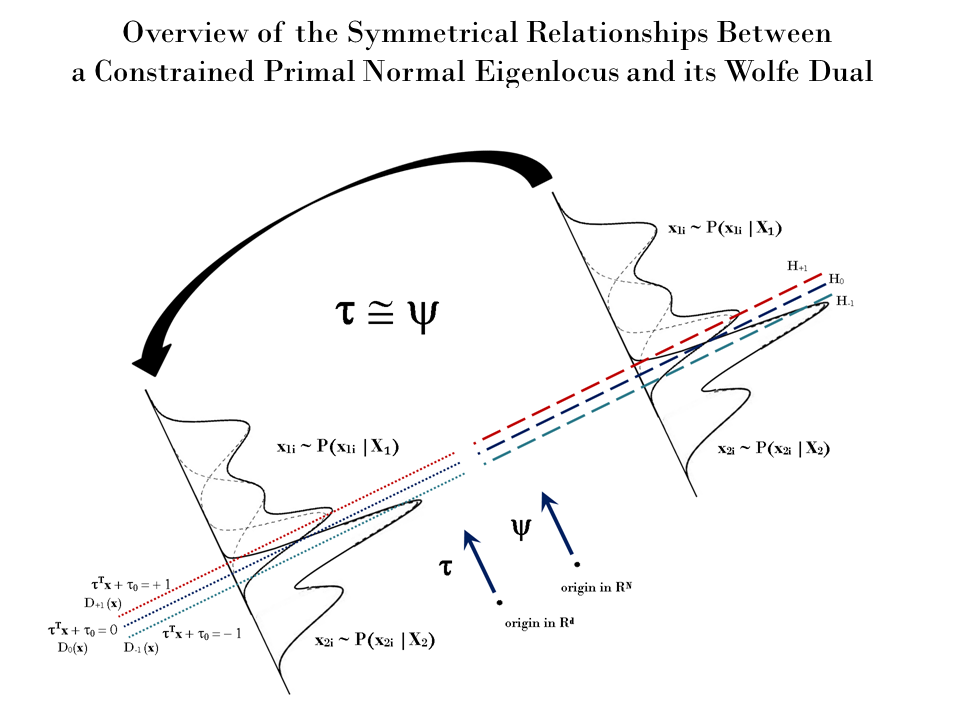}%
\end{center}
}
\end{center}

\begin{flushleft}
Figure $12$: Illustration of geometric and topological symmetries between
correlated linear partitioning systems of a constrained primal normal
eigenlocus $\mathbf{\tau}$ and its Wolfe dual $\mathbf{\psi}$, all of which
are created by the strong duality relationships between the algebraic systems
of $\min\Psi\left(  \mathbf{\tau}\right)  $ and $\max\Xi\left(  \mathbf{\psi
}\right)  $.
\end{flushleft}

\subsection{Strong Duality Relationships Between a Constrained Primal and a
Wolfe Dual Normal Eigenlocus}

All of the constrained primal normal eigenaxis components on a strong dual
normal eigenlocus $\mathbf{\tau}$ possess magnitudes and directions that
jointly determine the constrained eigen-coordinate locations of an unknown
normal eigenaxis of a symmetrical set of linear partitioning curves or
surfaces in $%
\mathbb{R}
^{d}$. A comprehensive examination of the statistical decision system of a
strong dual normal eigenlocus will reveal how this statistical balancing feat
is routinely accomplished by solving the inequality constrained optimization
problem of Eq. (\ref{Primal Normal Eigenlocus}). Sections $7$ $- $ $12$ will
identify strong duality relationships between the algebraic systems of the
constrained primal and the Wolfe dual normal eigenaxis components of $\min
\Psi\left(  \mathbf{\tau}\right)  $ and $\max\Xi\left(  \mathbf{\psi}\right)
$. Sections $14$ and $15$ will demonstrate how the eigenloci of the normal
eigenaxis components on the constrained primal normal eigenlocus
$\mathbf{\tau}$ of $\min\Psi\left(  \mathbf{\tau}\right)  $ are completely
specified by the eigenloci of the Wolfe dual normal eigenaxis components of
$\max\Xi\left(  \mathbf{\psi}\right)  $. Sections $14$ and $15$ will also
identify geometric and statistical properties which are jointly exhibited by
all of the normal eigenaxis components on $\mathbf{\tau}$ and its symmetrical
Wolfe dual $\mathbf{\psi}$.

\subsection{Uniform Geometric and Statistical Properties Jointly Exhibited by
Correlated Normal Eigenaxis Components on $\mathbf{\tau}$ and $\mathbf{\psi}$}

The strong duality relationships between the constrained primal normal
eigenlocus of $\min\Psi\left(  \mathbf{\tau}\right)  $ and the Wolfe dual
normal eigenlocus of $\max\Xi\left(  \mathbf{\psi}\right)  $ ensure that
correlated normal eigenaxis components on $\mathbf{\tau}$ and $\mathbf{\psi}$
exhibit symmetrical geometric and statistical properties. Sections $14$ and
$15$ will demonstrate how the geometric locations of the Wolfe dual normal
eigenaxis components are symmetrically correlated to the geometric locations
of their constrained primal counterparts, such that all of the constrained
primal normal eigenaxis components on $\mathbf{\tau}$ determine principal
locations of large covariance. Sections $14$ and $15$ will also demonstrate
that correlated normal eigenaxis components on $\mathbf{\tau}$ and
$\mathbf{\psi}$ exhibit directional symmetry. Sections $14$ and $15$ will
examine how eigen-balanced, symmetrical relationships between all of
the\ normal eigenaxis components on $\mathbf{\tau}$ and $\mathbf{\psi}$
determine suitable magnitudes and geometric locations for each of the
constrained primal and Wolfe dual normal eigenaxis components. Thereby, it
will be shown that the eigenlocus of each constrained primal normal eigenaxis
component on $\mathbf{\tau}$ is jointly delineated by the eigenloci of a
constrained primal and a Wolfe dual normal eigenaxis component, both of which
symmetrically encode magnitudes and directions of large covariance in $%
\mathbb{R}
^{d}$ and $%
\mathbb{R}
^{N}$ respectively. Thereby, it will be demonstrated that the geometric and
statistical properties which are jointly exhibited by $\mathbf{\tau}$ and
$\mathbf{\psi}$ involve similarities in magnitudes and directions of
correlated constrained primal and Wolfe dual normal eigenaxis components, all
of which determine elegant correlations between the total allowed
eigenenergies of $\mathbf{\tau}$ and $\mathbf{\psi}$ and the statistical
equilibrium point of $\mathbf{\tau}$. Accordingly, it will be shown that
$\mathbf{\tau}$ and $\mathbf{\psi}$ delineate interconnected, dual geometric
architectures of symmetrical linear partitions, which jointly determine
probabilistic linear discriminant functions. Moreover, it will be shown that
the regularized, data-driven geometric architectures, which are jointly
delineated by $\mathbf{\psi}$ and $\mathbf{\tau}$, determine statistical
decision systems that provide a robust means for recognizing unknown objects.

\subsection{Fundamental Relationships Between Joint Statistical Estimates of
$\mathbf{\tau}$ and $\mathbf{\psi}$}

It is claimed that the fundamental geometric and statistical property of a
strong dual normal eigenlocus $\mathbf{\tau}$ is its total allowed
eigenenergy. Furthermore, it is claimed that $\mathbf{\tau}$ exhibits a
critical minimum eigenenergy$\left\Vert \mathbf{\tau}\right\Vert _{\min_{c}%
}^{2}$ which effectively characterizes the geometric configuration of a linear
decision boundary and the widths of its decision borders. It is also claimed
that $\mathbf{\tau}$ satisfies a linear decision boundary and its decision
borders in terms of its critical minimum eigenenergy.

Given the strong duality relationships between the joint statistical estimates
of $\mathbf{\tau}$ and $\mathbf{\psi}$, it follows that a constrained primal
normal eigenlocus $\mathbf{\tau}$ of $\min\Psi\left(  \mathbf{\tau}\right)  $
exhibits a statistical equilibrium point which is symmetrically related to and
determined by the eigenlocus of its Wolfe dual $\mathbf{\psi}$ of $\max
\Xi\left(  \mathbf{\psi}\right)  $. Therefore, the total allowed eigenenergies
$\left\Vert \mathbf{\tau}\right\Vert _{\min_{c}}^{2}$ and $\left\Vert
\mathbf{\psi}\right\Vert _{\min_{c}}^{2}$ of $\mathbf{\tau}$ and
$\mathbf{\psi}$ are symmetrically related to each other%
\[
\left\Vert \mathbf{\tau}\right\Vert _{\min_{c}}^{2}\cong\left\Vert
\mathbf{\psi}\right\Vert _{\min_{c}}^{2}\text{,}%
\]
in a manner that determines the critical minimum eigenenergy $\left\Vert
\mathbf{\tau}\right\Vert _{\min_{c}}^{2}$ of $\mathbf{\tau}$. Section $17$
will develop algebraic and statistical expressions that describe symmetrical
relationships between the total allowed eigenenergies $\left\Vert
\mathbf{\tau}\right\Vert _{\min_{c}}^{2}$ of $\mathbf{\tau}$ and the
magnitudes or lengths of the Wolfe dual normal eigenaxis components on
$\mathbf{\psi}$. Section $17$ will develop an identity for which the total
allowed eigenenergies $\left\Vert \mathbf{\tau}\right\Vert _{\min_{c}}%
^{2}\cong\left\Vert \mathbf{\tau}_{1}-\mathbf{\tau}_{2}\right\Vert _{\min_{c}%
}^{2}$ of a strong dual normal eigenlocus $\mathbf{\tau=\tau}_{1}%
-\mathbf{\tau}_{2}$ satisfy the law of cosines in a surprisingly elegant and
symmetric manner. Thereby, Section $17$ will show that the total allowed
eigenenergies $\left\Vert \mathbf{\tau}_{1}-\mathbf{\tau}_{2}\right\Vert
_{\min_{c}}^{2}$ of $\mathbf{\tau}$ are consistent with the conservation of
energy. Section $18$ will show that the squares $\left\Vert \psi_{1_{i\ast}%
}\mathbf{x}_{1_{i\ast}}\right\Vert _{\min_{c}}^{2}$ and $\left\Vert
\psi_{2_{i\ast}}\mathbf{x}_{2_{i\ast}}\right\Vert _{\min_{c}}^{2}$ of the
constrained primal normal eigenaxis components\textbf{\ }$\psi_{1_{i\ast}%
}\mathbf{x}_{1_{i\ast}}$\textbf{\ }and $\psi_{2_{i\ast}}\mathbf{x}_{2_{i\ast}%
}$\textbf{\ }determine the probabilities of finding extreme data points
$\mathbf{x}_{1_{i\ast}}$ and $\mathbf{x}_{2_{i\ast}}$ in particular regions of
$\mathbf{%
\mathbb{R}
}^{d}$, where $\left\Vert \psi_{1_{i\ast}}\mathbf{x}_{1_{i\ast}}\right\Vert
_{\min_{c}}^{2}$ is the total allowed eigenenergy of $\psi_{1_{i\ast}%
}\mathbf{x}_{1_{i\ast}}$ and $\left\Vert \psi_{2_{i\ast}}\mathbf{x}_{2_{i\ast
}}\right\Vert _{\min_{c}}^{2}$ is the total allowed eigenenergy of
$\psi_{2_{i\ast}}\mathbf{x}_{2_{i\ast}}$. All of these results will be used to
demonstrate how the strong duality relationships between $\mathbf{\tau}$ and
$\mathbf{\psi}$ enable joint statistical estimates of the constrained
eigen-coordinate locations of an unknown normal eigenaxis\emph{\ }$\mathbf{v}%
$\emph{\ }in $%
\mathbb{R}
^{d}$.

The regularized Wolfe dual for the strong dual normal eigenlocus problem will
be derived by means of the Lagrangian described in the next section. Several
strong dual normal eigenlocus equations will be introduced and developed, all
of which jointly determine a statistical decision system for probabilistic
linear classification.

\section{The Lagrangian of the Primal Normal Eigenlocus}

The inequality constrained optimization problem in Eq.
(\ref{Primal Normal Eigenlocus}) is solved by using Lagrange multipliers
$\psi_{i}\geq0$ and the Lagrangian:%
\begin{align}
L_{\Psi\left(  \mathbf{\tau}\right)  }\left(  \mathbf{\tau,}\tau
_{0},\mathbf{\xi},\mathbf{\psi}\right)   &  =\left\Vert \mathbf{\tau
}\right\Vert ^{2}/2\label{Lagrangian Normal Eigenlocus}\\
&  +C/2\sum\nolimits_{i=1}^{N}\xi_{i}^{2}\nonumber\\
&  -\sum\nolimits_{i=1}^{N}\psi_{i}\nonumber\\
&  \times\left\{  y_{i}\left(  \mathbf{x}_{i}^{T}\mathbf{\tau}+\tau
_{0}\right)  -1+\xi_{i}\right\}  \text{.}\nonumber
\end{align}
The Karush-Kuhn-Tucker (KKT) constraints on the Lagrangian $L_{\Psi\left(
\mathbf{\tau}\right)  }$ specify a statistical decision system for
probabilistic linear discrimination. It will be shown that the constrained
Lagrangian functional $L_{\Psi\left(  \mathbf{\tau}\right)  }$ of Eq.
(\ref{Lagrangian Normal Eigenlocus}) returns the minimum number of normal
eigenaxis components that are necessary to symmetrically partition a two class
feature space. The KKT constraints on $L_{\Psi\left(  \mathbf{\tau}\right)  }$
are summarized below
\citet{Cristianini2000}%
,
\citet{Scholkopf2002}%
.

\subsection{A Statistical Decision System for Probabilistic Binary Linear
Classification}

The KKT constraints on the Lagrangian functional $L_{\Psi\left(  \mathbf{\tau
}\right)  }$:%
\begin{equation}
\mathbf{\tau}-\sum\nolimits_{i=1}^{N}\psi_{i}y_{i}\mathbf{x}_{i}=0,\text{
\ }i=1,...N\text{,} \label{KKTE1}%
\end{equation}%
\begin{equation}
\sum\nolimits_{i=1}^{N}\psi_{i}y_{i}=0,\text{ \ }i=1,...,N\text{,}
\label{KKTE2}%
\end{equation}%
\begin{equation}
C\sum\nolimits_{i=1}^{N}\xi_{i}-\sum\nolimits_{i=1}^{N}\psi_{i}=0\text{,}
\label{KKTE3}%
\end{equation}%
\begin{equation}
y_{i}\left(  \mathbf{x}_{i}^{T}\mathbf{\tau}+\tau_{0}\right)  -1+\xi_{i}%
\geq0,\text{ \ }i=1,...,N\text{,} \label{KKTE4}%
\end{equation}%
\begin{equation}
\psi_{i}\geq0,\text{ \ }i=1,...,N\text{,} \label{KKTE5}%
\end{equation}%
\begin{equation}
\psi_{i}\left\{  y_{i}\left(  \mathbf{x}_{i}^{T}\mathbf{\tau}+\tau_{0}\right)
-1+\xi_{i}\right\}  \geq0,\ i=1,...,N\text{,} \label{KKTE6}%
\end{equation}
determine a statistical discriminant function%
\begin{equation}
D\left(  \mathbf{x}\right)  =\mathbf{\tau}^{T}\mathbf{x}+\tau_{0}\text{,}
\label{Discriminant Function}%
\end{equation}
which satisfies the set of constraints:%
\begin{align*}
D_{0}\left(  \mathbf{x}\right)   &  =0\text{,}\\
D_{+1}\left(  \mathbf{x}\right)   &  =+1\text{,}\\
D_{-1}\left(  \mathbf{x}\right)   &  =-1\text{.}%
\end{align*}
It will now be shown that the above set of constraints on Eq.
(\ref{Discriminant Function}) determine three strong dual normal eigenlocus
equations of symmetrical linear partitioning curves or surfaces, where each of
the points on all three linear loci reference $\mathbf{\tau}$. Returning to
Eq. (\ref{Normal Form Normal Eigenaxis}), recall that the locus equation of a
normal eigenaxis $\mathbf{v}$ can be written as:%
\[
\frac{\mathbf{x}^{T}\mathbf{v}}{\left\Vert \mathbf{v}\right\Vert }=\left\Vert
\mathbf{v}\right\Vert \text{,}%
\]
where the normal eigenaxis $\mathbf{v}/\left\Vert \mathbf{v}\right\Vert $ has
length $1$ and points in the direction of the principal eigenvector
$\mathbf{v}$, such that $\left\Vert \mathbf{v}\right\Vert $ is the distance of
a specified line, plane, or hyperplane to the origin. Any point $\mathbf{x}$
that satisfies the above locus equation is on the linear locus of points
specified by $\mathbf{v}$, where all of the points $\mathbf{x}$ on the linear
locus reference $\mathbf{v}$.

Equation (\ref{Normal Form Normal Eigenaxis}) and the set of constraints
satisfied by the discriminant function $D\left(  \mathbf{x}\right)  $ of Eq.
(\ref{Discriminant Function}) are now used to obtain the set of strong dual
normal eigenlocus equations that delineate a linear decision boundary and its
bilaterally symmetrical linear decision borders.

\subsubsection{Eigenlocus Equation of the Linear Decision Boundary}

Using Eq. (\ref{Normal Form Normal Eigenaxis}) and assuming that $D\left(
\mathbf{x}\right)  =0$, the discriminant function%
\[
D\left(  \mathbf{x}\right)  =\mathbf{\tau}^{T}\mathbf{x}+\tau_{0}\text{,}%
\]
can be rewritten as:%
\begin{equation}
\frac{\mathbf{x}^{T}\mathbf{\tau}}{\left\Vert \mathbf{\tau}\right\Vert
}=-\frac{\tau_{0}}{\left\Vert \mathbf{\tau}\right\Vert }\text{,}
\label{Decision Boundary}%
\end{equation}
where $\frac{\left\vert \tau_{0}\right\vert }{\left\Vert \mathbf{\tau
}\right\Vert }$ is the distance of a linear decision boundary $D_{0}\left(
\mathbf{x}\right)  $ to the origin. Any point $\mathbf{x}$ that satisfies Eq.
(\ref{Decision Boundary}) is on the linear decision boundary $D_{0}\left(
\mathbf{x}\right)  $. All of the points $\mathbf{x}$ on $D_{0}\left(
\mathbf{x}\right)  $ reference $\mathbf{\tau}$. It has been demonstrated by
analyses and simulation studies that the linear decision boundary of Eq.
(\ref{Decision Boundary}) optimally partitions the normally distributed
training data described by Eq.
(\ref{Linearly Separable Classification Problem})
\citet{Reeves2009}%
.

\subsubsection{Eigenlocus Equation of the $D_{+1}\left(  \mathbf{x}\right)  $
Decision Border}

Using Eq. (\ref{Normal Form Normal Eigenaxis}) and assuming that $D\left(
\mathbf{x}\right)  =1$, the discriminant function of Eq.
(\ref{Discriminant Function}) can be rewritten as:%
\begin{equation}
\frac{\mathbf{x}^{T}\mathbf{\tau}}{\left\Vert \mathbf{\tau}\right\Vert
}=-\frac{\tau_{0}}{\left\Vert \mathbf{\tau}\right\Vert }+\frac{1}{\left\Vert
\mathbf{\tau}\right\Vert }\text{,} \label{Decision Border One}%
\end{equation}
where $\frac{\left\vert 1-\tau_{0}\right\vert }{\left\Vert \mathbf{\tau
}\right\Vert }$ is the distance of the linear decision border $D_{+1}\left(
\mathbf{x}\right)  $ to the origin. Any point $\mathbf{x}$ that satisfies Eq.
(\ref{Decision Border One}) is on the linear decision border $D_{+1}\left(
\mathbf{x}\right)  $. All of the points $\mathbf{x}$ on $D_{+1}\left(
\mathbf{x}\right)  $ reference $\mathbf{\tau}$.

\subsubsection{Eigenlocus Equation of the $D_{-1}\left(  \mathbf{x}\right)  $
Decision Border}

Using Eq. (\ref{Normal Form Normal Eigenaxis}) and assuming that $D\left(
\mathbf{x}\right)  =-1$, the discriminant function of Eq.
(\ref{Discriminant Function}) can be rewritten as:%

\begin{equation}
\frac{\mathbf{x}^{T}\mathbf{\tau}}{\left\Vert \mathbf{\tau}\right\Vert
}=-\frac{\tau_{0}}{\left\Vert \mathbf{\tau}\right\Vert }-\frac{1}{\left\Vert
\mathbf{\tau}\right\Vert }\text{,} \label{Decision Border Two}%
\end{equation}
where $\frac{\left\vert -1-\tau_{0}\right\vert }{\left\Vert \mathbf{\tau
}\right\Vert }$ is the distance of the linear decision border $D_{-1}\left(
\mathbf{x}\right)  $ to the origin. Any point $\mathbf{x}$ that satisfies Eq.
(\ref{Decision Border Two}) is on the linear decision border $D_{-1}\left(
\mathbf{x}\right)  $. All of the points $\mathbf{x}$ on $D_{-1}\left(
\mathbf{x}\right)  $ reference $\mathbf{\tau}$.

It is concluded that the constrained discriminant function $D\left(
\mathbf{x}\right)  $ of Eq. (\ref{Discriminant Function}) determines three,
symmetrical linear partitioning curves or surfaces, where all of the points on
$D_{0}\left(  \mathbf{x}\right)  $, $D_{+1}\left(  \mathbf{x}\right)  $, and
$D_{-1}\left(  \mathbf{x}\right)  $ exclusively reference $\mathbf{\tau}$. The
eigenlocus equations of the linear decision borders are now used to obtain an
algebraic expression for the distance between the linear decision borders.

\subsubsection*{Distance Between the Linear Decision Borders}

Using Eqs (\ref{Decision Border One}) and (\ref{Decision Border Two}), the
distance between the linear decision borders $D_{+1}\left(  \mathbf{x}\right)
$ and $D_{-1}\left(  \mathbf{x}\right)  $:%
\begin{align}
D_{\left(  D_{+1}\left(  \mathbf{x}\right)  -D_{-1}\left(  \mathbf{x}\right)
\right)  }  &  =\left(  -\frac{\tau_{0}}{\left\Vert \mathbf{\tau}\right\Vert
}+\frac{1}{\left\Vert \mathbf{\tau}\right\Vert }\right)
\label{Distance Between Decision Borders}\\
&  -\left(  -\frac{\tau_{0}}{\left\Vert \mathbf{\tau}\right\Vert }-\frac
{1}{\left\Vert \mathbf{\tau}\right\Vert }\right)  \text{,}\nonumber\\
&  =\frac{2}{\left\Vert \mathbf{\tau}\right\Vert }\text{,}\nonumber
\end{align}
is inversely proportional to the length of $\mathbf{\tau}$. It is concluded
that the distance between the linear decision borders is regulated by the term
$\frac{2}{\left\Vert \mathbf{\tau}\right\Vert }$, which is proportional to the
inverted length of a strong dual normal eigenlocus $\mathbf{\tau}$.

Algebraic expressions are now obtained for the distances between the linear
decision boundary and the linear decision borders.

\subsubsection*{Distances Between the Linear Decision Boundary and its
Borders}

Using Eqs (\ref{Decision Boundary}) and (\ref{Decision Border One}), the
distance between the linear decision border $D_{+1}\left(  \mathbf{x}\right)
$ and the linear decision boundary $D_{0}\left(  \mathbf{x}\right)  $ is
$\frac{1}{\left\Vert \mathbf{\tau}\right\Vert }$:%
\begin{align}
D_{\left(  D_{+1}\left(  \mathbf{x}\right)  -D_{0}\left(  \mathbf{x}\right)
\right)  }  &  =\left(  -\frac{\tau_{0}}{\left\Vert \mathbf{\tau}\right\Vert
}+\frac{1}{\left\Vert \mathbf{\tau}\right\Vert }\right)
\label{Symmetrical Distance Between Border One and Boundary}\\
&  -\left(  -\frac{\tau_{0}}{\left\Vert \mathbf{\tau}\right\Vert }\right)
\text{,}\nonumber\\
&  =\frac{1}{\left\Vert \mathbf{\tau}\right\Vert }\text{.}\nonumber
\end{align}
Using Eqs (\ref{Decision Boundary}) and (\ref{Decision Border Two}), the
distance between the linear decision boundary $D_{0}\left(  \mathbf{x}\right)
$ and the linear decision border $D_{-1}\left(  \mathbf{x}\right)  $ is also
$\frac{1}{\left\Vert \mathbf{\tau}\right\Vert }$:%
\begin{align}
D_{\left(  D_{0}\left(  \mathbf{x}\right)  -D_{-1}\left(  \mathbf{x}\right)
\right)  }  &  =\left(  -\frac{\tau_{0}}{\left\Vert \mathbf{\tau}\right\Vert
}\right) \label{Symmetrical Distance Between Border Two and Boundary}\\
&  -\left(  -\frac{\tau_{0}}{\left\Vert \mathbf{\tau}\right\Vert }-\frac
{1}{\left\Vert \mathbf{\tau}\right\Vert }\right)  \text{,}\nonumber\\
&  =\frac{1}{\left\Vert \mathbf{\tau}\right\Vert }\text{.}\nonumber
\end{align}
The equivalent distance of $\frac{1}{\left\Vert \mathbf{\tau}\right\Vert }$
between each linear decision border and the linear decision boundary reveals
that the algebraic and geometric source of the bilateral symmetry of the
linear decision borders is the constrained strong dual normal eigenlocus
$\mathbf{\tau}$. It is concluded that the equivalent and constant widths of
the bipartite, congruent geometric regions delineated by the linear decision
boundary of Eq. (\ref{Decision Boundary}) and the linear decision borders of
Eqs (\ref{Decision Border One}) and (\ref{Decision Border Two}) are regulated
by the\ inverted length\textit{\ }$\frac{1}{\left\Vert \mathbf{\tau
}\right\Vert }$ of $\mathbf{\tau}$.

\subsection{Axis of Symmetry for Bilateral Linear Partitions}

Equations (\ref{Distance Between Decision Borders}),
(\ref{Symmetrical Distance Between Border One and Boundary}), and
(\ref{Symmetrical Distance Between Border Two and Boundary}) show that a
strong dual normal eigenlocus $\mathbf{\tau}$ determines an axis of symmetry
that delineates congruent geometric regions between a linear decision boundary
and the bilaterally symmetrical decision borders which bound it. Section $9$
will demonstrate that the linear decision borders of Eqs
(\ref{Decision Border One}) and (\ref{Decision Border Two}) span $\left(
1\right)  $ geometric regions of data distribution overlap for overlapping
data distributions, and $\left(  2\right)  $ geometric regions of large
covariance between non-overlapping data distributions. Given this assumption,
it is remarkable that a strong dual normal eigenlocus $\mathbf{\tau}$
describes regions of data distribution overlap that exhibit symmetrical widths
of $\frac{1}{\left\Vert \mathbf{\tau}\right\Vert }$. The sections that follow
will determine the manner in which this feat is accomplished.

The next section of the paper will begin to identify geometric and statistical
properties exhibited by the\ primal normal eigenlocus represented within the
Wolfe dual eigenspace. The statistical representation of the primal normal
eigenlocus within the Wolfe dual eigenspace will be shown to specify a highly
interconnected set of constrained primal and Wolfe dual normal eigenaxis
components, which are organized in a symmetric manner that encodes essential
geometric underpinnings and statistical machinery for a statistical decision
system. Section $9$ will demonstrate that the statistical representation of
the primal normal eigenlocus within the Wolfe dual eigenspace $\left(
1\right)  $ determines a regularized, data-driven geometric architecture that
encodes a robust likelihood ratio, and $\left(  2\right)  $ delineates an
elegant curve and coordinate system that symmetrically partitions any given
feature space.

\section{Statistical Representation of $\mathbf{\tau}$ Within the Wolfe Dual
Eigenspace}

This section of the paper will begin the process of describing the primal
normal eigenlocus within the Wolfe dual eigenspace. Accordingly, the
Lagrangian $L_{\Psi\left(  \mathbf{\tau}\right)  }$ is minimized with respect
to the primal variables $\mathbf{\tau}$\textbf{, }$\tau_{0}$, and $\xi_{i}$,
and is maximized with respect to the dual variables $\psi_{i}$
\citet{Cristianini2000}%
,
\citet{Scholkopf2002}%
. The extrema obtained by representing the primal normal eigenlocus within the
Wolfe dual eigenspace are summarized below.

\subsection{The Constrained Primal Normal Eigenlocus}

The KKT constraint of Eq. (\ref{KKTE5}) restricts the length $\psi_{i}$ of any
Wolfe dual normal eigenaxis component $\psi_{i}\overrightarrow{\mathbf{e}}%
_{i}$ on $\mathbf{\psi}$ to either satisfy or exceed zero: $\psi_{i}\geq0$.
Any Wolfe dual normal eigenaxis component $\psi_{i}\overrightarrow{\mathbf{e}%
}_{i}$ which has the length $\psi_{i}=0$ is \emph{not on} the Wolfe dual
normal eigenlocus $\mathbf{\psi}$. It follows that the constrained primal
normal eigenaxis component $\psi_{i}\mathbf{x}_{i}$ which has the length
$\left\Vert \psi_{i}\mathbf{x}_{i}\right\Vert =0$ is \emph{not on} the
constrained primal normal eigenlocus $\mathbf{\tau}$. The KKT\ constraints of
Eqs (\ref{KKTE1}) and (\ref{KKTE5}) jointly determine the primal normal
eigenlocus $\mathbf{\tau}$ within the Wolfe dual eigenspace, so that an
estimate for $\mathbf{\tau}$ satisfies the following strong dual normal
eigenlocus equation:%
\begin{equation}
\mathbf{\tau}=\sum\nolimits_{i=1}^{N}y_{i}\psi_{i}\mathbf{x}_{i}\text{,}
\label{Normal Eigenlocus Estimate}%
\end{equation}
where the $y_{i}$ terms are training set labels (if $\mathbf{x}_{i}$ is a
member of pattern class one, assign $y_{i}=1$; otherwise, assign $y_{i}=-1$)
and the magnitude $\psi_{i}$ of each Wolfe dual normal eigenaxis component
$\psi_{i}\overrightarrow{\mathbf{e}}_{i}$ is greater than or equal to zero:
$\psi_{i}\geq0$. Training points $\mathbf{x}_{i}$ which are correlated with
Wolfe dual normal eigenaxes $\psi_{i}\overrightarrow{\mathbf{e}}_{i}$ that
have non-zero magnitudes or lengths $\psi_{i}>0$ are termed \emph{extreme}
training vectors. Accordingly, extreme training vectors are essentially
unconstrained primal normal eigenaxis components. Extreme training points
are\ innermost data points of large covariance that are located between
overlapping or non-overlapping data distributions. Given these assumptions,
Eq. (\ref{Normal Eigenlocus Estimate}) determines a dual statistical
eigenlocus of normal eigenaxis components formed by eigen-scaled extreme
training points, all of which encode principal locations of large covariance.
The location properties of extreme training points are defined next.

\subsection*{Location Properties of Extreme Training Points}

Take a collection of training data drawn from any two probability
distributions. An extreme training point is defined to be a data point which
exhibits a high variability of geometric location, that is, possesses a large
covariance, such that it is located $(1)$ relatively far from its distribution
mean, $(2)$ relatively close to the mean of the other distribution, and $(2)$
relatively close to other extreme points. Accordingly, an extreme data point
is located somewhere between a pair of overlapping or non-overlapping data
distributions. Given the location properties exhibited by the geometric locus
of an extreme data point, it follows that a set of extreme vectors determine
principal directions of large covariance for a given collection of training
data. Likewise, the geometric loci of a set of extreme vectors span a
geometric region of large covariance. Therefore, a set of extreme training
points span a geometric region of large covariance that is located between two
distributions of training data. It follows that the geometric loci of any
given set of extreme vectors span a particular region of $\mathbf{%
\mathbb{R}
}^{d}$.

It will now be argued that extreme training vectors are unconstrained primal
normal eigenaxis components used to form $\mathbf{\tau}$. Section $18$ will
demonstrate that each constrained primal normal eigenaxis component describes
the probability of finding an extreme data point in a particular region of
$\mathbf{%
\mathbb{R}
}^{d}$. Thereby, Section $18$ will show that the integrated set of constrained
primal normal eigenaxis components on $\mathbf{\tau}_{1}$ and $\mathbf{\tau
}_{2}$, i.e., on $\mathbf{\tau=\tau}_{1}-\mathbf{\tau}_{2}$, describes the
probabilities of finding each of the extreme data points in particular regions
of $\mathbf{%
\mathbb{R}
}^{d}$, where all of the extreme data points are located in regions of large
covariance between either overlapping or non-overlapping data distributions.

The location properties of extreme data points for overlapping and
non-overlapping data distributions are defined next.

\subsubsection{Extreme Data Points of Overlapping Data Distributions}

For overlapping data distributions, the geometric loci of the extreme data
points from each pattern class are distributed within bipartite, joint
geometric regions of large covariance, both of which span the region of data
distribution overlap. Figure $13$ depicts bipartite, joint geometric regions
of large variance that are located between two overlapping data distributions.

\begin{center}%
\begin{center}
\includegraphics[
natheight=7.499600in,
natwidth=9.999800in,
height=3.2897in,
width=4.3777in
]%
{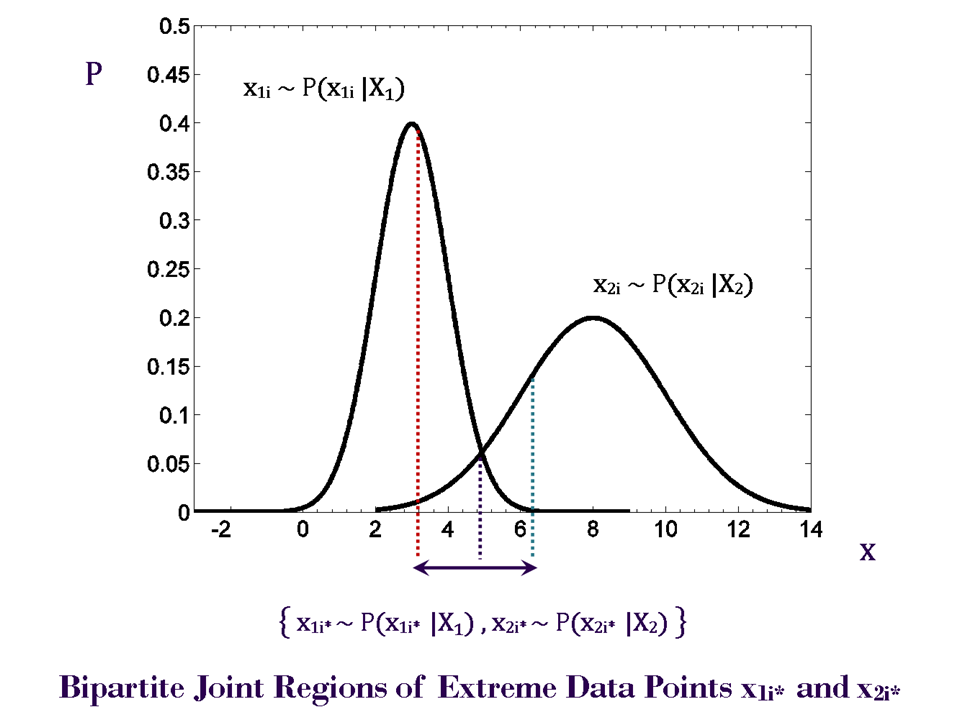}%
\end{center}

\end{center}

\begin{flushleft}
FIGURE $13$. Illustration of extreme data points, denoted by $\mathbf{x}%
_{1_{i\ast}}$ and $\mathbf{x}_{2_{i\ast}}$, which are located in bipartite,
joint geometric regions of large variance that are positioned between two
overlapping data distributions.
\end{flushleft}

\subsubsection{Extreme Data Points of Non-overlapping Data Distributions}

For non-overlapping data distributions, the geometric loci of the extreme data
points are distributed within bipartite, disjoint geometric regions of large
covariance, i.e., separate tail regions, that are located between the data
distributions. Because tail regions of distributions determine intervals of
low probability, it follows that relatively few extreme data points are
located within tail regions. Therefore, relatively few extreme data points are
located between non-overlapping data distributions. Figure $14$ illustrates
how a small number of extreme data points are located within the tail regions
of non-overlapping Gaussian data distributions.

\begin{center}%
\begin{center}
\includegraphics[
natheight=6.105600in,
natwidth=12.973100in,
height=2.3895in,
width=5.047in
]%
{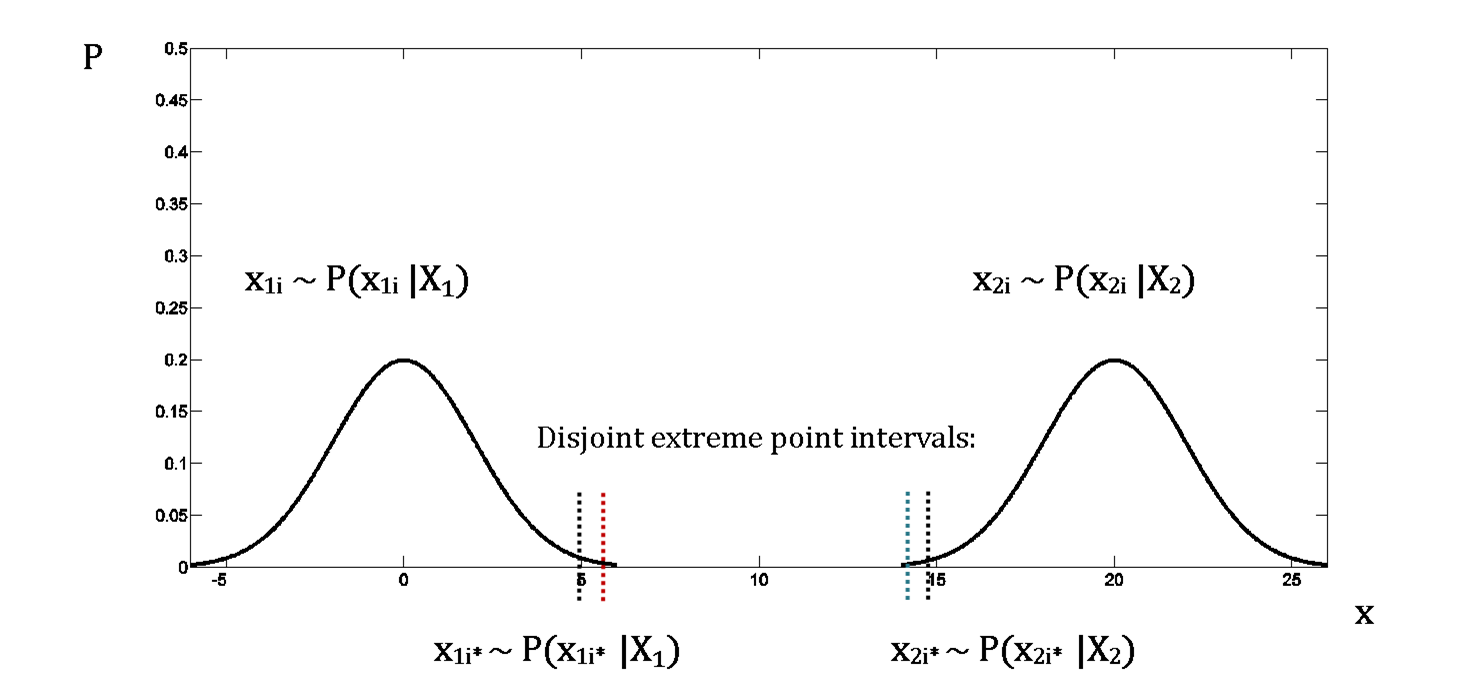}%
\end{center}

\end{center}

\begin{flushleft}
FIGURE $14$. Illustration of how relatively few extreme data points, denoted
by $\mathbf{x}_{1_{i\ast}}$ and $\mathbf{x}_{2_{i\ast}}$, are located in the
tail regions of non-overlapping Gaussian data distributions.
\end{flushleft}

Sections $14$ and $15$ will demonstrate that each Wolfe dual normal eigenaxis
component encodes an eigen-balanced, unidirectional pointwise covariance
estimate for an extreme data point, which specifies an eigen-scale for that
extreme training vector. The next section will consider how a constrained
primal normal eigenlocus $\mathbf{\tau}$ is formed by a pair of strong dual,
i.e., constrained primal, normal eigenlocus components.

\subsection{The Pair of Strong Dual Normal Eigenlocus Components}

All of the constrained primal normal eigenaxis components on a strong dual
normal eigenlocus $\mathbf{\tau}$ are labeled, eigen-scaled extreme training
points in $%
\mathbb{R}
^{d}$. Denote the eigen-scaled extreme training vectors that belong to pattern
classes one and two by $\psi_{1_{i\ast}}\mathbf{x}_{1_{i_{\ast}}}$ and
$\psi_{2_{i\ast}}\mathbf{x}_{2_{i\ast}}$, with eigen-scales $\psi_{1_{i\ast}}$
and $\psi_{2_{i\ast}}$, extreme training vectors $\mathbf{x}_{1_{i\ast}}$ and
$\mathbf{x}_{2_{i\ast}}$, and training labels $y_{i}=1$ and $y_{i}=-1$
respectively. Let there be $l_{1}$ eigen-scaled extreme training points
$\left\{  \psi_{1_{i\ast}}\mathbf{x}_{1_{i\ast}}\right\}  _{i=1}^{l_{1}}$ and
$l_{2}$ eigen-scaled extreme training points $\left\{  \psi_{2_{i\ast}%
}\mathbf{x}_{2_{i\ast}}\right\}  _{i=1}^{l_{2}}$.

Given Eq. (\ref{Normal Eigenlocus Estimate}) and the assumptions outlined
above, it follows that an estimate for a strong dual normal eigenlocus
$\mathbf{\tau}$ is based on the vector difference between a pair of
constrained primal normal eigenlocus components:%
\begin{align}
\mathbf{\tau}  &  =\sum\nolimits_{i=1}^{l_{1}}\psi_{1_{i\ast}}\mathbf{x}%
_{1_{i\ast}}-\sum\nolimits_{i=1}^{l_{2}}\psi_{2_{i\ast}}\mathbf{x}_{2_{i\ast}%
}\text{,}\label{Pair of Normal Eigenlocus Components}\\
&  =\mathbf{\tau}_{1}-\mathbf{\tau}_{2}\text{,}\nonumber
\end{align}
where the constrained primal normal eigenlocus components $\sum\nolimits_{i=1}%
^{l_{1}}\psi_{1_{i\ast}}\mathbf{x}_{1_{i\ast}}$ and $\sum\nolimits_{i=1}%
^{l_{2}}\psi_{2_{i\ast}}\mathbf{x}_{2_{i\ast}}$ are denoted by $\mathbf{\tau
}_{1}$ and $\mathbf{\tau}_{2}$. The eigen-scaled extreme training points
$\left\{  \psi_{1_{1\ast}}\mathbf{x}_{1_{i\ast}}\right\}  _{i=1}^{l_{1}}$ and
$\left\{  \psi_{2_{1\ast}}\mathbf{x}_{2_{i\ast}}\right\}  _{i=1}^{l_{2}}$ on
$\mathbf{\tau}_{1}$ and $\mathbf{\tau}_{2}$ determine the eigenloci of
$\mathbf{\tau}_{1}$ and $\mathbf{\tau}_{2}$, and thereby determine the
eigenlocus of $\mathbf{\tau}=\mathbf{\tau}_{1}-\mathbf{\tau}_{2}$. Figure $15$
depicts how the geometric configurations of the $\mathbf{\tau}_{1}$ and
$\mathbf{\tau}_{2}$ strong dual normal eigenlocus components of $\mathbf{\tau
}$ effectively determine the geometric configuration of $\mathbf{\tau}$.

\begin{center}%
\begin{center}
\includegraphics[
natheight=7.499600in,
natwidth=9.999800in,
height=3.2897in,
width=4.3777in
]%
{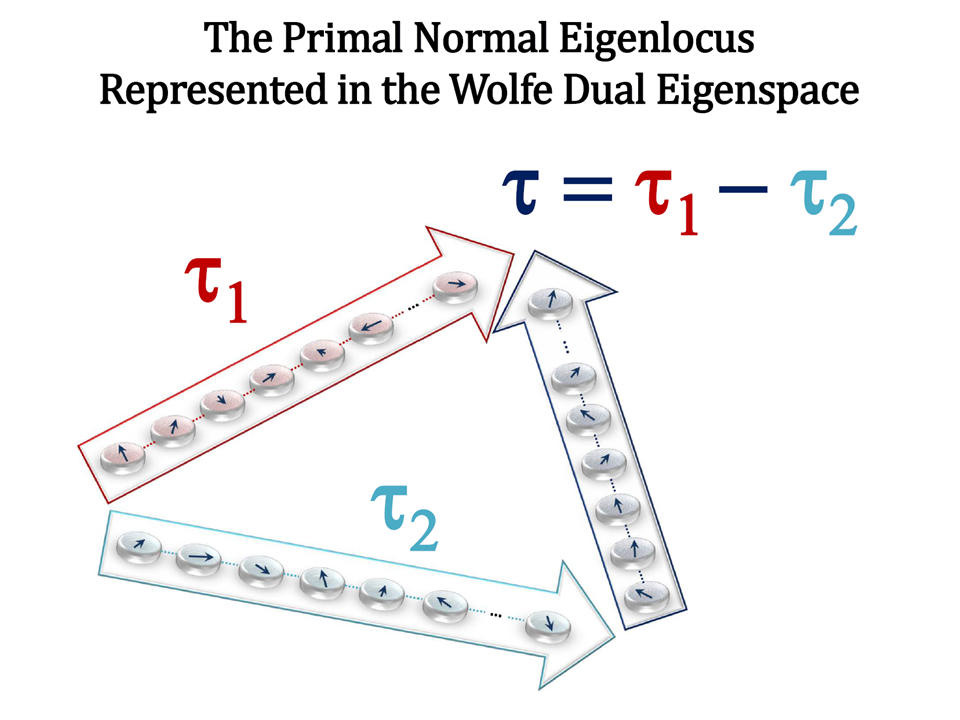}%
\end{center}

\end{center}

\begin{flushleft}
Figure $15$: Illustration of how the primal normal eigenlocus $\mathbf{\tau} $
represented in the Wolfe dual eigenspace is formed by the vector difference
$\mathbf{\tau}_{1}-\mathbf{\tau}_{2}$ between a pair of constrained primal
normal eigenlocus components $\mathbf{\tau}_{1}$ and $\mathbf{\tau}_{2}$. The
eigen-scaled extreme points on $\mathbf{\tau}_{1}$, $\mathbf{\tau}_{2}$, and
$\mathbf{\tau}$ are depicted by variable length arrows pointing in various
directions, which illustrate eigen-scaled extreme training vectors that
possess unchanged directions and eigen-balanced lengths.
\end{flushleft}

It will now be demonstrated how the eigenloci of the constrained primal normal
eigenlocus components $\mathbf{\tau}_{1}$ and $\mathbf{\tau}_{2}$ regulate the
geometric width, i.e., the breadth of the geometric region, between the linear
decision borders $D_{1}\left(  \mathbf{x}\right)  $ and $D_{-1}\left(
\mathbf{x}\right)  $.

\section{Width Regulation of Linear Decision Regions}

Substitution of the expression for $\mathbf{\tau}$ in Eq.
(\ref{Pair of Normal Eigenlocus Components}) into Eq.
(\ref{Distance Between Decision Borders}) provides a new expression for the
width of the geometric region between the linear decision borders
$D_{1}\left(  \mathbf{x}\right)  $ and $D_{-1}\left(  \mathbf{x}\right)  $:%
\begin{equation}
D_{\left(  D_{1}\left(  \mathbf{x}\right)  -D_{-1}\left(  \mathbf{x}\right)
\right)  }=\frac{2}{\left\Vert \mathbf{\tau}_{1}-\mathbf{\tau}_{2}\right\Vert
}\text{,} \label{Width of Linear Decision Region}%
\end{equation}
where the constrained width of the geometric region between the linear
decision borders is inversely proportional to the magnitude of the vector
difference of $\mathbf{\tau}_{1}$ and $\mathbf{\tau}_{2}$. Equation
(\ref{Width of Linear Decision Region}) shows that the span of the geometric
region between the linear decision borders is regulated by the magnitudes and
the directions of the constrained primal normal eigenlocus components
$\mathbf{\tau}_{1}$ and $\mathbf{\tau}_{2}$.

The eigenloci of the constrained primal normal eigenlocus components
$\mathbf{\tau}_{1}$ and $\mathbf{\tau}_{2}$ also regulate the span of the
congruent geometric regions between the linear decision boundary $D_{0}\left(
\mathbf{x}\right)  $ and the linear decision borders $D_{+1}\left(
\mathbf{x}\right)  $ and $D_{-1}\left(  \mathbf{x}\right)  $. Substitution of
the expression for $\mathbf{\tau}$ in Eq.
(\ref{Pair of Normal Eigenlocus Components}) into Eq.
(\ref{Symmetrical Distance Between Border One and Boundary}) provides a new
expression for the width of the geometric region between the linear decision
border $D_{+1}\left(  \mathbf{x}\right)  $ and the linear decision boundary
$D_{0}\left(  \mathbf{x}\right)  $:
\begin{align}
D_{\left(  D_{+1}\left(  \mathbf{x}\right)  -D_{0}\left(  \mathbf{x}\right)
\right)  }  &  =\left(  -\frac{\tau_{0}}{\left\Vert \mathbf{\tau}%
_{1}-\mathbf{\tau}_{2}\right\Vert }+\frac{1}{\left\Vert \mathbf{\tau}%
_{1}-\mathbf{\tau}_{2}\right\Vert }\right) \label{Large Covariance Region One}%
\\
&  -\left(  -\frac{\tau_{0}}{\left\Vert \mathbf{\tau}_{1}-\mathbf{\tau}%
_{2}\right\Vert }\right)  \text{,}\nonumber\\
&  =\frac{1}{\left\Vert \mathbf{\tau}_{1}-\mathbf{\tau}_{2}\right\Vert
}\text{,}\nonumber
\end{align}
where the width of the geometric region between $D_{0}\left(  \mathbf{x}%
\right)  $ and $D_{+1}\left(  \mathbf{x}\right)  $ satisfies $\frac
{1}{\left\Vert \mathbf{\tau}_{1}-\mathbf{\tau}_{2}\right\Vert }$.

Likewise, the span of the geometric region between the linear decision
boundary $D_{0}\left(  \mathbf{x}\right)  $ and the linear decision border
$D_{-1}\left(  \mathbf{x}\right)  $:%
\begin{align}
D_{\left(  D_{0}\left(  \mathbf{x}\right)  -D_{-1}\left(  \mathbf{x}\right)
\right)  }  &  =\left(  -\frac{\tau_{0}}{\left\Vert \mathbf{\tau}%
_{1}-\mathbf{\tau}_{2}\right\Vert }\right) \label{Large Covariance Region Two}%
\\
&  -\left(  -\frac{\tau_{0}}{\left\Vert \mathbf{\tau}_{1}-\mathbf{\tau}%
_{2}\right\Vert }-\frac{1}{\left\Vert \mathbf{\tau}_{1}-\mathbf{\tau}%
_{2}\right\Vert }\right)  \text{,}\nonumber\\
&  =\frac{1}{\left\Vert \mathbf{\tau}_{1}-\mathbf{\tau}_{2}\right\Vert
}\text{,}\nonumber
\end{align}
also satisfies $\frac{1}{\left\Vert \mathbf{\tau}_{1}-\mathbf{\tau}%
_{2}\right\Vert }$.

It is concluded that the width of the bipartite, congruent geometric regions
between the linear decision boundary and the linear decision borders is
inversely proportional to the magnitude of the vector difference of
$\mathbf{\tau}_{1}$ and $\mathbf{\tau}_{2}$:%
\[
\frac{1}{\left\Vert \mathbf{\tau}_{1}-\mathbf{\tau}_{2}\right\Vert }\text{,}%
\]
which indicates that the balanced geometric widths of the symmetrical decision
regions of the constrained Lagrangian of Eq.
(\ref{Lagrangian Normal Eigenlocus}) are regulated by the magnitudes and the
directions of the constrained primal normal eigenlocus components
$\mathbf{\tau}_{1}$ and $\mathbf{\tau}_{2}$.

\subsection{Bipartite Symmetric Partitions of Large Covariance Regions}

It will now be argued that the bipartite, symmetrical decision regions
delineated by the linear decision boundary of Eq. (\ref{Decision Boundary})
and the linear decision borders of Eqs (\ref{Decision Border One}) and
(\ref{Decision Border Two}), describe symmetric regions of large covariance
for both overlapping and non-overlapping data distributions. Recall that, by
definition, extreme training points exhibit a high variability of geometric
location and therefore posses a large covariance, such that the geometric
locations of a set of extreme training points span a geometric region of large
covariance that is located between two distributions of training data.
Therefore, by definition, the width of any large covariance geometric region
depends on the geometric loci of the extreme vectors of the distributions.

\subsubsection*{Assumptions}

At this stage of the analysis, it is necessary to develop more of the
geometric underpinnings and statistical machinery that is produced by the
constrained Lagrangian of Eq. (\ref{Lagrangian Normal Eigenlocus}). For this
reason, several significant results will be assumed that will be substantiated
later on. Section $18$ will show that the geometric configuration of the
linear decision boundary, and the widths of the bipartite, congruent geometric
regions located between the linear decision boundary of Eq.
(\ref{Decision Boundary}) and the linear decision borders of Eqs
(\ref{Decision Border One}) and (\ref{Decision Border Two}), are regulated by
the probability of finding extreme data points in particular regions of
$\mathbf{%
\mathbb{R}
}^{d}$.

For now, it is assumed that the integrated set of constrained primal normal
eigenaxis components on $\mathbf{\tau=\tau}_{1}-\mathbf{\tau}_{2}$ describes
the probabilities of finding the extreme data points in particular regions of
$\mathbf{%
\mathbb{R}
}^{d}$, where all of the extreme data points are located in regions of large
covariance between either overlapping or non-overlapping data distributions.
Thereby, it is assumed that the constrained primal normal eigenaxis components
on $\mathbf{\tau}_{1}$ and $\mathbf{\tau}_{2}$ describe disjoint tail regions
between non-overlapping data distributions, and bipartite, joint geometric
regions of large covariance\ between overlapping data distributions. The next
section will examine strong dual normal eigenlocus transforms for
non-overlapping data distributions.

\subsection{Strong Dual Normal Eigenlocus Transforms for Non-overlapping Data
Distributions}

It will now be argued that the linear decision boundary of Eq.
(\ref{Decision Boundary}) and the linear decision borders of Eqs
(\ref{Decision Border One}) and (\ref{Decision Border Two}) delineate
symmetric, non-overlapping regions of large covariance for any two
non-overlapping data distributions.

Take a collection of training data generated by any two non-overlapping
probability distributions, where all of the extreme data points are located
within the bipartite, disjoint tail regions of the distributions. Given these
assumptions and Eq. (\ref{Pair of Normal Eigenlocus Components}), it follow
that the strong dual normal eigenlocus components $\mathbf{\tau}_{1}$ and
$\mathbf{\tau}_{2}$ on $\mathbf{\tau=\tau}_{1}-\mathbf{\tau}_{2}$ are formed
by relatively few eigen-scaled extreme data points, i.e., $\mathbf{\tau}%
_{1}=\sum\nolimits_{i=1}^{l_{1}}\psi_{1_{i\ast}}\mathbf{x}_{1_{i\ast}}$ and
$\mathbf{\tau}_{2}=\sum\nolimits_{i=1}^{l_{2}}\psi_{2_{i\ast}}\mathbf{x}%
_{2_{i\ast}}$, all of which describe the probabilities of finding the extreme
data points in the tail regions of two data distributions in $\mathbf{%
\mathbb{R}
}^{d}$. Given Eqs (\ref{Pair of Normal Eigenlocus Components}) and
(\ref{Width of Linear Decision Region}), it follows that the width $\frac
{2}{\left\Vert \mathbf{\tau}_{1}-\mathbf{\tau}_{2}\right\Vert }$ of the
geometric region located between the linear decision borders of Eqs
(\ref{Decision Border One}) and (\ref{Decision Border Two}) is regulated by
the eigen-transformed locations of the extreme training points on
$\mathbf{\tau}_{1}-\mathbf{\tau}_{2}$, where each eigen-transformed location
of an extreme data point describes the probability of finding the extreme data
point in the tail region of a data distribution in $\mathbf{%
\mathbb{R}
}^{d}$. Given Eqs (\ref{Pair of Normal Eigenlocus Components}),
(\ref{Large Covariance Region One}), and (\ref{Large Covariance Region Two}),
it follows that the equivalent widths $\frac{1}{\left\Vert \mathbf{\tau}%
_{1}-\mathbf{\tau}_{2}\right\Vert }$ of the congruent geometric regions, which
are located between the linear decision boundary of Eq.
(\ref{Decision Boundary}) and the linear decision borders of Eqs
(\ref{Decision Border One}) and (\ref{Decision Border Two}), are regulated by
the eigen-transformed locations of the extreme training points on
$\mathbf{\tau}_{1}-\mathbf{\tau}_{2}$.

Given the above assumptions and chain of arguments, it is concluded that the
bipartite, congruent geometric regions located between the linear decision
boundary of Eq. (\ref{Decision Boundary}) and the linear decision borders of
Eqs (\ref{Decision Border One}) and (\ref{Decision Border Two}) delineate
bipartite, congruent, non-overlapping geometric regions of large covariance
for non-overlapping data distributions. It is also concluded that the linear
decision borders of Eqs (\ref{Decision Border One}) and
(\ref{Decision Border Two}) delineate a geometric region of large covariance
that spans a geometric region between the tails of two data distributions.

\subsection{Beyond Classical Interpolation Methods}

It is well known that the components of the extreme training vectors from each
of the pattern classes satisfy their respective decision borders for
non-overlapping data distributions
\citet{Cortes1995}%
,
\citet{Cristianini2000}%
,
\citet{Hastie2001}%
,
\citet{Scholkopf2002}%
. However, the mathematical machinery behind classical interpolation or
regression methods provides no real insight into what is actually going on.

Given non-overlapping data distributions and two extreme training points
$\mathbf{x}_{1_{\ast}}$and $\mathbf{x}_{2_{\ast}}$, it can be shown that the
symmetrical eigen-scale $\psi_{1_{\ast}}=\psi_{2_{\ast}}$ for each extreme
training vector is the reciprocal of the inner product of the vector
difference $\mathbf{x}_{1_{\ast}}-\mathbf{x}_{2_{\ast}}$
\[
\psi_{1_{i\ast}}=\psi_{2_{i\ast}}=\frac{2}{\left\Vert \mathbf{x}_{1_{i\ast}%
}-\mathbf{x}_{2_{i\ast}}\right\Vert ^{2}}\text{,}%
\]
of $\mathbf{x}_{1_{\ast}}$and $\mathbf{x}_{2_{\ast}}$, which indicates that
the magnitudes of the Wolfe dual normal eigenaxis components involve
second-order distance statistics between the locations of the extreme points.

Using the above expression and Eq. (\ref{Width of Linear Decision Region}), it
follows that the width of the geometric region between the linear decision
borders $D_{1}\left(  \mathbf{x}\right)  $ and $D_{-1}\left(  \mathbf{x}%
\right)  $ is determined by the eigen-transformed locations of the extreme
training points $\mathbf{x}_{1_{\ast}}$and $\mathbf{x}_{2_{\ast}}$%
\begin{align*}
\frac{2}{\left\Vert \mathbf{\tau}_{1}-\mathbf{\tau}_{2}\right\Vert }  &
=2\left\Vert \frac{2\mathbf{x}_{1_{\ast}}}{\left\Vert \mathbf{x}_{1_{\ast}%
}-\mathbf{x}_{2_{\ast}}\right\Vert ^{2}}-\frac{2\mathbf{x}_{2_{\ast}}%
}{\left\Vert \mathbf{x}_{1_{\ast}}-\mathbf{x}_{2_{\ast}}\right\Vert ^{2}%
}\right\Vert ^{-1}\text{,}\\
&  =2\left\Vert \frac{2\left(  \mathbf{x}_{1_{\ast}}-\mathbf{x}_{2_{\ast}%
}\right)  }{\left\Vert \mathbf{x}_{1_{\ast}}\right\Vert ^{2}+\left\Vert
\mathbf{x}_{2_{\ast}}\right\Vert ^{2}-2\left\Vert \mathbf{x}_{1_{\ast}%
}\right\Vert \left\Vert \mathbf{x}_{2_{\ast}}\right\Vert \cos\theta
_{\mathbf{x}_{1_{\ast}}\mathbf{x}_{2_{\ast}}}}\right\Vert ^{-1}\text{,}%
\end{align*}
which reduces to%
\begin{align*}
\frac{2}{\left\Vert \mathbf{\tau}_{1}-\mathbf{\tau}_{2}\right\Vert }  &
=2\left\Vert \frac{2}{\left(  \mathbf{x}_{1_{\ast}}-\mathbf{x}_{2_{\ast}%
}\right)  }\right\Vert ^{-1}\text{,}\\
&  =\left\Vert \mathbf{x}_{1_{\ast}}-\mathbf{x}_{2_{\ast}}\right\Vert \text{.}%
\end{align*}
Returning to Eq. (\ref{Inner Product Statistic}) and Fig. $9$ in Section $5$,
it follows that the span of the geometric region between the linear decision
borders $D_{1}\left(  \mathbf{x}\right)  $ and $D_{-1}\left(  \mathbf{x}%
\right)  $ is determined by the distance between the geometric loci of the
extreme training points $\mathbf{x}_{1_{\ast}}$and $\mathbf{x}_{2_{\ast}}$.
Thereby, the geometric loci of the extreme training vectors from each of the
pattern classes satisfy their respective decision borders.

Using the expression for the symmetrical eigen-scale $\psi_{1_{\ast}}%
=\psi_{2_{\ast}}$ and Eqs (\ref{Large Covariance Region One}) and
(\ref{Large Covariance Region Two}), it follows that the symmetrical widths of
the non-overlapping regions of large covariance located between the linear
decision border $D_{1}\left(  \mathbf{x}\right)  $ or $D_{-1}\left(
\mathbf{x}\right)  $ and the linear decision boundary $D_{0}\left(
\mathbf{x}\right)  $ are determined by equally proportioned eigen-transformed
locations of $\mathbf{x}_{1_{\ast}}$and $\mathbf{x}_{2_{\ast}}$%
\begin{align*}
\frac{1}{\left\Vert \mathbf{\tau}_{1}-\mathbf{\tau}_{2}\right\Vert }  &
=\left\Vert \frac{2\left(  \mathbf{x}_{1_{\ast}}-\mathbf{x}_{2_{\ast}}\right)
}{\left\Vert \mathbf{x}_{1_{\ast}}\right\Vert ^{2}+\left\Vert \mathbf{x}%
_{2_{\ast}}\right\Vert ^{2}-2\left\Vert \mathbf{x}_{1_{\ast}}\right\Vert
\left\Vert \mathbf{x}_{2_{\ast}}\right\Vert \cos\theta_{\mathbf{x}_{1_{\ast}%
}\mathbf{x}_{2_{\ast}}}}\right\Vert ^{-1}\\
&  =\frac{1}{2}\left\Vert \mathbf{x}_{1_{\ast}}-\mathbf{x}_{2_{\ast}%
}\right\Vert \text{,}%
\end{align*}
which indicates that the symmetrical widths of the decision regions are
determined by the half the distance between the geometric loci of the extreme
training points $\mathbf{x}_{1_{\ast}}$and $\mathbf{x}_{2_{\ast}}$.

For non-overlapping data distributions, simulation studies show that the
extreme training points from each of the pattern classes lie on their
respective decision borders. Figure $16$ illustrates how the constrained
discriminant function of Eq. (\ref{Discriminant Function}) delineates
bipartite, congruent, non-overlapping geometric regions of large covariance
for two non-overlapping Gaussian data distributions. Figure $16$ also shows
that the symmetrical widths of the non-overlapping regions of large covariance
are determined by the locations of the extreme vectors of the data distributions.

\begin{center}%
\begin{center}
\includegraphics[
natheight=7.656200in,
natwidth=14.781400in,
height=2.2857in,
width=4.3881in
]%
{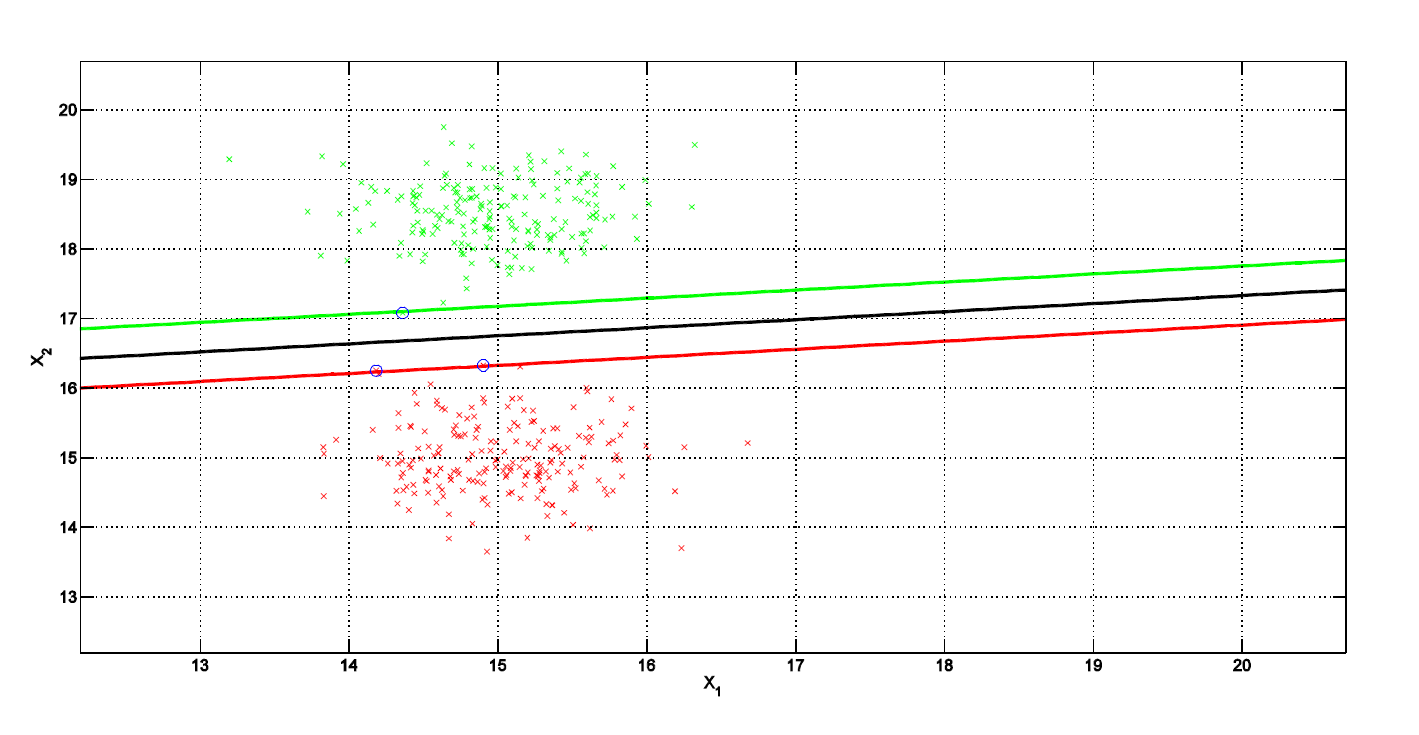}%
\end{center}

\end{center}

\begin{flushleft}
Figure $16$: Illustration of the geometric configurations of a linear decision
boundary and bilaterally symmetrical linear decision borders, for two
non-overlapping data distributions, which are determined by eigen-transformed
locations of three extreme training points, all of which lie on their
respective decision borders. Each extreme training point is enclosed in a blue circle.
\end{flushleft}

In general, for any given pair of non-overlapping data distributions, the
symmetrical widths of the bipartite, congruent, non-overlapping geometric
regions of large covariance delineated by Eqs (\ref{Decision Boundary}),
(\ref{Decision Border One}), and (\ref{Decision Border Two}), are a function
of eigen-balanced distances between the geometric loci of the extreme points
of the data distributions. Figure $17$ depicts the symmetrical decision
regions of large covariance that are delineated by a constrained strong dual
normal eigenlocus discriminate function $\mathbf{\tau}^{T}\mathbf{x}+\tau_{0}$
for non-overlapping data distributions.

\begin{center}%
\begin{center}
\includegraphics[
natheight=7.499600in,
natwidth=9.999800in,
height=3.2897in,
width=4.3777in
]%
{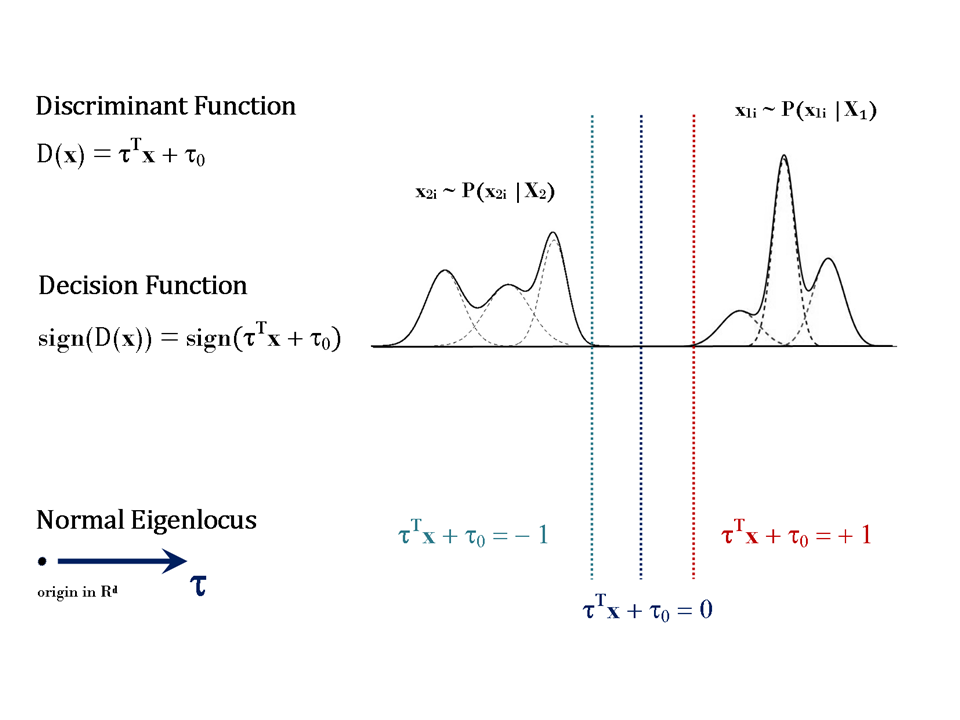}%
\end{center}

\end{center}

\begin{flushleft}
Figure $17$: Depiction of the bipartite, symmetrical decision regions of large
covariance that are delineated by a constrained strong dual normal eigenlocus
discriminant function for non-overlapping data distributions.
\end{flushleft}

The next section will examine strong dual normal eigenlocus transforms for
overlapping data distributions.

\subsection{Strong Dual Normal Eigenlocus Transforms for Overlapping Data
Distributions}

It will now be argued that the linear decision boundary of Eq.
(\ref{Decision Boundary}) and the linear decision borders of Eqs
(\ref{Decision Border One}) and (\ref{Decision Border Two}) delineate
symmetric regions of data distribution overlap for any two overlapping data distributions.

Take a collection of training data generated by any two overlapping
probability distributions, where all of the extreme data points are located
within the bipartite, joint (overlapping) geometric regions of large
covariance that span the region of data distribution overlap. For any given
collection of training data, both the number and the locations of the extreme
data points are determined by the probability density functions of the
training data. Therefore, the geometric shape or configuration of the data
distribution overlap is determined by the probability density functions of the
training data.

Now take a pair of overlapping data distributions. Let there be $l_{1}$
extreme training points $\left\{  \mathbf{x}_{1_{i\ast}}\right\}
_{i=1}^{l_{1}}$ and $l_{2}$ extreme training points $\left\{  \mathbf{x}%
_{2_{i\ast}}\right\}  _{i=1}^{l_{2}}$. Given the above assumptions and Eq.
(\ref{Pair of Normal Eigenlocus Components}), it follows that the strong dual
normal eigenlocus components $\mathbf{\tau}_{1}$ and $\mathbf{\tau}_{2}$ on
$\mathbf{\tau=\tau}_{1}-\mathbf{\tau}_{2}$ are formed by $l_{1}+l_{2}$
eigen-scaled extreme data points, i.e., $\mathbf{\tau}_{1}=\sum\nolimits_{i=1}%
^{l_{1}}\psi_{1_{i\ast}}\mathbf{x}_{1_{i\ast}}$ and $\mathbf{\tau}_{2}%
=\sum\nolimits_{i=1}^{l_{2}}\psi_{2_{i\ast}}\mathbf{x}_{2_{i\ast}}$, all of
which jointly describe the bipartite, joint geometric regions of large
covariance\ between the overlapping data distributions. It is assumed that
each eigen-scaled extreme data point $\psi_{1_{i\ast}}\mathbf{x}_{1_{i\ast}}$
or $\psi_{2_{i\ast}}\mathbf{x}_{2_{i\ast}}$ describes the probability of
finding that extreme data point $\mathbf{x}_{1_{i\ast}}$ or $\mathbf{x}%
_{2_{i\ast}}$ within a specific region between the overlapping data distributions.

Given Eqs (\ref{Pair of Normal Eigenlocus Components}) and
(\ref{Width of Linear Decision Region}), it follows that the width $\frac
{2}{\left\Vert \mathbf{\tau}_{1}-\mathbf{\tau}_{2}\right\Vert }$ of the
geometric region located between the linear decision borders of Eqs
(\ref{Decision Border One}) and (\ref{Decision Border Two}) is regulated by
the eigen-transformed locations of the extreme training points on
$\mathbf{\tau}_{1}-\mathbf{\tau}_{2}$:%
\[
\frac{2}{\left\Vert \sum\nolimits_{i=1}^{l_{1}}\psi_{1_{i\ast}}\mathbf{x}%
_{1_{i\ast}}-\sum\nolimits_{i=1}^{l_{2}}\psi_{2_{i\ast}}\mathbf{x}_{2_{i\ast}%
}\right\Vert }\text{,}%
\]
where each eigen-scaled extreme training point describes the probability of
finding the extreme point within a region of data distribution overlap.

It is concluded that the total width $\frac{2}{\left\Vert \mathbf{\tau}%
_{1}-\mathbf{\tau}_{2}\right\Vert }$ of the geometric region between the
linear decision borders delineated by Eq.
(\ref{Width of Linear Decision Region}) is regulated by the eigen-transformed
locations of the extreme vectors of the data distributions, where the
eigen-transformed location of an extreme training point describes the
probability of finding that extreme data point within a specific region
between the overlapping data distributions. Therefore, it is concluded that
the geometric region located between the linear decision borders of Eqs
(\ref{Decision Border One}) and (\ref{Decision Border Two}) spans the region
of data distribution overlap.

Given Eqs (\ref{Pair of Normal Eigenlocus Components}),
(\ref{Large Covariance Region One}), and (\ref{Large Covariance Region Two}),
it follows that the equivalent widths $\frac{1}{\left\Vert \mathbf{\tau}%
_{1}-\mathbf{\tau}_{2}\right\Vert }$ of the bipartite, congruent geometric
regions located between the linear decision boundary of Eq.
(\ref{Decision Boundary}) and the linear decision borders of Eqs
(\ref{Decision Border One}) and (\ref{Decision Border Two}), are regulated by
the eigen-transformed locations of all of the extreme training points on
$\mathbf{\tau}_{1}-\mathbf{\tau}_{2}$:%
\[
\frac{1}{\left\Vert \sum\nolimits_{i=1}^{l_{1}}\psi_{1_{i\ast}}\mathbf{x}%
_{1_{i\ast}}-\sum\nolimits_{i=1}^{l_{2}}\psi_{2_{i\ast}}\mathbf{x}_{2_{i\ast}%
}\right\Vert }\text{.}%
\]
This implies that the balanced widths $\frac{1}{\left\Vert \mathbf{\tau}%
_{1}-\mathbf{\tau}_{2}\right\Vert }$ of the congruent geometric regions of
distribution overlap delineated by Eqs (\ref{Large Covariance Region One}) and
(\ref{Large Covariance Region Two}), are determined by eigen-transformed
locations of the extreme vectors of the data distributions, where the
geometric loci of the extreme vectors determine the amount of data
distribution overlap, and the eigen-transformed locations of the extreme
vectors describe the probabilities of finding the extreme data points within
specific regions between the overlapping data distributions. It is concluded
that the linear decision boundary of Eq. (\ref{Decision Boundary}) and the
linear decision borders of Eqs (\ref{Decision Border One}) and
(\ref{Decision Border Two}) delineate bipartite, congruent, large covariance
geometric regions of data distribution overlap for any given pair of
overlapping data distributions.

It has been demonstrated by simulation studies that the constrained
discriminate function $\mathbf{\tau}^{T}\mathbf{x}+\tau_{0}$ of Eq.
(\ref{Discriminant Function}) does indeed delineate bipartite, congruent
geometric regions of data distribution overlap
\citet{Reeves2007}%
,
\citet{Reeves2009}%
. For example, consider Figs. $5$ and $7$ of Section $3$. In general, strong
dual normal eigenlocus decision systems delineate bipartite, symmetrical
decision regions for any given pair of overlapping data distributions. Figure
$18$ depicts the symmetrical decision regions delineated by a constrained
discriminant function $\mathbf{\tau}^{T}\mathbf{x}+\tau_{0}$ for overlapping
data distributions with different covariance structures.

\begin{center}%
\begin{center}
\includegraphics[
natheight=7.499600in,
natwidth=9.999800in,
height=3.2897in,
width=4.3777in
]%
{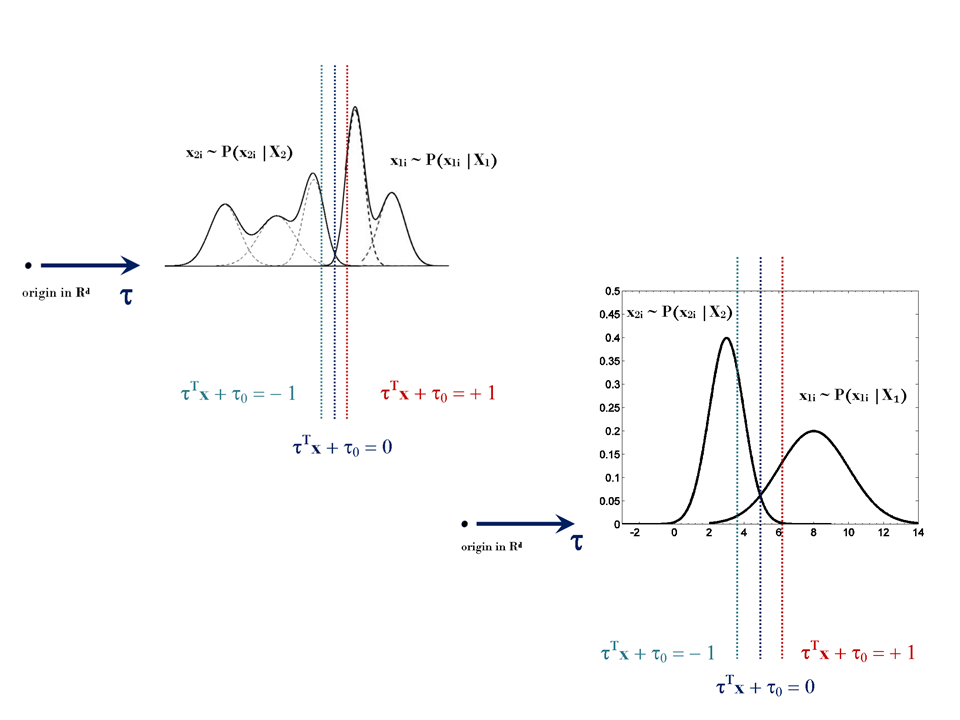}%
\end{center}

\end{center}

\begin{flushleft}
Figure $18$: Depiction of the bipartite, symmetrical decision regions
delineated by a constrained strong dual normal eigenlocus discriminant
function for overlapping data distributions with different covariance structures.
\end{flushleft}

Consider the following example of overlapping Gaussian data distributions with
different covariance structures. Figure $19$ illustrates the bipartite,
symmetrical decision regions delineated by a constrained discriminant function
$\mathbf{\tau}^{T}\mathbf{x}+\tau_{0}$ for two Gaussian data sets that have
the covariance matrices%
\[
\mathbf{\Sigma}_{1}=%
\begin{pmatrix}
1 & 0\\
0 & 1
\end{pmatrix}
\text{ and }\mathbf{\Sigma}_{2}=%
\begin{pmatrix}
0.25 & 0\\
0 & 5
\end{pmatrix}
\text{,}%
\]
and the mean vectors $\mathbf{\mu}_{1}=%
\begin{pmatrix}
1, & 2
\end{pmatrix}
^{T}$ and $\mathbf{\mu}_{2}=%
\begin{pmatrix}
0, & 2
\end{pmatrix}
^{T}$. The constrained discriminant function determines a centrally located
linear decision boundary that symmetrically partitions the feature space.

\begin{center}%
\begin{center}
\includegraphics[
natheight=7.728800in,
natwidth=15.041700in,
height=2.2866in,
width=4.4235in
]%
{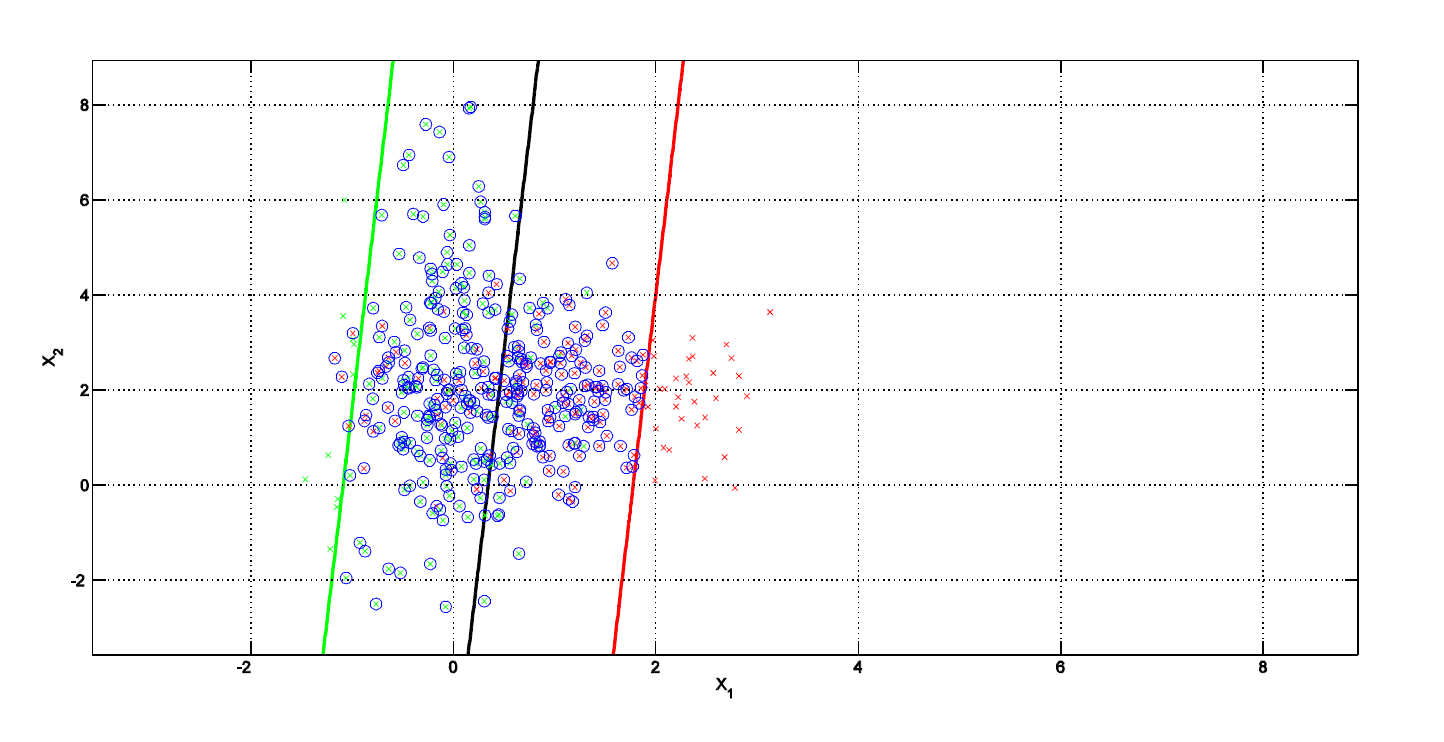}%
\end{center}

\end{center}

\begin{flushleft}
Figure $19$: Illustration of the bipartite, symmetrical decision regions
delineated by a constrained discriminant function $\mathbf{\tau}^{T}%
\mathbf{x}+\tau_{0}$ for overlapping data distributions with different
covariance structures. The centrally located linear decision boundary
symmetrically partitions the feature space. The linear decision borders
delineate a geometric region of large covariance that spans bipartite,
congruent geometric regions of data distribution overlap.
\end{flushleft}

\subsection{Regularized and Customized Geometric Architectures}

Take any given pair of data distributions whose expected values and covariance
structures do not vary over time. This paper will demonstrate how strong dual
normal eigenlocus transforms produce \emph{customized and regularized
geometric architectures that encode robust decision statistics for the binary
classification task}.

Furthermore, Figs $15$ and $20$ illustrate that the data-driven
directions\ and eigen-balanced magnitudes which can be realized by the strong
dual normal eigenlocus components $\mathbf{\tau}_{1}$ and $\mathbf{\tau}_{2}$
on $\mathbf{\tau=\tau}_{1}-\mathbf{\tau}_{2}$, determine an unlimited number
of customized, regularized geometric architectures that can be implemented by
the strong dual decision system of Eqs ( \ref{Discriminant Function}),
(\ref{Decision Boundary}), (\ref{Decision Border One}), and
(\ref{Decision Border Two}).

Figure $20$ depicts the regularized, data-driven geometric architecture of a
strong dual normal eigenlocus in the Euclidean plane.

\begin{center}%
\begin{center}
\includegraphics[
natheight=7.499600in,
natwidth=9.999800in,
height=3.2897in,
width=4.3777in
]%
{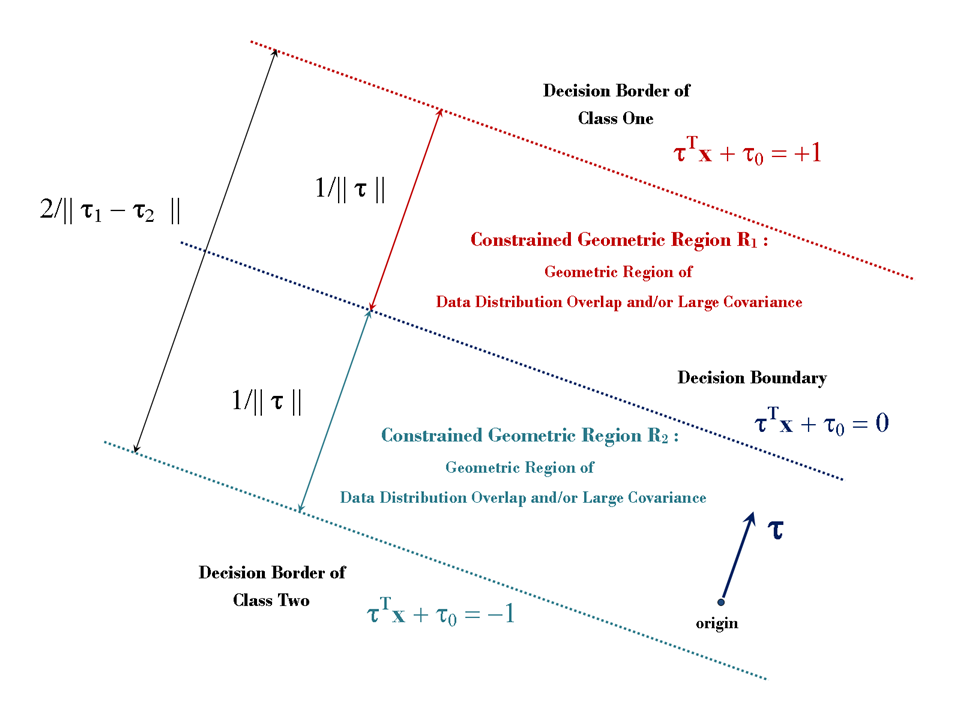}%
\end{center}

\end{center}

\begin{flushleft}
Figure $20$: Illustration of a statistical decision system of a strong dual
normal eigenlocus in the Euclidean plane. The algebraic system of strong dual
normal eigenlocus equations satisfied by $\mathbf{\tau}$, $\mathbf{\tau}_{1}$,
$\mathbf{\tau}_{2}$, and $\tau_{0}$ specifies the geometric configurations of
$\mathbf{\tau}_{1}$, $\mathbf{\tau}_{2}$, and $\mathbf{\tau}$, which jointly
specify the configurations of the constrained geometric regions of large
covariance denoted by $R_{1}$ and $R_{2}$.
\end{flushleft}

So far, this paper has argued that the KKT constraints on the Lagrangian
functional $L_{\Psi\left(  \mathbf{\tau}\right)  }$ jointly specify a set of
symmetrical eigen-scales for a set of extreme training points that are located
between two given distributions of training data. Later on, Sections $14$ and
$15$ will examine how the eigenlocus of each constrained primal normal
eigenaxis component $\psi_{1_{i\ast}}\mathbf{x}_{1_{i\ast}}$ or $\psi
_{2_{i\ast}}\mathbf{x}_{2_{i\ast}}$ on $\mathbf{\tau}_{1}$ or $\mathbf{\tau
}_{2}$ specifies a principal region of high variability, which contributes to
the symmetrical partitioning of a region of large covariance located between
two data distributions.

Now, consider again the regularized, data-driven geometric architecture
depicted in Fig. $20$. Given that $\left(  1\right)  $ the total allowed
eigenenergy of $\left\Vert \mathbf{\tau}\right\Vert _{\min_{c}}^{2}$ of
$\mathbf{\tau}$ satisfies the linear decision boundaries and decision borders
depicted in Figs $17$, $18$, and $20$, and that $\left(  2\right)  $ the
magnitudes and directions of $\mathbf{\tau}_{1}$ and $\mathbf{\tau}_{2}$
regulate the symmetrical configurations of the constrained geometric regions
of large covariance depicted in Figs $17$, $18$, and $20$, it is claimed that
the total allowed eigenenergies of $\mathbf{\tau}_{1}$ and $\mathbf{\tau}_{2}$
must be symmetrically balanced with each other. It will now be argued that the
critical minimum eigenenergies of $\mathbf{\tau}_{1}$ and $\mathbf{\tau}_{2}$
must be balanced in a symmetric manner.

\subsection{Balancing the Total Allowed Eigenenergies of $\mathbf{\tau}_{1}$
and $\mathbf{\tau}_{2}$}

Substitution of the expression for the pair of constrained primal normal
eigenlocus components $\mathbf{\tau}_{1}-\mathbf{\tau}_{2}$ on $\mathbf{\tau}$
in Eq. (\ref{Pair of Normal Eigenlocus Components}) into the critical minimum
eigenenergy constraint for $\mathbf{\tau}$ in Eq.
(\ref{Minimum Total Eigenenergy Primal Normal Eigenlocus}) produces an
expression%
\[
\left\Vert \mathbf{\tau}_{1}-\mathbf{\tau}_{2}\right\Vert _{\min_{c}}^{2}%
\cong\left(  \left\Vert \mathbf{\tau}_{1}\right\Vert _{_{\min_{c}}}%
^{2}+\left\Vert \mathbf{\tau}_{2}\right\Vert _{_{\min_{c}}}^{2}-\mathbf{\tau
}_{2}^{T}\mathbf{\tau}_{1}-\mathbf{\tau}_{1}^{T}\mathbf{\tau}_{2}\right)
\text{,}%
\]
which shows that the normal eigenlocus components $\mathbf{\tau}_{1}$ and
$\mathbf{\tau}_{2}$ on the constrained primal normal eigenlocus $\mathbf{\tau
}_{1}-\mathbf{\tau}_{2}$ must satisfy the critical minimum eigenenergy
constraint%
\[
\left\Vert \mathbf{\tau}_{1}-\mathbf{\tau}_{2}\right\Vert _{\min_{c}}^{2}%
\cong\left\Vert \mathbf{\tau}_{1}\right\Vert _{\min_{c}}^{2}+\left\Vert
\mathbf{\tau}_{2}\right\Vert _{\min_{c}}^{2}-2\mathbf{\tau}_{1}^{T}%
\mathbf{\tau}_{2}\text{.}%
\]
It is claimed that the total allowed eigenenergies of $\mathbf{\tau}_{1}$ and
$\mathbf{\tau}_{2}$ are symmetrically balanced with each other by means of a
symmetric equalizer statistic $\nabla_{eq}$ in relation to a centrally located
statistical fulcrum $f_{s}$. Given the assumption that the eigenenergies of
$\mathbf{\tau}_{1}$ and $\mathbf{\tau}_{2}$ are balanced in a symmetric
manner, it is claimed that the total allowed eigenenergy $\left\Vert
\mathbf{\tau}_{1}-\mathbf{\tau}_{2}\right\Vert _{\min_{c}}^{2}$ of
$\mathbf{\tau}$
\[
\left\Vert \mathbf{\tau}_{1}-\mathbf{\tau}_{2}\right\Vert _{\min_{c}}^{2}%
\cong\left\{  \left\Vert \mathbf{\tau}_{1}\right\Vert _{\min_{c}}%
^{2}-\mathbf{\tau}_{1}^{T}\mathbf{\tau}_{2}\right\}  +\left\{  \left\Vert
\mathbf{\tau}_{2}\right\Vert _{\min_{c}}^{2}-\mathbf{\tau}_{2}^{T}%
\mathbf{\tau}_{1}\right\}  \text{,}%
\]
is minimized when the total allowed eigenenergy of $\mathbf{\tau}_{1}$
satisfies the expression
\[
\left\Vert \mathbf{\tau}_{1}\right\Vert _{\min_{c}}^{2}-\left\Vert
\mathbf{\tau}_{1}\right\Vert \left\Vert \mathbf{\tau}_{2}\right\Vert
\cos\theta_{\mathbf{\tau}_{1}\mathbf{\tau}_{2}}\cong f_{s}-\nabla_{eq}\text{,}%
\]
and the total allowed eigenenergy of $\mathbf{\tau}_{2}$ satisfies the
expression%
\[
\left\Vert \mathbf{\tau}_{2}\right\Vert _{\min_{c}}^{2}-\left\Vert
\mathbf{\tau}_{1}\right\Vert \left\Vert \mathbf{\tau}_{2}\right\Vert
\cos\theta_{\mathbf{\tau}_{1}\mathbf{\tau}_{2}}\cong f_{s}+\nabla_{eq}\text{,}%
\]
where $\nabla_{eq}$ denotes a symmetric equalizer statistic and $f_{s}$
denotes a centrally located statistical fulcrum.

Section $17$ will examine how the total allowed eigenenergies of
$\mathbf{\tau}_{1}$ and $\mathbf{\tau}_{2}$ are balanced by means of a
symmetric equalizer statistic $\nabla_{eq}$%
\begin{equation}
\left(  \left\Vert \mathbf{\tau}_{1}\right\Vert _{\min_{c}}^{2}-\left\Vert
\mathbf{\tau}_{1}\right\Vert \left\Vert \mathbf{\tau}_{2}\right\Vert
\cos\theta_{\mathbf{\tau}_{1}\mathbf{\tau}_{2}}\right)  +\nabla_{eq}%
\Leftrightarrow\left(  \left\Vert \mathbf{\tau}_{2}\right\Vert _{\min_{c}}%
^{2}-\left\Vert \mathbf{\tau}_{1}\right\Vert \left\Vert \mathbf{\tau}%
_{2}\right\Vert \cos\theta_{\mathbf{\tau}_{1}\mathbf{\tau}_{2}}\right)
-\nabla_{eq}\text{,} \label{Balancing Factor for Eigenlocus Components}%
\end{equation}
in relation to a centrally located statistical fulcrum $f_{s}$. Section $17$
will develop statistical expressions for the symmetric equalizer statistic
$\nabla_{eq}$ and the statistical fulcrum $f_{s}$. Section $17$ will also
develop the statistical machinery behind a strong dual normal eigenlocus
equilibrium point. The KKT complementary conditions on a strong dual normal
eigenlocus $\mathbf{\tau}$ are examined next.

\subsection{KKT Complementary Conditions}

The KKT complementary conditions of optimization theory require that for all
constraints that are not active (are not precisely met as equalities, i.e.,
$y_{i}\left(  \mathbf{x}_{i}^{T}\mathbf{\tau}+\tau_{0}\right)  -1+\xi_{i}>0$),
the corresponding magnitudes $\psi_{i}$ of the Wolfe dual normal eigenaxis
components $\psi_{i}\overrightarrow{\mathbf{e}}_{i}$ must be $0$: $\psi
_{i}=0$
\citet{Cristianini2000}%
,
\citet{Scholkopf2002}%
. It follows that Eqs (\ref{KKTE4}) and (\ref{KKTE6}) must be satisfied as
equalities. Accordingly, let there be $l$ Wolfe dual normal eigenaxis
components $\psi_{i\ast}\overrightarrow{\mathbf{e}}_{i}$ that have non-zero
magnitudes $\left\{  \psi_{i\ast}\overrightarrow{\mathbf{e}}_{i}|\psi_{i\ast
}>0\right\}  _{i=1}^{l}$ for all constraints that are precisely met as equalities.

The next section will consider the manner in which Eq. (\ref{KKTE4})
determines the primal normal eigenlocus within the Wolf dual eigenspace.
Equation (\ref{KKTE4}) will be used to derive an expression for the $\tau_{0}$
term in Eq. (\ref{Discriminant Function}). The expression for $\tau_{0}$ will
then be used to obtain a normal eigenlocus test statistic for classifying
unknown pattern vectors.

\subsection{Statistical Functionality of the $\tau_{0}$ Term}

Given Eq. (\ref{KKTE4}), the following set of constrained primal normal
eigenlocus equations must be satisfied as strict equalities:%
\[
y_{i}\left(  \mathbf{x}_{i\ast}^{T}\mathbf{\tau}+\tau_{0}\right)  -1+\xi
_{i}=0,\ i=1,...,l\text{,}%
\]
so that an estimate for $\tau_{0}$ satisfies the strong dual normal eigenlocus
equation:%
\begin{align}
\tau_{0}  &  =\frac{1}{l}\sum\nolimits_{i=1}^{l}y_{i}\left(  1-\xi_{i}\right)
-\frac{1}{l}\sum\nolimits_{i=1}^{l}\mathbf{x}_{i\ast}^{T}\mathbf{\tau}%
\text{,}\label{Normal Eigenlocus Projection Factor}\\
&  =\frac{1}{l}\sum\nolimits_{i=1}^{l}y_{i}\left(  1-\xi_{i}\right)  -\left(
\frac{1}{l}\sum\nolimits_{i=1}^{l}\mathbf{x}_{i\ast}\right)  ^{T}\mathbf{\tau
}\text{,}\nonumber
\end{align}
where the expression for $\tau_{0}$ is comprised of class training labels and
inner product statistics between the extreme training points and
$\mathbf{\tau}$. Section $17$ will show that the $\tau_{0}$ term plays a large
role in balancing the total allowed eigenenergies of $\mathbf{\tau}_{1}$ and
$\mathbf{\tau}_{2}$; Section $17$ will examine how $\tau_{0}$ determines the
symmetric equalizer statistic $\nabla_{eq}$ in Eq.
(\ref{Balancing Factor for Eigenlocus Components}).

The expression for $\tau_{0}$ in Eq.
(\ref{Normal Eigenlocus Projection Factor}) is now used to obtain a normal
eigenlocus test statistic that is used to classify unknown pattern vectors.

\subsection{The Normal Eigenlocus Test Statistic}

Substitution of the expression for $\tau_{0}$ in Eq.
(\ref{Normal Eigenlocus Projection Factor}) into the expression for the
discriminant function $D\left(  \mathbf{x}\right)  $ in Eq.
(\ref{Discriminant Function}) provides a normal eigenlocus test statistic
$\Lambda_{\mathbf{\tau}}\left(  \mathbf{x}\right)  \overset{H_{1}%
}{\underset{H_{2}}{\gtrless}}0$ for classifying an unknown pattern vector
$\mathbf{x}$:%
\begin{align}
\Lambda_{\mathbf{\tau}}\left(  \mathbf{x}\right)   &  =\mathbf{x}%
^{T}\mathbf{\tau}-\frac{1}{l}\sum\nolimits_{i=1}^{l}\mathbf{x}_{i\ast}%
^{T}\mathbf{\tau}\label{NormalEigenlocusTestStatistic}\\
&  \mathbf{+}\frac{1}{l}\sum\nolimits_{i=1}^{l}y_{i}\left(  1-\xi_{i}\right)
\text{,}\nonumber\\
&  =\left(  \mathbf{x}-\overline{\mathbf{x}}_{i\ast}\right)  ^{T}\mathbf{\tau
}\nonumber\\
&  \mathbf{+}\frac{1}{l}\sum\nolimits_{i=1}^{l}y_{i}\left(  1-\xi_{i}\right)
\text{,}\nonumber
\end{align}
where the statistic $\overline{\mathbf{x}}_{i\ast}$ determines the expected
locus of a set of extreme training points and the statistic $\frac{1}{l}%
\sum\nolimits_{i=1}^{l}y_{i}\left(  1-\xi_{i}\right)  $ accounts for the class
membership of the normal eigenaxis components on $\mathbf{\tau}_{1}$ and
$\mathbf{\tau}_{2}$. An expression is now obtained for a statistical decision
locus which specifies the likelihood that an unknown pattern vector belongs to
a pattern category. The expression provides geometric insight into the
statistical machinery of the normal eigenlocus discriminant function.

\subsubsection{Statistical Decision Locus}

Denote the unit normal eigenlocus $\mathbf{\tau/}\left\Vert \mathbf{\tau
}\right\Vert $ by $\widehat{\mathbf{\tau}}$. Letting $\mathbf{\tau=\tau
/}\left\Vert \mathbf{\tau}\right\Vert $ in Eq.
(\ref{NormalEigenlocusTestStatistic}) provides an expression for a statistical
decision locus%
\begin{align}
\Lambda_{\widehat{\mathbf{\tau}}}\left(  \mathbf{x}\right)   &  =\left(
\mathbf{x}-\overline{\mathbf{x}}_{i\ast}\right)  ^{T}\mathbf{\tau/}\left\Vert
\mathbf{\tau}\right\Vert \nonumber\\
&  \mathbf{+}\frac{1}{l\left\Vert \mathbf{\tau}\right\Vert }\sum
\nolimits_{i=1}^{l}y_{i}\left(  1-\xi_{i}\right)  \text{,}\nonumber
\end{align}
which is determined by the scalar projection of $\mathbf{x}-\overline
{\mathbf{x}}_{i\ast}$ onto $\widehat{\mathbf{\tau}}$. More specifically, the
component of $\mathbf{x}-\overline{\mathbf{x}}_{i\ast}$ along
$\widehat{\mathbf{\tau}}$ determines a signed magnitude $\left\Vert
\mathbf{x}-\overline{\mathbf{x}}_{i\ast}\right\Vert \cos\theta$ along the axis
of $\widehat{\mathbf{\tau}}$, where $\theta$ is the angle between the vector
$\mathbf{x}-\overline{\mathbf{x}}_{i\ast}$ and $\widehat{\mathbf{\tau}}$.
Accordingly, the component $\operatorname{comp}%
_{\overrightarrow{\widehat{\mathbf{\tau}}}}\left(  \overrightarrow{\left(
\mathbf{x}-\overline{\mathbf{x}}_{i\ast}\right)  }\right)  $ of the vector
transform $\mathbf{x}-\overline{\mathbf{x}}_{i\ast}$ of an unknown pattern
vector $\mathbf{x}$ along the axis of the unit normal eigenlocus
$\widehat{\mathbf{\tau}}$%
\[
\operatorname{comp}_{\overrightarrow{\widehat{\mathbf{\tau}}}}\left(
\overrightarrow{\left(  \mathbf{x}-\overline{\mathbf{x}}_{i\ast}\right)
}\right)  =\left\Vert \mathbf{x}-\overline{\mathbf{x}}_{i\ast}\right\Vert
\cos\theta\text{,}%
\]
determines a statistical locus $P_{D\left(  \mathbf{x}\right)  }$ of a
category decision, where $P_{D\left(  \mathbf{x}\right)  }$ is at a distance
of $\left\Vert \mathbf{x}-\overline{\mathbf{x}}_{i\ast}\right\Vert \cos\theta$
from the origin, along the axis of the strong dual normal eigenlocus
$\mathbf{\tau}$. Given Eqs (\ref{Decision Boundary}),
(\ref{Decision Border One}), and (\ref{Decision Border Two}), it is concluded
that the scaled $1\mathbf{/}\left\Vert \mathbf{\tau}\right\Vert $ discriminant
function $\Lambda_{\widehat{\mathbf{\tau}}}\left(  \mathbf{x}\right)  $
generates a statistical decision locus $P_{D\left(  \mathbf{x}\right)  }$
which lies in one of the decision regions delineated by the constrained
discriminant function $D\left(  \mathbf{x}\right)  $ in Eq.
(\ref{Discriminant Function}).

The expression for a statistical decision locus $P_{D\left(  \mathbf{x}%
\right)  }$ provides insight into how the discriminant function $D\left(
\mathbf{x}\right)  $ in Eq. (\ref{Discriminant Function}) assigns an unknown
pattern vector to a pattern class. Given Eqs (\ref{Decision Boundary}),
(\ref{Decision Border One}), and (\ref{Decision Border Two}), it follows that
the statistic $\operatorname{comp}_{\overrightarrow{\widehat{\mathbf{\tau}}}%
}\left(  \overrightarrow{\left(  \mathbf{x}-\overline{\mathbf{x}}_{i\ast
}\right)  }\right)  $ delineates a statistical decision locus $P_{D\left(
\mathbf{x}\right)  }$ which lies in a geometric region that is either $\left(
1\right)  $ inside one of the symmetrical decision regions of large covariance
depicted in Figs $17$, $19$, and $20$, $\left(  2\right)  $ on the other side
of the linear decision border $D_{1}\left(  \mathbf{x}\right)  $, where
$\mathbf{\tau}^{T}\mathbf{x}+\tau_{0}=+1$, or $\left(  3\right)  $ on the
other side of the linear decision border $D_{-1}\left(  \mathbf{x}\right)  $,
where $\mathbf{\tau}^{T}\mathbf{x}+\tau_{0}=-1$. It is concluded that the
statistic $\operatorname{comp}_{\overrightarrow{\widehat{\mathbf{\tau}}}%
}\left(  \overrightarrow{\left(  \mathbf{x}-\overline{\mathbf{x}}_{i\ast
}\right)  }\right)  $ generates a statistical decision locus $P_{D\left(
\mathbf{x}\right)  }$ which specifies a likelihood that an unknown pattern
vector belongs to category one or category two.

Again, letting $\mathbf{\tau=\tau/}\left\Vert \mathbf{\tau}\right\Vert $ in
Eq. (\ref{NormalEigenlocusTestStatistic}), the scaled $1\mathbf{/}\left\Vert
\mathbf{\tau}\right\Vert $ discriminant function $\Lambda
_{\widehat{\mathbf{\tau}}}\left(  \mathbf{x}\right)  $%
\[
\Lambda_{\widehat{\mathbf{\tau}}}\left(  \mathbf{x}\right)
=\operatorname{comp}_{\overrightarrow{\widehat{\mathbf{\tau}}}}\left(
\overrightarrow{\left(  \mathbf{x}-\overline{\mathbf{x}}_{i\ast}\right)
}\right)  +\frac{1}{l\left\Vert \mathbf{\tau}\right\Vert }\sum\nolimits_{i=1}%
^{l}y_{i}\left(  1-\xi_{i}\right)  \text{,}%
\]
generates an output based on the statistic $\operatorname{comp}%
_{\overrightarrow{\widehat{\mathbf{\tau}}}}\left(  \mathbf{x}-\overline
{\mathbf{x}}_{i\ast}\right)  $ and the class membership statistic $\frac
{1}{l\left\Vert \mathbf{\tau}\right\Vert }\sum\nolimits_{i=1}^{l}y_{i}\left(
1-\xi_{i}\right)  $. It follows that the likelihood that an unknown pattern
vector $\mathbf{x}$ belongs to a pattern class is a function of a statistical
locus $\operatorname{comp}_{\overrightarrow{\widehat{\mathbf{\tau}}}}\left(
\mathbf{x}-\overline{\mathbf{x}}_{i\ast}\right)  $ of a category decision and
a class membership statistic $\frac{1}{l\left\Vert \mathbf{\tau}\right\Vert
}\sum\nolimits_{i=1}^{l}y_{i}\left(  1-\xi_{i}\right)  $. Figure $21$ depicts
a statistical decision locus generated by the discriminant function
$\Lambda_{\widehat{\mathbf{\tau}}}\left(  \mathbf{x}\right)  $.

\begin{center}%
\begin{center}
\includegraphics[
natheight=7.499600in,
natwidth=9.999800in,
height=3.2897in,
width=4.3777in
]%
{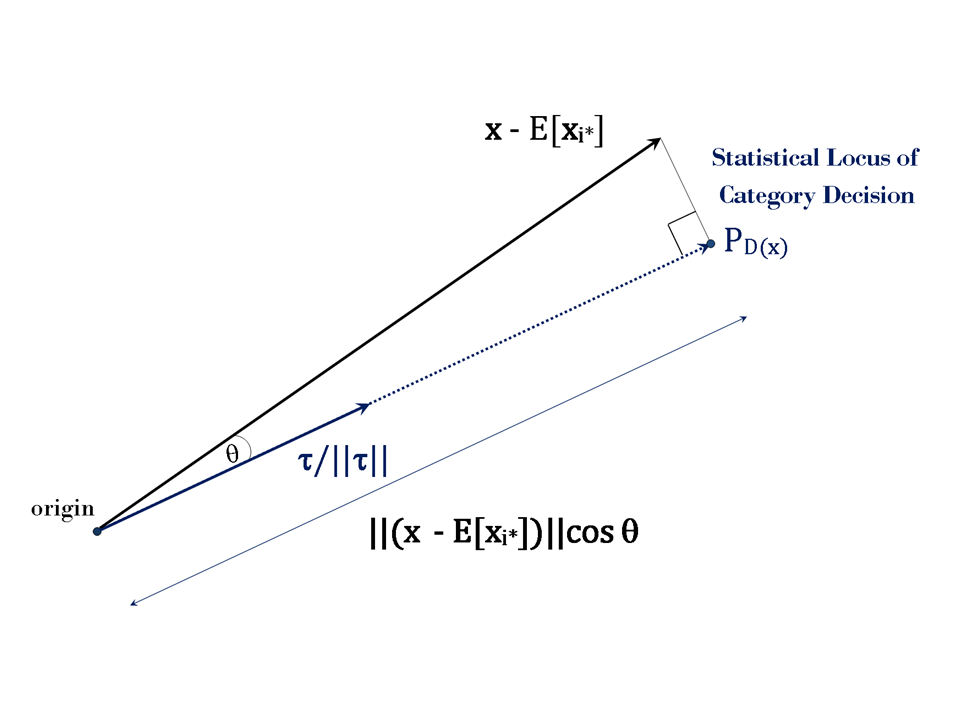}%
\end{center}

\end{center}

\begin{flushleft}
Figure $21$: Illustration of a statistical decision locus $P_{D\left(
\mathbf{x}\right)  }$ generated by a scaled $1\mathbf{/}\left\Vert
\mathbf{\tau}\right\Vert $ discriminant function $\Lambda
_{\widehat{\mathbf{\tau}}}\left(  \mathbf{x}\right)  $ for an unknown
transformed pattern vector $\mathbf{x}-E[\mathbf{x}_{i\ast}]$ that is
projected onto $\mathbf{\tau/}\left\Vert \mathbf{\tau}\right\Vert $. Values of
the statistical decision locus $\operatorname{comp}%
_{\overrightarrow{\widehat{\mathbf{\tau}}}}\left(  \mathbf{x}-\overline
{\mathbf{x}}_{i\ast}\right)  $ and the class membership statistic $\frac
{1}{l\left\Vert \mathbf{\tau}\right\Vert }\sum\nolimits_{i=1}^{l}y_{i}\left(
1-\xi_{i}\right)  $ specify the likelihood that the unknown vector
$\mathbf{x}$ belongs to class one or class two.
\end{flushleft}

Using the expression for the discriminant function in Eq.
(\ref{NormalEigenlocusTestStatistic}), the strong dual statistical decision
function $\operatorname{sign}\left(  \Lambda_{\mathbf{\tau}}\left(
\mathbf{x}\right)  \right)  $%
\begin{align*}
\operatorname{sign}\left(  \Lambda_{\mathbf{\tau}}\left(  \mathbf{x}\right)
\right)   &  =\operatorname{sign}\left[  \left(  \mathbf{x}-\frac{1}{l}%
\sum\nolimits_{i=1}^{l}\mathbf{x}_{i\ast}\right)  ^{T}\mathbf{\tau+\cdots
}\right] \\
&  \operatorname{sign}\left[  \cdots+\frac{1}{l}\sum\nolimits_{i=1}^{l}%
y_{i}\left(  1-\xi_{i}\right)  \right]  \text{,}%
\end{align*}
where $\operatorname{sign}\left(  x\right)  \equiv\frac{x}{\left\vert
x\right\vert }$ for $x\neq0$, classifies an unknown pattern vector
$\mathbf{x}_{1_{i}}$ or $\mathbf{x}_{2_{i}}$ into category one if
$\operatorname{sign}\left(  \Lambda_{\mathbf{\tau}}\left(  \mathbf{x}\right)
\right)  =1$ and into category two if $\operatorname{sign}\left(
\Lambda_{\mathbf{\tau}}\left(  \mathbf{x}\right)  \right)  =-1$.

Substitution of the expression for $\mathbf{\tau}$ of Eq.
(\ref{Pair of Normal Eigenlocus Components}) into the normal eigenlocus test
statistic in Eq. (\ref{NormalEigenlocusTestStatistic}) provides a normal
eigenlocus test statistic in terms of the strong dual normal eigenlocus
components $\mathbf{\tau}_{1}$ and $\mathbf{\tau}_{2}$:%
\begin{align*}
\Lambda_{\mathbf{\tau}_{1}\mathbf{-\tau}_{2}}\left(  \mathbf{x}\right)   &
=\left(  \mathbf{x}-\frac{1}{l}\sum\nolimits_{i=1}^{l}\mathbf{x}_{i\ast
}\right)  ^{T}\mathbf{\tau}_{1}\\
&  -\left(  \mathbf{x}-\frac{1}{l}\sum\nolimits_{i=1}^{l}\mathbf{x}_{i\ast
}\right)  ^{T}\mathbf{\tau}_{2}\\
&  \mathbf{+}\frac{1}{l}\sum\nolimits_{i=1}^{l}y_{i}\left(  1-\xi_{i}\right)
\overset{H_{1}}{\underset{H_{2}}{\gtrless}}0\text{.}%
\end{align*}
Section $18$ will examine the robust likelihood ratio that is encoded within
the normal eigenlocus test statistic $\Lambda_{\mathbf{\tau}_{1}\mathbf{-\tau
}_{2}}\left(  \mathbf{x}\right)  $.

Strong dual normal eigenlocus transforms generate regularized, data-driven
geometric architectures that encode robust decision statistics for any two
data distributions. It will now be demonstrated that strong dual normal
eigenlocus transforms determine unforeseen optimal decision systems for
completely overlapping data distributions.

\subsection{Strong Dual Normal Eigenlocus Transforms for Homogeneous
Distributions}

Take two Gaussian data sets that are characterized by identical means and
covariance matrices. Let both pattern classes have the covariance matrix:%
\[
\mathbf{\Sigma}_{1}=\mathbf{\Sigma}_{2}=%
\begin{pmatrix}
1 & 0\\
0 & 1
\end{pmatrix}
\text{,}%
\]
and the mean vector%
\[
\mathbf{\mu}_{1}=\mathbf{\mu}_{2}=%
\begin{pmatrix}
0, & 0
\end{pmatrix}
^{T}\text{,}%
\]
where the Bayes' discriminant function has an error rate of $50\%$. Before the
strong dual decision system for the homogeneous data sets outlined above is
revealed, a few remarks are in order.

Recall that the constrained Lagrangian functional $L_{\Psi\left(
\mathbf{\tau}\right)  }$ of Eq. (\ref{Lagrangian Normal Eigenlocus}) returns
the minimum number of normal eigenaxis components that are necessary to
symmetrically partition a two class feature space. By definition, an extreme
data point is located somewhere between a pair of data distributions. Given
that the above data distributions are completely overlapping, it follows that
all of the training vectors are extreme data points. Therefore, almost
identical sets of eigen-scales will be determined for each pattern class,
resulting in similar constrained primal normal eigenaxis components on
$\mathbf{\tau}_{1}$ and $\mathbf{\tau}_{2}.$ Given that $\mathbf{\tau}_{1}$
and $\mathbf{\tau}_{2}$ are formed by similar normal eigenaxis components, it
follows that $\mathbf{\tau}_{1}$ $\mathbf{\thickapprox\tau}_{2}$. A strong
dual decision system was obtained for the homogeneous data sets outlined
above. The results are summarized below.

$300$ training vectors were obtained for each identical data category. The
complete data set of $600$ training vectors were transformed into a strong
dual normal eigenlocus of constrained primal normal eigenaxis components by
solving the inequality constrained optimization problem of Eq.
(\ref{Primal Normal Eigenlocus}). As expected, all $600$ training vectors were
transformed into constrained primal normal eigenaxis components on
$\mathbf{\tau}_{1}$ and $\mathbf{\tau}_{2}$:%
\begin{align*}
\mathbf{\tau}  &  =\mathbf{\tau}_{1}-\mathbf{\tau}_{2}\text{,}\\
&  =\sum\nolimits_{i=1}^{300}\psi_{1_{i\ast}}\mathbf{x}_{1_{i\ast}}%
-\sum\nolimits_{i=1}^{300}\psi_{2_{i\ast}}\mathbf{x}_{2_{i\ast}}\text{,}%
\end{align*}
where $\mathbf{\tau}_{1}$ $\mathbf{\thickapprox\tau}_{2}$. Given that
$\mathbf{\tau}_{1}$ $\mathbf{\thickapprox\tau}_{2}$, the width $\frac
{1}{\left\Vert \mathbf{\tau}_{1}-\mathbf{\tau}_{2}\right\Vert }$ of the
geometric regions between the linear decision boundary and the linear decision
borders is extremely large; the distance $\frac{2}{\left\Vert \mathbf{\tau
}_{1}-\mathbf{\tau}_{2}\right\Vert }$ between the linear decision borders is
also extremely large. Figure $22$ illustrates that strong dual decision
systems determine symmetrical linear partitions of completely overlapping data
distributions, where each extreme data point is enclosed in a blue circle.

\begin{center}%
\begin{center}
\includegraphics[
natheight=7.416600in,
natwidth=14.635200in,
height=2.2857in,
width=4.4841in
]%
{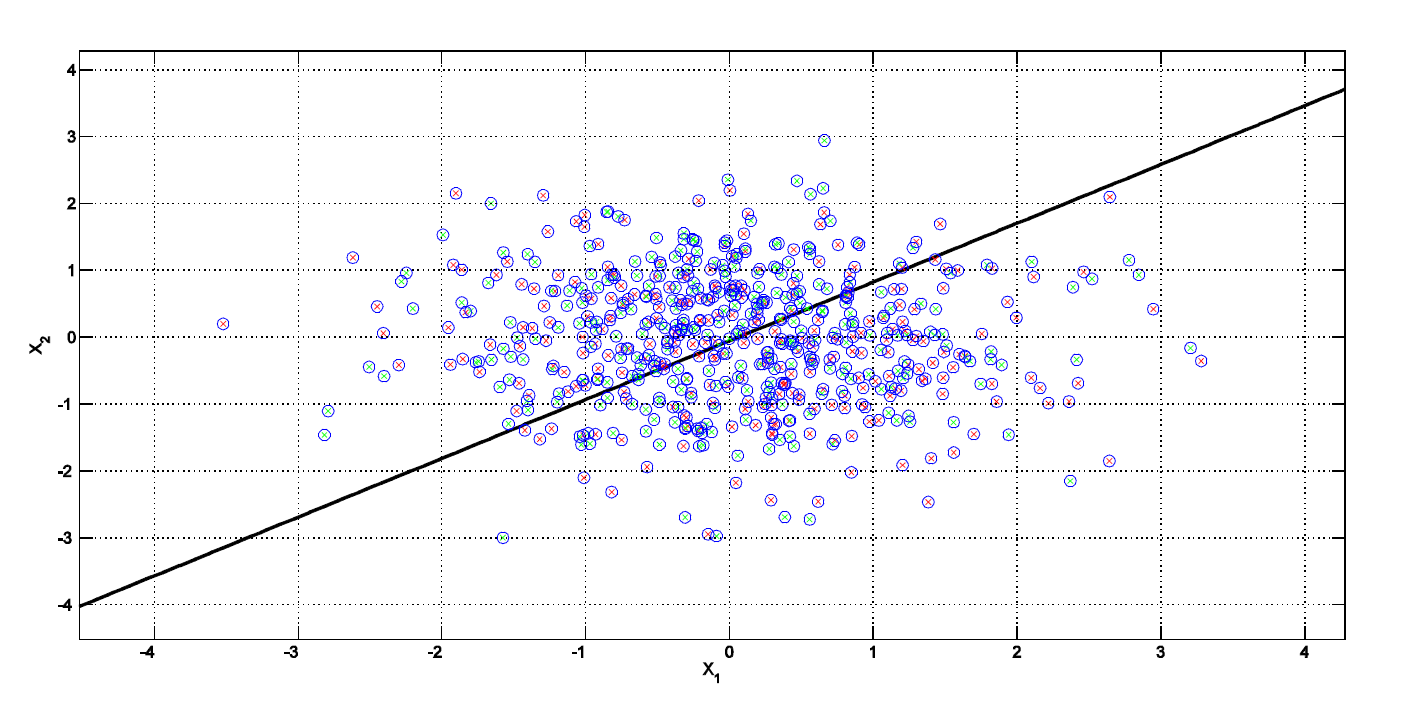}%
\end{center}

\end{center}

\begin{flushleft}
FIGURE $22$. Illustration that strong dual normal eigenlocus transforms
determine symmetrical linear partitions of homogeneous data distributions
which are completely overlapping with each other.
\end{flushleft}

Because the breadth of the geometric regions between the linear decision
boundary and its borders is too large to be depicted, only the linear decision
boundary can be seen in Fig. $22$. The strong dual statistical decision system
depicted in Fig. $22$ achieves the Bayes error rate of $50\%$.

\subsection{Equilibrium States of Strong Dual Decision Systems}

The next two sections will outline two interrelated KKT constraints that will
play a large role in determining the regularized geometric configurations of
$\mathbf{\psi}$ and $\mathbf{\tau}$. Later on, Sections $14$ and $15$ will
examine how the KKT constraint of Eq.
(\ref{Equilibrium Constraint on Dual Eigen-components}) determines a state of
statistical equilibrium in which the normal eigenaxis components on
$\mathbf{\psi}$ and $\mathbf{\tau}$ are jointly and symmetrically distributed
over the constrained primal normal eigenaxis components on $\mathbf{\tau}$.
Section $17$ will demonstrate that, given any strong dual decision system in a
state of statistical equilibrium, the magnitudes $\psi_{i\ast}$ of the Wolfe
dual normal eigenaxis components $\psi_{i\ast}\overrightarrow{\mathbf{e}}_{i}$
\ determine symmetrical eigen-scales, for which joint eigenenergies of
$\mathbf{\psi}$ and $\mathbf{\tau}$ are symmetrically distributed over the
eigen-scaled extreme points on $\mathbf{\tau=\tau}_{1}-\mathbf{\tau}_{2}$,
such that the total allowed eigenenergies of $\mathbf{\tau}_{1}$ and
$\mathbf{\tau}_{2}$ are symmetrically balanced with each other. Therefore,
given a strong dual decision system in a state of statistical equilibrium, all
of the normal eigenaxis components on $\mathbf{\psi}$ and $\mathbf{\tau}$
exhibit critical lengths which satisfy the state of statistical equilibrium.
Section $17$ will develop and use all of these results to define the
statistical equilibrium point of a strong dual decision system.

\subsection{Critical Minimum Eigenenergy Constraints on $\mathbf{\tau}_{1}$,
$\mathbf{\tau}_{2}$, and $\mathbf{\tau}$}

Returning to the KKT constraint of Eq. (\ref{KKTE6}), the following algebraic
system of constrained primal normal eigenlocus equations must be satisfied as
strict equalities:%
\begin{equation}
\psi_{i\ast}\left\{  y_{i}\left(  \mathbf{x}_{i\ast}^{T}\mathbf{\tau}+\tau
_{0}\right)  -1+\xi_{i}\right\}  =0,\ i=1,...,l\text{.}
\label{Minimum Eigenenergy Constraints}%
\end{equation}
Section $17$ will examine how the algebraic system of $l$ strong dual normal
eigenlocus equations in Eq. (\ref{Minimum Eigenenergy Constraints}) determine
critical minimum eigenenergy constraints that are satisfied by the constrained
primal normal eigenlocus components $\mathbf{\tau}_{1}-\mathbf{\tau}_{2}$ on
$\mathbf{\tau}$. Equation (\ref{Minimum Eigenenergy Constraints}) will be used
to develop expressions for the symmetric equalizer statistic $\nabla_{eq}$ and
the implicit statistical fulcrum $f$ in Eq.
(\ref{Balancing Factor for Eigenlocus Components}). The KKT constraint of Eq.
(\ref{KKTE2}) is examined next.

\subsection{Equilibrium Constraints on Wolfe Dual Normal Eigenaxis Components}

The KKT constraint of Eq. (\ref{KKTE2}) specifies that the magnitudes of all
of the Wolfe dual normal eigenaxis components on $\mathbf{\psi}$ must satisfy
the strong dual normal eigenlocus equation:%
\[
\left(  y_{i}=1\right)  \sum\nolimits_{i=1}^{l_{1}}\psi_{1_{i\ast}}+\left(
y_{i}=-1\right)  \sum\nolimits_{i=1}^{l_{2}}\psi_{2_{i\ast}}=0\text{,}%
\]
so that the integrated lengths of the Wolfe dual normal eigenaxes correlated
with each pattern category:%
\[
\sum\nolimits_{i=1}^{l_{1}}\psi_{1_{i\ast}}-\sum\nolimits_{i=1}^{l_{2}}%
\psi_{2_{i\ast}}=0\text{,}%
\]
must balance each other:%
\begin{equation}
\sum\nolimits_{i=1}^{l_{1}}\psi_{1_{i\ast}}=\sum\nolimits_{i=1}^{l_{2}}%
\psi_{2_{i\ast}}\text{.}
\label{Equilibrium Constraint on Dual Eigen-components}%
\end{equation}
Section $17$ will demonstrate that Eq.
(\ref{Equilibrium Constraint on Dual Eigen-components}) determines a state of
statistical equilibrium, for which correlated normal eigenaxis components on
$\mathbf{\psi}$ and $\mathbf{\tau}$ possess critical magnitudes or lengths. It
will be shown that the lengths of the Wolfe dual normal eigenaxis components
must be selected so that the total allowed eigenenergies of $\mathbf{\tau}%
_{1}$ and $\mathbf{\tau}_{2}$ are balanced by means of a symmetric equalizer
statistic $\nabla_{eq}$ in relation to a centrally located statistical fulcrum
$f_{s}$. Section $18$ will demonstrate how this statistical balancing feat
enables the constrained discriminant function $D\left(  \mathbf{x}\right)
=\mathbf{x}^{T}\mathbf{\tau}+\tau_{0}$ of Eq. (\ref{Discriminant Function}) to
delineate centrally located, bipartite, congruent geometric regions of large
covariance for a wide variety of data distributions. Sections $14$, $15$, and
$16$ will demonstrate how the KKT constraint of Eq.
(\ref{Equilibrium Constraint on Dual Eigen-components}) enforces joint
symmetrical distributions of the components of $\mathbf{\psi}$ and
$\mathbf{\tau}$ over each of the $l$ constrained primal normal eigenaxis
components on $\mathbf{\tau}$, whereby Section $17$ will demonstrate that each
constrained primal normal eigenaxis component on $\mathbf{\tau}$ encodes an
eigen-balanced eigenlocus of an extreme data point, such that the total
allowed eigenenergies of $\mathbf{\tau}_{1}$ and $\mathbf{\tau}_{2}$ are
symmetrically balanced with each other.

The next stage of the analysis will examine the strong dual normal eigenlocus
problem within the context of the eigenlocus equation of a Wolfe dual normal
eigenlocus. Section $10$ will define the eigenlocus equation of a Wolfe dual
normal eigenlocus. Section $10$ will also examine the geometric essence and
the fundamental properties of a Wolfe dual normal eigenlocus.

\section{Eigenlocus Equation of a Wolfe Dual Normal Eigenlocus}

This stage of the analysis returns to the six KKT constraints on the
Lagrangian functional $L_{\Psi\left(  \mathbf{\tau}\right)  }$ of Eq.
(\ref{Lagrangian Normal Eigenlocus}) which are specified by Eqs (\ref{KKTE1}),
(\ref{KKTE2}), (\ref{KKTE3}), (\ref{KKTE4}), (\ref{KKTE5}), and (\ref{KKTE6}).
The resulting expressions for a primal normal eigenlocus $\mathbf{\tau}$ and
regularization parameters $\xi_{i}$ and $C$, in terms of a Wolfe dual normal
eigenlocus $\mathbf{\psi}$, are substituted into the Lagrangian functional
$L_{\Psi\left(  \mathbf{\tau}\right)  }$ of Eq.
(\ref{Lagrangian Normal Eigenlocus}) and simplified. This produces the
eigenlocus equation of a Wolfe dual normal eigenlocus:%
\begin{equation}
\max\Xi\left(  \mathbf{\psi}\right)  =\sum\nolimits_{i=1}^{N}\psi_{i}%
-\sum\nolimits_{i,j=1}^{N}\psi_{i}\psi_{j}y_{i}y_{j}\frac{\left[
\mathbf{x}_{i}^{T}\mathbf{x}_{j}+\delta_{ij}/C\right]  }{2}\text{,}
\label{Wolfe Dual Normal Eigenlocus}%
\end{equation}
which is subject to the algebraic constraints that $\sum\nolimits_{i=1}%
^{N}y_{i}\psi_{i}=0$ and $\psi_{i}\geq0$, where $\delta_{ij}$ is the Kronecker
$\delta$ defined as unity for $i=j$ and $0$ otherwise.

Equation (\ref{Wolfe Dual Normal Eigenlocus}) can be written in vector
notation by letting $\mathbf{Q}\triangleq\epsilon\mathbf{I}%
+\widetilde{\mathbf{X}}\widetilde{\mathbf{X}}^{T}$ and $\widetilde{\mathbf{X}%
}\triangleq\mathbf{D}_{y}\mathbf{X}$, where $\mathbf{D}_{y}$ is an $N\times N$
diagonal matrix of training labels $y_{i}$ and the $N\times d$ data matrix is
$\mathbf{X}$ $=%
\begin{pmatrix}
\mathbf{x}_{1}, & \mathbf{x}_{2}, & \ldots, & \mathbf{x}_{N}%
\end{pmatrix}
^{T}$. This produces the matrix version of an equation of a primal normal
eigenlocus in a Wolfe dual eigenspace:%
\begin{equation}
\max\Xi\left(  \mathbf{\psi}\right)  =\mathbf{1}^{T}\mathbf{\psi}%
-\frac{\mathbf{\psi}^{T}\mathbf{Q\psi}}{2}\text{,}
\label{Vector Form Wolfe Dual}%
\end{equation}
which is subject to the algebraic constraints $\mathbf{\psi}^{T}\mathbf{y}=0$
and $\psi_{i}\geq0$
\citet{Reeves2009}%
. It will be assumed that the $N$-dimensional vector $\mathbf{\psi}$ whose
components $\psi_{i\ast}\overrightarrow{\mathbf{e}}_{i}$ satisfy Eqs
(\ref{Wolfe Dual Normal Eigenlocus}) and (\ref{Vector Form Wolfe Dual}) is the
Wolfe dual normal eigenlocus of a hyperplane decision surface in $%
\mathbb{R}
^{N}$ that is bounded by bilaterally symmetrical hyperplane borders. Sections
$14$ and $15$ will consider how symmetrical Wolfe dual normal eigenaxis
components $\psi_{i\ast}\overrightarrow{\mathbf{e}}_{i}$ on a Wolfe dual
normal eigenlocus $\mathbf{\psi}$ determine the locus of a separating
hyperplane $D_{h_{0}}\left(  \mathbf{x}\right)  $ that is bounded by a pair of
bilaterally symmetrical hyperplane borders $D_{h_{1}}\left(  \mathbf{x}%
\right)  $ and $D_{h_{-1}}\left(  \mathbf{x}\right)  $. Section $11$ will
examine how the geometric configurations of $D_{h_{0}}\left(  \mathbf{x}%
\right)  $, $D_{h_{1}}\left(  \mathbf{x}\right)  $, and $D_{h_{-1}}\left(
\mathbf{x}\right)  $ are determined by the eigenspectrum of $\mathbf{Q}$. It
will shortly be demonstrated how the constraint $\mathbf{\psi}^{T}%
\mathbf{y}=0$ effectively determines the eigenlocus of $\mathbf{\psi}$.

Now consider any Wolfe dual normal eigenaxis component $\psi_{i\ast
}\overrightarrow{\mathbf{e}}_{i}\ $on $\mathbf{\psi}$, where $\psi_{i\ast}>0$.
It will be assumed that each Wolfe dual normal eigenaxis component
$\psi_{i\ast}\overrightarrow{\mathbf{e}}_{i}$ on $\mathbf{\psi}$ is correlated
with a $d$-dimensional extreme training vector $\mathbf{x}_{i\ast}$, which
determines the direction of a constrained primal normal eigenaxis component
$\psi_{i\ast}\mathbf{x}_{i\ast}$ on $\mathbf{\tau}$. Later on, Sections $14$
and $15$ will examine uniform geometric and statistical properties which are
jointly exhibited by the Wolfe dual normal eigenaxis components $\psi_{i\ast
}\overrightarrow{\mathbf{e}}_{i}$ on $\mathbf{\psi} $ and the constrained
primal normal eigenaxis components $\psi_{i\ast}\mathbf{x}_{i\ast}$ on
$\mathbf{\tau}$. Sections $14$ and $15$ will demonstrate how the length
$\psi_{i\ast}$ of each Wolfe dual normal eigenaxis component $\psi_{i\ast
}\overrightarrow{\mathbf{e}}_{i}$ on $\mathbf{\psi}$ determines the length
$\psi_{i\ast}\left\Vert \mathbf{x}_{i\ast}\right\Vert $ of a correlated,
constrained primal normal eigenaxis component $\psi_{i\ast}\mathbf{x}_{i\ast}$
on $\mathbf{\tau}$. Sections $14$ and $15$ will also demonstrate that the
direction of each Wolfe dual normal eigenaxis component $\psi_{i\ast
}\overrightarrow{\mathbf{e}}_{i}$ is identical to the direction of a
correlated, constrained primal normal eigenaxis component $\psi_{i\ast
}\mathbf{x}_{i\ast}$. Thereby, the eigenloci of the Wolfe dual normal
eigenaxis components $\psi_{1_{i\ast}}\overrightarrow{\mathbf{e}}_{i}$ and
$\psi_{2_{i\ast}}\overrightarrow{\mathbf{e}}_{i}$ will be shown to determine
well-proportioned eigen-scales\emph{\ }$\psi_{1_{i\ast}}\mathbf{\ }$and
$\psi_{2_{i\ast}}\mathbf{\ }$for the constrained primal normal eigenaxis
components $\psi_{1_{i\ast}}\mathbf{x}_{1_{i\ast}}$ and $\psi_{2_{i\ast}%
}\mathbf{x}_{2_{i\ast}}$ on the strong dual normal eigenlocus components
$\mathbf{\tau}_{1}$ and $\mathbf{\tau}_{2}$ respectively. It will be shown
that each eigen-scaled extreme training point $\psi_{1_{i\ast}}\mathbf{x}%
_{1_{i\ast}}$ or $\psi_{2_{i\ast}}\mathbf{x}_{2_{i\ast}}$ specifies an
eigen-balanced geometric location of a constrained primal normal eigenaxis
component on $\mathbf{\tau}$.

Thus far, this paper has demonstrated that strong dual normal eigenlocus
transforms generate robust statistical decision systems for a wide variety of
data distributions, including completely overlapping distributions. Section
$17$ will show that the regularized, data-driven geometric architecture
depicted in Fig. $20$ is configured by enforcing joint symmetrical
distributions of the eigenenergies of $\mathbf{\psi}$ and $\mathbf{\tau}$ over
the eigen-scaled extreme training vectors on $\mathbf{\tau}_{1}$ and
$\mathbf{\tau}_{2}$, whereby the eigenenergies of the strong dual normal
eigenlocus components $\mathbf{\tau}_{1}$ and $\mathbf{\tau}_{2}$ on
$\mathbf{\tau}$ are symmetrically balanced with each other.

The chain of arguments outlined above will be used to demonstrate how
integrated, eigen-balanced sets of constrained primal normal eigenaxis
components on a strong dual normal eigenlocus $\mathbf{\tau=\tau}%
_{1}-\mathbf{\tau}_{2}$ provide an estimate of the real unknowns, which are
the constrained eigen-coordinate locations of an unknown normal eigenaxis
$\mathbf{v}$ that provides an axis of symmetry for a statistical decision
system of linear partitions.

\subsection{The Wolfe Dual Normal Eigenlocus}

The geometric and statistical properties exhibited by a Wolfe dual normal
eigenlocus are examined next. These properties are specified by Eqs
(\ref{Wolfe Dual Normal Eigenlocus}) and (\ref{Vector Form Wolfe Dual}),
strong duality relationships between the algebraic systems of $\max\Xi\left(
\mathbf{\psi}\right)  $ and $\min\Psi\left(  \mathbf{\tau}\right)  $, and the
KKT constraints on the Lagrangian functional $L_{\Psi\left(  \mathbf{\tau
}\right)  }$. The first property concerns the geometric nature of the
second-degree homogeneous polynomial surface in Eq.
(\ref{Wolfe Dual Normal Eigenlocus}).

\subsubsection{Second-degree Homogeneous Polynomial Surfaces}

The Wolfe dual normal eigenlocus $\mathbf{\psi}$ of Eq.
(\ref{Wolfe Dual Normal Eigenlocus}) is determined by a constrained polynomial
equation of the form%
\begin{equation}
\sum\nolimits_{i,j=1}^{N}q_{ij}x_{i}x_{j}\text{,}
\label{Second-degree Homogeneous Polynomial Surface}%
\end{equation}
which is a second-degree homogeneous polynomial in $N$ variables $\left(
x_{1},x_{2},\ldots,x_{N}\right)  ^{T}$. Second-degree polynomials written in
vector notation $\mathbf{x}^{T}\mathbf{Qx}$ are commonly known as quadratic
forms. Quadratic forms describe six classes of second-order surfaces that
include $N$-dimensional circles, ellipses, hyperbolae, parabolas, lines, and
points
\citet{Hewson2009}%
. Second-order surfaces are also known as quadratic or quadric surfaces.

It has previously been argued that the constrained quadratic form
$\mathbf{\psi}^{T}\mathbf{Q\psi}$ denoted in Eq. (\ref{Vector Form Wolfe Dual}%
) describes three, symmetrically positioned $N$-dimensional hyperplane
partitioning surfaces, where the distance from each hyperplane surface to the
origin is determined by a correlated constraint on the discriminant function
of Eq. (\ref{Discriminant Function}). Rayleigh's principle is now used to
precisely define the geometric essence of $\mathbf{\psi}$. Rayleigh's
principle can be found in
\citet{Strang1986}%
.

\subsubsection{Geometric Essence of $\mathbf{\psi}$}

Rayleigh's principle guarantees that the quadratic ratio%
\[
r\left(  \mathbf{Q},\mathbf{x}\right)  =\frac{\mathbf{x}^{T}\mathbf{Qx}%
}{\mathbf{x}^{T}\mathbf{x}}\text{,}%
\]
where $\mathbf{Q}$ is an $N\times N$ real symmetric matrix, is minimized by
the last and smallest eigenvector $\mathbf{x}_{N}$, with its minimal value
equal to the smallest eigenvalue $\lambda_{N}$:
\[
\lambda_{N}=\underset{0\neq\mathbf{x\in}%
\mathbb{R}
^{N}}{\min}r\left(  \mathbf{Q},\mathbf{x}\right)  \text{,}%
\]
and is maximized by the first and largest eigenvector $\mathbf{x}_{1}$, with
its maximal value equal to the largest eigenvalue $\lambda_{1}$:%
\[
\lambda_{1}=\underset{0\neq\mathbf{x\in}%
\mathbb{R}
^{N}}{\max}r\left(  \mathbf{Q},\mathbf{x}\right)  \text{,}%
\]
where the inner product term $\mathbf{x}^{T}\mathbf{x}$ in the quadratic ratio
$r\left(  \mathbf{Q},\mathbf{x}\right)  /$ $\mathbf{x}^{T}\mathbf{x}$
evaluates to a scalar. Raleigh's principle is also used to find principal
eigenvectors $\mathbf{x}_{1}$ which satisfy additional constraints such as%
\[
a_{1}x_{1}+\cdots a_{N}x_{N}=c\text{,}%
\]
for which
\begin{equation}
\lambda_{1}=\underset{a_{1}x_{1}+\cdots a_{N}x_{N}=c}{\max}r\left(
\mathbf{Q},\mathbf{x}\right)  \text{.} \label{Rayleigh's Principle}%
\end{equation}
Raleigh's principle and the theorem for convex duality jointly show that Eq.
(\ref{Vector Form Wolfe Dual}) provides an estimate of the largest eigenvector
$\mathbf{\psi}$ of a Gram matrix $\mathbf{Q}$, where $\mathbf{\psi}$ is a
principal eigenaxis of three, symmetrical\textit{\ }hyperplane partitioning
surfaces associated with the constrained quadratic form $\mathbf{\psi}%
^{T}\mathbf{Q\psi}$, such that $\mathbf{\psi}$ satisfies the constraints
$\mathbf{\psi}^{T}\mathbf{y}=0$ and $\psi_{i}\geq0$. Sections $14$, $15$,
$16$, and $17$ will examine how the strong duality relationships between the
algebraic systems of $\min\Psi\left(  \mathbf{\tau}\right)  $ and $\max
\Xi\left(  \mathbf{\psi}\right)  $ constrain the Wolfe dual normal eigenaxis
components on $\mathbf{\psi}$ to be suitably proportioned so that the total
allowed eigenenergies of $\mathbf{\tau}_{1}$ and $\mathbf{\tau}_{2}$ are
balanced in a symmetric manner. The geometric and statistical properties of
$\mathbf{\psi}$ are now summarized.

\subsection{Fundamental Properties of $\mathbf{\psi}$}

The fundamental properties of $\mathbf{\psi}$ are now defined. It will first
be argued that a Wolfe dual normal eigenlocus $\mathbf{\psi}$ satisfies a
critical minimum eigenenergy constraint that is symmetrically related to the
critical minimum eigenenergy constraint on $\mathbf{\tau}$.

\subsubsection{The Critical Minimum Eigenenergy Constraint on $\mathbf{\psi}$}

Equation (\ref{Minimum Total Eigenenergy Primal Normal Eigenlocus}) and the
theorem for convex duality indicate that $\mathbf{\psi}$ satisfies a critical
minimum eigenenergy constraint $\left\Vert \mathbf{\psi}\right\Vert _{\min
_{c}}^{2}$ that is symmetrically related to the critical minimum eigenenergy
constraint $\left\Vert \mathbf{\tau}\right\Vert _{\min_{c}}^{2}$ on
$\mathbf{\tau}$
\[
\left\Vert \mathbf{\psi}\right\Vert _{\min_{c}}^{2}\cong\left\Vert
\mathbf{\tau}\right\Vert _{\min_{c}}^{2}\text{.}%
\]
Accordingly, the functional $\mathbf{1}^{T}\mathbf{\psi}-\mathbf{\psi}%
^{T}\mathbf{Q\psi/}2$ in Eq. (\ref{Vector Form Wolfe Dual}) is maximized by a
set of eigen-balanced magnitudes
\[
\sum\nolimits_{i=1}^{l_{1}}\psi_{1_{i\ast}}=\sum\nolimits_{i=1}^{l_{2}}%
\psi_{2_{i\ast}}%
\]
of Wolfe dual normal eigenaxis components $\left\{  \psi_{1_{i\ast}%
}\overrightarrow{\mathbf{e}}_{1i\ast}\right\}  _{i=1}^{l_{1}}$ and $\left\{
\psi_{2_{i\ast}}\overrightarrow{\mathbf{e}}_{2i\ast}\right\}  _{i=1}^{l_{2}}$,
for which the quadratic form%
\[
\mathbf{\psi}^{T}\mathbf{Q\psi/}2\text{,}%
\]
reaches its smallest possible value. This indicates that the eigen-balanced
magnitudes of the Wolfe dual normal eigenaxis components on $\mathbf{\psi}$
are constrained to have the smallest possible lengths, such that the
eigen-scaled extreme training points on $\mathbf{\tau}$ posses suitable
locations for which the total allowed eigenenergies of $\mathbf{\tau}_{1}$ and
$\mathbf{\tau}_{2}$ are symmetrically balanced. Sections $12$ - $16$ will
demonstrate that a Wolfe dual normal eigenlocus $\mathbf{\psi}$ satisfies a
critical minimum eigenenergy constraint:%
\[
\max\mathbf{\psi}^{T}\mathbf{Q\psi}=\lambda_{\max\mathbf{\psi}}\left\Vert
\mathbf{\psi}\right\Vert _{\min_{c}}^{2}\text{,}%
\]
which is symmetrically related to the restriction of the primal normal
eigenlocus to the Wolfe dual eigenspace. Section $17$ will examine how the
magnitudes of the Wolfe dual and the constrained primal normal eigenaxis
components on $\mathbf{\psi}$ and $\mathbf{\tau}$ are properly proportioned,
such that the eigenloci of the constrained primal normal eigenaxis components
on $\mathbf{\tau=\tau}_{1}-\mathbf{\tau}_{2}$ possess locations which
effectively balance the eigenenergies of $\mathbf{\tau}_{1}$ and
$\mathbf{\tau}_{2}$.

It will now be demonstrated how the KKT constraint $\mathbf{\psi}%
^{T}\mathbf{y}=0$ effectively determines the statistical eigenlocus of
$\mathbf{\psi}$.

\subsection{Wolfe Dual Statistical Systems of Partitioning Hyperplanes}

The statistical representation of the primal normal eigenlocus $\mathbf{\tau}$
within the Wolfe dual eigenspace involves the KKT condition of Eq.
(\ref{KKTE2}):%
\[
\sum\limits_{i=1}^{N}\psi_{i}y_{i}=0,\ i=1,...,N\text{,}%
\]
which constrains the Wolfe dual normal eigenlocus $\mathbf{\psi}$ and the
vector of training labels $\mathbf{y}$ to be orthogonal $\mathbf{\psi}%
\perp\mathbf{y}$ so that
\[
\mathbf{\psi}^{T}\mathbf{y}=0\text{.}%
\]
The orthogonality $\mathbf{\psi}\perp\mathbf{y}$ of $\mathbf{\psi}$ and
$\mathbf{y}$ indicates that the vector of training labels $\mathbf{y}$
provides an implicit statistical directrix which determines an intrinsic
reference axis in $%
\mathbb{R}
^{N}$. Thereby, the statistical eigenlocus of $\mathbf{\psi}$ is uniquely
specified by the distance from the statistical directrix $\mathbf{y}$ to the
endpoint of $\mathbf{\psi}$. Given the strong duality relationships between
the algebraic systems of $\min\Psi\left(  \mathbf{\tau}\right)  $ and $\max
\Xi\left(  \mathbf{\psi}\right)  $, it follows that a Wolfe dual normal
eigenlocus $\mathbf{\psi}$ is an implicit, intrinsic reference axis for a
separating hyperplane $H_{0}\left(  \mathbf{x}\right)  \in%
\mathbb{R}
^{N}$ that is bounded by bilaterally symmetrical hyperplane decision borders
$H_{_{+1}}\in%
\mathbb{R}
^{N}$ and $H_{_{-1}}\left(  \mathbf{x}\right)  \in%
\mathbb{R}
^{N}$. Figure $23$ depicts a high level overview of a Wolfe dual statistical
system of partitioning hyperplanes which is implicitly described by the
orthogonality relationship $\mathbf{\psi}\perp\mathbf{y}$ between a
statistical directrix $\mathbf{y}$ and a Wolfe dual normal eigenlocus
$\mathbf{\psi}$.

\begin{center}%
\begin{center}
\includegraphics[
natheight=7.499600in,
natwidth=9.999800in,
height=3.2897in,
width=4.3777in
]%
{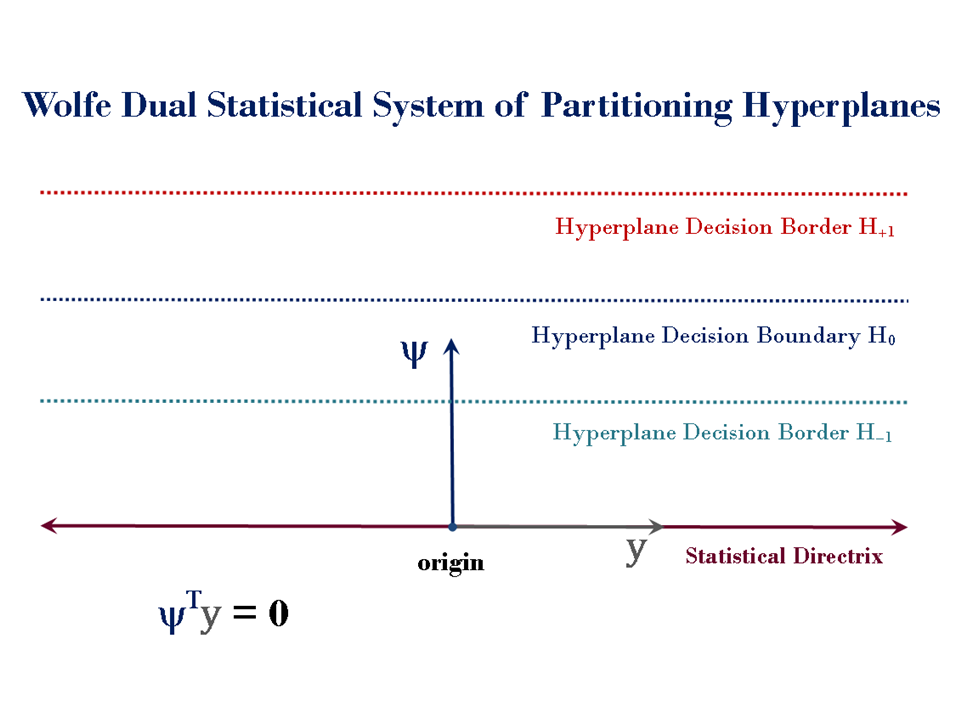}%
\end{center}

\end{center}

\begin{flushleft}
Figure $23$: Illustration of a Wolfe dual statistical eigen-coordinate system
of partitioning hyperplanes. All of the points on the hyperplane surfaces
$H_{0}$, $H_{_{+1}}$, and $H_{_{-1}}$, exclusively reference the Wolfe dual
normal eigenlocus $\mathbf{\psi}$, which satisfies all three hyperplane
surfaces in terms of its total allowed eigenenergy.
\end{flushleft}

The Wolfe dual statistical eigen-coordinate system depicted in Fig. $23$
illustrates that each of the $\psi_{i}$ terms returned by Eq.
(\ref{Vector Form Wolfe Dual}) specifies the magnitude of a normal eigenaxis
component on a Wolfe dual normal eigenlocus $\mathbf{\psi}$, where
$\mathbf{\psi}$ is exclusively referenced by all three of the hyperplane
surfaces specified by the Gram matrix $\mathbf{Q}$ associated with the
constrained quadratic form. The Wolfe dual normal eigenlocus $\mathbf{\psi}$
satisfies all three of the hyperplane surfaces in terms of its total allowed
eigenenergy $\lambda_{\max\mathbf{\psi}}\left\Vert \mathbf{\psi}\right\Vert
_{\min_{c}}^{2}$. Later on, Sections $14$ and $15$ will demonstrate that the
directions of the Wolfe dual normal eigenaxis components on $\mathbf{\psi}$
are determined by the directions of correlated extreme training vectors.

The next section of the paper will examine how the geometric and statistical
properties of strong dual normal eigenlocus transforms are sensitive to
eigenspectrums of Gram matrices. Section $11$ will define the principal
statistical state and the characteristic eigenstates of strong dual
statistical decision systems. Section $11$ will consider how low rank Gram
matrices cause principal statistical states and characteristic eigenstates to
be substantially diminished, resulting in irregular geometric architectures
which determine asymmetric linear partitions of feature spaces, resulting in
ill-formed decision regions. Section $11$ will also consider how the
eigenspectrum of Gram matrices determines the shapes of the quadratic surfaces
described by the constrained quadratic form in Eq.
(\ref{Vector Form Wolfe Dual}).

\section{Weak Dual Normal Eigenlocus Transforms}

This section will demonstrate how the geometric and statistical properties of
strong dual normal eigenlocus transforms are sensitive to eigenspectrums of
Gram matrices. It will be demonstrated that both the number and the locations
of the constrained primal normal eigenaxis components on $\mathbf{\tau}$ are
considerably affected by the rank and eigenspectrum of $\mathbf{Q}$. It will
be shown that incomplete eigenspectrums of low rank Gram matrices $\mathbf{Q}$
result in weak dual normal eigenlocus transforms that determine ill-formed
linear decision boundaries which exhibit substandard generalization
performance. It will also be shown that the geometric configurations of the
dual, symmetrical linear partitioning systems in $%
\mathbb{R}
^{N}$ and $%
\mathbb{R}
^{d}$ depicted in Figs $12$, $20$, and $23$ are largely shaped by the
eigenspectrum of the Gram matrix $\mathbf{Q}$ associated with the constrained
quadratic form in Eq. (\ref{Vector Form Wolfe Dual}).

\subsection{Eigenspectrums of Gram Matrices}

For pattern recognition applications where the training vector dimension $d$
exceeds the number $N$ of training vectors ($d>N$), the solution for the Wolfe
dual normal eigenlocus of Eq. (\ref{Vector Form Wolfe Dual}) is well-posed,
because the Gram matrix $\mathbf{Q}$ has full rank. Machine learning solutions
with eigenstructure deficiencies are generally ill-posed and ill-conditioned,
and must be constrained in some manner. Numerical techniques that constrain
matrix based solutions to mitigate eigenstructure deficiencies are called
regularization methods
\citet{Linz2003}%
. For example, the Tikhonov method of regularization addresses the problem of
small singular values
\citet{Tikhovov1977}%
. Regularization methods such as ridge regression and diagonal loading
recondition covariance or correlation matrices
\citet{Hoerl1962}%
.

Regularization components are essential numerical ingredients in machine
learning algorithms that involve inversions of data matrices
\citet{Linz1979}%
,
\citet{Groetsch1984}%
,
\citet{Wahba1987}%
,
\citet{Groetsch1993}%
,
\citet{Hansen1998}%
,
\citet{Engl2000}%
,
\citet{Linz2003}%
. The machine learning algorithm for strong dual normal eigenlocus transforms
involves an inversion of the Gram matrix $\mathbf{Q}$ in Eq.
(\ref{Vector Form Wolfe Dual}), so some type of regularization is required for
low rank Gram matrices
\citet{Reeves2009}%
,
\citet{Reeves2011}%
.

\subsection{Incomplete Eigenspectrums of Low Rank Gram Matrices}

The solution for the Wolfe dual normal eigenlocus of Eq.
(\ref{Vector Form Wolfe Dual}) is ill-posed for low rank Gram matrices
$\mathbf{Q}$, because $\mathbf{Q}$ is singular and noninvertable. In general,
learning machines that learn $N$ parameters with $d$ eigenfunctions have
insufficient learning capacity whenever $N>d$. For low rank Gram matrices
$\mathbf{Q}$, where the number $N$ of training vectors exceeds the dimension
$d$ of the training vectors, it has been shown that a Wolfe dual normal
eigenlocus $\mathbf{\psi\in%
\mathbb{R}
}^{N}$ is spanned by an incomplete set of $d$ eigenvectors
\citet{Reeves2011}%
. It will now be demonstrated that low rank Gram matrices $\mathbf{Q}$ cause
the principal statistical state and the characteristic eigenstates of strong
dual decision systems to be substantially diminished, resulting in irregular
geometric architectures and ill-formed decision regions. It is said that
diminished principal statistical states and characteristic eigenstates of
strong dual decision systems produce \emph{weak dual normal eigenlocus
transforms}. Principal statistical states and characteristic eigenstates of
strong dual decision systems are defined next.

\subsection{Principal Statistical States and Characteristic Eigenstates of
Strong Dual Decision Systems}

Denote the principal characteristic root, i.e., the principal eigenvalue,
associated with the Wolfe dual normal eigenlocus $\mathbf{\psi}$ of Eq.
(\ref{Vector Form Wolfe Dual}) by $\lambda_{\max_{\mathbf{\psi}}}$. Define the
principal characteristic root $\lambda_{\max_{\mathbf{\psi}}}$ to be the
principal statistical state of a strong dual decision system. Define the
characteristic eigenstates%
\[
\Psi_{\mathbf{\tau}}\left(
\mathbb{R}
^{d}\right)  =\left\{  \left\{  \Psi_{\mathbf{\tau}_{1}}\left(
\mathbb{R}
^{d}\right)  \right\}  _{i=1}^{l_{1}},\left\{  \Psi_{\mathbf{\tau}_{2}}\left(
%
\mathbb{R}
^{d}\right)  \right\}  _{i=1}^{l_{2}}\right\}  \text{,}%
\]
of a strong dual decision system to be the eigen-scaled extreme training
points on $\mathbf{\tau}_{1}$%
\begin{equation}
\left\{  \Psi_{\mathbf{\tau}_{1}}\left(
\mathbb{R}
^{d}\right)  \right\}  _{i=1}^{l_{1}}\triangleq\left\{  \psi_{1i\ast
}\mathbf{x}_{1_{i\ast}}|\mathbf{x}_{1_{i\ast}}\sim p_{\mathbf{x}_{1_{i}%
}|\boldsymbol{X}_{1}}\left(  \mathbf{x}_{1_{i}}|\boldsymbol{X}_{1}\right)
\right\}  _{i=1}^{l_{1}}\text{,} \label{Characteristic Eigenstates Class One}%
\end{equation}
and on $\mathbf{\tau}_{2}$%
\begin{equation}
\left\{  \Psi_{\mathbf{\tau}_{2}}\left(
\mathbb{R}
^{d}\right)  \right\}  _{i=1}^{l_{2}}\triangleq\left\{  \psi_{2i\ast
}\mathbf{x}_{2_{i\ast}}|\mathbf{x}_{2_{i\ast}}\sim p_{\mathbf{x}_{2_{i}%
}|\boldsymbol{X}_{2}}\left(  \mathbf{x}_{2_{i}}|\boldsymbol{X}_{2}\right)
\right\}  _{i=1}^{l_{2}}\text{.} \label{Characteristic Eigenstates Class Two}%
\end{equation}
Later on, Section $18$ will demonstrate that the characteristic eigenstates in
Eqs (\ref{Characteristic Eigenstates Class One}) and
(\ref{Characteristic Eigenstates Class Two}) encode the likelihoods of finding
extreme data points in particular regions of $%
\mathbb{R}
^{d}$.

The principal statistical state $\lambda_{\max_{\mathbf{\psi}}}$ of a strong
dual decision system is extensively diminished for low rank Gram matrices
$\mathbf{Q}$. In particular, low rank Gram matrices $\mathbf{Q}$ provide
insufficient estimates of principal statistical states $\lambda_{\max
_{\mathbf{\psi}}}$, resulting in incomplete and/or defective sets of
characteristic eigenstates $\left\{  \left\{  \Psi_{\mathbf{\tau}_{1}}\left(
\mathbb{R}
^{d}\right)  \right\}  _{i=1}^{l_{1}},\left\{  \Psi_{\mathbf{\tau}_{2}}\left(
%
\mathbb{R}
^{d}\right)  \right\}  _{i=1}^{l_{2}}\right\}  $ of strong dual decision
systems. Thereby, low rank Gram matrices $\mathbf{Q}$ of linear kernel SVMs
generate weak dual normal eigenlocus transforms that produce ill-formed linear
decision boundaries which exhibit substandard generalization performance for
overlapping data distributions. For non-overlapping data distributions, low
rank Gram matrices $\mathbf{Q}$ of linear kernel SVMs determine ill-formed
linear decision boundaries which exhibit optimal generalization performance at
the expense of unnecessary sets of characteristic eigenstates.

Both the number and the locations of the constrained primal normal eigenaxis
components on $\mathbf{\tau}$ are considerably affected by the rank and
eigenspectrum of $\mathbf{Q}$. For example, given non-overlapping data
distributions and low rank Gram matrices, \emph{all} of the training data are
transformed into normal eigenaxis components
\citet{Reeves2009}%
. In general, perturbations of the principal statistical states and
characteristic eigenstates of strong dual decision systems produce irregular
geometric architectures which determine asymmetric linear partitions of
feature spaces, resulting in ill-formed decision regions. As an example, Fig.
$24$ depicts an asymmetric partitioning produced by a weak dual normal
eigenlocus transform of training data described by the covariance matrix:%
\[
\mathbf{\Sigma}_{1}=\mathbf{\Sigma}_{2}=%
\begin{pmatrix}
7 & 0\\
0 & 0.5
\end{pmatrix}
\text{,}%
\]
and the mean vectors $\mathbf{\mu}_{1}=%
\begin{pmatrix}
3, & 7
\end{pmatrix}
^{T}$ and $\mathbf{\mu}_{2}=%
\begin{pmatrix}
2, & 6
\end{pmatrix}
^{T}$, for which the linear decision boundary and its borders are badly skewed
and poorly positioned.

\begin{center}%
\begin{center}
\includegraphics[
natheight=7.749600in,
natwidth=14.958700in,
height=2.2866in,
width=4.3872in
]%
{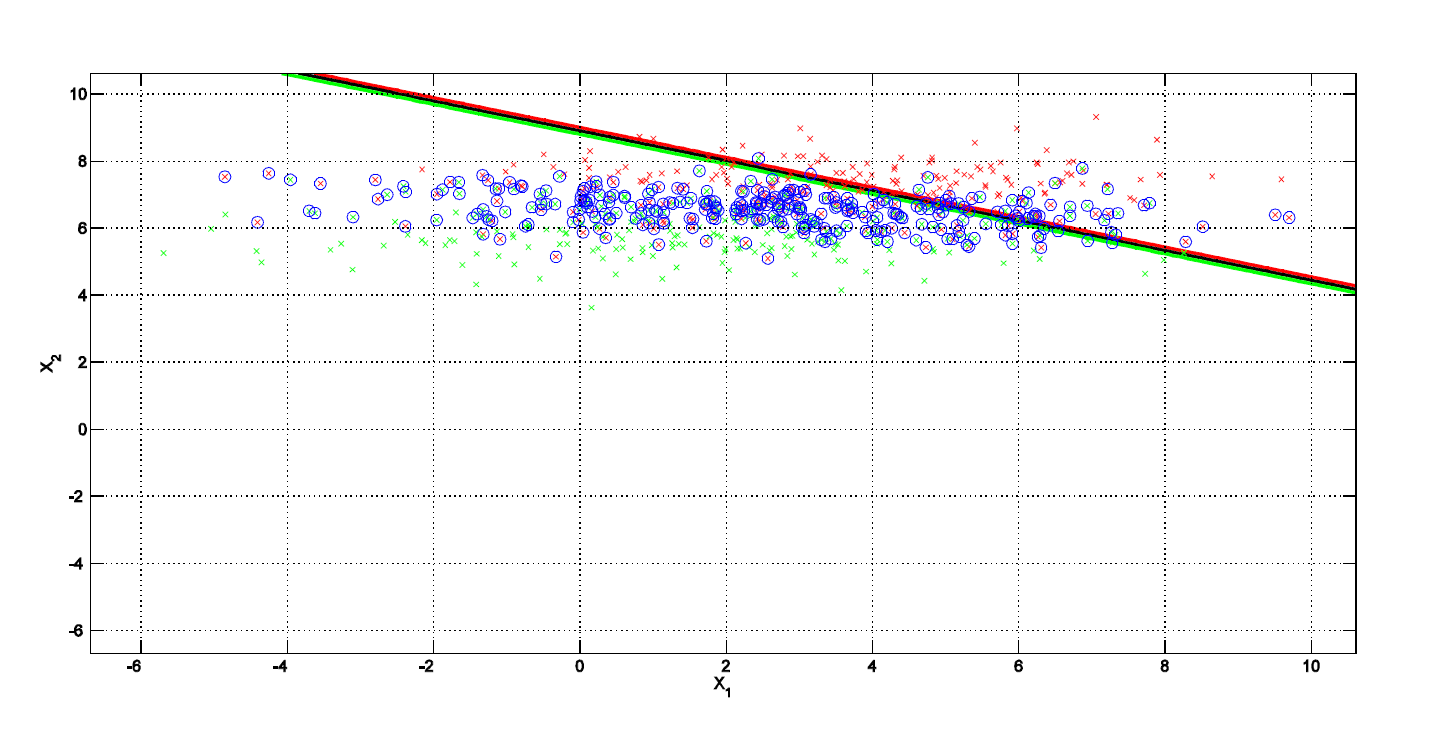}%
\end{center}

\end{center}

\begin{flushleft}
Figure $24$: Illustration that weak dual normal eigenlocus transforms based on
insufficient eigenstates result in asymmetric linear partitions and poorly
positioned decision regions.
\end{flushleft}

Only $53\%$ of the training data are transformed into normal eigenaxis
components, whereas properly regularized linear SVM transforms $\approx86\%$
of training data to learn this optimal partitioning
\citet{Reeves2009}%
. For this example, low rank Gram matrices cause asymmetrical distributions of
principal eigenenergies over insufficient sets of eigen-scaled extreme data
points. On the other hand, given non-overlapping data distributions and low
rank Gram matrices, all of the training data are transformed into normal
eigenaxis components. In both instances, low rank Gram matrices generate weak
dual normal eigenlocus transforms. Additional examples of ill-formed linear
decision regions resulting from weak dual normal eigenlocus transforms can be
found in
\citet{Reeves2009}
and
\citet{Reeves2011}%
.

Given a previous analysis of $\mathbf{\psi}$ for low rank Gram matrices
$\mathbf{Q}$, which can be found in
\citet{Reeves2011}%
, within the context of strong dual normal eigenlocus transforms, it is
concluded that incomplete sets of eigenvectors generate incomplete
eigenspectrums and insufficient eigenstates for strong dual normal eigenlocus
transforms. Overall, it is concluded that Wolfe dual normal eigenlocus
estimates of $\mathbf{\psi}$ that are based on incomplete eigenspectrums of
low rank Gramian matrices $\mathbf{Q}$ produce weak dual normal eigenlocus
estimates of $\mathbf{\tau}$ that are based on perturbed principal statistical
states and insufficient eigenstates. Generally speaking, low rank Gramian
matrices $\mathbf{Q}$ provide insufficient estimates of principal statistical
states $\lambda_{\max_{\mathbf{\psi}}}$, resulting in asymmetrical
distributions of the eigenenergies of $\mathbf{\psi}$ and $\mathbf{\tau}$ over
the eigen-scaled extreme data points on $\mathbf{\tau}$. Figure $25$ depicts
the geometric and statistical connections between the\ joint statistical
contents and the symmetrical geometric configurations of $\mathbf{\psi}$ and
$\mathbf{\tau}$.

\begin{center}%
\begin{center}
\includegraphics[
natheight=7.499600in,
natwidth=9.999800in,
height=3.2897in,
width=4.3777in
]%
{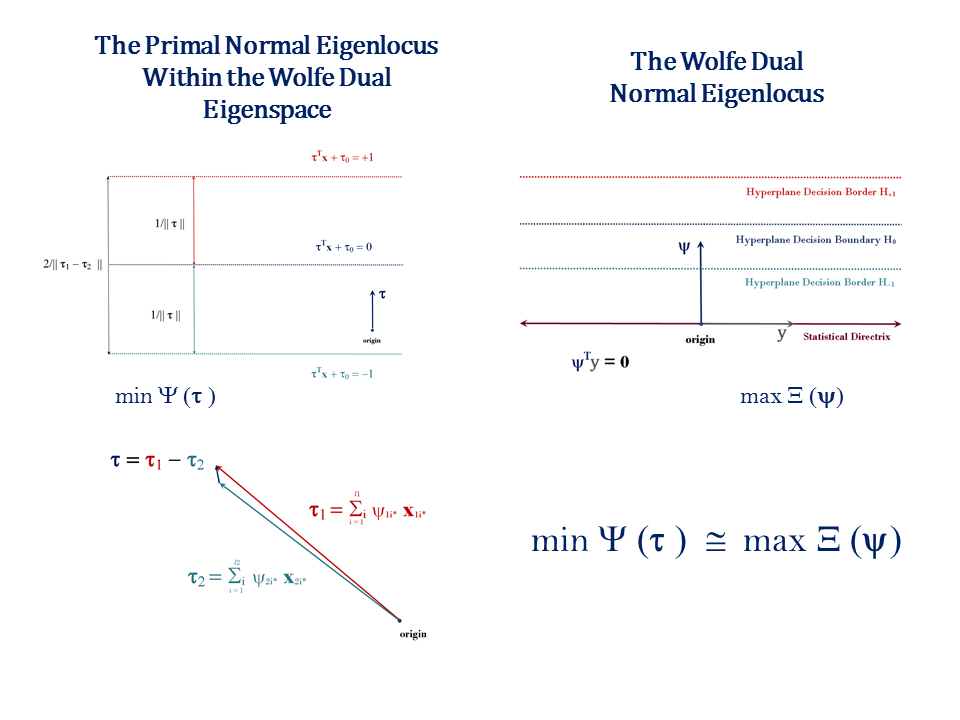}%
\end{center}

\end{center}

\begin{flushleft}
FIGURE $25$. Illustration of the symmetrical geometric relationships between
the constrained primal normal eigenaxis components on $\mathbf{\tau}$ and the
Wolfe dual normal eigenaxis components on $\mathbf{\psi}$. The constrained
primal normal eigenlocus $\mathbf{\tau=\tau}_{1}-\mathbf{\tau}_{2}$ possesses
a geometric configuration which is determined by the statistical contents of
symmetrical normal eigenaxis components on its Wolfe dual $\mathbf{\psi}$.
\end{flushleft}

\subsection{Generating Sufficient Eigenspectrums for Low Rank Gram Matrices}

Take any collection of $N$ training vectors of dimension $d$, for which $d<N$
and $\mathbf{Q}$ has low rank. It has been shown that the regularized form of
Eq. (\ref{Vector Form Wolfe Dual}), for which $\epsilon\ll1$ and
$\mathbf{Q}\triangleq\epsilon\mathbf{I}+\widetilde{\mathbf{X}}%
\widetilde{\mathbf{X}}^{T}$, ensures that $\mathbf{Q}$ has full rank, and
thereby ensures that $\mathbf{Q}$ has a complete eigenspectrum and eigenvector
set. Thus, the auxiliary functional $C/2\sum\nolimits_{i=1}^{N}\xi_{i}^{2}$ in
Eq. (\ref{Lagrangian Normal Eigenlocus}) ensures that the matrix-based
estimate of the hyperplane surfaces determined by Eq.
(\ref{Vector Form Wolfe Dual}) is based on a full rank Gram matrix
$\mathbf{Q}$, so that the statistical contents of $\mathbf{\psi}$ are based on
a complete eigenspectrum and eigenvector set. The regularization constant $C$
in Eq. (\ref{Wolfe Dual Normal Eigenlocus}) is related to the regularization
parameter $\epsilon$\ by $\frac{1}{C}$
\citet{Reeves2011}%
.

Therefore, given any collection of $N$ training vectors of dimension $d$, for
which $N<d$, the Gram matrix $\mathbf{Q}$ in Eq. (\ref{Vector Form Wolfe Dual}%
) has full rank, and the regularization parameters $\xi_{i}$ in the primal
normal eigenlocus of Eq. (\ref{Primal Normal Eigenlocus}) and all of its
derivatives are set equal to zero: $\xi_{i}=0$.

The eigenspectrum of $\mathbf{Q}$ plays a fundamental role in describing the
hyperplane surfaces which are implicitly delineated by $\mathbf{\psi}$. The
next section will demonstrate that the eigenspectrum of $\mathbf{Q}$
determines the shapes of the quadratic surfaces described by the constrained
quadratic form in Eq. (\ref{Vector Form Wolfe Dual}).

\subsection{Eigenspectrum Shaping of Quadratic Surfaces}

Take the standard equation of a quadratic form: $\mathbf{x}^{T}\mathbf{Qx}=1$.
Write $\mathbf{x}$ in terms of an orthogonal basis of unit eigenvectors
$\left\{  \mathbf{v}_{1},\ldots,\mathbf{v}_{N}\right\}  $ so that
$\mathbf{x=}\sum\nolimits_{i=1}^{N}x_{i}\mathbf{v}_{i}$. Substitution of this
expression into $\mathbf{x}^{T}\mathbf{Qx}$%
\[
\mathbf{x}^{T}\mathbf{Qx}=\left(  \sum\nolimits_{i=1}^{N}x_{i}\mathbf{v}%
_{i}\right)  ^{T}\mathbf{Q}\left(  \sum\nolimits_{j=1}^{N}x_{j}\mathbf{v}%
_{j}\right)
\]
produces a simple coordinate form expression of a second-order surface:%
\begin{equation}
\lambda_{1}x_{1}^{2}+\lambda_{2}x_{2}^{2}+\ldots+\lambda_{N}x_{N}%
^{2}=1\text{,} \label{Eigenvalue Coordinate Form Second Order Loci}%
\end{equation}
solely in terms of the eigenvalues $\lambda_{N}\leq$ $\lambda_{N-1}\ldots
\leq\lambda_{1}$ of the matrix $\mathbf{Q}$
\citet{Hewson2009}%
. Equation (\ref{Eigenvalue Coordinate Form Second Order Loci}) reveals that
\emph{the geometric shape of a quadratic surface is completely determined by
the eigenvalues of the matrix associated with a quadratic form}. This general
property of quadratic forms will lead to far reaching consequences for the
strong dual normal eigenlocus method for linear decision boundary estimates
and the strong dual principal eigenlocus method for second-order decision
boundary estimates.

It will now be argued that the inner product statistics of a training data
collection effectively determine the geometric shapes of the quadratic
surfaces described by the constrained quadratic form in Eq.
(\ref{Vector Form Wolfe Dual}).

Consider a Gram or kernel matrix $\mathbf{Q}$ associated with the constrained
quadratic form in Eq. (\ref{Vector Form Wolfe Dual}). Denote the elements of
the Gram or kernel matrix $\mathbf{Q}$ by $\varphi\left(  \mathbf{x}%
_{i},\mathbf{x}_{j}\right)  $, where $\varphi\left(  \mathbf{x}_{i}%
,\mathbf{x}_{j}\right)  $ denotes an inner product relationship between the
training vectors $\mathbf{x}_{i}$ and $\mathbf{x}_{j}$. The Cayley-Hamilton
theorem provides the result that the eigenvalues $\left\{  \lambda
_{i}\right\}  _{i=1}^{N}\in\Re$ of $\mathbf{Q}$ satisfy the characteristic
equation%
\[
\det\left(  \mathbf{Q-\lambda I}\right)  =0\text{,}%
\]
which is a polynomial of degree $N$. The roots $p\left(  \lambda\right)  =0$
of the characteristic polynomial $p\left(  \lambda\right)  $ of $\mathbf{Q}$:%
\[
\det\left(
\begin{bmatrix}
\varphi\left(  \mathbf{x}_{1},\mathbf{x}_{1}\right)  -\lambda_{1} & \cdots &
\varphi\left(  \mathbf{x}_{1},\mathbf{x}_{N}\right) \\
\varphi\left(  \mathbf{x}_{2},\mathbf{x}_{1}\right)  & \cdots & \varphi\left(
\mathbf{x}_{2},\mathbf{x}_{N}\right) \\
\vdots & \ddots & \vdots\\
\varphi\left(  \mathbf{x}_{N},\mathbf{x}_{1}\right)  & \cdots & \varphi\left(
\mathbf{x}_{N},\mathbf{x}_{N}\right)  -\lambda_{N}%
\end{bmatrix}
\right)  =0\text{,}%
\]
are also the eigenvalues $\lambda_{N}\leq$ $\lambda_{N-1}\leq\ldots\leq
\lambda_{1}$ of $\mathbf{Q}$
\citet{Lathi1998}%
. Therefore, given that $(1)$ the roots of a characteristic polynomial
$p\left(  \lambda\right)  $ vary continuously with its coefficients, and that
$(2)$ the coefficients of $p\left(  \lambda\right)  $ can be expressed in
terms of sums of principal minors
\citet{Meyer2000}%
, it follows that the coefficients of $p\left(  \lambda\right)  $, and
therefore the eigenvalues of $\mathbf{Q}$, vary continuously with the inner
product elements $\varphi\left(  \mathbf{x}_{i},\mathbf{x}_{j}\right)  $ of
$\mathbf{Q}$. It is concluded that the eigenvalues $\lambda_{N}\leq$
$\lambda_{N-1}\leq\ldots\leq\lambda_{1}$ of a Gram or kernel matrix
$\mathbf{Q}$ are actually determined by its inner product elements
$\varphi\left(  \mathbf{x}_{i},\mathbf{x}_{j}\right)  $.

Given Eq. (\ref{Eigenvalue Coordinate Form Second Order Loci}) and the
continuous functional relationship between the inner product elements and the
eigenvalues of a Gram or kernel matrix, it follows that the geometric shapes
of the three, symmetrical quadratic partitioning surfaces described by Eqs
(\ref{Wolfe Dual Normal Eigenlocus}) or (\ref{Vector Form Wolfe Dual}) are an
inherent\textit{\ }function of inner product statistics $\varphi\left(
\mathbf{x}_{i},\mathbf{x}_{j}\right)  $ between training vectors.

It is concluded that the algebraic form of the inner product statistics
encoded within Gram or kernel matrices effectively determines the shapes of
the three, symmetrical quadratic partitioning surfaces described by Eqs
(\ref{Wolfe Dual Normal Eigenlocus}) or (\ref{Vector Form Wolfe Dual}). For
strong dual normal eigenlocus transforms, given that coordinate form
expressions of hyperplane surfaces involve first-degree vector components
$x_{i}$, it is claimed that the algebraic form of an inner product statistic
must encode first-degree vector components for effective descriptions of
hyperplane surfaces.

Alternatively, given that coordinate form expressions for nonlinear
second-order surfaces involve first $x_{i}$ and second-degree, i.e.,
$x_{i}^{2}$ or $x_{i}x_{j}$, vector components, it is claimed that the
algebraic form of an inner product statistic must encode both first and
second-degree vector components for effective descriptions of quadratic surfaces.

Given the chain of arguments outlined above, it is concluded that the
algebraic form of the inner product statistics encoded within a Gram or kernel
matrix determine the geometric shapes of the three, symmetrical quadratic
partitioning surfaces described by the constrained quadratic form in Eq.
(\ref{Vector Form Wolfe Dual}). It follows that, given a suitable algebraic
form for an inner product statistic, the eigenvalues of a Gram or kernel
matrix associated with the constrained quadratic form in Eq.
(\ref{Vector Form Wolfe Dual}) describe either\ $N$-dimensional circles,
ellipses, hyperbolae, parabolas, or lines. Section $12$ will argue that a Gram
matrix $\mathbf{Q}$ associated with the constrained quadratic form in Eq.
(\ref{Vector Form Wolfe Dual}), whose inner product elements $\varphi\left(
\mathbf{x}_{i},\mathbf{x}_{j}\right)  $ have the algebraic form of
$\mathbf{x}_{i}^{T}\mathbf{x}_{j}$, encodes descriptive statistics for three,
symmetrical hyperplane partitioning surfaces.

It will now be demonstrated that kernel matrices $\mathbf{Q}$ associated with
the constrained quadratic form in Eq. (\ref{Vector Form Wolfe Dual}), whose
inner product elements $\varphi\left(  \mathbf{x}_{i},\mathbf{x}_{j}\right)  $
have the algebraic form of $\left(  \mathbf{x}_{i}^{T}\mathbf{x}_{j}+1\right)
^{2}$, encode descriptive statistics for three, symmetrical, $N $-dimensional
partitioning circles, ellipses, hyperbolae, or parabolas, which are correlated
to three, symmetrical, $d$-dimensional partitioning circles, ellipses,
hyperbolae, or parabolas. The claim is demonstrated by applying second-order
polynomial kernel SVMs to two sets of Gaussian data.

Second-order polynomial kernel SVMs are first applied to the overlapping
Gaussian data sets of classification example two. Figure $26$ illustrates a
second-order decision boundary that is determined by three, symmetrical,
$2$-dimensional partitioning parabolas, all of which are delineated by the
constrained discriminant function of a strong dual principal eigenlocus
transform. All three parabolas are positioned in symmetrical locations that
delineate geometric regions of data distribution overlap. Moreover, the strong
dual principal decision system achieves the Bayes' error rate of $25\% $ for
this classification problem. All of the points that lie on each $2
$-dimensional parabola exclusively reference a common principal eigenaxis. The
principal eigenaxis estimate, which is specified by the primal and Wolfe dual
eigenlocus equations of a \emph{strong dual principal eigenlocus}, involves
solving an inequality constrained optimization problem that is similar in
nature to Eq. (\ref{Primal Normal Eigenlocus}).

\begin{center}%
\begin{center}
\includegraphics[
natheight=7.478900in,
natwidth=14.708700in,
height=2.2866in,
width=4.4693in
]%
{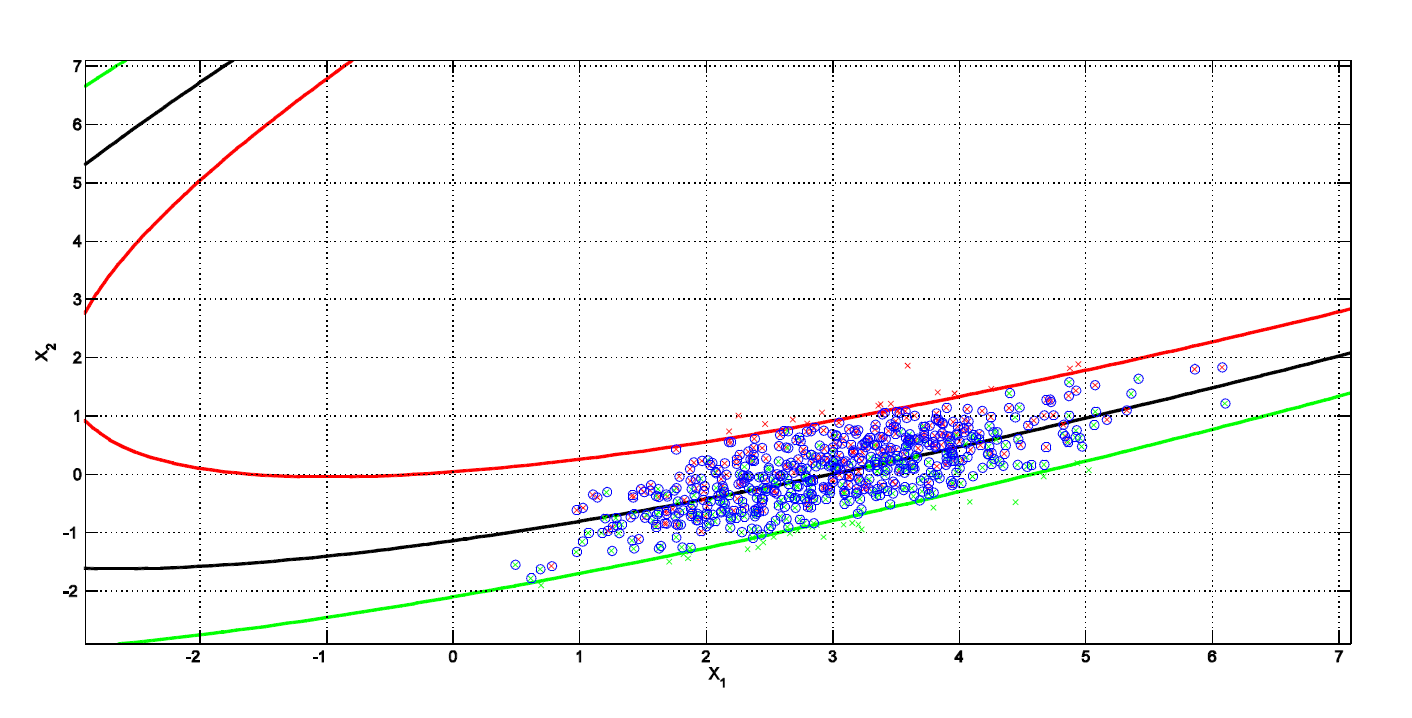}%
\end{center}

\end{center}

\begin{flushleft}
Figure $26$: Illustration that a second-order polynomial kernel matrix encodes
descriptive statistics for three, symmetrically positioned, $N$-dimensional
partitioning parabolas. Thereby, polynomial kernel SVM\ estimates a principal
eigenaxis which is exclusively referenced by all of the points on each
symmetrically positioned $2$-dimensional parabola, such that all three
$2$-dimensional parabolas jointly delineate a symmetrical partitioning of
overlapping Gaussian data.
\end{flushleft}

Next, second-order polynomial kernel SVMs are applied to the completely
overlapping Gaussian data sets considered in Section $9$. Figure $27$
illustrates a second-order decision boundary that is determined by three,
symmetrical, $2$-dimensional partitioning hyperbolae, all of which are
delineated by the constrained discriminant function of a strong dual principal
eigenlocus transform. All three hyperbolae are positioned in symmetrical
locations that delineate geometric regions of data distribution overlap. All
of the points that lie on each $2$-dimensional hyperbola exclusively reference
a common principal eigenaxis. The strong dual principal eigenlocus transform
is specified by the primal and Wolfe dual eigenlocus equations of a strong
dual principal eigenlocus. The strong dual principal decision system achieves
the Bayes' error rate of $50\%$ for this classification problem.

\begin{center}%
\begin{center}
\includegraphics[
natheight=7.771200in,
natwidth=14.854000in,
height=2.2866in,
width=4.3448in
]%
{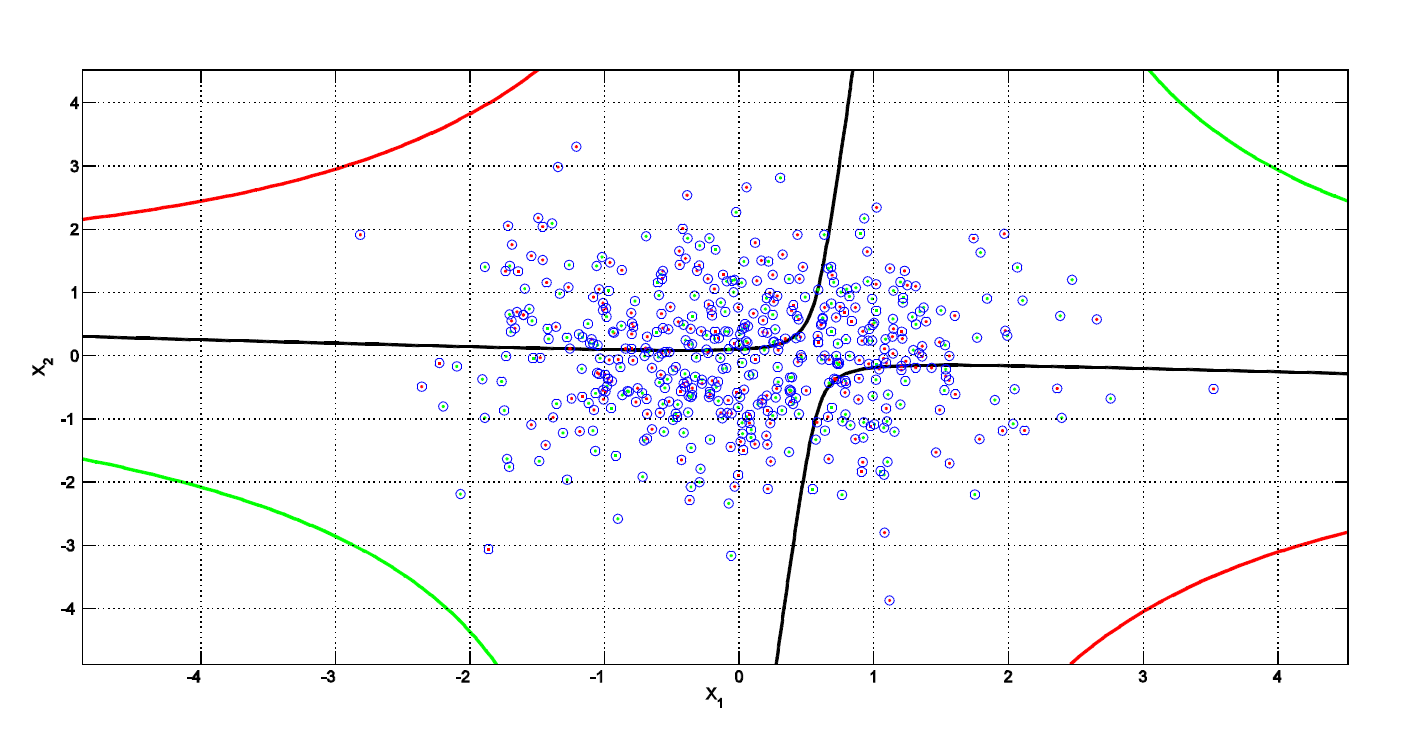}%
\end{center}

\end{center}

\begin{flushleft}
Figure $27$: Illustration that a second-order polynomial kernel matrix encodes
descriptive statistics for three, symmetrically positioned, $N$-dimensional
partitioning hyperbolae. Thereby, polynomial kernel SVM\ estimates a principal
eigenaxis, which is exclusively referenced by all of the points on each
symmetrically positioned $2$-dimensional hyperbola, such that all three
$2$-dimensional hyperbolae jointly delineate a symmetrical partitioning of
completely overlapping Gaussian data.
\end{flushleft}

\subsection{Descriptive Statistics Encoded Within $\mathbf{\psi}$}

Consider the Gram matrix $\mathbf{Q}$ associated with the constrained
quadratic form in Eq. (\ref{Vector Form Wolfe Dual}). The eigenvectors
$\mathbf{\upsilon}$ of $\mathbf{Q}$%
\[
\mathbf{Q\upsilon}_{i}=\lambda_{i}\mathbf{\upsilon}_{i},\ i=1,...,N\text{,}%
\]
correspond to directions left unchanged by the action of the Gram matrix
$\mathbf{Q}$
\citet{Meyer2000}%
. This implies that the directions of the Wolfe dual normal eigenaxis
components $\psi_{i\ast}\overrightarrow{\mathbf{e}}_{i\ast}\ $on
$\mathbf{\psi}$ are left unchanged by the inner product elements
$\mathbf{x}_{i}^{T}\mathbf{x}_{j}$ of $\mathbf{Q}$.

Suppose that $\mathbf{Q}$ contains descriptive statistics $\mathbf{\Sigma
}\left(  \mathbf{x}_{i},\mathbf{x}_{j}\right)  $ for a hyperplane decision
surface $h_{D_{0}}\left(  \mathbf{x}\right)  $ that is bounded by bilaterally
symmetrical hyperplane decision borders $h_{D_{h_{+1}}}\left(  \mathbf{x}%
\right)  $ and $h_{D_{h_{-1}}}\left(  \mathbf{x}\right)  $. Consider
transforming the statistics $\mathbf{\Sigma}\left(  \mathbf{x}_{i}%
,\mathbf{x}_{j}\right)  $ embedded within $\mathbf{Q}$:%
\[
\mathbf{Q\psi}=\lambda\mathbf{_{\max\mathbf{\psi}}\psi}\text{,}%
\]
into a data-driven, non-orthogonal set of Wolfe dual normal eigenaxis
components%
\[
\mathbf{Q\sum\nolimits_{i=1}^{l}}\psi_{i\ast}\overrightarrow{\mathbf{e}%
}_{i\ast}=\lambda\mathbf{_{\max\mathbf{\psi}}}\sum\nolimits_{i=1}^{l}%
\psi_{i\ast}\overrightarrow{\mathbf{e}}_{i\ast}\text{,}%
\]
formed by $l$ eigen-scaled $\psi_{i\ast}$ non-orthogonal unit vectors
$\left\{  \overrightarrow{\mathbf{e}}_{1\ast},\ldots
,\overrightarrow{\mathbf{e}}_{l\ast}\right\}  $, where the eigenlocus of each
Wolfe dual normal eigenaxis component $\psi_{i\ast}\overrightarrow{\mathbf{e}%
}_{i\ast}\ $is determined by the direction and eigen-balanced magnitude of a
correlated extreme vector $\mathbf{x}_{i\ast}$.

Given the above assumptions, it will shortly be demonstrated how a Wolfe dual
normal eigenlocus $\mathbf{\psi}$ provides an estimate of a distinctive normal
eigenaxis in $%
\mathbb{R}
^{N}$ that shapes and complements the constrained primal estimate
$\mathbf{\tau}$ of a similar normal eigenaxis in $%
\mathbb{R}
^{d}$. An expression will be developed for a Wolfe dual normal eigenlocus
$\mathbf{\psi}$ that contains descriptive statistics for three, symmetrical
hyperplane partitioning surfaces in $%
\mathbb{R}
^{N}$. The same expression describes point and coordinate relationships
between the eigen-scaled extreme points on $\mathbf{\tau}$ and the Wolfe dual
normal eigenaxis components on $\mathbf{\psi}$. The expression will be used to
identify uniform geometric and statistical properties which are jointly
exhibited by correlated normal eigenaxis components on $\mathbf{\psi}$ and
$\mathbf{\tau}$.

The next section will motivate the examination of point and coordinate
relationships between the constrained primal and the Wolfe dual normal
eigenaxis components. Section $12$ will define pointwise covariance statistics
for individual training points, and will demonstrate how pointwise covariance
statistics can be used to find extreme data points which possess large
pointwise covariances. Section $12$ will also consider the total allowed
eigenenergies of a strong dual normal eigenlocus.

\section{Point and Coordinate Relationships Between Constrained Primal and
Wolfe Dual Normal Eigenaxis Components}

A geometric object is assumed to be independent of the coordinate system that
is used to describe it
\citet{Hewson2009}%
. On the contrary, this paper considers major intrinsic coordinate axes of
conic sections and quadratic surfaces to be an inherent part of second-order
geometric loci. An upcoming paper will substantiate the claim that the locus
of a principal eigenaxis is a distinctive, invariant, and hardwired geometric
property of second-order curves and surfaces, which effectively determines the
points on a second-order locus.

This paper has rigorously demonstrated that the locus of a normal eigenaxis is
a distinctive, invariant, and hardwired geometric property of linear curves
and surfaces, which effectively determines the points on a linear locus. It
has been argued that the locations of the constrained primal normal eigenaxis
components on the\ constrained primal normal eigenlocus $\mathbf{\tau}$ of Eq.
(\ref{Primal Normal Eigenlocus}) provide estimates for the constrained
eigen-coordinate locations of a normal eigenaxis $\mathbf{v}$ of a linear
decision boundary. It has been demonstrated that the\ constrained normal
eigenlocus $\mathbf{\tau}$ of Eq. (\ref{Primal Normal Eigenlocus}) delineates
a linear decision boundary that is bounded by bilaterally symmetrical linear
decision borders. It has also been demonstrated that the statistical
eigen-coordinate system of Eqs (\ref{Discriminant Function}),
(\ref{Decision Boundary}), (\ref{Decision Border One}), and
(\ref{Decision Border Two}), depicted in Fig. $20$, delineates bipartite,
symmetric regions of large covariance located between two data distributions
in $%
\mathbb{R}
^{d}$,\textbf{\ }which describe regions of data distribution overlap for
overlapping distributions and bipartite symmetric partitions between the tail
regions of non-overlapping data distributions.

Thus far, this paper has argued that the scaling parameters $\psi_{i\ast}$
returned by Eq. (\ref{Vector Form Wolfe Dual}) determine symmetrical lengths
of Wolfe dual normal eigenaxis components $\psi_{i\ast}%
\mathbf{\overrightarrow{\mathbf{e}}}_{i\ast}$ on a Wolfe dual normal
eigenlocus $\mathbf{\psi}$. Additional insights can be obtained by
investigating the algebraic, geometric, and statistical\ nature of the point
and coordinate relationships between the eigen-scaled extreme points on
$\mathbf{\tau}$ and the Wolfe dual normal eigenaxis components on
$\mathbf{\psi}$. In order to obtain these insights, it will be necessary to
develop algebraic expressions which describe algebraic, geometric, and
statistical relationships between the Wolfe dual normal eigenaxis components
and the eigen-scaled extreme training points. The expressions must also
describe point and coordinate relationships between the extreme training points.

Sections $13$ - $15$ will develop an algebraic expression for $\mathbf{\psi}$
that describes the point and coordinate relationships outlined above. The
expression will be used to examine how each Wolfe dual normal eigenaxis
component is formed by an eigen-balanced set of eigen-scaled scalar
projections of extreme training vectors, along the common axis of an extreme
training vector which is correlated with the Wolfe dual normal eigenaxis
component. Thereby, the expression will be used to identify uniform
geometrical and statistical properties which are exhibited by Wolfe dual
normal eigenaxis components $\psi_{i\ast}\mathbf{\overrightarrow{\mathbf{e}}%
}_{i}$ on $\mathbf{\psi}$ and correlated, constrained primal normal eigenaxis
components $\psi_{i\ast}\mathbf{x}_{i\ast}$ on $\mathbf{\tau}$. It will be
demonstrated that each Wolfe dual normal eigenaxis component $\psi_{i\ast
}\mathbf{\overrightarrow{\mathbf{e}}}_{i}$ $\ $on $\mathbf{\psi}\in%
\mathbb{R}
^{N}$ has an eigenlocus which stores an eigen-balanced, pointwise covariance
estimate\textit{\ }$\psi_{i\ast}$ of a correlated extreme data
point\textit{\ }$\mathbf{x}_{i\ast}\in%
\mathbb{R}
^{d}$, such that each eigen-balanced, pointwise covariance estimate
$\psi_{i\ast}$ encodes an eigen-balanced first and second order statistical
moment about the locus of an extreme data point $\mathbf{x}_{i\ast}$, which
determines a suitable length $\psi_{i\ast}\left\Vert \mathbf{x}_{i\ast
}\right\Vert $ for a constrained primal normal eigenaxis component
$\psi_{i\ast}\mathbf{x}_{i\ast}$ on $\mathbf{\tau}\in%
\mathbb{R}
^{d}$.

The notion of a first and second order statistical moment about the locus of a
data point will be defined next, along with the notion of a pointwise
covariance estimate, both of which are shown to provide a maximum\textit{\ }%
covariance estimate in a principal location.

\subsection{Joint Statistical Underpinnings of $\mathbf{\psi}$ and
$\mathbf{\tau}$}

An algebraic expression has been obtained for $\mathbf{\psi}$
\[
\mathbf{\psi}=\mathbf{Q}^{-1}\left[  \mathbf{Q}-\frac{\mathbf{yy}^{T}%
}{\mathbf{y}^{T}\mathbf{Q}^{-1}\mathbf{y}}\right]  \mathbf{Q}^{-1}\left(
\mathbf{1}+\mathbf{\lambda}\right)  \text{,}%
\]
that relates the Wolfe dual normal eigenaxis components to inner product
statistics between the training vectors stored within $\mathbf{Q}$
\citet{Reeves2009}%
,
\citet{Reeves2011}%
. The above expression for $\mathbf{\psi}$ clearly illustrates that the Wolfe
dual normal eigenlocus solution of Eq. (\ref{Vector Form Wolfe Dual}) is
ill-posed for singular and noninvertable $\mathbf{Q}$. The expression is a
nonlinear functional of $\mathbf{y}$, $\mathbf{Q}$, and $\mathbf{Q}^{-1}$ that
generally involves intractable point and coordinate relationships between the
training data. Therefore, the above algebraic expression cannot be used to
investigate the algebraic, geometric, or statistical nature of the point and
coordinate relationships between the eigen-scaled extreme points and the Wolfe
dual normal eigenaxis components.

However, it can be investigated how the magnitudes and the directions of the
Wolfe dual normal eigenaxis components on $\mathbf{\psi}$ are selected to
minimize the value of the quadratic form $\mathbf{\psi}^{T}\mathbf{Q\psi}$ in
Eq. (\ref{Vector Form Wolfe Dual}). To accomplish this, an algebraic
connection will be exploited, between the quadratic form $\mathbf{\psi}%
^{T}\mathbf{Q\psi}$ in Eq. (\ref{Vector Form Wolfe Dual}) and the critical
minimum eigenenergies of $\mathbf{\psi}$ and $\mathbf{\tau}$, where the
algebraic connection involves a principal eigen-decomposition of $\mathbf{Q}$.

An algebraic expression for a principal eigen-decomposition of $\mathbf{Q}$
will be developed that offers tractable point and coordinate relationships
between the eigen-scaled extreme training points on $\mathbf{\tau}$ and the
Wolfe dual normal eigenaxis components on $\mathbf{\psi}$. The expression will
be used to demonstrate how eigenloci of Wolfe dual normal eigenaxis components
and constrained primal normal eigenaxis components are determined by
eigen-balanced vector components along the axes of extreme vectors. The
expression will also be used to demonstrate that Wolfe dual normal eigenaxis
components on $\mathbf{\psi}$ and correlated constrained primal normal
eigenaxis components on $\mathbf{\tau}$ possess symmetrical lengths and
exhibit directional symmetry, which jointly describe principal locations of
large covariance, whereby the constrained discriminant function $D\left(
\mathbf{x}\right)  =\mathbf{x}^{T}\mathbf{\tau}+\tau_{0}$ delineates centrally
located, bipartite, symmetric regions of large covariance between two data distributions.

The next section will examine how first and second order statistical moments
of data points are encoded within Gramian matrices. The section begins with
distributions of first degree vector coordinates.

\subsection{Distributions of First Degree Vector Coordinates}

Consider again the Gramian matrix $\mathbf{Q}$ associated with the constrained
quadratic form in Eq. (\ref{Vector Form Wolfe Dual})
\begin{equation}
\mathbf{Q}=%
\begin{pmatrix}
\mathbf{x}_{1}^{T}\mathbf{x}_{1} & \mathbf{x}_{1}^{T}\mathbf{x}_{2} & \cdots &
-\mathbf{x}_{1}^{T}\mathbf{x}_{N}\\
\mathbf{x}_{2}^{T}\mathbf{x}_{1} & \mathbf{x}_{2}^{T}\mathbf{x}_{2} & \cdots &
-\mathbf{x}_{2}^{T}\mathbf{x}_{N}\\
\vdots & \vdots & \ddots & \vdots\\
-\mathbf{x}_{N}^{T}\mathbf{x}_{1} & -\mathbf{x}_{N}^{T}\mathbf{x}_{2} & \cdots
& \mathbf{x}_{N}^{T}\mathbf{x}_{N}%
\end{pmatrix}
\text{,} \label{Autocorrelation Matrix}%
\end{equation}
where $\mathbf{Q}\triangleq\widetilde{\mathbf{X}}\widetilde{\mathbf{X}}^{T}$,
$\widetilde{\mathbf{X}}\triangleq\mathbf{D}_{y}\mathbf{X}$, $\mathbf{D}_{y}$
is a $N\times N$ diagonal matrix of training labels $y_{i}$ and the $N\times
d$ data matrix is $\mathbf{X}$ $=%
\begin{pmatrix}
\mathbf{x}_{1}, & \mathbf{x}_{2}, & \ldots, & \mathbf{x}_{N}%
\end{pmatrix}
^{T}$. Without loss of generality, let $N$ be an even number. Let the first
$N/2$ vectors have the training label $y_{i}=1$ and the last $N/2$ vectors
have the training label $y_{i}=-1$.

Given the above assumptions, the Gramian matrix $\mathbf{Q}$ stores a highly
structured collection of inner product statistics $\mathbf{x}_{i}%
^{T}\mathbf{x}_{j}$ between the geometric loci of the $N$ training points
stored within $\widetilde{\mathbf{X}}$. Take the training point $\mathbf{x}%
_{i}$ or $\mathbf{x}_{j}$, along with the constraint that index $i=j$. It
follow that row $\mathbf{Q}\left(  i,:\right)  $ or column $\mathbf{Q}\left(
:,j\right)  $ encodes sample inner product statistics between the vector
$\mathbf{x}_{i}$ or $\mathbf{x}_{j}$ and all of the vectors $\left(
\mathbf{x}_{1},\cdots,\mathbf{x}_{N}\right)  $ in a training data collection.
It will now be shown that inner product statistics encoded within Gram
matrices determine distributions of first degree vector coordinates. At this
stage of the analysis, training labels will not be taken into account.

Take the training points $\mathbf{x}_{i}$ and $\mathbf{x}_{j}$, along with the
constraint that index $i=j$. Using the algebraic relationship%
\[
\mathbf{x}_{i}^{T}\mathbf{x}_{j}=\left\Vert \mathbf{x}_{i}\right\Vert
\left\Vert \mathbf{x}_{j}\right\Vert \cos\theta_{\mathbf{x}_{i}\mathbf{x}_{j}%
}\text{,}%
\]
satisfied by the inner product statistic $\mathbf{x}_{i}^{T}\mathbf{x}_{j}$,
it follows that row $\mathbf{Q}\left(  i,:\right)  $ in Eq.
(\ref{Autocorrelation Matrix}) encodes uniformly weighted $\left\Vert
\mathbf{x}_{i}\right\Vert $ scalar projections $\left\Vert \mathbf{x}%
_{j}\right\Vert \cos\theta_{\mathbf{x}_{i}\mathbf{x}_{j}}$ for each of the $N$
vectors $\left\{  \mathbf{x}_{j}\right\}  _{j=1}^{N}$ onto the vector
$\mathbf{x}_{i}$:%
\begin{equation}
\widetilde{\mathbf{Q}}=%
\begin{pmatrix}
\left\Vert \mathbf{x}_{1}\right\Vert \left\Vert \mathbf{x}_{1}\right\Vert
\cos\theta_{\mathbf{x_{1}x}_{1}} & \cdots & -\left\Vert \mathbf{x}%
_{1}\right\Vert \left\Vert \mathbf{x}_{N}\right\Vert \cos\theta_{\mathbf{x}%
_{1}\mathbf{x}_{N}}\\
\left\Vert \mathbf{x}_{2}\right\Vert \left\Vert \mathbf{x}_{1}\right\Vert
\cos\theta_{\mathbf{x}_{2}\mathbf{x}_{1}} & \cdots & -\left\Vert
\mathbf{x}_{2}\right\Vert \left\Vert \mathbf{x}_{N}\right\Vert \cos
\theta_{\mathbf{x}_{2}\mathbf{x}_{N}}\\
\vdots & \ddots & \vdots\\
-\left\Vert \mathbf{x}_{N}\right\Vert \left\Vert \mathbf{x}_{1}\right\Vert
\cos\theta_{\mathbf{x}_{N}\mathbf{x}_{1}} & \cdots & \left\Vert \mathbf{x}%
_{N}\right\Vert \left\Vert \mathbf{x}_{N}\right\Vert \cos\theta_{\mathbf{x}%
_{N}\mathbf{x}_{N}}%
\end{pmatrix}
\text{,} \label{Inner Product Matrix}%
\end{equation}
where $0<\theta_{\mathbf{x}_{i}\mathbf{x}_{j}}\leq\frac{\pi}{2}$ or $\frac
{\pi}{2}<\theta_{\mathbf{x}_{i}\mathbf{x}_{j}}\leq\pi$. Alternatively, column
$\mathbf{Q}\left(  :,j\right)  $ in Eq. (\ref{Autocorrelation Matrix}) encodes
weighted $\left\Vert \mathbf{x}_{i}\right\Vert $ scalar projections$\left\Vert
\mathbf{x}_{j}\right\Vert \cos\theta_{\mathbf{x}_{i}\mathbf{x}_{j}}$ for the
vector $\mathbf{x}_{j}$ onto each of the $N$ vectors $\left\{  \mathbf{x}%
_{i}\right\}  _{i=1}^{N}$.

\subsubsection{Signed Magnitudes of Vector Projections}

Now consider the $i$th row $\widetilde{\mathbf{Q}}\left(  i,:\right)  $ of
$\widetilde{\mathbf{Q}}$ in Eq. (\ref{Inner Product Matrix}). Given Eq.
(\ref{Scalar Projection}), it follows that element $\widetilde{\mathbf{Q}%
}\left(  i,j\right)  $ of row $\widetilde{\mathbf{Q}}\left(  i,:\right)  $
encodes the length of the vector $\mathbf{x}_{i}$ multiplied by the scalar
projection of\ $\mathbf{x}_{j}$ onto $\mathbf{x}_{i}$:%
\[
\widetilde{\mathbf{Q}}\left(  i,j\right)  =\left\Vert \mathbf{x}%
_{i}\right\Vert \left[  \left\Vert \mathbf{x}_{j}\right\Vert \cos
\theta_{\mathbf{x}_{i}\mathbf{x}_{j}}\right]  \text{,}%
\]
where the signed magnitude of the vector projection of $\mathbf{x}_{j}$ along
the axis of $\mathbf{x}_{i}$%
\begin{align*}
\operatorname{comp}_{\overrightarrow{\mathbf{x}}_{i}}\left(
\overrightarrow{\mathbf{x}}_{j}\right)   &  =\left\Vert \mathbf{x}%
_{j}\right\Vert \cos\theta_{\mathbf{x}_{i}\mathbf{x}_{j}}\\
&  =\left(  \frac{\mathbf{x}_{i}}{\left\Vert \mathbf{x}_{i}\right\Vert
}\right)  ^{T}\mathbf{x}_{j}\text{,}%
\end{align*}
provides a measure of the first-degree components (point coordinates) of the
vector $\mathbf{x}_{j}$%
\[
\mathbf{x}_{j}=\left(  x_{j_{1}},x_{j_{2}},\cdots,x_{j_{d}}\right)
^{T}\text{,}%
\]
along the axis of the vector $\mathbf{x}_{i}$%
\[
\mathbf{x}_{i}=\left(  x_{i_{1}},x_{i_{2}},\cdots,x_{i_{d}}\right)
^{T}\text{.}%
\]
Note that $\operatorname{comp}_{\overrightarrow{\mathbf{x}}_{i}}\left(
\overrightarrow{\mathbf{x}}_{j}\right)  $ is positive or negative if
$0<\theta\leq\frac{\pi}{2}$ or $\frac{\pi}{2}<\theta\leq\pi$ respectively.
Also, if $\theta=\frac{\pi}{2}$, then $\operatorname{comp}%
_{\overrightarrow{\mathbf{x}}_{i}}\left(  \overrightarrow{\mathbf{x}}%
_{j}\right)  =0$.

Using the above assumptions and notation, given any row $\widetilde{\mathbf{Q}%
}\left(  i,:\right)  $ of Eq. (\ref{Inner Product Matrix}), it follows that
the statistic denoted by $E_{\mathbf{x}_{i}}\left[  \mathbf{x}_{i}|\left\{
\mathbf{x}_{j}\right\}  _{j=1}^{N}\right]  $
\begin{align}
E_{\mathbf{x}_{i}}\left[  \mathbf{x}_{i}|\left\{  \mathbf{x}_{j}\right\}
_{j=1}^{N}\right]   &  =\left\Vert \mathbf{x}_{i}\right\Vert
{\displaystyle\sum\nolimits_{j}}
\operatorname{comp}_{\overrightarrow{\mathbf{x}}_{i}}\left(
\overrightarrow{\mathbf{x}}_{j}\right)
\label{Row Distribution First Order Vector Coordinates}\\
&  =\left\Vert \mathbf{x}_{i}\right\Vert
{\displaystyle\sum\nolimits_{j}}
\left\Vert \mathbf{x}_{j}\right\Vert \cos\theta_{\mathbf{x}_{i}\mathbf{x}_{j}%
}\text{,}\nonumber
\end{align}
provides an estimate $E_{\mathbf{x}_{i}}\left[  \mathbf{x}_{i}|\left\{
\mathbf{x}_{j}\right\}  _{j=1}^{N}\right]  $ for the amount of first degree
components of the vector $\mathbf{x}_{i}$ that are contained in a set of
training vectors $\left\{  \mathbf{x}_{j}\right\}  _{j=1}^{N}$, where training
labels have not been taken into account. It is concluded that Eq.
(\ref{Row Distribution First Order Vector Coordinates}) describes a
distribution of first degree coordinates for the pattern vector $\mathbf{x}%
_{i}$ in a training data collection.

Given that Eq. (\ref{Row Distribution First Order Vector Coordinates})
involves signed magnitudes of vector projections along the axis of a fixed
vector $\mathbf{x}_{i}$, the distribution of first degree vector coordinates
described by Eq. (\ref{Row Distribution First Order Vector Coordinates}) is
said to \emph{determine a first order statistical moment about the geometric
locus of a data point }$\mathbf{x}_{i}$. Because the statistic $E_{\mathbf{x}%
_{i}}\left[  \mathbf{x}_{i}|\left\{  \mathbf{x}_{j}\right\}  _{j=1}%
^{N}\right]  $ depends on the uniform direction of $\mathbf{x}_{i}$, the
statistic $E_{\mathbf{x}_{i}}\left[  \mathbf{x}_{i}|\left\{  \mathbf{x}%
_{j}\right\}  _{j=1}^{N}\right]  $ is said to be unidirectional.

Alternatively, element $\widetilde{\mathbf{Q}}\left(  i,j\right)  $ in the
$j$th column $\widetilde{\mathbf{Q}}\left(  :,j\right)  $ of Eq.
(\ref{Inner Product Matrix}) encodes the length of the vector $\mathbf{x}_{i}$
multiplied by the scalar projection of\ $\mathbf{x}_{j}$ onto $\mathbf{x}_{i}
$%
\[
\widetilde{\mathbf{Q}}\left(  i,j\right)  =\left\Vert \mathbf{x}%
_{i}\right\Vert \left[  \left\Vert \mathbf{x}_{j}\right\Vert \cos
\theta_{\mathbf{x}_{i}\mathbf{x}_{j}}\right]  \text{,}%
\]
where the signed magnitude of the vector projection of $\mathbf{x}_{j}$, along
each axis of a given training vector $\mathbf{x}_{i}$, provides an estimate of
how much of the first degree components of the training vector $\mathbf{x}%
_{i}$ are contained in the vector $\mathbf{x}_{j}$. It follows that the
statistic denoted by $E_{\mathbf{x}_{j}}\left[  \mathbf{x}_{j}|\left\{
\mathbf{x}_{i}\right\}  _{j=1}^{N}\right]  $%
\begin{align}
E_{\mathbf{x}_{j}}\left[  \mathbf{x}_{j}|\left\{  \mathbf{x}_{i}\right\}
_{j=1}^{N}\right]   &  =\left\Vert \mathbf{x}_{i}\right\Vert
{\displaystyle\sum\nolimits_{i}}
\operatorname{comp}_{\overrightarrow{\mathbf{x}}_{i}}\left(
\overrightarrow{\mathbf{x}}_{j}\right)  \text{,}%
\label{Column Distribution First Order Vector Coordinates}\\
&  =%
{\displaystyle\sum\nolimits_{i}}
\left\Vert \mathbf{x}_{i}\right\Vert \left\Vert \mathbf{x}_{j}\right\Vert
\cos\theta_{\mathbf{x}_{i}\mathbf{x}_{j}}\text{,}\nonumber\\
&  =\left\Vert \mathbf{x}_{j}\right\Vert
{\displaystyle\sum\nolimits_{i}}
\left\Vert \mathbf{x}_{i}\right\Vert \cos\theta_{\mathbf{x}_{i}\mathbf{x}_{j}%
}\text{,}\nonumber
\end{align}
provides an estimate $E_{\mathbf{x}_{j}}\left[  \mathbf{x}_{j}|\left\{
\mathbf{x}_{i}\right\}  _{j=1}^{N}\right]  $ for the amount of first degree
vector coordinates of a training data collection $\left\{  \mathbf{x}%
_{i}\right\}  _{i=1}^{N}$ that are contained in the pattern vector
$\mathbf{x}_{j}$, where training labels have not been taken into account.
Because the statistic $E_{\mathbf{x}_{j}}\left[  \mathbf{x}_{j}|\left\{
\mathbf{x}_{i}\right\}  _{j=1}^{N}\right]  $ depends on the directions of all
of the training vectors of $\left\{  \mathbf{x}_{i}\right\}  _{i=1}^{N}$, the
statistic $E_{\mathbf{x}_{j}}\left[  \mathbf{x}_{j}|\left\{  \mathbf{x}%
_{i}\right\}  _{j=1}^{N}\right]  $ is said to be omnidirectional.

It is concluded that row $\widetilde{\mathbf{Q}}\left(  i,:\right)  $ of Eq.
(\ref{Inner Product Matrix}) encodes distributions of first degree vector
coordinates for a training vector $\mathbf{x}_{i}$ within a training data
collection $\left\{  \mathbf{x}_{j}\right\}  _{j=1}^{N}$, and that column
$\widetilde{\mathbf{Q}}\left(  :,j\right)  $ of Eq.
(\ref{Inner Product Matrix}) encodes distributions of first degree vector
coordinates for a training data collection $\left\{  \mathbf{x}_{i}\right\}
_{i=1}^{N}$ within a training vector $\mathbf{x}_{j}$. The next section will
develop unidirectional (pointwise) covariance statistics, which encode
distributions of first degree vector coordinates for individual pattern
vectors $\mathbf{x}_{i}$ within training data collections $\left\{
\mathbf{x}_{j}\right\}  _{j=1}^{N}$.

\subsection{Omnidirectional and Unidirectional Covariance Statistics}

It will first be argued that classical covariance statistics provide
omnidirectional, and therefore non-coherent, estimates of the joint variation
of the random variables of a collection of training vectors about their common
mean. Pointwise covariance statistics will then be developed. Pointwise
covariance statistics provide a unidirectional estimate of how much a group of
data and their common mean varies from a given vector, where the axis of the
given vector is a fixed reference axis. Omnidirectional covariance statistics
are considered next.

\subsubsection{Omnidirectional Covariance Statistics}

Take the data matrix $\mathbf{X}$ $=%
\begin{pmatrix}
\mathbf{x}_{1}, & \mathbf{x}_{2}, & \ldots, & \mathbf{x}_{N}%
\end{pmatrix}
^{T}$ and consider the classical covariance statistic:%
\begin{align}
\widehat{\operatorname{cov}}\left(  \mathbf{X}\right)   &  =\frac{1}{N}%
{\displaystyle\sum\nolimits_{i}}
\left(  \mathbf{x}_{i}-\overline{\mathbf{x}}\right)  ^{2}\text{,}%
\label{Classical Covariance Statistic}\\
&  =\frac{1}{N}%
{\displaystyle\sum\nolimits_{i}}
\left(  \mathbf{x}_{i}-\left(  \frac{1}{N}%
{\displaystyle\sum\nolimits_{i}}
\mathbf{x}_{i}\right)  \right)  ^{2}\text{,}\nonumber
\end{align}
written in vector notation. The statistic $\widehat{\operatorname{cov}}\left(
\mathbf{X}\right)  $ measures the Euclidean distance between a common mean
vector $\overline{\mathbf{x}}$ and each of the training vectors $\mathbf{x}%
_{i}$ in a collection of training data $\left\{  \mathbf{x}_{i}\right\}
_{i=1}^{N}$
\citet{Ash1993}%
,
\citet{Flury1997}%
. Because the statistic $\widehat{\operatorname{cov}}\left(  \mathbf{X}%
\right)  $ depends on $N$ directions of $N$ training vectors, the statistic
$\widehat{\operatorname{cov}}\left(  \mathbf{X}\right)  $ is said to be
omnidirectional. The statistic $\widehat{\operatorname{cov}}\left(
\mathbf{X}\right)  $ provides an omnidirectional estimate of the joint
variation of the $d\times N$ random variables of a collection of $N$ pattern
vectors $\left\{  \mathbf{x}_{i}\right\}  _{i=1}^{N}$ about the geometric
locus of the mean vector $\overline{\mathbf{x}}$, where training labels are
not taken into account.

The statistic $\widehat{\operatorname{cov}}\left(  \mathbf{X}\right)  $ in Eq.
(\ref{Classical Covariance Statistic}) produces a scalar quantity of a
covariance estimate. A statistic is now developed that produces a vector
quantity of a covariance estimate, where the statistic encodes a magnitude and
a direction. The statistic provides a measure of how much a group of data and
its common mean varies from a given vector, where the measure involves signed
magnitudes of vector projections along the axis of the given vector. More
specifically, a pointwise covariance statistic $\widehat{\operatorname{cov}%
}_{up}\left(  \mathbf{x}_{i}\right)  $ provides a unidirectional estimate of
how much a group of data $\left\{  \mathbf{x}_{j}\right\}  _{j=1}^{N}$ and its
common mean $\overline{\mathbf{x}}$ varies from a given vector $\mathbf{x}%
_{i}$, where the axis of the vector $\mathbf{x}_{i}$ is a fixed reference
axis, and the Euclidean distance $\left\Vert \mathbf{x}_{i}\right\Vert
\left\Vert \mathbf{x}_{j}\right\Vert \cos\theta_{\mathbf{x}_{i}\mathbf{x}_{j}%
}$ between $\mathbf{x}_{i}$ and each of the training vectors $\left\{
\mathbf{x}_{j}\right\}  _{j=1}^{N}$ encodes the signed magnitude of the vector
projection%
\[
\left\Vert \mathbf{x}_{j}\right\Vert \cos\theta_{\mathbf{x}_{i}\mathbf{x}_{j}%
}\text{,}%
\]
along the axis of $\mathbf{x}_{i}$, where $\theta_{\mathbf{x}_{i}%
\mathbf{x}_{j}}$ is the angle between $\mathbf{x}_{i}$ and $\mathbf{x}_{j}$.
Likewise, the Euclidean distance $\left\Vert \mathbf{x}_{i}\right\Vert
\left\Vert \overline{\mathbf{x}}\right\Vert \cos\theta_{\mathbf{x}%
_{i}\overline{\mathbf{x}}}$ between $\mathbf{x}_{i}$ and the mean vector
$\overline{\mathbf{x}}$ encodes the signed magnitude of the vector projection%
\[
\left\Vert \overline{\mathbf{x}}\right\Vert \cos\theta_{\mathbf{x}%
_{i}\overline{\mathbf{x}}}\text{,}%
\]
along the axis of $\mathbf{x}_{i}$, where $\cos\theta_{\mathbf{x}_{i}%
\overline{\mathbf{x}}}$ is the angle between $\mathbf{x}_{i}$ and
$\overline{\mathbf{x}}$. Because the statistic $\operatorname{cov}_{up}\left(
\mathbf{x}_{i}\right)  $ depends on the uniform direction of $\mathbf{x}_{i}$,
the statistic $\operatorname{cov}_{up}\left(  \mathbf{x}_{i}\right)  $ is said
to be unidirectional.

Pointwise covariance statistics $\widehat{\operatorname{cov}}_{up}\left(
\mathbf{x}_{i}\right)  $ are now developed which are shown to determine first
and second order statistical moments about the geometric loci of individual
training vectors.

\subsubsection{Pointwise Covariance Statistics}

Take any row $\widetilde{\mathbf{Q}}\left(  i,:\right)  $ of the matrix
$\widetilde{\mathbf{Q}}$ in Eq. (\ref{Inner Product Matrix}) and consider the
inner product statistic $\left\Vert \mathbf{x}_{i}\right\Vert \left\Vert
\mathbf{x}_{j}\right\Vert \cos\theta_{\mathbf{x}_{i}\mathbf{x}_{j}}$ in
element $\widetilde{\mathbf{Q}}\left(  i,j\right)  $. Given Eqs
(\ref{Geometric Locus of Vector}) and (\ref{Inner Product Statistic}), it
follows that element $\widetilde{\mathbf{Q}}\left(  i,j\right)  $ in row
$\widetilde{\mathbf{Q}}\left(  i,:\right)  $ encodes the joint variation
$\operatorname{cov}\left(  \mathbf{x}_{i},\mathbf{x}_{j}\right)  $%
\[
\operatorname{cov}\left(  \mathbf{x}_{i},\mathbf{x}_{j}\right)  =\left\Vert
\mathbf{x}_{i}\right\Vert \left\Vert \mathbf{x}_{j}\right\Vert \cos
\theta_{\mathbf{x}_{i}\mathbf{x}_{j}}\text{,}%
\]
between the vector components (point coordinates) of the geometric locus of
the vector $\mathbf{x}_{i}$%
\[%
\begin{pmatrix}
\left\Vert \mathbf{x}_{i}\right\Vert \cos\mathbb{\alpha}_{\mathbf{x}_{i1}1}, &
\left\Vert \mathbf{x}_{i}\right\Vert \cos\mathbb{\alpha}_{\mathbf{x}_{i2}2}, &
\cdots, & \left\Vert \mathbf{x}_{i}\right\Vert \cos\mathbb{\alpha}%
_{\mathbf{x}_{id}d}%
\end{pmatrix}
\text{,}%
\]
and the vector components (point coordinates) of the geometric locus of the
vector $\mathbf{x}_{j}$%
\[%
\begin{pmatrix}
\left\Vert \mathbf{x}_{j}\right\Vert \cos\mathbb{\alpha}_{\mathbf{x}_{j1}1}, &
\left\Vert \mathbf{x}_{j}\right\Vert \cos\mathbb{\alpha}_{\mathbf{x}_{j2}2}, &
\cdots, & \left\Vert \mathbf{x}_{j}\right\Vert \cos\mathbb{\alpha}%
_{\mathbf{x}_{jd}d}%
\end{pmatrix}
\text{,}%
\]
so that the $j$th element $\widetilde{\mathbf{Q}}\left(  i,j\right)  $ of row
$\widetilde{\mathbf{Q}}\left(  i,:\right)  $ encodes the joint variation of
the $d$ variables of a training vector $\mathbf{x}_{j}$ about the $d$
variables of the training vector $\mathbf{x}_{i}$. Thus, row
$\widetilde{\mathbf{Q}}\left(  i,:\right)  $ encodes the joint variations
between a vector $\mathbf{x}_{i}$ and an entire collection of training data.

Again, take any row $\widetilde{\mathbf{Q}}\left(  i,:\right)  $ of the matrix
$\widetilde{\mathbf{Q}}$ in Eq. (\ref{Inner Product Matrix}). Given Eq.
(\ref{Scalar Projection}), it follows that the statistic
$\widehat{\operatorname{cov}}_{up}\left(  \mathbf{x}_{i}\right)  $:%
\begin{align}
\widehat{\operatorname{cov}}_{up}\left(  \mathbf{x}_{i}\right)   &  =%
{\displaystyle\sum\nolimits_{j=1}^{N}}
\left\Vert \mathbf{x}_{i}\right\Vert \left\Vert \mathbf{x}_{j}\right\Vert
\cos\theta_{\mathbf{x}_{i}\mathbf{x}_{j}}\text{,}%
\label{Pointwise Covariance Statistic}\\
&  =%
{\displaystyle\sum\nolimits_{j=1}^{N}}
\mathbf{x}_{i}^{T}\mathbf{x}_{j}\text{,}\nonumber\\
&  =\mathbf{x}_{i}^{T}\left(
{\displaystyle\sum\nolimits_{j=1}^{N}}
\mathbf{x}_{j}\right)  \text{,}\nonumber\\
&  =\left\Vert \mathbf{x}_{i}\right\Vert
{\displaystyle\sum\nolimits_{j=1}^{N}}
\left\Vert \mathbf{x}_{j}\right\Vert \cos\theta_{\mathbf{x}_{i}\mathbf{x}_{j}%
}\text{,}\nonumber
\end{align}
provides a unidirectional estimate of the joint variation of the $d$ variables
of each of the $N$ training vectors of a training data collection $\left\{
\mathbf{x}_{j}\right\}  _{j=1}^{N}$ and the $d$ variables of the common mean $%
{\displaystyle\sum\nolimits_{j=1}^{N}}
\mathbf{x}_{j}$ of the training data, about the $d$ variables of the vector
$\mathbf{x}_{i}$, along the axis of $\mathbf{x}_{i}$. Note that Eq.
(\ref{Pointwise Covariance Statistic}) does not take training labels into
account. The statistic $\widehat{\operatorname{cov}}_{up}\left(
\mathbf{x}_{i}\right)  $ encodes the direction of the vector $\mathbf{x}_{i}$
and a signed magnitude along the axis of $\mathbf{x}_{i}$.

The statistic $\widehat{\operatorname{cov}}_{up}\left(  \mathbf{x}_{i}\right)
$ in Eq. (\ref{Pointwise Covariance Statistic}) is defined to be a pointwise
covariance estimate for the data point $\mathbf{x}_{i}$, where the statistic
$\widehat{\operatorname{cov}}_{up}\left(  \mathbf{x}_{i}\right)  $ provides a
unidirectional estimate of the joint variations between the geometric locus of
each training vector $\mathbf{x}_{j}$ and the geometric locus of the vector
$\mathbf{x}_{i}$, which includes a unidirectional estimate of the joint
variations between the locus of the mean vector $%
{\displaystyle\sum\nolimits_{j=1}^{N}}
\mathbf{x}_{j}$ and the locus of the given vector $\mathbf{x}_{i}$. Given that
the joint variations estimated by the statistic $\widehat{\operatorname{cov}%
}_{up}\left(  \mathbf{x}_{i}\right)  $ are derived from second order distance
statistics $\left\Vert \mathbf{x}_{i}-\mathbf{x}_{j}\right\Vert ^{2}$, which
involve signed magnitudes of vector projections along the common axis of the
vector $\mathbf{x}_{i}$, a pointwise covariance estimate
$\widehat{\operatorname{cov}}_{up}\left(  \mathbf{x}_{i}\right)  $ is said to
determine \emph{a second order statistical moment about the geometric locus of
the data point} $\mathbf{x}_{i}$.

Returning to Eq. (\ref{Row Distribution First Order Vector Coordinates}), it
follows that Eq. (\ref{Pointwise Covariance Statistic}) also encodes a
distribution of first order coordinates for the training vector $\mathbf{x}%
_{i}$, which determines a first order statistical moment about the geometric
locus of $\mathbf{x}_{i}$. The distribution of first order coordinates for
$\mathbf{x}_{i}$ describes how the components of $\mathbf{x}_{i}$ are
distributed within a training data collection. It is concluded that Eq.
(\ref{Pointwise Covariance Statistic}) determines a first and second order
statistical moment about the geometric locus of the training point
$\mathbf{x}_{i}$. It will now be demonstrated how pointwise covariance
statistics can be used to find extreme data points which possess large
pointwise covariances.

\subsection{Discovery of Extreme Data Points with Pointwise Covariance
Statistics}

The Gramian matrix associated with the constrained quadratic form in Eq.
(\ref{Vector Form Wolfe Dual}) encodes inner product statistics for\ two
labeled collections of training data. Denote those data points that belong to
pattern class $\boldsymbol{X}_{1}$ by $\mathbf{x}_{1_{i}}$ and those that
belong to pattern class $\boldsymbol{X}_{2}$ by $\mathbf{x}_{2_{i}}$. Let
$\overline{\mathbf{x}}_{1}$ and $\overline{\mathbf{x}}_{2}$ denote the mean
vectors of pattern class $\boldsymbol{X}_{1}$ and pattern class
$\boldsymbol{X}_{2}$ respectively. Let $i=1:n_{1}$ where the pattern vector
$\mathbf{x}_{1_{i}}$ has the training label $y_{i}=1$, and let $i=n_{1}%
+1:n_{2}$ where the pattern vector $\mathbf{x}_{2_{i}}$ has the training label
$y_{i}=-1$. Using training label information, Eq.
(\ref{Pointwise Covariance Statistic}) can be rewritten as%
\[
\widehat{\operatorname{cov}}_{up}\left(  \mathbf{x}_{1_{i}}\right)
=\mathbf{x}_{1_{i}}^{T}\left(  \sum\nolimits_{j=1}^{n_{1}}\mathbf{x}_{1_{j}%
}-\sum\nolimits_{j=n_{1}+1}^{n_{2}}\mathbf{x}_{2_{j}}\right)  \text{,}%
\]
and%
\[
\widehat{\operatorname{cov}}_{up}\left(  \mathbf{x}_{2_{i}}\right)
=\mathbf{x}_{2_{i}}^{T}\left(  \sum\nolimits_{j=n_{1}+1}^{n_{2}}%
\mathbf{x}_{2_{j}}-\sum\nolimits_{j=1}^{n_{1}}\mathbf{x}_{1_{j}}\right)
\text{.}%
\]
It will now be shown that extreme training points possess large pointwise
covariances relative to the non-extreme training points in each respective
pattern class. Denote an extreme training point by $\mathbf{x}_{1_{i\ast}}$ or
$\mathbf{x}_{2_{i\ast}}$ and a non-extreme training point by $\mathbf{x}%
_{1_{i}}$ or $\mathbf{x}_{2_{i}}$.

Take any extreme training vector $\mathbf{x}_{1_{i\ast}}$ and any non-extreme
training vector $\mathbf{x}_{1_{i}}$ that belong to the $\boldsymbol{X}_{1}$
pattern class and consider the pointwise covariance estimates of the extreme
data point $\mathbf{x}_{1_{i\ast}}$:%
\begin{align*}
\widehat{\operatorname{cov}}_{up}\left(  \mathbf{x}_{1_{i\ast}}\right)   &
=\mathbf{x}_{1_{i\ast}}^{T}\left(  \sum\nolimits_{j=1}^{n_{1}}\mathbf{x}%
_{1_{j}}-\sum\nolimits_{j=n_{1}+1}^{n_{2}}\mathbf{x}_{2_{j}}\right)
\text{,}\\
&  =\mathbf{x}_{1_{i\ast}}^{T}\overline{\mathbf{x}}_{1}-\mathbf{x}_{1_{i\ast}%
}^{T}\overline{\mathbf{x}}_{2}\text{,}%
\end{align*}
and the non-extreme data point $\mathbf{x}_{1_{i}}$:
\begin{align*}
\widehat{\operatorname{cov}}_{up}\left(  \mathbf{x}_{1}\right)   &
=\mathbf{x}_{1_{i}}^{T}\left(  \sum\nolimits_{j=1}^{n_{1}}\mathbf{x}_{1_{j}%
}-\sum\nolimits_{j=n_{1}+1}^{n_{2}}\mathbf{x}_{2_{j}}\right)  \text{,}\\
&  =\mathbf{x}_{1_{i}}^{T}\overline{\mathbf{x}}_{1}-\mathbf{x}_{1_{i}}%
^{T}\overline{\mathbf{x}}_{2}\text{.}%
\end{align*}
Given that $\mathbf{x}_{1_{i\ast}}$ is an extreme data point, it follows that
$\mathbf{x}_{1_{i\ast}}^{T}\overline{\mathbf{x}}_{1}>$ $\mathbf{x}_{1_{i}}%
^{T}\overline{\mathbf{x}}_{1}$ and that $\mathbf{x}_{1_{i\ast}}^{T}%
\overline{\mathbf{x}}_{2}<$ $\mathbf{x}_{1_{i}}^{T}\overline{\mathbf{x}}_{2}$,
which shows that $\widehat{\operatorname{cov}}_{up}\left(  \mathbf{x}%
_{1_{i\ast}}\right)  >\widehat{\operatorname{cov}}_{up}\left(  \mathbf{x}%
_{1_{i}}\right)  $.

Now take any extreme training vector $\mathbf{x}_{2_{i\ast}}$ and any
non-extreme training vector $\mathbf{x}_{2_{i}}$ that belong to the
$\boldsymbol{X}_{2}$ pattern class and consider the pointwise covariance
estimates of the extreme data point $\mathbf{x}_{2_{i\ast}}$:%
\begin{align*}
\widehat{\operatorname{cov}}_{up}\left(  \mathbf{x}_{2_{i\ast}}\right)   &
=\mathbf{x}_{2_{i\ast}}^{T}\left(  \sum\nolimits_{j=n_{1}+1}^{n_{2}}%
\mathbf{x}_{2_{j}}-\sum\nolimits_{j=1}^{n_{1}}\mathbf{x}_{1_{j}}\right)
\text{,}\\
&  =\mathbf{x}_{2_{i\ast}}^{T}\overline{\mathbf{x}}_{2}-\mathbf{x}_{2_{i\ast}%
}^{T}\overline{\mathbf{x}}_{1}\text{,}%
\end{align*}
and the non-extreme data point $\mathbf{x}_{1_{i}}$:
\begin{align*}
\widehat{\operatorname{cov}}_{up}\left(  \mathbf{x}_{2}\right)   &
=\mathbf{x}_{2_{i}}^{T}\left(  \sum\nolimits_{j=n_{1}+1}^{n_{2}}%
\mathbf{x}_{2_{j}}-\sum\nolimits_{j=1}^{n_{1}}\mathbf{x}_{1_{j}}\right)
\text{,}\\
&  =\mathbf{x}_{2_{i}}^{T}\overline{\mathbf{x}}_{2}-\mathbf{x}_{2_{i}}%
^{T}\overline{\mathbf{x}}_{1}\text{.}%
\end{align*}
Given that $\mathbf{x}_{2_{i\ast}}$ is an extreme data point, it follows that
$\mathbf{x}_{2_{i\ast}}^{T}\overline{\mathbf{x}}_{2}>$ $\mathbf{x}_{2_{i}}%
^{T}\overline{\mathbf{x}}_{2}$ and that $\mathbf{x}_{2_{i\ast}}^{T}%
\overline{\mathbf{x}}_{1}<$ $\mathbf{x}_{2_{i}}^{T}\overline{\mathbf{x}}_{1}$,
which shows that $\widehat{\operatorname{cov}}_{up}\left(  \mathbf{x}%
_{2_{i\ast}}\right)  >\widehat{\operatorname{cov}}_{up}\left(  \mathbf{x}%
_{2_{i}}\right)  $.

It is concluded that extreme training points possess large pointwise
covariances relative to the non-extreme training points in their respective
pattern class.

\subsection{Eigen-scaled Pointwise Covariance Statistics}

Consider again the Gramian matrix associated with the constrained quadratic
form in Eq. (\ref{Vector Form Wolfe Dual}) which encodes inner product
statistics for\ two labeled collections of training data. This section will
define eigen-scaled, pointwise covariance statistics for two collections of
labeled training data. Denote those data points that belong to pattern class
$\boldsymbol{X}_{1}$ by $\mathbf{x}_{1_{i}}$ and those that belong to pattern
class $\boldsymbol{X}_{2}$ by $\mathbf{x}_{2_{i}}$. Let $\overline{\mathbf{x}%
}_{1}$ and $\overline{\mathbf{x}}_{2}$ denote the mean vectors of pattern
class $\boldsymbol{X}_{1}$ and pattern class $\boldsymbol{X}_{2}$
respectively. Let $i=1:n_{1}$ where the pattern vector $\mathbf{x}_{1_{i}}$
has the training label $y_{i}=1$ and let $i=n_{1}+1:n_{2}$ where the pattern
vector $\mathbf{x}_{2_{i}}$ has the training label $y_{i}=-1$.

Suppose that a principal eigen-decomposition of $\mathbf{Q}$ provides two
distinct types of eigen-scales $\psi_{1j}$ and $\psi_{2j}$ for the column
vectors $\mathbf{Q}\left(  :,j\right)  $ of $\mathbf{Q}$ which define
eigen-scales for elements in the rows $\mathbf{Q}\left(  i,:\right)  $ of
$\mathbf{Q}$, where the eigen-scales denoted by $\psi_{1j}$ are correlated
with the pattern class $\boldsymbol{X}_{1}$, the eigen-scales denoted by
$\psi_{2j}$ are correlated with the pattern class $\boldsymbol{X}_{2}$, and
not all of the eigen-scales exceed zero.

Given Eq. (\ref{Pointwise Covariance Statistic}) and the notation and
assumptions outlined above, it follows that summation over the eigen-scaled
$\psi_{1j}$ and $\psi_{2j}$ elements of row $\widetilde{\mathbf{Q}}\left(
i,:\right)  $ denoted in Eq. (\ref{Inner Product Matrix}) provides
eigen-scaled pointwise covariance estimates for the training vectors
$\mathbf{x}_{1_{i}}$:%
\begin{align}
\widehat{\operatorname{cov}}_{up}\left(  \mathbf{x}_{1_{i}}\right)   &
=\left\Vert \mathbf{x}_{1_{i}}\right\Vert \sum\nolimits_{j=1}^{n_{1}}%
\psi_{1_{j}}\left\Vert \mathbf{x}_{1_{j}}\right\Vert \cos\theta_{\mathbf{x}%
_{1_{i}}\mathbf{x}_{1_{j}}}\label{Pointwise Covariance Estimate Class One}\\
&  -\left\Vert \mathbf{x}_{1_{i}}\right\Vert \sum\nolimits_{j=n_{1}+1}^{n_{2}%
}\psi_{2_{j}}\left\Vert \mathbf{x}_{2_{j}}\right\Vert \cos\theta
_{\mathbf{x}_{1_{i}}\mathbf{x}_{2_{j}}}\text{,}\nonumber
\end{align}
and $\mathbf{x}_{2_{i}}$:%
\begin{align}
\widehat{\operatorname{cov}}_{up}\left(  \mathbf{x}_{2_{i}}\right)   &
=\left\Vert \mathbf{x}_{2_{i}}\right\Vert \sum\nolimits_{j=n_{1}+1}^{n_{2}%
}\psi_{2_{j}}\left\Vert \mathbf{x}_{2_{j}}\right\Vert \cos\theta
_{\mathbf{x}_{2_{i}}\mathbf{x}_{2_{j}}}%
\label{Pointwise Covariance Estimate Class Two}\\
&  -\left\Vert \mathbf{x}_{2_{i}}\right\Vert \sum\nolimits_{j=1}^{n_{1}}%
\psi_{1_{j}}\left\Vert \mathbf{x}_{1_{j}}\right\Vert \cos\theta_{\mathbf{x}%
_{2_{i}}\mathbf{x}_{1_{j}}}\text{.}\nonumber
\end{align}
Given Eqs (\ref{Equilibrium Constraint on Dual Eigen-components}) and
(\ref{Pointwise Covariance Statistic}), it follows that Eqs
(\ref{Pointwise Covariance Estimate Class One}) and
(\ref{Pointwise Covariance Estimate Class Two}) determine eigen-balanced first
and second order statistical moments about the geometric locus of a training point.

It has been demonstrated that extreme training points possess large pointwise
covariances relative to the non-extreme training points in their respective
pattern class. Therefore, it will be assumed that any given extreme data point
$\mathbf{x}_{1_{i_{\ast}}}$ or $\mathbf{x}_{2_{i_{\ast}}}$ exhibits a critical
first and second order statistical moment that exceeds some threshold for
which $\psi_{1i\ast}>0$ or $\psi_{2i\ast}>0$. This implies that first and
second order statistical moments $\operatorname{cov}_{up}\left(
\mathbf{x}_{1_{i}}\right)  $ and $\operatorname{cov}_{up}\left(
\mathbf{x}_{2_{i}}\right)  $ about the loci of non-extreme data points
$\mathbf{x}_{1_{i}}$ and $\mathbf{x}_{2_{i}}$ do not exceed the threshold, so
that $\psi_{1i}=0$ and $\psi_{2i}=0$. This indicates that the eigen-scaled
pointwise covariance estimates in Eqs
(\ref{Pointwise Covariance Estimate Class One}) and
(\ref{Pointwise Covariance Estimate Class Two}) are a function of eigen-scaled
extreme training points. The next section will consider descriptive statistics
for separating hyperplanes.

\subsection{Descriptive Statistics for Separating Hyperplanes}

It has previously been argued that the inner product statistics encoded within
a matrix associated with a quadratic form determine the geometric shapes of
second-order surfaces. It will now be argued that inner product statistics
which have the algebraic form of $\mathbf{x}_{i}^{T}\mathbf{x}_{j}$ describe
hyperplane surfaces.

\subsubsection*{The Coordinate Equation Version of a Hyperplane Surface}

Every equation of the first degree specifies the locus of a linear curve or
surface
\citet{Nichols1893}%
,
\citet{Tanner1898}%
,
\citet{Eisenhart1939}%
. Indeed, the coordinate equation version of an $N$-dimensional hyperplane
surface:%
\[
Ax_{i_{1}}+Bx_{i_{2}}+\ldots+Nx_{i_{N}}=P\text{,}%
\]
contains $N$ first degree vector coordinates $x_{i_{j}}$. Now take the Gram
matrix $\mathbf{Q}$ associated with the constrained quadratic form in Eq.
(\ref{Vector Form Wolfe Dual}). Given that $\left(  1\right)  $ the elements
$\mathbf{x}_{i}^{T}\mathbf{x}_{j}$ of $\mathbf{Q}$ describe the geometric
shapes of three, symmetric quadratic partitioning surfaces, and that $\left(
2\right)  $ any row $\mathbf{Q}\left(  i,:\right)  $ of $\mathbf{Q}$ encodes a
distribution of first degree vector coordinates for a training vector
$\mathbf{x}_{i}$, it follows that $\mathbf{Q}$ contains descriptive statistics
for three, symmetrical hyperplane partitioning surfaces. It has been
demonstrated by simulation studies that inner product statistics, which have
the algebraic form of $\mathbf{x}_{i}^{T}\mathbf{x}_{j}$, do indeed provide
descriptive statistics for three, symmetrical hyperplane partitioning
surfaces
\citet{Reeves2007}%
,
\citet{Reeves2009}%
,
\citet{Reeves2011}%
.

For the analyses that follow, it will be assumed that the Gram matrix
$\mathbf{Q}$ associated with the constrained quadratic form in Eq.
(\ref{Vector Form Wolfe Dual}) describes three, symmetrical hyperplane
partitioning surfaces. Equations
(\ref{Pointwise Covariance Estimate Class One}) and
(\ref{Pointwise Covariance Estimate Class Two}) will be used, in connection
with a principal eigen-decomposition of the Gram matrix $\mathbf{Q}$
associated with the constrained quadratic form in Eq.
(\ref{Vector Form Wolfe Dual}), to demonstrate how each Wolfe dual normal
eigenaxis component on $\mathbf{\psi}$ encodes an eigen-balanced, pointwise
covariance estimate for an extreme data point, which determines a properly
proportioned eigen-scale that determines a suitable magnitude, and therefore a
suitable eigenlocus (location), for a constrained primal normal eigenaxis
component on $\mathbf{\tau}$. Thereby, the analyses will demonstrate how the
eigenlocus and corresponding eigenstate of each eigen-scaled extreme point on
$\mathbf{\tau}$ is determined by the eigenlocus of a Wolfe dual normal
eigenaxis component. Later on, an expression will be derived for the total
allowed eigenenergy of an eigen-scaled extreme point and its corresponding
eigenstate. The total allowed eigenenergies of a strong dual normal eigenlocus
are revisited next.

\subsection{Symmetrical Relationships Between the Total Allowed Eigenenergies
of a Strong Dual Normal Eigenlocus}

It has previously been claimed that strong duality relationships between the
algebraic systems of $\min\Psi\left(  \mathbf{\tau}\right)  $ and $\max
\Xi\left(  \mathbf{\psi}\right)  $ impose some type of symmetrical
relationships between the total allowed eigenenergies of $\mathbf{\tau}$ and
$\mathbf{\psi}$%
\[
\left\Vert \mathbf{\tau}\right\Vert _{\min_{c}}^{2}\cong\left\Vert
\mathbf{\psi}\right\Vert _{\min_{c}}^{2}\text{.}%
\]
It has also been claimed that the lengths of the Wolfe dual normal eigenaxis
components must be selected so that the total allowed eigenenergies of
$\mathbf{\tau}_{1}$ and $\mathbf{\tau}_{2}$ are balanced by means of a
symmetric equalizer statistic $\nabla_{eq}$%
\[
\left(  \left\Vert \mathbf{\tau}_{1}\right\Vert _{\min_{c}}^{2}-\left\Vert
\mathbf{\tau}_{1}\right\Vert \left\Vert \mathbf{\tau}_{2}\right\Vert
\cos\theta_{\mathbf{\tau}_{1}\mathbf{\tau}_{2}}\right)  +\nabla_{eq}%
\Leftrightarrow\left(  \left\Vert \mathbf{\tau}_{2}\right\Vert _{\min_{c}}%
^{2}-\left\Vert \mathbf{\tau}_{1}\right\Vert \left\Vert \mathbf{\tau}%
_{2}\right\Vert \cos\theta_{\mathbf{\tau}_{1}\mathbf{\tau}_{2}}\right)
-\nabla_{eq}\text{,}%
\]
in relation to a centrally located statistical fulcrum $f_{s}$, whereby the
principal statistical state $\lambda\mathbf{_{\max_{\mathbf{\psi}}}}$ and the
characteristic eigenstates
\[
\left\{  \Psi_{\mathbf{\tau}_{1}}\left(
\mathbb{R}
^{d}\right)  \right\}  _{i=1}^{l_{1}}\triangleq\left\{  \psi_{1i\ast
}\mathbf{x}_{1_{i\ast}}|\mathbf{x}_{1_{i\ast}}\sim p_{\mathbf{x}_{1_{i}%
}|\boldsymbol{X}_{1}}\left(  \mathbf{x}_{1_{i}}|\boldsymbol{X}_{1}\right)
\right\}  _{i=1}^{l_{1}}\text{,}%
\]
and
\[
\left\{  \Psi_{\mathbf{\tau}_{2}}\left(
\mathbb{R}
^{d}\right)  \right\}  _{i=1}^{l_{2}}\triangleq\left\{  \psi_{2i\ast
}\mathbf{x}_{2_{i\ast}}|\mathbf{x}_{2_{i\ast}}\sim p_{\mathbf{x}_{2_{i}%
}|\boldsymbol{X}_{2}}\left(  \mathbf{x}_{2_{i}}|\boldsymbol{X}_{2}\right)
\right\}  _{i=1}^{l_{2}}\text{,}%
\]
determine a point $\mathbf{\tau}$ of statistical equilibrium. Algebraic and
statistical equations that determine the symmetrical relationships between the
critical minimum eigenenergies $\left\Vert \mathbf{\tau}\right\Vert _{\min
_{c}}^{2}\ $and $\left\Vert \mathbf{\psi}\right\Vert _{\min_{c}}^{2}$ of
$\mathbf{\tau}$ and $\mathbf{\psi}$ will be developed in Sections $14$ - $17$.
The findings presented in Sections $14$ and $15$ will show how the geometric
configuration of a constrained primal normal eigenlocus $\mathbf{\tau}$ is
symmetrically shaped by the eigen-balanced vector components of a Wolfe dual
normal eigenlocus $\mathbf{\psi}$, all of which satisfy critical length
constraints. Section $17$ will demonstrate that the eigenenergies of the
eigen-scaled extreme points on $\mathbf{\mathbf{\tau}}_{1}$ and
$\mathbf{\mathbf{\tau}}_{2}$ are distributed in a symmetrical manner which
symmetrically balances the critical minimum eigenenergies of\textit{\ }%
$\mathbf{\mathbf{\tau}}_{1}$ and $\mathbf{\mathbf{\tau}}_{2}$.

The paper will now motivate and introduce the form of an algebraic expression
that will be used to examine eigenlocus relationships between the eigen-scaled
extreme data points on $\mathbf{\tau}$ and the Wolfe dual normal eigenaxis
components on $\mathbf{\psi}$. Recall that the Wolfe dual normal eigenlocus
$\mathbf{\psi}$ of Eq. (\ref{Vector Form Wolfe Dual})%
\[
\max\Xi\left(  \mathbf{\psi}\right)  =\mathbf{1}^{T}\mathbf{\psi}%
-\frac{\mathbf{\psi}^{T}\mathbf{Q\psi}}{2}\text{,}%
\]
which satisfies the inner product statistic $\mathbf{\psi}^{T}\mathbf{y}=0$
and magnitude constraints $\psi_{i}\geq0$, provides an estimate of the
principal eigenvector $\mathbf{\psi}$ of the Gram matrix $\mathbf{Q}$
associated with the constrained quadratic form $\mathbf{\psi}^{T}%
\mathbf{Q\psi}$, whereby $\mathbf{\psi}$ is the principal eigenaxis of three,
symmetrical\textit{\ }hyperplane partitioning surfaces described by the
constrained quadratic form $\mathbf{\psi}^{T}\mathbf{Q\psi}$, and $\mathbf{y}$
is a directrix which is orthogonal to $\mathbf{\psi:}$ $\mathbf{\psi}%
\perp\mathbf{y}$. The strong duality relationships between the algebraic
systems of $\min\Psi\left(  \mathbf{\tau}\right)  $ and $\max\Xi\left(
\mathbf{\psi}\right)  $ indicate that the principal eigenvector $\mathbf{\psi
}$ of $\mathbf{Q}$ satisfies a critical minimum eigenenergy constraint
$\left\Vert \mathbf{\psi}\right\Vert _{\min_{c}}^{2}$ that is symmetrically
related to the critical minimum eigenenergy constraint $\left\Vert
\mathbf{\tau}\right\Vert _{\min_{c}}^{2}$ satisfied by a strong dual normal
eigenlocus $\mathbf{\tau}$.

Suppose that an expression is obtained for a principal eigen-decomposition of
a Gram matrix $\mathbf{Q}$%
\[
\max\mathbf{Q\psi}=\lambda_{\max_{\mathbf{\psi}}}\mathbf{\psi}\text{,}%
\]
which is associated with an estimate of a Wolfe dual normal eigenlocus
$\mathbf{\psi}$. Multiplying both sides of the above expression by the vector
transpose $\mathbf{\psi}^{T}$ of a Wolfe dual normal eigenlocus $\mathbf{\psi
}$ provides an expression which relates the constrained quadratic form
$\mathbf{\psi}^{T}\mathbf{Q\psi}$ in Eq. (\ref{Vector Form Wolfe Dual}) to the
total allowed (critical minimum) eigenenergy $\left\Vert \mathbf{\psi
}\right\Vert _{\min}^{2}$ of a Wolfe dual normal eigenlocus $\mathbf{\psi}$:%
\begin{align}
\max\mathbf{\psi}^{T}\mathbf{Q\psi}  &  \mathbf{=\psi}^{T}\lambda
_{\max_{\mathbf{\psi}}}\mathbf{\psi}\text{,}%
\label{Critical Minimum Eigenenergy of a Wolf Dual Normal Eigenlocus}\\
&  =\lambda_{\max_{\mathbf{\psi}}}\mathbf{\psi}^{T}\mathbf{\psi}%
\text{,}\nonumber\\
&  =\lambda_{\max_{\mathbf{\psi}}}\left\Vert \mathbf{\psi}\right\Vert
_{\min_{c}}^{2}\text{,}\nonumber
\end{align}
where the critical minimum eigenenergy $\lambda_{\max_{\mathbf{\psi}}%
}\left\Vert \mathbf{\psi}\right\Vert _{\min}^{2}$ of $\mathbf{\psi}$ is
symmetrically correlated with the critical minimum eigenenergy $\left\Vert
\mathbf{\tau}\right\Vert _{\min}^{2}$ of a constrained primal normal
eigenlocus $\mathbf{\tau}$%
\[
\max\mathbf{\psi}^{T}\mathbf{Q\psi}\cong\left\Vert \mathbf{\tau}\right\Vert
_{\min_{c}}^{2}\text{.}%
\]

Given the algebraic relationships outlined directly above, this paper will
examine the geometric and statistical properties exhibited by a Wolfe dual
normal eigenlocus $\mathbf{\psi}$ by analyzing a principal eigen-decomposition
of the Gram matrix $\mathbf{Q}$ denoted in Eqs (\ref{Autocorrelation Matrix})
and (\ref{Inner Product Matrix}). The next part of the paper will examine a
general expression of a principal eigen-decomposition that offers tractable
point and coordinate relationships between the eigen-scaled extreme data
points on $\mathbf{\tau}_{1}$ and $\mathbf{\tau}_{2}$ and their correlated
Wolfe dual normal eigenaxis components on $\mathbf{\mathbf{\psi}}$. Analysis
of the expression will provide significant insights into geometric and
statistical interconnections between the constrained primal and the Wolfe dual
normal eigenaxis components. Sections $14$ and $15$ will use the general
expression for the principal eigen-decomposition to develop algebraic
expressions for the eigenloci (the geometric locations) of the $\psi_{1i\ast
}\overrightarrow{\mathbf{e}}_{1i\ast}$ and the $\psi_{2i\ast}%
\overrightarrow{\mathbf{e}}_{2i\ast}$ Wolfe dual normal eigenaxis components.
These expressions will be used to define uniform geometric and statistical
properties which are jointly exhibited by the Wolfe dual normal eigenaxis
components on $\mathbf{\mathbf{\psi}}$ and the constrained primal normal
eigenaxis components on $\mathbf{\tau}$. The general expression for the
principal eigen-decomposition is obtained next.

\section{Underneath the Hood of a Wolfe Dual Normal Eigenlocus}

Take the Gram matrix $\mathbf{Q}$ associated with the quadratic form in Eq
(\ref{Vector Form Wolfe Dual}). Let $\mathbf{q}_{j\text{ }}$denote the $j$th
column of $\mathbf{Q}$, which is an $N$-vector. Let $\lambda_{\max
_{\mathbf{\psi}}}$ and $\mathbf{\psi}$ denote the largest eigenvalue and
largest eigenvector of $\mathbf{Q}$ respectively. Using this notation
\citet{Trefethen1998}%
, the principal eigen-decomposition of $\mathbf{Q}$:%
\[
\mathbf{Q\mathbf{\mathbf{\psi}}}=\lambda\mathbf{_{\max_{\mathbf{\psi}}%
}\mathbf{\psi}}\text{,}%
\]
can be rewritten as%
\[
\lambda_{\max_{\mathbf{\psi}}}\mathbf{\mathbf{\psi}}=%
{\displaystyle\sum\nolimits_{j=1}^{N}}
\psi_{_{j}}\mathbf{q}_{j\text{ }}\text{,}%
\]
where the Wolfe dual normal eigenaxis $\mathbf{\mathbf{\psi}}$ of $\mathbf{Q}$
is expressed as a linear combination of the eigen-transformed vectors
$\psi_{j}\mathbf{q}_{j\text{ }}$:%
\begin{equation}
\lambda_{\max_{\mathbf{\psi}}}\left[
\begin{array}
[c]{c}%
\\
\mathbf{\mathbf{\psi}_{1}}\\
\\
\end{array}
\right]  =\psi_{1}\left[
\begin{array}
[c]{c}%
\\
\mathbf{q}_{1\text{ }}\\
\\
\end{array}
\right]  +\psi_{2}\left[
\begin{array}
[c]{c}%
\\
\mathbf{q}_{2\text{ }}\\
\\
\end{array}
\right]  +\cdots+\psi_{N}\left[
\begin{array}
[c]{c}%
\\
\mathbf{q}_{N\text{ }}\\
\\
\end{array}
\right]  \text{,} \label{Alternate Eigendecomposition Equation}%
\end{equation}
where the $i$th element of the vector $\mathbf{q}_{j\text{ }}$ encodes an
inner product statistic $\mathbf{x}_{i}^{T}\mathbf{x}_{j}$ between the vectors
$\mathbf{x}_{i}$ and $\mathbf{x}_{j}$.

Equation (\ref{Alternate Eigendecomposition Equation}) is now used to examine
how eigen-transformed, inner product statistical relationships between the
extreme training vectors and the Wolfe dual normal eigenaxis components
specify the eigenloci of the Wolfe dual normal eigenaxis components on
$\mathbf{\psi}$.

Using Eqs (\ref{Autocorrelation Matrix}) and
(\ref{Alternate Eigendecomposition Equation}), a Wolfe dual normal eigenlocus
$\mathbf{\psi} $%
\[
\mathbf{\psi}=\left(  \psi_{1},\psi_{2},\cdots,\psi_{N}\right)  ^{T}\text{,}%
\]
can be written as:%
\begin{align}
\mathbf{\psi}  &  =\frac{\psi_{1}}{\lambda_{\max_{\mathbf{\psi}}}}%
\begin{pmatrix}
\mathbf{x}_{1}^{T}\mathbf{x}_{1}\\
\mathbf{x}_{2}^{T}\mathbf{x}_{1}\\
\vdots\\
-\mathbf{x}_{N}^{T}\mathbf{x}_{1}%
\end{pmatrix}
\label{Dual Normal Eigenlocus Components}\\
&  +\frac{\psi_{2}}{\lambda_{\max_{\mathbf{\psi}}}}%
\begin{pmatrix}
\mathbf{x}_{1}^{T}\mathbf{x}_{2}\\
\mathbf{x}_{2}^{T}\mathbf{x}_{2}\\
\vdots\\
-\mathbf{x}_{N}^{T}\mathbf{x}_{2}%
\end{pmatrix}
+\cdots\nonumber\\
&  \cdots+\frac{\psi_{N}}{\lambda_{\max_{\mathbf{\psi}}}}%
\begin{pmatrix}
-\mathbf{x}_{1}^{T}\mathbf{x}_{N}\\
-\mathbf{x}_{2}^{T}\mathbf{x}_{N}\\
\vdots\\
\mathbf{x}_{N}^{T}\mathbf{x}_{N}%
\end{pmatrix}
\text{,}\nonumber
\end{align}
which illustrates that the magnitude $\psi_{j}$ of the $j^{th}$ Wolfe dual
normal eigenaxis component is correlated with joint variations of the training
data about the training vector $\mathbf{x}_{j}$.

Alternatively, using Eqs (\ref{Inner Product Matrix}) and
(\ref{Alternate Eigendecomposition Equation}), a Wolfe dual normal eigenlocus
$\mathbf{\psi} $ can be written as:%
\begin{align}
\mathbf{\psi}  &  =\frac{\psi_{1}}{\lambda_{\max_{\mathbf{\psi}}}}\left(
\begin{array}
[c]{c}%
\left\Vert \mathbf{x}_{1}\right\Vert \left\Vert \mathbf{x}_{1}\right\Vert
\cos\theta_{\mathbf{x}_{1}^{T}\mathbf{x}_{1}}\\
\left\Vert \mathbf{x}_{2}\right\Vert \left\Vert \mathbf{x}_{1}\right\Vert
\cos\theta_{\mathbf{x}_{2}^{T}\mathbf{x}_{1}}\\
\vdots\\
-\left\Vert \mathbf{x}_{N}\right\Vert \left\Vert \mathbf{x}_{1}\right\Vert
\cos\theta_{\mathbf{x}_{N}^{T}\mathbf{x}_{1}}%
\end{array}
\right)  +\cdots\label{Dual Normal Eigenlocus Component Projections}\\
&  \cdots+\frac{\psi_{N}}{\lambda_{\max_{\mathbf{\psi}}}}\left(
\begin{array}
[c]{c}%
-\left\Vert \mathbf{x}_{1}\right\Vert \left\Vert \mathbf{x}_{N}\right\Vert
\cos\theta_{\mathbf{x}_{1}^{T}\mathbf{x}_{N}}\\
-\left\Vert \mathbf{x}_{2}\right\Vert \left\Vert \mathbf{x}_{N}\right\Vert
\cos\theta_{\mathbf{x}_{2}^{T}\mathbf{x}_{N}}\\
\vdots\\
\left\Vert \mathbf{x}_{N}\right\Vert \left\Vert \mathbf{x}_{N}\right\Vert
\cos\theta_{\mathbf{x}_{N}^{T}\mathbf{x}_{N}}%
\end{array}
\right)  \text{,}\nonumber
\end{align}
which illustrates that the magnitude $\psi_{j}$ of the $j^{th}$ Wolfe dual
normal eigenaxis component on $\mathbf{\mathbf{\psi}}$ is correlated with
scalar projections $\left\Vert \mathbf{x}_{j}\right\Vert \cos\theta
_{\mathbf{x}_{i}\mathbf{x}_{j}}$ of the training vector $\mathbf{x}_{j}$ onto
the training data.

\subsection{Non-Orthogonal Eigenaxes of $\mathbf{\psi}$}

Express a Wolfe dual normal eigenlocus $\mathbf{\psi}$ in terms of $l$
non-orthogonal unit vectors $\left\{  \overrightarrow{\mathbf{e}}_{1\ast
},\ldots,\overrightarrow{\mathbf{e}}_{l\ast}\right\}  $%
\begin{align}
\mathbf{\mathbf{\psi}}  &  =\sum\nolimits_{i=1}^{l}\psi_{i\ast}%
\overrightarrow{\mathbf{e}}_{i\ast}\text{,}%
\label{Non-orthogonal Eigenaxes of Dual Normal Eigenlocus}\\
&  =\sum\nolimits_{i=1}^{l_{1}}\psi_{1i\ast}\overrightarrow{\mathbf{e}%
}_{1i\ast}+\sum\nolimits_{i=1}^{l_{2}}\psi_{2i\ast}\overrightarrow{\mathbf{e}%
}_{2i\ast}\text{,}\nonumber
\end{align}
where the eigen-scaled, non-orthogonal unit vector denoted by $\psi_{1i\ast
}\overrightarrow{\mathbf{e}}_{1i\ast}$ or $\psi_{2i\ast}%
\overrightarrow{\mathbf{e}}_{2i\ast}$ is correlated with the extreme training
vector $\mathbf{x}_{1_{i\ast}}$ or $\mathbf{x}_{2_{i\ast}}$ respectively.
Accordingly, a Wolfe dual normal eigenaxis component $\psi_{1i\ast
}\overrightarrow{\mathbf{e}}_{1i\ast}$ or $\psi_{2i\ast}%
\overrightarrow{\mathbf{e}}_{2i\ast}$ is an eigen-scaled, non-orthogonal unit
vector that contributes to the estimation of $\mathbf{\mathbf{\psi}}$.

Equations (\ref{Dual Normal Eigenlocus Component Projections}) and
(\ref{Non-orthogonal Eigenaxes of Dual Normal Eigenlocus}) will be used to
identify the geometric and statistical properties which determine the
magnitudes and the directions of the Wolfe dual $\psi_{1i\ast}%
\overrightarrow{\mathbf{e}}_{1i\ast}$ and $\psi_{2i\ast}%
\overrightarrow{\mathbf{e}}_{2i\ast}$ and the magnitudes of the constrained
primal $\psi_{1i\ast}\mathbf{x}_{1_{\ast}}$ and $\psi_{2i\ast}\mathbf{x}%
_{2_{\ast}}$ normal eigenaxis components. It will be demonstrated that each
Wolfe dual normal eigenaxis component stores an eigen-balanced first and
second order statistical moment about the locus of an extreme data point,
which determines an eigen-scale for a constrained primal normal eigenaxis
component, such that each correlated Wolfe dual and constrained primal normal
eigenaxis component exhibit symmetrical magnitudes and symmetrical directions.

The next two sections will examine the geometric and statistical properties of
the eigen-scales used to form the constrained primal normal eigenaxis
components. It will be shown that the eigenlocus of each Wolfe dual normal
eigenaxis component on $\mathbf{\psi}$ encodes an eigen-balanced, positive
magnitude along the axis of a correlated extreme vector, which is determined
by an eigen-balanced, first and second order statistical moment about the
locus of the extreme vector, such that the eigenlocus of each Wolfe dual
normal eigenaxis component on $\mathbf{\psi}$ provides a symmetrical
eigen-scale which determines the critical length of a correlated extreme
training vector on $\mathbf{\tau}$. The direction of each non-orthogonal unit
vector $\overrightarrow{\mathbf{e}}_{1i\ast}$ or $\overrightarrow{\mathbf{e}%
}_{2i\ast}$ will be shown to be identical to the direction of an extreme
training vector $\mathbf{x}_{1_{i\ast}}$ or $\mathbf{x}_{2_{i\ast}} $. It will
also be shown that the eigenloci of the constrained normal eigenaxis
components on $\mathbf{\mathbf{\psi}}$ and $\mathbf{\tau}$ delineate an
implicit estimate of a separating hyperplane that is bounded by bilaterally
symmetrical hyperplane borders. Sections $16$, $17$, and $18$ will use results
from the analysis to examine the geometric and statistical characteristics of
the statistical equilibrium state implied by the KKT condition of Eq.
(\ref{KKTE2}). The analysis begins with extreme point notation and assumptions.

\subsection*{Extreme Point Notation and Assumptions}

Pattern category information defines a distinct pair of pattern classes.
Denote the pattern classes one and two by $\boldsymbol{X}_{1}$ and
$\boldsymbol{X}_{2}$ respectively. Denote those extreme points that belong to
pattern class $\boldsymbol{X}_{1}=\left\{  \mathbf{x}_{1_{i_{\ast}}%
}|\mathbf{x}_{1_{i_{\ast}}}\in\boldsymbol{X}_{1},y_{i}=+1\right\}  $ by
$\mathbf{x}_{1_{i_{\ast}}}$ and those that belong to pattern class
$\boldsymbol{X}_{2}=\left\{  \mathbf{x}_{2_{i_{\ast}}}|\mathbf{x}_{2_{i_{\ast
}}}\in\boldsymbol{X}_{2},y_{i}=-1\right\}  $ by $\mathbf{x}_{2_{i_{\ast}}}$.
Let $l_{1}$ denote the number of extreme data points that belong to the
pattern class $\boldsymbol{X}_{1}$ and let $l_{2}$ denote the number that
belong to the pattern class $\boldsymbol{X}_{2}$. Let the extreme training
point $\mathbf{x}_{1_{i_{\ast}}}$ associated with the $\psi_{1i\ast
}\overrightarrow{\mathbf{e}}_{1i\ast}$ normal eigenaxis component have the
training label $y_{i}=1$, and let the extreme training point $\mathbf{x}%
_{2_{i_{\ast}}}$ associated with the $\psi_{2i\ast}\overrightarrow{\mathbf{e}%
}_{2i\ast}$ normal eigenaxis component have the training label $y_{i}=-1$.
Denote the number of $\psi_{1i\ast}\overrightarrow{\mathbf{e}}_{1i\ast}$ and
$\psi_{2i\ast}\overrightarrow{\mathbf{e}}_{2i\ast}$ normal eigenaxis
components by $l_{1}$ and $l_{2}$ respectively. Assume that $l_{1}+l_{2}=l$.

Given the above assumptions and Eqs
(\ref{Pointwise Covariance Estimate Class One}) and
(\ref{Pointwise Covariance Estimate Class Two}), it follows that an extreme
data point $\mathbf{x}_{1_{i_{\ast}}}$ possesses a large pointwise covariance%
\begin{align}
\widehat{\operatorname{cov}}_{up_{\updownarrow}}\left(  \mathbf{x}%
_{1_{i_{\ast}}}\right)   &  =\left\Vert \mathbf{x}_{1_{i_{\ast}}}\right\Vert
\sum\nolimits_{j=1}^{l_{1}}\psi_{1_{j\ast}}\left\Vert \mathbf{x}_{1_{j\ast}%
}\right\Vert \cos\theta_{\mathbf{x}_{1_{i\ast}}\mathbf{x}_{1_{j\ast}}%
}\label{Large Unidirectional Pointwise Covariance Estimate Class One}\\
&  -\left\Vert \mathbf{x}_{1_{i_{\ast}}}\right\Vert \sum\nolimits_{j=1}%
^{l_{2}}\psi_{2_{j\ast}}\left\Vert \mathbf{x}_{2_{j\ast}}\right\Vert
\cos\theta_{\mathbf{x}_{1_{i\ast}}\mathbf{x}_{2_{j\ast}}}\text{,}\nonumber
\end{align}
relative to the $\boldsymbol{X}_{1}$ training vectors, where
$\widehat{\operatorname{cov}}_{up_{\updownarrow}}\left(  \mathbf{x}_{1_{i}%
}\right)  =0$ for all non-extreme training vectors that belong to pattern
class $\boldsymbol{X}_{1} $. Likewise, an extreme data point $\mathbf{x}%
_{2_{i_{\ast}}}$ possesses a large pointwise covariance%
\begin{align}
\widehat{\operatorname{cov}}_{up_{\updownarrow}}\left(  \mathbf{x}_{2_{i\ast}%
}\right)   &  =\left\Vert \mathbf{x}_{2_{i\ast}}\right\Vert \sum
\nolimits_{j=1}^{l_{2}}\psi_{2_{j\ast}}\left\Vert \mathbf{x}_{2_{j\ast}%
}\right\Vert \cos\theta_{\mathbf{x}_{2_{i\ast}}\mathbf{x}_{2_{j\ast}}%
}\label{Large Unidirectional Pointwise Covariance Estimate Class Two}\\
&  -\left\Vert \mathbf{x}_{2_{i_{\ast}}}\right\Vert \sum\nolimits_{j=1}%
^{l_{2}}\psi_{1_{j\ast}}\left\Vert \mathbf{x}_{1_{j\ast}}\right\Vert
\cos\theta_{\mathbf{x}_{1_{i\ast}}\mathbf{x}_{2_{j\ast}}}\text{,}\nonumber
\end{align}
relative to the $\boldsymbol{X}_{2}$ training vectors, where
$\widehat{\operatorname{cov}}_{up_{\updownarrow}}\left(  \mathbf{x}_{2_{i}%
}\right)  =0$ for all non-extreme training vectors that belong to pattern
class $\boldsymbol{X}_{2} $.

The next section will examine the eigenloci of the $\psi_{1i\ast
}\overrightarrow{\mathbf{e}}_{1i\ast}$ Wolfe dual normal eigenaxis components.

\section{Eigenloci of the $\psi_{1i\ast}\protect\overrightarrow{\mathbf{e}%
}_{1i\ast}$ Wolfe Dual Normal Eigenaxis Components}

Let $i=1:l_{1}$, where the extreme training vector $\mathbf{x}_{1_{i_{\ast}}}$
has the training label $y_{i}=1$. Using Eqs
(\ref{Dual Normal Eigenlocus Component Projections}) and
(\ref{Non-orthogonal Eigenaxes of Dual Normal Eigenlocus}), the eigenlocus of
the $i^{th}$ Wolfe dual normal eigenaxis component $\psi_{1i\ast
}\overrightarrow{\mathbf{e}}_{1i\ast}$ on $\mathbf{\psi}$ is a function of the
expression:%
\begin{align}
\psi_{1i\ast}  &  =\lambda_{\max_{\mathbf{\psi}}}^{-1}\left\Vert
\mathbf{x}_{1_{i_{\ast}}}\right\Vert \sum\nolimits_{j=1}^{l_{1}}\psi
_{1_{j\ast}}\left\Vert \mathbf{x}_{1_{j\ast}}\right\Vert \cos\theta
_{\mathbf{x}_{1_{i\ast}}\mathbf{x}_{1_{j\ast}}}%
\label{Dual Eigen-coordinate Locations Component One}\\
&  -\lambda_{\max_{\mathbf{\psi}}}^{-1}\left\Vert \mathbf{x}_{1_{i_{\ast}}%
}\right\Vert \sum\nolimits_{j=1}^{l_{2}}\psi_{2_{j\ast}}\left\Vert
\mathbf{x}_{2_{j\ast}}\right\Vert \cos\theta_{\mathbf{x}_{1_{i\ast}}%
\mathbf{x}_{2_{j\ast}}}\text{,}\nonumber
\end{align}
where $\psi_{1i\ast}$ provides an eigen-scaling for the non-orthogonal unit
vector $\overrightarrow{\mathbf{e}}_{1i\ast}$. Geometric and statistical
explanations for the eigenlocus statistics%
\begin{equation}
\psi_{1_{j\ast}}\left\Vert \mathbf{x}_{1_{j\ast}}\right\Vert \cos
\theta_{\mathbf{x}_{1_{i\ast}}\mathbf{x}_{1_{j\ast}}}\text{ and }%
\psi_{2_{j\ast}}\left\Vert \mathbf{x}_{2_{j\ast}}\right\Vert \cos
\theta_{\mathbf{x}_{1_{i\ast}}\mathbf{x}_{2_{j\ast}}}\text{,}
\label{Projection Statistics psi1}%
\end{equation}
in Eq. (\ref{Dual Eigen-coordinate Locations Component One}) are considered next.

\subsubsection*{Geometric and Statistical Interpretations of $\psi_{1i\ast
}\protect\overrightarrow{\mathbf{e}}_{1i\ast}$ Eigenlocus Statistics}

The first geometric interpretation of the eigenlocus statistics in Eq.
(\ref{Projection Statistics psi1}) defines the terms:%
\[
\psi_{1_{j\ast}}\text{ and }\psi_{2_{j\ast}}\text{,}%
\]
to be eigen-scales for the signed magnitudes of the vector projections%
\[
\left\Vert \mathbf{x}_{1_{j\ast}}\right\Vert \cos\theta_{\mathbf{x}_{1_{i\ast
}}\mathbf{x}_{1_{j\ast}}}\text{ and }\left\Vert \mathbf{x}_{2_{j\ast}%
}\right\Vert \cos\theta_{\mathbf{x}_{1_{i\ast}}\mathbf{x}_{2_{j\ast}}}\text{,}%
\]
of the eigen-scaled extreme vectors $\psi_{1_{j\ast}}\mathbf{x}_{1_{j\ast}}$
and $\psi_{2_{j\ast}}\mathbf{x}_{2_{j\ast}}$ along the axis of the extreme
vector $\mathbf{x}_{1_{i\ast}}$, where $\cos\theta_{\mathbf{x}_{1_{i\ast}%
}\mathbf{x}_{1_{j\ast}}}$ and $\cos\theta_{\mathbf{x}_{1_{i\ast}}%
\mathbf{x}_{2_{j\ast}}}$ are the angles between the axes of the eigen-scaled
extreme vectors $\psi_{1_{j\ast}}\mathbf{x}_{1_{j\ast}}$ and $\psi_{2_{j\ast}%
}\mathbf{x}_{2_{j\ast}}$ and the axis of the extreme vector $\mathbf{x}%
_{1_{i\ast}}$. Figure $28$ illustrates the geometric and statistical nature of
the eigenlocus statistics in Eqs (\ref{Projection Statistics psi1}).

\begin{center}%
\begin{center}
\includegraphics[
natheight=7.499600in,
natwidth=9.999800in,
height=3.2897in,
width=4.3777in
]%
{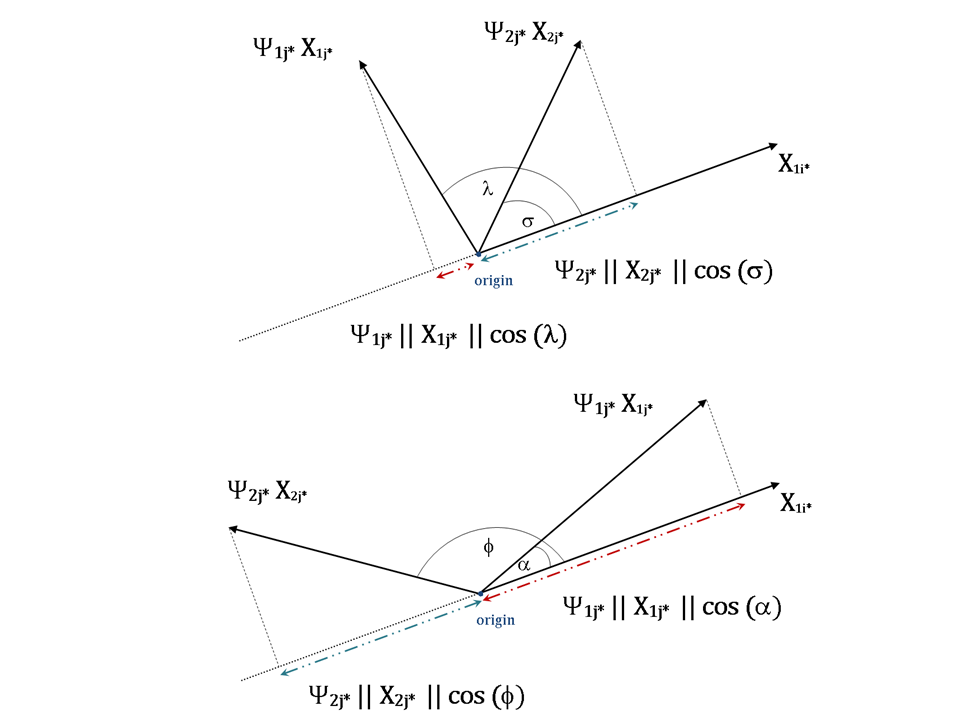}%
\end{center}

\end{center}

\begin{flushleft}
Figure $28$: Illustration of eigen-scaled signed magnitudes of vector
projections of eigen-scaled extreme vectors $\psi_{1_{j\ast}}\mathbf{x}%
_{1_{j\ast}}$ and $\psi_{2_{j\ast}}\mathbf{x}_{2_{j\ast}}$ along the axis of
an extreme vector $\mathbf{x}_{1_{i\ast}}$ which is correlated with a Wolfe
dual normal eigenaxis component $\psi_{1i\ast}\overrightarrow{\mathbf{e}%
}_{1i\ast}$. Any given eigen-scaled signed magnitude $\psi_{1_{j\ast}%
}\left\Vert \mathbf{x}_{1_{j\ast}}\right\Vert \cos\theta_{\mathbf{x}%
_{1_{i\ast}}\mathbf{x}_{1_{j\ast}}}$ or $\psi_{2_{j\ast}}\left\Vert
\mathbf{x}_{2_{j\ast}}\right\Vert \cos\theta_{\mathbf{x}_{1_{i\ast}}%
\mathbf{x}_{2_{j\ast}}}$ may be positive or negative.
\end{flushleft}

\subsubsection*{An Alternative Geometric Interpretation}

An alternative geometric explanation for the eigenlocus statistics in Eq.
(\ref{Projection Statistics psi1}) accounts for the representation of the
$\mathbf{\tau}_{1}$ and $\mathbf{\tau}_{2}$ primal normal eigenlocus
components within the Wolfe dual eigenspace. Consider the algebraic
relationships%
\[
\psi_{1_{j\ast}}\left\Vert \mathbf{x}_{1_{j\ast}}\right\Vert =\left\Vert
\psi_{1_{j\ast}}\mathbf{x}_{1_{j\ast}}\right\Vert =\left\Vert \mathbf{\tau
}_{1}(j)\right\Vert \text{,}%
\]
and%
\[
\psi_{2_{j\ast}}\left\Vert \mathbf{x}_{2_{j\ast}}\right\Vert =\left\Vert
\psi_{2_{j\ast}}\mathbf{x}_{2_{j\ast}}\right\Vert =\left\Vert \mathbf{\tau
}_{2}(j)\right\Vert \text{,}%
\]
where $\mathbf{\tau}_{1}(j)$ and $\mathbf{\tau}_{2}(j)$ are the $j$th
constrained primal normal eigenaxis components on $\mathbf{\tau}_{1}$ and
$\mathbf{\tau}_{2}$. Given the above expressions, it follows that the
eigen-scaled $\psi_{1_{j\ast}}$ signed magnitude $\left\Vert \mathbf{x}%
_{1_{j\ast}}\right\Vert \cos\theta_{\mathbf{x}_{1_{i\ast}}\mathbf{x}%
_{1_{j\ast}}}$ of the vector projection of the eigen-scaled extreme vector
$\psi_{1_{j\ast}}\mathbf{x}_{1_{j\ast}}$ along the axis of the extreme vector
$\mathbf{x}_{1_{i\ast}}$%
\[
\psi_{1_{j\ast}}\left\Vert \mathbf{x}_{1_{j\ast}}\right\Vert \cos
\theta_{\mathbf{x}_{1_{i\ast}}\mathbf{x}_{1_{j\ast}}}\text{,}%
\]
encodes the cosine-scaled $\cos\theta_{\mathbf{x}_{1_{i\ast}}\mathbf{x}%
_{1_{j\ast}}}$ length of the $j$th constrained primal normal eigenaxis
component $\mathbf{\tau}_{1}(j)$ on $\mathbf{\tau}_{1}$%
\[
\cos\theta_{\mathbf{x}_{1_{i\ast}}\mathbf{x}_{1_{j\ast}}}\left\Vert
\mathbf{\tau}_{1}(j)\right\Vert \text{,}%
\]
where $\psi_{1_{j\ast}}$ is the magnitude of the $\psi_{1j\ast}%
\overrightarrow{\mathbf{e}}_{1j\ast}$ Wolfe dual normal eigenaxis component,
and $\cos\theta_{\mathbf{x}_{1_{i\ast}}\mathbf{x}_{1_{j\ast}}} $ encodes the
angle between the extreme training vectors $\mathbf{x}_{1_{i\ast}}$ and
$\mathbf{x}_{1_{j\ast}}$.

Likewise, the eigen-scaled $\psi_{2_{j\ast}}$ signed magnitude $\left\Vert
\mathbf{x}_{2_{j\ast}}\right\Vert \cos\theta_{\mathbf{x}_{1_{i\ast}}%
\mathbf{x}_{2_{j\ast}}}$ of the vector projection of the eigen-scaled extreme
vector $\psi_{2_{j\ast}}\mathbf{x}_{2_{j\ast}}$ along the axis of the extreme
vector $\mathbf{x}_{1_{i\ast}}$%
\[
\psi_{2_{j\ast}}\left\Vert \mathbf{x}_{2_{j\ast}}\right\Vert \cos
\theta_{\mathbf{x}_{1_{i\ast}}\mathbf{x}_{2_{j\ast}}}\text{,}%
\]
encodes the cosine-scaled $\cos\theta_{\mathbf{x}_{1_{i\ast}}\mathbf{x}%
_{2_{j\ast}}}$ length of the $j$th constrained primal normal eigenaxis
component $\mathbf{\tau}_{2}(j)$ on $\mathbf{\tau}_{2}$%
\[
\cos\theta_{\mathbf{x}_{1_{i\ast}}\mathbf{x}_{2_{j\ast}}}\left\Vert
\mathbf{\tau}_{2}(j)\right\Vert \text{,}%
\]
where $\psi_{2_{j\ast}}$ is the magnitude of the $\psi_{2j\ast}%
\overrightarrow{\mathbf{e}}_{2j\ast}$ Wolfe dual normal eigenaxis component,
and $\cos\theta_{\mathbf{x}_{1_{i\ast}}\mathbf{x}_{2_{j\ast}}} $ encodes the
angle between the extreme training vectors $\mathbf{x}_{1_{i\ast}}$ and
$\mathbf{x}_{2_{j\ast}}$.

Given the above analysis, it follows that the eigenlocus of the Wolfe dual
normal eigenaxis component $\psi_{1i\ast}\overrightarrow{\mathbf{e}}_{1i\ast}$
is a function of the constrained primal normal eigenaxis components on
$\mathbf{\tau}_{1}$ and $\mathbf{\tau}_{2}$:%
\begin{align}
\psi_{1i\ast}  &  =\lambda_{\max_{\mathbf{\psi}}}^{-1}\left\Vert
\mathbf{x}_{1_{i_{\ast}}}\right\Vert \sum\nolimits_{j=1}^{l_{1}}\cos
\theta_{\mathbf{x}_{1_{i\ast}}\mathbf{x}_{1_{j\ast}}}\left\Vert \mathbf{\tau
}_{1}(j)\right\Vert \label{Constrained Primal Eigenlocus psi1}\\
&  -\lambda_{\max_{\mathbf{\psi}}}^{-1}\left\Vert \mathbf{x}_{1_{i_{\ast}}%
}\right\Vert \sum\nolimits_{j=1}^{l_{2}}\cos\theta_{\mathbf{x}_{1_{i\ast}%
}\mathbf{x}_{2_{j\ast}}}\left\Vert \mathbf{\tau}_{2}(j)\right\Vert
\text{.}\nonumber
\end{align}
Previous analyses and simulation studies indicate that the constrained primal
normal eigenaxis components on $\mathbf{\tau}_{1}-\mathbf{\tau}_{2}$ account
for a bipartite symmetric partitioning of a region of large covariance that is
well-positioned between a pair of data distributions. The next analysis
examines how uniform geometric and statistical properties which are jointly
exhibited by the Wolfe dual $\psi_{1i\ast}\overrightarrow{\mathbf{e}}_{1i\ast
}$\textbf{\ }and the constrained primal $\psi_{1i\ast}\mathbf{x}_{1_{i\ast}}$
normal eigenaxis components account for this bipartite symmetric partitioning.

\subsection{Uniform Geometric and Statistical Properties Jointly Exhibited by
Normal Eigenaxis Components on $\mathbf{\psi}$ and $\mathbf{\tau}_{1}$}

Using results from the previous analysis, the KKT constraint $\sum
\nolimits_{i=1}^{l_{1}}\psi_{1_{i\ast}}=\sum\nolimits_{i=1}^{l_{2}}%
\psi_{2_{i\ast}}$ of Eq.
(\ref{Equilibrium Constraint on Dual Eigen-components}) indicates that Eq.
(\ref{Dual Eigen-coordinate Locations Component One}) determines an
eigen-balanced, signed magnitude along the axis of an extreme vector
$\mathbf{x}_{1_{i\ast}}$.

Let $\operatorname{comp}_{\overrightarrow{\mathbf{x}_{1i\ast}}}\left(
\overrightarrow{\widetilde{\psi}_{1i\ast}\left\Vert \widetilde{\mathbf{x}%
}_{\ast}\right\Vert _{_{1i_{\ast}}}}\right)  $ denote the eigen-balanced,
signed magnitude%
\begin{align}
\operatorname{comp}_{\overrightarrow{\mathbf{x}_{1i\ast}}}\left(
\overrightarrow{\widetilde{\psi}_{1i\ast}\left\Vert \widetilde{\mathbf{x}%
}_{\ast}\right\Vert _{_{1i_{\ast}}}}\right)   &  =\sum\nolimits_{j=1}^{l_{1}%
}\psi_{1_{j\ast}}\label{Unidirectional Scaling Term One1}\\
&  \times\left[  \left\Vert \mathbf{x}_{1_{j\ast}}\right\Vert \cos
\theta_{\mathbf{x}_{1_{i\ast}}\mathbf{x}_{1_{j\ast}}}\right] \nonumber\\
&  -\sum\nolimits_{j=1}^{l_{2}}\psi_{2_{j\ast}}\nonumber\\
&  \times\left[  \left\Vert \mathbf{x}_{2_{j\ast}}\right\Vert \cos
\theta_{\mathbf{x}_{1_{i\ast}}\mathbf{x}_{2_{j\ast}}}\right]  \text{,}%
\nonumber
\end{align}
along the axis of the extreme vector $\mathbf{x}_{1_{i\ast}}$ that is
correlated with the Wolfe dual normal eigenaxis component $\psi_{1i\ast
}\overrightarrow{\mathbf{e}}_{1i\ast}$.

Given Eqs (\ref{Equilibrium Constraint on Dual Eigen-components}) and
(\ref{Large Unidirectional Pointwise Covariance Estimate Class One}), it
follows that Eq. (\ref{Unidirectional Scaling Term One1}) determines an
eigen-balanced distribution of eigen-scaled first degree coordinates of
extreme training vectors along the axis of $\mathbf{x}_{1_{i\ast}}$; Eqs
(\ref{Equilibrium Constraint on Dual Eigen-components}) and
(\ref{Large Unidirectional Pointwise Covariance Estimate Class One}) also
indicate that Eq. (\ref{Unidirectional Scaling Term One1}) determines an
eigen-balanced first and second order statistical moment about the geometric
locus of $\mathbf{x}_{1_{i\ast}}$.

Furthermore, Eqs (\ref{Equilibrium Constraint on Dual Eigen-components}),
(\ref{Dual Eigen-coordinate Locations Component One}), and
(\ref{Unidirectional Scaling Term One1}) show that joint distributions of the
components of $\mathbf{\psi}$ and $\mathbf{\tau}$ are symmetrically
distributed over the axis of the extreme vector $\mathbf{x}_{1_{i\ast}}$. This
indicates that joint distributions of the components of $\mathbf{\psi}$ and
$\mathbf{\tau}$ are symmetrically distributed over the axis of the Wolfe dual
normal eigenaxis component $\psi_{1i\ast}\overrightarrow{\mathbf{e}}_{1i\ast}$.

Alternatively, given Eq. (\ref{Constrained Primal Eigenlocus psi1}), the
eigen-balanced, signed magnitude in Eq.
(\ref{Unidirectional Scaling Term One1}) depends upon the difference between
integrated cosine-scaled magnitudes of the constrained primal normal eigenaxis
components on $\mathbf{\tau}_{1}$ and $\mathbf{\tau}_{2}$:%
\begin{align}
\operatorname{comp}_{\overrightarrow{\mathbf{x}_{1i\ast}}}\left(
\overrightarrow{\widetilde{\psi}_{1i\ast}\left\Vert \widetilde{\mathbf{x}%
}_{\ast}\right\Vert _{_{1_{i\ast}}}}\right)   &  =\sum\nolimits_{j=1}^{l_{1}%
}\cos\theta_{\mathbf{x}_{1_{i\ast}}\mathbf{x}_{1_{j\ast}}}\left\Vert
\mathbf{\tau}_{1}(j)\right\Vert \label{Unidirectional Scaling Term One2}\\
&  -\sum\nolimits_{j=1}^{l_{2}}\cos\theta_{\mathbf{x}_{1_{i\ast}}%
\mathbf{x}_{2_{j\ast}}}\left\Vert \mathbf{\tau}_{2}(j)\right\Vert
\text{,}\nonumber
\end{align}
which also shows that joint distributions of the components of $\mathbf{\psi}$
and $\mathbf{\tau}$ are symmetrically distributed over the axes of both
$\psi_{1i\ast}\overrightarrow{\mathbf{e}}_{1i\ast}$ and $\mathbf{x}_{1_{i\ast
}}$.

Using Eqs (\ref{Dual Eigen-coordinate Locations Component One}) and
(\ref{Unidirectional Scaling Term One1}), it follows that the length
$\psi_{1i\ast}$ of the Wolfe dual normal eigenaxis component $\psi_{1i\ast
}\overrightarrow{\mathbf{e}}_{1i\ast}$ is determined by a weighted length of
the correlated extreme training vector $\mathbf{x}_{1_{i\ast}}$%
\begin{equation}
\psi_{1i\ast}=\left[  \lambda_{\max_{\mathbf{\psi}}}^{-1}\times
\operatorname{comp}_{\overrightarrow{\mathbf{x}_{1i\ast}}}\left(
\overrightarrow{\widetilde{\psi}_{1i\ast}\left\Vert \widetilde{\mathbf{x}%
}_{\ast}\right\Vert _{_{1i_{\ast}}}}\right)  \right]  \left\Vert
\mathbf{x}_{1_{i\ast}}\right\Vert \text{,}
\label{Magnitude Dual Normal Eigenaxis Component Class One}%
\end{equation}
where the weighting factor encodes an eigenvalue $\lambda_{\max_{\mathbf{\psi
}}}^{-1}$ scaling of an eigen-balanced, signed magnitude $\operatorname{comp}%
_{\overrightarrow{\mathbf{x}_{1i\ast}}}\left(  \overrightarrow{\widetilde{\psi
}_{1i\ast}\left\Vert \widetilde{\mathbf{x}}_{\ast}\right\Vert _{_{1i_{\ast}}}%
}\right)  $ along the axis of $\mathbf{x}_{1_{i\ast}}$.

Given that $\psi_{1i\ast}>0$, $\lambda_{\max_{\mathbf{\psi}}}^{-1}>0$, and
$\left\Vert \mathbf{x}_{1_{i\ast}}\right\Vert >0$, it follows that the
eigen-balanced, signed magnitude along the axis of $\mathbf{x}_{1_{i\ast}}$ is
positive%
\[
\operatorname{comp}_{\overrightarrow{\mathbf{x}_{1i\ast}}}\left(
\overrightarrow{\widetilde{\psi}_{1i\ast}\left\Vert \widetilde{\mathbf{x}%
}_{\ast}\right\Vert _{_{1i_{\ast}}}}\right)  >0\text{,}%
\]
which indicates that the weighting factor in Eq.
(\ref{Magnitude Dual Normal Eigenaxis Component Class One}) determines an
eigen-balanced length%
\[
\lambda_{\max_{\mathbf{\psi}}}^{-1}\operatorname{comp}%
_{\overrightarrow{\mathbf{x}_{1i\ast}}}\left(  \overrightarrow{\widetilde{\psi
}_{1i\ast}\left\Vert \widetilde{\mathbf{x}}_{\ast}\right\Vert _{_{1i_{\ast}}}%
}\right)  \left\Vert \mathbf{x}_{1_{i\ast}}\right\Vert \text{,}%
\]
for the extreme vector $\mathbf{x}_{1_{i\ast}}$. Therefore, Eq.
(\ref{Magnitude Dual Normal Eigenaxis Component Class One}) determines an
eigen-balanced length for both $\psi_{1i\ast}\overrightarrow{\mathbf{e}%
}_{1i\ast}$ and $\mathbf{x}_{1_{i\ast}}$.

Returning to Eqs
(\ref{Large Unidirectional Pointwise Covariance Estimate Class One}) and
(\ref{Dual Eigen-coordinate Locations Component One}), it follows that the
length $\psi_{1i\ast}$ of the Wolfe dual normal eigenaxis component
$\psi_{1i\ast}\overrightarrow{\mathbf{e}}_{1i\ast}$ on $\mathbf{\psi}$%
\[
\psi_{1i\ast}=\lambda_{\max_{\mathbf{\psi}}}^{-1}\operatorname{comp}%
_{\overrightarrow{\mathbf{x}_{1i\ast}}}\left(  \overrightarrow{\widetilde{\psi
}_{1i\ast}\left\Vert \widetilde{\mathbf{x}}_{\ast}\right\Vert _{_{1i_{\ast}}}%
}\right)  \left\Vert \mathbf{x}_{1_{i\ast}}\right\Vert \text{,}%
\]
is shaped by an eigen-balanced first and second order statistical moment about
the geometric locus of the correlated extreme vector $\mathbf{x}_{1_{i\ast}}$.

Now, take any given Wolfe dual $\psi_{1i\ast}\overrightarrow{\mathbf{e}%
}_{1i\ast}$ and correlated constrained primal $\psi_{1_{i\ast}}\mathbf{x}%
_{1_{i\ast}}$ normal eigenaxis component. It will now be shown that the
direction of $\psi_{1i\ast}\overrightarrow{\mathbf{e}}_{1i\ast}$ is identical
to the direction of $\psi_{1_{i\ast}}\mathbf{x}_{1_{i\ast}}$.

\subsection{Directional Symmetries Exhibited by Normal Eigenaxis Components on
$\mathbf{\psi}$ and $\mathbf{\tau}_{1}$}

The vector direction of the Wolfe dual $\psi_{1i\ast}%
\overrightarrow{\mathbf{e}}_{1i\ast}$ normal eigenaxis component is implicitly
specified by Eq. (\ref{Dual Eigen-coordinate Locations Component One}), where
it has been assumed that $\psi_{1i\ast}$ provides an eigen-scaling for a
non-orthogonal unit vector $\overrightarrow{\mathbf{e}}_{1i\ast}$. Given Eqs
(\ref{Pointwise Covariance Statistic}) and
(\ref{Large Unidirectional Pointwise Covariance Estimate Class One}), it
follows that the eigen-balanced pointwise covariance statistic in Eq.
(\ref{Dual Eigen-coordinate Locations Component One}) encodes the direction of
the extreme vector $\mathbf{x}_{1_{i_{\ast}}}$ and an eigen-balanced magnitude
along the axis of the extreme vector $\mathbf{x}_{1_{i_{\ast}}}$.

Returning to Eqs (\ref{Unidirectional Scaling Term One1}),
(\ref{Unidirectional Scaling Term One2}), and
(\ref{Magnitude Dual Normal Eigenaxis Component Class One}), take any given
Wolfe dual normal eigenaxis component $\psi_{1i\ast}\overrightarrow{\mathbf{e}%
}_{1i\ast}$ that is correlated with an extreme vector $\mathbf{x}_{1_{i_{\ast
}}}$. Given that the length $\psi_{1i\ast}$ of the Wolfe dual normal eigenaxis
component $\psi_{1i\ast}\overrightarrow{\mathbf{e}}_{1i\ast}$ is determined by
the eigen-balanced length of the extreme vector $\mathbf{x}_{1_{i_{\ast}}}$%
\[
\psi_{1i\ast}=\lambda_{\max_{\mathbf{\psi}}}^{-1}\operatorname{comp}%
_{\overrightarrow{\mathbf{x}_{1i\ast}}}\left(  \overrightarrow{\widetilde{\psi
}_{1i\ast}\left\Vert \widetilde{\mathbf{x}}_{\ast}\right\Vert _{_{1i_{\ast}}}%
}\right)  \left\Vert \mathbf{x}_{1_{i\ast}}\right\Vert \text{,}%
\]
it follows that the non-orthogonal unit vector $\overrightarrow{\mathbf{e}%
}_{1i\ast}$ has the same direction as the extreme vector $\mathbf{x}%
_{1_{i_{\ast}}}$%
\[
\overrightarrow{\mathbf{e}}_{1i\ast}\equiv\frac{\mathbf{x}_{1_{i_{\ast}}}%
}{\left\Vert \mathbf{x}_{1_{i_{\ast}}}\right\Vert }\text{.}%
\]
Thus, the direction of the Wolfe dual normal eigenaxis component $\psi
_{1i\ast}\overrightarrow{\mathbf{e}}_{1i\ast}$ is identical to the direction
of the constrained primal normal eigenaxis component $\psi_{1_{i\ast}%
}\mathbf{x}_{1_{i\ast}}$, which is determined by the direction of the
eigen-scaled $\psi_{1i\ast}$ extreme training vector $\mathbf{x}_{1_{i\ast}}$.
The Wolfe dual $\psi_{1i\ast}\overrightarrow{\mathbf{e}}_{1i\ast}$ and the
constrained primal $\psi_{1i\ast}\mathbf{x}_{1_{i\ast}}$ normal eigenaxis
components are said to exhibit directional symmetry. Accordingly, each Wolfe
dual $\psi_{1i\ast}\overrightarrow{\mathbf{e}}_{1i\ast}$ and correlated
constrained primal $\psi_{1i\ast}\mathbf{x}_{1_{i\ast}}$ normal eigenaxis
component exhibit directional symmetry.

It is concluded that the uniform directions of the Wolfe dual $\psi_{1i\ast
}\overrightarrow{\mathbf{e}}_{1i\ast}$ and the constrained primal
$\psi_{1_{i\ast}}\mathbf{x}_{1_{i\ast}}$ normal eigenaxis components determine
critical directions of large covariance, which contribute to a symmetric
partitioning of a minimal geometric region of constant width that spans a
region of large covariance between the distributions of two classes of
training data. It is also concluded that each of the correlated normal
eigenaxis components on $\mathbf{\psi}$ and $\mathbf{\tau}_{1}$ possess
critical lengths for which the constrained discriminant function $D\left(
\mathbf{x}\right)  =\mathbf{x}^{T}\mathbf{\tau}+\tau_{0}$ delineates centrally
located, bipartite, symmetric regions of large covariance between two data
distributions. Expressions for the integrated lengths $\sum\nolimits_{i=1}%
^{l_{1}}\psi_{1i\ast}$ of the $\psi_{1i\ast}\overrightarrow{\mathbf{e}%
}_{1i\ast}$ components are obtained next.

\subsection*{Integrated Lengths of $\psi_{1i\ast}%
\protect\overrightarrow{\mathbf{e}}_{1i\ast}$ Components on $\mathbf{\psi}$}

Using Eq. (\ref{Dual Eigen-coordinate Locations Component One}), an expression
is obtained for the integrated lengths $\sum\nolimits_{i=1}^{l_{1}}%
\psi_{1i\ast}$ of the $\psi_{1i\ast}\overrightarrow{\mathbf{e}}_{1i\ast}$
Wolfe dual normal eigenaxis components on $\mathbf{\psi}$ :%
\begin{align}
\sum\nolimits_{i=1}^{l_{1}}\psi_{1i\ast}  &  =\lambda_{\max_{\mathbf{\psi}}%
}^{-1}\sum\nolimits_{i=1}^{l_{1}}\left\Vert \mathbf{x}_{1_{i\ast}}\right\Vert
\label{integrated dual loci one1}\\
&  \times\sum\nolimits_{j=1}^{l_{1}}\psi_{1_{j\ast}}\left\Vert \mathbf{x}%
_{1_{j\ast}}\right\Vert \cos\theta_{\mathbf{x}_{1_{i\ast}}\mathbf{x}%
_{1_{j\ast}}}\nonumber\\
&  -\lambda_{\max_{\mathbf{\psi}}}^{-1}\sum\nolimits_{i=1}^{l_{1}}\left\Vert
\mathbf{x}_{1_{i\ast}}\right\Vert \nonumber\\
&  \times\sum\nolimits_{j=1}^{l_{2}}\psi_{2_{j\ast}}\left\Vert \mathbf{x}%
_{2_{j\ast}}\right\Vert \cos\theta_{\mathbf{x}_{1_{i\ast}}\mathbf{x}%
_{2_{j\ast}}}\text{,}\nonumber
\end{align}
where the direction of the Wolfe dual normal eigenaxis component $\psi
_{1i\ast}\overrightarrow{\mathbf{e}}_{1i\ast}$ on $\mathbf{\psi}\in%
\mathbb{R}
^{N}$ has the same direction as the correlated extreme training vector
$\mathbf{x}_{1_{i\ast}}\in%
\mathbb{R}
^{d}$.

Alternatively, Eq. (\ref{Constrained Primal Eigenlocus psi1}) provides the
expression%
\begin{align}
\sum\nolimits_{i=1}^{l_{1}}\psi_{1i\ast}  &  =\lambda_{\max_{\mathbf{\psi}}%
}^{-1}\sum\nolimits_{i=1}^{l_{1}}\left\Vert \mathbf{x}_{1_{i\ast}}\right\Vert
\label{integrated dual loci one2}\\
&  \times\sum\nolimits_{j=1}^{l_{1}}\cos\theta_{\mathbf{x}_{1_{i\ast}%
}\mathbf{x}_{1_{j\ast}}}\left\Vert \mathbf{\tau}_{1}(j)\right\Vert \nonumber\\
&  -\lambda_{\max_{\mathbf{\psi}}}^{-1}\sum\nolimits_{i=1}^{l_{1}}\left\Vert
\mathbf{x}_{1_{i\ast}}\right\Vert \nonumber\\
&  \times\sum\nolimits_{j=1}^{l_{2}}\cos\theta_{\mathbf{x}_{1_{i\ast}%
}\mathbf{x}_{2_{j\ast}}}\left\Vert \mathbf{\tau}_{2}(j)\right\Vert
\text{;}\nonumber
\end{align}
Eqs (\ref{integrated dual loci one1}) and (\ref{integrated dual loci one2})
will be used to examine the algebraic and geometric nature of the statistical
equilibrium state that is implied by Eq.
(\ref{Equilibrium Constraint on Dual Eigen-components}). The uniform geometric
and statistical properties which are jointly exhibited by the Wolfe dual
$\psi_{1i\ast}\overrightarrow{\mathbf{e}}_{1i\ast}$ and the constrained primal
$\psi_{1_{i\ast}}\mathbf{x}_{1_{i\ast}}$ normal eigenaxis components are
summarized below.

\subsection*{Summary of Uniform Geometric and Statistical Properties Jointly
Exhibited by Normal Eigenaxis Components on $\mathbf{\psi}$ and $\mathbf{\tau
}_{1}$}

Results of the previous analysis are now used to identify uniform geometric
and statistical properties which are jointly exhibited by the $\psi_{1i\ast
}\overrightarrow{\mathbf{e}}_{1i\ast}$ Wolfe dual normal eigenaxis components
on $\mathbf{\mathbf{\psi}}$ and the $\psi_{1_{i\ast}}\mathbf{x}_{1_{i\ast}}$
constrained primal normal eigenaxis components on $\mathbf{\tau}_{1}$. The
properties are summarized below.

\begin{conclusion}
The direction of each Wolfe dual normal eigenaxis component $\psi_{1i\ast
}\overrightarrow{\mathbf{e}}_{1i\ast}$ on $\mathbf{\mathbf{\psi}\in%
\mathbb{R}
}^{N}$ is identical to the direction of a constrained primal normal eigenaxis
component $\psi_{1_{i\ast}}\mathbf{x}_{1_{i\ast}}$ on$\mathbf{\ \mathbf{\tau
}_{1}\in%
\mathbb{R}
}^{d}$.
\end{conclusion}

\begin{conclusion}
The lengths of each Wolfe dual normal eigenaxis component $\psi_{1i\ast
}\overrightarrow{\mathbf{e}}_{1i\ast}$ on $\mathbf{\mathbf{\psi}\in%
\mathbb{R}
}^{N}$ and correlated constrained primal normal eigenaxis component
$\psi_{1_{i\ast}}\mathbf{x}_{1_{i\ast}}$ on$\mathbf{\ \mathbf{\tau}_{1}\in%
\mathbb{R}
}^{d}$ are shaped by identical joint symmetrical distributions of normal
eigenaxis components on $\mathbf{\psi}$ and $\mathbf{\tau}$.
\end{conclusion}

\begin{conclusion}
The length $\psi_{1i\ast}$ of each Wolfe dual normal eigenaxis component
$\psi_{1i\ast}\overrightarrow{\mathbf{e}}_{1i\ast}$ on $\mathbf{\mathbf{\psi
}\in%
\mathbb{R}
}^{N}$
\[
\psi_{1i\ast}=\lambda_{\max_{\mathbf{\psi}}}^{-1}\operatorname{comp}%
_{\overrightarrow{\mathbf{x}_{1i\ast}}}\left(  \overrightarrow{\widetilde{\psi
}_{1i\ast}\left\Vert \widetilde{\mathbf{x}}_{\ast}\right\Vert _{_{1i_{\ast}}}%
}\right)  \left\Vert \mathbf{x}_{1_{i\ast}}\right\Vert \text{,}%
\]
is shaped by an eigen-balanced pointwise covariance estimate%
\begin{align*}
\widehat{\operatorname{cov}}_{up_{\updownarrow}}\left(  \mathbf{x}%
_{1_{i_{\ast}}}\right)   &  =\lambda_{\max_{\mathbf{\psi}}}^{-1}\left\Vert
\mathbf{x}_{1_{i_{\ast}}}\right\Vert \\
&  \times\sum\nolimits_{j=1}^{l_{1}}\psi_{1_{j\ast}}\left\Vert \mathbf{x}%
_{1_{j\ast}}\right\Vert \cos\theta_{\mathbf{x}_{1_{i\ast}}\mathbf{x}%
_{1_{j\ast}}}\\
&  -\lambda_{\max_{\mathbf{\psi}}}^{-1}\left\Vert \mathbf{x}_{1_{i_{\ast}}%
}\right\Vert \\
&  \times\sum\nolimits_{j=1}^{l_{2}}\psi_{2_{j\ast}}\left\Vert \mathbf{x}%
_{2_{j\ast}}\right\Vert \cos\theta_{\mathbf{x}_{1_{i\ast}}\mathbf{x}%
_{2_{j\ast}}}\text{,}%
\end{align*}
for a correlated extreme training vector\textbf{\ }$\mathbf{x}_{1_{i\ast}%
}\mathbf{\in%
\mathbb{R}
}^{d}$, such that the eigenlocus of each constrained primal normal eigenaxis
component\textbf{\ }$\psi_{1_{i\ast}}\mathbf{x}_{1_{i\ast}}$\textbf{\ }on
$\mathbf{\mathbf{\tau}_{1}\in%
\mathbb{R}
}^{d}$\textbf{\ }provides a maximum\textit{\ }covariance estimate in a
principal location, in the form of an eigen-balanced first and second order
statistical moment about the geometric locus of an extreme data point
$\mathbf{x}_{1_{i\ast}}$.
\end{conclusion}

\begin{conclusion}
Each Wolfe dual normal eigenaxis component\textbf{\ }$\psi_{1i\ast
}\overrightarrow{\mathbf{e}}_{1i\ast}$ on $\mathbf{\mathbf{\psi}}$%
\textbf{\ }encodes an eigen-balanced first and second order statistical moment
about the locus of a correlated extreme data point $\mathbf{x}_{1_{i\ast}}$,
relative to the eigenloci of all of the eigen-scaled extreme training points,
which determines the eigenlocus of a constrained primal normal eigenaxis
component $\psi_{1_{i\ast}}\mathbf{x}_{1_{i\ast}}$ on $\mathbf{\tau}_{1}$.
\end{conclusion}

\begin{conclusion}
Any given eigen-balanced pointwise covariance estimate
$\widehat{\operatorname{cov}}_{up_{\updownarrow}}\left(  \mathbf{x}%
_{1_{i_{\ast}}}\right)  $ encodes a distribution of first order coordinates
for an extreme training vector $\mathbf{x}_{1_{i\ast}}$, relative to the
eigen-scaled extreme training vectors for a given data set. The distribution
of first order coordinates for $\mathbf{x}_{1_{i\ast}}$ describes how the
components of $\mathbf{x}_{1_{i\ast}}$ are distributed within the given
collection of eigen-scaled extreme vectors.
\end{conclusion}

\begin{conclusion}
Returning to Figs $12$ and $23$, the integrated eigenloci of the $\psi
_{1i\ast}\overrightarrow{\mathbf{e}}_{1i\ast}$ Wolfe dual normal eigenaxis
components jointly and implicitly specify the geometric locus of a hyperplane
decision border $H_{+1}$. Likewise, the integrated eigenloci of the
$\psi_{1i\ast}\overrightarrow{\mathbf{e}}_{1i\ast}$ Wolfe dual normal
eigenaxis components jointly and implicitly account for a symmetric
partitioning of a minimal area surface of large covariance that is delineated
by a hyperplane decision boundary $H_{0}$ which is symmetrically located
between a pair of hyperplane decision borders $H_{+1}$ and $H_{-1}$.

\begin{claim}
The square $\left\Vert \psi_{1_{i\ast}}\mathbf{x}_{1_{i\ast}}\right\Vert
_{\min_{c}}^{2}$ of the constrained primal normal eigenaxis
component\textbf{\ }$\psi_{1_{i\ast}}\mathbf{x}_{1_{i\ast}}$\textbf{\ }on
$\mathbf{\mathbf{\tau}}_{1}$\textbf{\ }$\mathbf{\in%
\mathbb{R}
}^{d}$ is the probability of finding the extreme data point $\mathbf{x}%
_{1_{i\ast}}$ in a particular region of $\mathbf{%
\mathbb{R}
}^{d}$, where $\left\Vert \psi_{1_{i\ast}}\mathbf{x}_{1_{i\ast}}\right\Vert
_{\min_{c}}^{2}$ is the total allowed eigenenergy of $\psi_{1_{i\ast}%
}\mathbf{x}_{1_{i\ast}}$.
\end{claim}
\end{conclusion}

The geometric and statistical properties of the eigenloci of the $\psi
_{2i\ast}\overrightarrow{\mathbf{e}}_{2i\ast}$ Wolfe dual normal eigenaxis
components on $\mathbf{\mathbf{\psi}}$ are examined in the next section.

\section{Eigenloci of the $\psi_{2i\ast}\protect\overrightarrow{\mathbf{e}%
}_{2i\ast}$ Wolfe Dual Normal Eigenaxis Components}

Let $i=1:l_{2}$, where the extreme pattern vector $\mathbf{x}_{2_{i_{\ast}}} $
has the training label $y_{i}=-1$. Using Eqs
(\ref{Dual Normal Eigenlocus Component Projections}) and
(\ref{Non-orthogonal Eigenaxes of Dual Normal Eigenlocus}), the eigenlocus of
the $i^{th}$ Wolfe dual normal eigenaxis component $\psi_{2i\ast
}\overrightarrow{\mathbf{e}}_{2i\ast}$ on $\mathbf{\psi}$ is a function of the
expression:%
\begin{align}
\psi_{2i\ast}  &  =\lambda_{\max_{\mathbf{\psi}}}^{-1}\left\Vert
\mathbf{x}_{2_{i\ast}}\right\Vert \sum\nolimits_{j=1}^{l_{2}}\psi_{2_{j\ast}%
}\left\Vert \mathbf{x}_{2_{j\ast}}\right\Vert \cos\theta_{\mathbf{x}%
_{2_{i\ast}}\mathbf{x}_{2_{j\ast}}}%
\label{Dual Eigen-coordinate Locations Component Two}\\
&  -\lambda_{\max_{\mathbf{\psi}}}^{-1}\left\Vert \mathbf{x}_{2_{i\ast}%
}\right\Vert \sum\nolimits_{j=1}^{l_{1}}\psi_{1_{j\ast}}\left\Vert
\mathbf{x}_{1_{j\ast}}\right\Vert \cos\theta_{\mathbf{x}_{2_{i\ast}}%
\mathbf{x}_{1_{j\ast}}}\text{,}\nonumber
\end{align}
where $\psi_{2i\ast}$ provides an eigen-scaling for the non-orthogonal unit
vector $\overrightarrow{\mathbf{e}}_{2i\ast}$. Geometric and statistical
explanations for the eigenlocus statistics:%
\begin{equation}
\psi_{2_{j\ast}}\left\Vert \mathbf{x}_{2_{j\ast}}\right\Vert \cos
\theta_{\mathbf{x}_{2_{i\ast}}\mathbf{x}_{2_{j\ast}}}\text{ and }%
\psi_{1_{j\ast}}\left\Vert \mathbf{x}_{1_{j\ast}}\right\Vert \cos
\theta_{\mathbf{x}_{2_{i\ast}}\mathbf{x}_{1_{j\ast}}}
\label{Projection Statistics psi2}%
\end{equation}
in Eq. (\ref{Dual Eigen-coordinate Locations Component Two}) are considered next.

\subsubsection*{Geometric and Statistical Explanations of $\psi_{2i\ast
}\protect\overrightarrow{\mathbf{e}}_{2i\ast}$ Eigenlocus Statistics}

The first geometric interpretation of the eigenlocus statistics in Eq.
(\ref{Projection Statistics psi2}) defines the terms:%
\[
\psi_{2_{j\ast}}\text{ and }\psi_{1_{j\ast}}\text{,}%
\]
to be eigen-scales for the signed magnitudes of the vector projections:
\[
\left\Vert \mathbf{x}_{2_{j\ast}}\right\Vert \cos\theta_{\mathbf{x}_{2_{i\ast
}}\mathbf{x}_{2_{j\ast}}}\text{ and }\left\Vert \mathbf{x}_{1_{j\ast}%
}\right\Vert \cos\theta_{\mathbf{x}_{2_{i\ast}}\mathbf{x}_{1_{j\ast}}}\text{,}%
\]
of the eigen-scaled extreme vectors $\psi_{2_{j\ast}}\mathbf{x}_{2_{j\ast}}$
and $\psi_{1_{j\ast}}\mathbf{x}_{1_{j\ast}}$ along the axis of the extreme
vector $\mathbf{x}_{2_{i\ast}}$, where $\cos\theta_{\mathbf{x}_{2_{i\ast}%
}\mathbf{x}_{2_{j\ast}}}$ and $\cos\theta_{\mathbf{x}_{2_{i\ast}}%
\mathbf{x}_{1_{j\ast}}}$ are the angles between the axes of the eigen-scaled
extreme vectors $\psi_{2_{j\ast}}\mathbf{x}_{2_{j\ast}}$ and $\psi_{1_{j\ast}%
}\mathbf{x}_{1_{j\ast}}$ and the axis of the extreme vector $\mathbf{x}%
_{2_{i\ast}}$.

\subsubsection*{An Alternative Geometric Interpretation}

An alternative geometric explanation for the eigenlocus statistics in Eq.
(\ref{Projection Statistics psi2}) accounts for the representation of the
$\mathbf{\tau}_{1}$ and $\mathbf{\tau}_{2}$ primal normal eigenlocus
components within the Wolfe dual eigenspace. Consider the algebraic
relationships%
\[
\psi_{1_{j\ast}}\left\Vert \mathbf{x}_{1_{j\ast}}\right\Vert =\left\Vert
\psi_{1_{j\ast}}\mathbf{x}_{1_{j\ast}}\right\Vert =\left\Vert \mathbf{\tau
}_{1}(j)\right\Vert \text{,}%
\]
and%
\[
\psi_{2_{j\ast}}\left\Vert \mathbf{x}_{2_{j\ast}}\right\Vert =\left\Vert
\psi_{2_{j\ast}}\mathbf{x}_{2_{j\ast}}\right\Vert =\left\Vert \mathbf{\tau
}_{2}(j)\right\Vert \text{,}%
\]
where $\mathbf{\tau}_{1}(j)$ and $\mathbf{\tau}_{2}(j)$ are the $j$th
constrained primal normal eigenaxis components on $\mathbf{\tau}_{1}$ and
$\mathbf{\tau}_{2}$. Given the above expressions, it follows that the
eigen-scaled $\psi_{2_{j\ast}}$ signed magnitude $\left\Vert \mathbf{x}%
_{2_{j\ast}}\right\Vert \cos\theta_{\mathbf{x}_{2_{i\ast}}\mathbf{x}%
_{2_{j\ast}}}$ of the vector projection of the eigen-scaled extreme vector
$\psi_{2_{j\ast}}\mathbf{x}_{2_{j\ast}}$ along the axis of the extreme vector
$\mathbf{x}_{2_{i\ast}}$%
\[
\psi_{2_{j\ast}}\left\Vert \mathbf{x}_{2_{j\ast}}\right\Vert \cos
\theta_{\mathbf{x}_{2_{i\ast}}\mathbf{x}_{2_{j\ast}}}\text{,}%
\]
encodes the cosine-scaled $\cos\theta_{\mathbf{x}_{2_{i\ast}}\mathbf{x}%
_{2_{j\ast}}}$ length of the $j$th constrained primal normal eigenaxis
component $\mathbf{\tau}_{2}(j)$ on $\mathbf{\tau}_{2}$%
\[
\cos\theta_{\mathbf{x}_{2_{i\ast}}\mathbf{x}_{2_{j\ast}}}\left\Vert
\mathbf{\tau}_{2}(j)\right\Vert \text{,}%
\]
where $\psi_{2_{j\ast}}$ is the magnitude of the $\psi_{2j\ast}%
\overrightarrow{\mathbf{e}}_{2j\ast}$ Wolfe dual normal eigenaxis component,
and $\cos\theta_{\mathbf{x}_{2_{i\ast}}\mathbf{x}_{2_{j\ast}}} $ encodes the
angle between the extreme training vectors $\mathbf{x}_{2_{i\ast}}$ and
$\mathbf{x}_{2_{j\ast}}$. Likewise, the eigen-scaled $\psi_{1_{j\ast}}$ signed
magnitude $\left\Vert \mathbf{x}_{1_{j\ast}}\right\Vert \cos\theta
_{\mathbf{x}_{2_{i\ast}}\mathbf{x}_{1_{j\ast}}}$ of the vector projection of
the the eigen-scaled extreme vector $\psi_{1_{j\ast}}\mathbf{x}_{1_{j\ast}}$
along the axis of the extreme vector $\mathbf{x}_{2_{i\ast}}$%
\[
\psi_{1_{j\ast}}\left\Vert \mathbf{x}_{1_{j\ast}}\right\Vert \cos
\theta_{\mathbf{x}_{2_{i\ast}}\mathbf{x}_{1_{j\ast}}}\text{,}%
\]
encodes the cosine-scaled $\cos\theta_{\mathbf{x}_{2_{i\ast}}\mathbf{x}%
_{1_{j\ast}}}$ length of the $j$th constrained primal normal eigenaxis
component $\mathbf{\tau}_{1}(j)$ on $\mathbf{\tau}_{1}$%
\[
\cos\theta_{\mathbf{x}_{2_{i\ast}}\mathbf{x}_{1_{j\ast}}}\left\Vert
\mathbf{\tau}_{1}(j)\right\Vert \text{,}%
\]
where $\psi_{1_{j\ast}}$ is the magnitude of the $\psi_{1j\ast}%
\overrightarrow{\mathbf{e}}_{1j\ast}$ Wolfe dual normal eigenaxis component,
and $\cos\theta_{\mathbf{x}_{2_{i\ast}}\mathbf{x}_{1_{j\ast}}} $ encodes the
angle between the extreme training vectors $\mathbf{x}_{2_{i\ast}}$ and
$\mathbf{x}_{1_{j\ast}}$.

It follows that the eigenlocus of the Wolfe dual normal eigenaxis component
$\psi_{2i\ast}\overrightarrow{\mathbf{e}}_{2i\ast}$ is a function of the
constrained primal normal eigenaxis components on $\mathbf{\tau}_{1}$ and
$\mathbf{\tau}_{2}$:%
\begin{align}
\psi_{2i\ast}  &  =\lambda_{\max_{\mathbf{\psi}}}^{-1}\left\Vert
\mathbf{x}_{2_{i\ast}}\right\Vert \sum\nolimits_{j=1}^{l_{2}}\cos
\theta_{\mathbf{x}_{2_{i\ast}}\mathbf{x}_{2_{j\ast}}}\left\Vert \mathbf{\tau
}_{2}(j)\right\Vert \label{Constrained Primal Eigenlocus psi2}\\
&  -\lambda_{\max_{\mathbf{\psi}}}^{-1}\left\Vert \mathbf{x}_{2_{i\ast}%
}\right\Vert \sum\nolimits_{j=1}^{l_{1}}\cos\theta_{\mathbf{x}_{2_{i\ast}%
}\mathbf{x}_{1_{j\ast}}}\left\Vert \mathbf{\tau}_{1}(j)\right\Vert
\text{.}\nonumber
\end{align}
The next analysis considers how uniform geometric and statistical properties
which are jointly exhibited by the Wolfe dual $\psi_{2i\ast}%
\overrightarrow{\mathbf{e}}_{2i\ast}$\textbf{\ }and the constrained primal
$\psi_{2i\ast}\mathbf{x}_{2_{i\ast}}$ normal eigenaxis components account for
a bipartite symmetric partitioning of a region of large covariance that is
well-positioned between a pair of data distributions.

\subsection{Uniform Geometric and Statistical Properties Jointly Exhibited by
Normal Eigenaxis Components on $\mathbf{\psi}$ and $\mathbf{\tau}_{2}$}

Using results from the previous analysis, the KKT constraint $\sum
\nolimits_{i=1}^{l_{1}}\psi_{1_{i\ast}}=\sum\nolimits_{i=1}^{l_{2}}%
\psi_{2_{i\ast}}$ of Eq.
(\ref{Equilibrium Constraint on Dual Eigen-components}) indicates that Eq.
(\ref{Dual Eigen-coordinate Locations Component Two}) determines an
eigen-balanced, signed magnitude along the axis of an extreme vector
$\mathbf{x}_{2_{i\ast}}$.

Let $\operatorname{comp}_{\overrightarrow{\mathbf{x}_{2i\ast}}}\left(
\overrightarrow{\widetilde{\psi}_{2i\ast}\left\Vert \widetilde{\mathbf{x}%
}_{\ast}\right\Vert _{_{2i_{\ast}}}}\right)  $ denote the eigen-balanced,
signed magnitude
\begin{align}
\operatorname{comp}_{\overrightarrow{\mathbf{x}_{2i\ast}}}\left(
\overrightarrow{\widetilde{\psi}_{2i\ast}\left\Vert \widetilde{\mathbf{x}%
}_{\ast}\right\Vert _{_{2i_{\ast}}}}\right)   &  =\sum\nolimits_{j=1}^{l_{2}%
}\psi_{2_{j\ast}}\label{Unidirectional Scaling Term Two1}\\
&  \times\left[  \left\Vert \mathbf{x}_{2_{j\ast}}\right\Vert \cos
\theta_{\mathbf{x}_{2_{i\ast}}\mathbf{x}_{2_{j\ast}}}\right] \nonumber\\
&  -\sum\nolimits_{j=1}^{l_{1}}\psi_{1_{j\ast}}\nonumber\\
&  \times\left[  \left\Vert \mathbf{x}_{1_{j\ast}}\right\Vert \cos
\theta_{\mathbf{x}_{2_{i\ast}}\mathbf{x}_{1_{j\ast}}}\right]  \text{,}%
\nonumber
\end{align}
along the axis of the extreme training vector $\mathbf{x}_{2_{i\ast}}$ that is
correlated with the Wolfe dual normal eigenaxis component $\psi_{2i\ast
}\overrightarrow{\mathbf{e}}_{2i\ast}$.

Given Eqs (\ref{Equilibrium Constraint on Dual Eigen-components}) and
(\ref{Large Unidirectional Pointwise Covariance Estimate Class Two}), it
follows that Eq. (\ref{Unidirectional Scaling Term Two1}) determines an
eigen-balanced distribution of eigen-scaled first degree coordinates of
extreme training vectors along the axis of $\mathbf{x}_{2_{i\ast}}$; Eqs
(\ref{Equilibrium Constraint on Dual Eigen-components}) and
(\ref{Large Unidirectional Pointwise Covariance Estimate Class Two}) also
indicate that Eq. (\ref{Unidirectional Scaling Term Two1}) determines an
eigen-balanced first and second order statistical moment about the geometric
locus of $\mathbf{x}_{2_{i\ast}}$.

Furthermore, Eqs (\ref{Equilibrium Constraint on Dual Eigen-components}),
(\ref{Dual Eigen-coordinate Locations Component Two}), and
(\ref{Unidirectional Scaling Term Two1}) show that joint distributions of the
components of $\mathbf{\psi}$ and $\mathbf{\tau}$ are symmetrically
distributed over the axis of the extreme vector $\mathbf{x}_{2_{i\ast}}$. This
indicates that joint distributions of the components of $\mathbf{\psi}$ and
$\mathbf{\tau}$ are symmetrically distributed over the axis of the Wolfe dual
normal eigenaxis component $\psi_{2i\ast}\overrightarrow{\mathbf{e}}_{2i\ast}$.

Alternatively, returning to Eq. (\ref{Constrained Primal Eigenlocus psi2}),
the eigen-balanced, signed magnitude in Eq.
(\ref{Unidirectional Scaling Term Two1}) depends upon the difference between
integrated, cosine-scaled magnitudes of the constrained primal normal
eigenaxis components on $\mathbf{\tau}_{2}$ and $\mathbf{\tau}_{1}$:%
\begin{align}
\operatorname{comp}_{\overrightarrow{\mathbf{x}_{2i\ast}}}\left(
\overrightarrow{\widetilde{\psi}_{2i\ast}\left\Vert \widetilde{\mathbf{x}%
}_{\ast}\right\Vert _{_{2i_{\ast}}}}\right)   &  =\sum\nolimits_{j=1}^{l_{2}%
}\cos\theta_{\mathbf{x}_{2_{i\ast}}\mathbf{x}_{2_{j\ast}}}\left\Vert
\mathbf{\tau}_{2}(j)\right\Vert \label{Unidirectional Scaling Term Two2}\\
&  -\sum\nolimits_{j=1}^{l_{1}}\cos\theta_{\mathbf{x}_{2_{i\ast}}%
\mathbf{x}_{1_{j\ast}}}\left\Vert \mathbf{\tau}_{1}(j)\right\Vert
\text{,}\nonumber
\end{align}
which also shows that joint distributions of the components of $\mathbf{\psi}$
and $\mathbf{\tau}$ are symmetrically distributed over the axes of both
$\psi_{2i\ast}\overrightarrow{\mathbf{e}}_{2i\ast}$ and $\mathbf{x}_{2_{i\ast
}}$.

Using Eqs (\ref{Dual Eigen-coordinate Locations Component Two}) and
(\ref{Unidirectional Scaling Term Two1}), it follows that the length
$\psi_{2i\ast}$ of the Wolfe dual normal eigenaxis component $\psi_{2i\ast
}\overrightarrow{\mathbf{e}}_{2i\ast}$ is determined by the weighted length of
the correlated extreme training vector $\mathbf{x}_{2_{i\ast}}$%
\begin{equation}
\psi_{2i\ast}=\left[  \lambda_{\max_{\mathbf{\psi}}}^{-1}\times
\operatorname{comp}_{\overrightarrow{\mathbf{x}_{2i\ast}}}\left(
\overrightarrow{\widetilde{\psi}_{2i\ast}\left\Vert \widetilde{\mathbf{x}%
}_{\ast}\right\Vert _{_{2i_{\ast}}}}\right)  \right]  \left\Vert
\mathbf{x}_{2_{i\ast}}\right\Vert \text{,}
\label{Magnitude Dual Normal Eigenaxis Component Class Two}%
\end{equation}
where the weighting factor encodes an eigenvalue $\lambda_{\max_{\mathbf{\psi
}}}^{-1}$ scaling of an eigen-balanced, signed magnitude $\operatorname{comp}%
_{\overrightarrow{\mathbf{x}_{2i\ast}}}\left(  \overrightarrow{\widetilde{\psi
}_{2i\ast}\left\Vert \widetilde{\mathbf{x}}_{\ast}\right\Vert _{_{2i_{\ast}}}%
}\right)  $ along the axis of $\mathbf{x}_{2_{i\ast}}$.

Given that $\psi_{2i\ast}>0$, $\lambda_{\max_{\mathbf{\psi}}}^{-1}>0$, and
$\left\Vert \mathbf{x}_{2_{i\ast}}\right\Vert >0$, it follows that the
eigen-balanced, signed magnitude along the axis of $\mathbf{x}_{2_{i\ast}}$ is
positive%
\[
\operatorname{comp}_{\overrightarrow{\mathbf{x}_{2i\ast}}}\left(
\overrightarrow{\widetilde{\psi}_{2i\ast}\left\Vert \widetilde{\mathbf{x}%
}_{\ast}\right\Vert _{_{2i_{\ast}}}}\right)  >0\text{,}%
\]
which indicates that the weighting factor in Eq.
(\ref{Magnitude Dual Normal Eigenaxis Component Class Two}) determines an
eigen-balanced length%
\[
\lambda_{\max_{\mathbf{\psi}}}^{-1}\operatorname{comp}%
_{\overrightarrow{\mathbf{x}_{2i\ast}}}\left(  \overrightarrow{\widetilde{\psi
}_{2i\ast}\left\Vert \widetilde{\mathbf{x}}_{\ast}\right\Vert _{_{2i_{\ast}}}%
}\right)  \left\Vert \mathbf{x}_{2_{i\ast}}\right\Vert \text{,}%
\]
for the extreme vector $\mathbf{x}_{2_{i\ast}}$. It follows that Eq.
(\ref{Magnitude Dual Normal Eigenaxis Component Class Two}) determines an
eigen-balanced length for $\psi_{2i\ast}\overrightarrow{\mathbf{e}}_{2i\ast}$
and $\mathbf{x}_{2_{i\ast}}$.

Returning to Eqs
(\ref{Large Unidirectional Pointwise Covariance Estimate Class Two}) and
(\ref{Dual Eigen-coordinate Locations Component Two}), it follows that the
length $\psi_{2i\ast}$ of the Wolfe dual normal eigenaxis component
$\psi_{2i\ast}\overrightarrow{\mathbf{e}}_{2i\ast}$ on $\mathbf{\psi}$%
\[
\psi_{2i\ast}=\lambda_{\max_{\mathbf{\psi}}}^{-1}\operatorname{comp}%
_{\overrightarrow{\mathbf{x}_{2i\ast}}}\left(  \overrightarrow{\widetilde{\psi
}_{2i\ast}\left\Vert \widetilde{\mathbf{x}}_{\ast}\right\Vert _{_{2i_{\ast}}}%
}\right)  \left\Vert \mathbf{x}_{2_{i\ast}}\right\Vert \text{,}%
\]
is shaped by an eigen-balanced first and second order statistical moment about
the geometric locus of the correlated extreme vector $\mathbf{x}_{2_{i\ast}}$.

Now, take any given Wolfe dual $\psi_{2i\ast}\overrightarrow{\mathbf{e}%
}_{2i\ast}$ and correlated constrained primal $\psi_{2_{i\ast}}\mathbf{x}%
_{2_{i\ast}}$ normal eigenaxis component. It will now be shown that the
direction of $\psi_{2i\ast}\overrightarrow{\mathbf{e}}_{2i\ast}$ is identical
to the direction of $\psi_{2_{i\ast}}\mathbf{x}_{2_{i\ast}}$.

\subsection{Directional Symmetries Exhibited by Normal Eigenaxis Components on
$\mathbf{\psi}$ and $\mathbf{\tau}_{2}$}

The vector direction of the $\psi_{2i\ast}\overrightarrow{\mathbf{e}}_{2i\ast
}$ component is implicitly specified by Eq.
(\ref{Dual Eigen-coordinate Locations Component Two}), where it has been
assumed that $\psi_{2i\ast}$ provides an eigen-scale for a non-orthogonal unit
vector $\overrightarrow{\mathbf{e}}_{2i\ast}$. Given Eqs
(\ref{Pointwise Covariance Statistic}) and
(\ref{Large Unidirectional Pointwise Covariance Estimate Class Two}), it
follows that the eigen-balanced pointwise covariance statistic in Eq.
(\ref{Dual Eigen-coordinate Locations Component Two}) encodes the direction of
the extreme vector $\mathbf{x}_{2_{i_{\ast}}}$ and an eigen-balanced magnitude
along the axis of the extreme vector $\mathbf{x}_{2_{i_{\ast}}}$.

Returning to Eqs (\ref{Unidirectional Scaling Term Two1}),
(\ref{Unidirectional Scaling Term Two2}), and
(\ref{Magnitude Dual Normal Eigenaxis Component Class Two}), take any given
Wolfe dual normal eigenaxis component $\psi_{2i\ast}\overrightarrow{\mathbf{e}%
}_{2i\ast}$ that is correlated with an extreme vector $\mathbf{x}_{2_{i_{\ast
}}}$. Given that the length $\psi_{2i\ast}$ of the Wolfe dual normal eigenaxis
component $\psi_{2i\ast}\overrightarrow{\mathbf{e}}_{2i\ast}$ is determined by
the eigen-balanced length of the extreme vector $\mathbf{x}_{2_{i_{\ast}}}$%
\[
\psi_{2i\ast}=\lambda_{\max_{\mathbf{\psi}}}^{-1}\operatorname{comp}%
_{\overrightarrow{\mathbf{x}_{2i\ast}}}\left(  \overrightarrow{\widetilde{\psi
}_{2\ast}\left\Vert \widetilde{\mathbf{x}}_{\ast}\right\Vert _{_{2_{\ast}}}%
}\right)  \left\Vert \mathbf{x}_{2_{i\ast}}\right\Vert \text{,}%
\]
it follows that the non-orthogonal unit vector $\overrightarrow{\mathbf{e}%
}_{2i\ast}$ has the same direction as the extreme vector $\mathbf{x}%
_{2_{i_{\ast}}}$%
\[
\overrightarrow{\mathbf{e}}_{2i\ast}\equiv\frac{\mathbf{x}_{2_{i_{\ast}}}%
}{\left\Vert \mathbf{x}_{2_{i_{\ast}}}\right\Vert }\text{.}%
\]
Therefore, the direction of the Wolfe dual normal eigenaxis component
$\psi_{2i\ast}\overrightarrow{\mathbf{e}}_{2i\ast}$ is identical to the
direction of the constrained primal normal eigenaxis component $\psi
_{2_{i\ast}}\mathbf{x}_{2_{i\ast}}$, which is determined by the direction of
the eigen-scaled $\psi_{2_{i\ast}}$ extreme training vector $\mathbf{x}%
_{2_{i\ast}}$. The Wolfe dual $\psi_{2i\ast}\overrightarrow{\mathbf{e}%
}_{1i\ast}$ and the constrained primal $\psi_{2_{i\ast}}\mathbf{x}_{2_{i\ast}%
}$ normal eigenaxis components are said to exhibit directional symmetry.
Accordingly, each Wolfe dual $\psi_{2i\ast}\overrightarrow{\mathbf{e}}%
_{2i\ast}$ and correlated constrained primal $\psi_{2_{i\ast}}\mathbf{x}%
_{2_{i\ast}}$ normal eigenaxis component exhibit directional symmetry.

It is concluded that the uniform directions of the Wolfe dual $\psi_{2i\ast
}\overrightarrow{\mathbf{e}}_{1i\ast}$ and the constrained primal
$\psi_{2_{i\ast}}\mathbf{x}_{2_{i\ast}}$ normal eigenaxis components determine
critical directions of large covariance, which contribute to a symmetric
partitioning of a minimal geometric region of constant width that spans a
region of large covariance between the distributions of two classes of
training data. It is also concluded that each of the correlated normal
eigenaxis components on $\mathbf{\psi}$ and $\mathbf{\tau}_{2}$ possess
critical lengths for which the constrained discriminant function $D\left(
\mathbf{x}\right)  =\mathbf{x}^{T}\mathbf{\tau}+\tau_{0}$ delineates centrally
located, bipartite, symmetric regions of large covariance between two data
distributions. Equations for the integrated lengths $\sum\nolimits_{i=1}%
^{l_{2}}\psi_{2i\ast}$ of the $\psi_{2i\ast}\overrightarrow{\mathbf{e}%
}_{2i\ast}$ components are obtained next.

\subsection*{Integrated Lengths of $\psi_{2i\ast}%
\protect\overrightarrow{\mathbf{e}}_{2i\ast}$ Components on $\mathbf{\psi}$}

Using Eq. (\ref{Dual Eigen-coordinate Locations Component Two}), an equation
is obtained for the integrated lengths $\sum\nolimits_{i=l_{1}+1}^{l_{2}}%
\psi_{2i\ast}$ of the $\psi_{2i\ast}\overrightarrow{\mathbf{e}}_{2i\ast} $
components:%
\begin{align}
\sum\nolimits_{i=1}^{l_{2}}\psi_{2i\ast}  &  =\lambda_{\max_{\mathbf{\psi}}%
}^{-1}\sum\nolimits_{i=1}^{l_{2}}\left\Vert \mathbf{x}_{2_{i\ast}}\right\Vert
\label{integrated dual loci two1}\\
&  \times\sum\nolimits_{j=1}^{l_{2}}\psi_{2_{j\ast}}\left\Vert \mathbf{x}%
_{2_{j\ast}}\right\Vert \cos\theta_{\mathbf{x}_{2_{i\ast}}\mathbf{x}%
_{2_{j\ast}}}\nonumber\\
&  -\lambda_{\max_{\mathbf{\psi}}}^{-1}\sum\nolimits_{i=1}^{l_{2}}\left\Vert
\mathbf{x}_{2_{i\ast}}\right\Vert \nonumber\\
&  \times\sum\nolimits_{j=1}^{l_{1}}\psi_{1_{j\ast}}\left\Vert \mathbf{x}%
_{1_{j\ast}}\right\Vert \cos\theta_{\mathbf{x}_{2_{i\ast}}\mathbf{x}%
_{1_{j\ast}}}\text{,}\nonumber
\end{align}
where the vector direction of the Wolfe dual normal eigenaxis $\psi_{2i\ast
}\overrightarrow{\mathbf{e}}_{1i\ast}$ on $\mathbf{\psi}\in%
\mathbb{R}
^{N}$ has the same direction as the correlated extreme training vector
$\mathbf{x}_{2_{i\ast}}\in%
\mathbb{R}
^{d}$.

Alternatively, Eq. (\ref{Constrained Primal Eigenlocus psi2}) provides the
expression:%
\begin{align}
\sum\nolimits_{i=1}^{l_{2}}\psi_{2i\ast}  &  =\lambda_{\max_{\mathbf{\psi}}%
}^{-1}\sum\nolimits_{i=1}^{l_{2}}\left\Vert \mathbf{x}_{2_{i\ast}}\right\Vert
\label{integrated dual loci two2}\\
&  \times\sum\nolimits_{j=1}^{l_{2}}\cos\theta_{\mathbf{x}_{2_{i\ast}%
}\mathbf{x}_{2_{j\ast}}}\left\Vert \mathbf{\tau}_{2}(j)\right\Vert \nonumber\\
&  -\lambda_{\max_{\mathbf{\psi}}}^{-1}\sum\nolimits_{i=1}^{l_{2}}\left\Vert
\mathbf{x}_{2_{i\ast}}\right\Vert \nonumber\\
&  \times\sum\nolimits_{j=1}^{l_{1}}\cos\theta_{\mathbf{x}_{2_{i\ast}%
}\mathbf{x}_{1_{j\ast}}}\left\Vert \mathbf{\tau}_{1}(j)\right\Vert
\text{;}\nonumber
\end{align}
Eqs (\ref{integrated dual loci two1}) and (\ref{integrated dual loci two2})
will be used to examine the algebraic and geometric nature of the statistical
equilibrium state that is implied by Eq.
(\ref{Equilibrium Constraint on Dual Eigen-components}). The uniform geometric
and statistical properties which are jointly exhibited by the $\psi_{2i\ast
}\overrightarrow{\mathbf{e}}_{2i\ast}$ and the $\psi_{2_{i\ast}}%
\mathbf{x}_{2_{i\ast}}$ normal eigenaxis components are summarized next.

\subsection*{Summary of Uniform Geometric and Statistical Properties Jointly
Exhibited by Normal Eigenaxis Components on $\mathbf{\psi}$ and $\mathbf{\tau
}_{2}$}

Results of the previous analysis are now used to identify uniform geometric
and statistical properties which are jointly exhibited by the $\psi_{2i\ast
}\overrightarrow{\mathbf{e}}_{2i\ast}$ Wolfe dual normal eigenaxis components
on $\mathbf{\mathbf{\psi}}$ and the $\psi_{2_{i\ast}}\mathbf{x}_{2_{i\ast}}$
constrained primal normal eigenaxis components on $\mathbf{\tau}_{2}$. The
properties are summarized below.

\begin{conclusion}
The direction of each Wolfe dual normal eigenaxis component $\psi_{2i\ast
}\overrightarrow{\mathbf{e}}_{2i\ast}$ on $\mathbf{\mathbf{\psi}\in%
\mathbb{R}
}^{N}$ is identical to the direction of a constrained primal normal eigenaxis
component $\psi_{2_{i\ast}}\mathbf{x}_{2_{i\ast}}$ on$\mathbf{\ \mathbf{\tau
}_{2}\in%
\mathbb{R}
}^{d}$.
\end{conclusion}

\begin{conclusion}
The lengths of each Wolfe dual normal eigenaxis component $\psi_{2i\ast
}\overrightarrow{\mathbf{e}}_{2i\ast}$ on $\mathbf{\mathbf{\psi}\in%
\mathbb{R}
}^{N}$ and correlated constrained primal normal eigenaxis component
$\psi_{2_{i\ast}}\mathbf{x}_{2_{i\ast}}$ on$\mathbf{\ \mathbf{\tau}_{2}\in%
\mathbb{R}
}^{d}$ are shaped by identical joint symmetrical distributions of normal
eigenaxis components on $\mathbf{\psi}$ and $\mathbf{\tau}$.
\end{conclusion}

\begin{conclusion}
The length $\psi_{2i\ast}$ of each Wolfe dual normal eigenaxis component
$\psi_{2i\ast}\overrightarrow{\mathbf{e}}_{2i\ast}$ on $\mathbf{\mathbf{\psi
}\in%
\mathbb{R}
}^{N}$
\[
\psi_{2i\ast}=\lambda_{\max_{\mathbf{\psi}}}^{-1}\operatorname{comp}%
_{\overrightarrow{\mathbf{x}_{2i\ast}}}\left(  \overrightarrow{\widetilde{\psi
}_{2i\ast}\left\Vert \widetilde{\mathbf{x}}_{\ast}\right\Vert _{_{2i_{\ast}}}%
}\right)  \left\Vert \mathbf{x}_{2_{i\ast}}\right\Vert \text{,}%
\]
is shaped by an eigen-balanced pointwise covariance estimate%
\begin{align*}
\widehat{\operatorname{cov}}_{up_{\updownarrow}}\left(  \mathbf{x}%
_{2_{i_{\ast}}}\right)   &  =\lambda_{\max_{\mathbf{\psi}}}^{-1}\left\Vert
\mathbf{x}_{2_{i\ast}}\right\Vert \\
&  \times\sum\nolimits_{j=1}^{l_{2}}\psi_{2_{j\ast}}\left\Vert \mathbf{x}%
_{2_{j\ast}}\right\Vert \cos\theta_{\mathbf{x}_{2_{i\ast}}\mathbf{x}%
_{2_{j\ast}}}\\
&  -\lambda_{\max_{\mathbf{\psi}}}^{-1}\left\Vert \mathbf{x}_{2_{i\ast}%
}\right\Vert \\
&  \times\sum\nolimits_{j=1}^{l_{1}}\psi_{1_{j\ast}}\left\Vert \mathbf{x}%
_{1_{j\ast}}\right\Vert \cos\theta_{\mathbf{x}_{2_{i\ast}}\mathbf{x}%
_{1_{j\ast}}}\text{,}%
\end{align*}
for a correlated extreme training vector\textbf{\ }$\mathbf{x}_{2_{i\ast}%
}\mathbf{\in%
\mathbb{R}
}^{d}$, such that the eigenlocus of each constrained primal normal eigenaxis
component\textbf{\ }$\psi_{2_{i\ast}}\mathbf{x}_{2_{i\ast}}$\textbf{\ }on
$\mathbf{\mathbf{\tau}_{2}\in%
\mathbb{R}
}^{d}$\textbf{\ }provides a maximum\textit{\ }covariance estimate in a
principal location, in the form of an eigen-balanced first and second order
statistical moment about the geometric locus of an extreme point
$\mathbf{x}_{2_{i\ast}}$.
\end{conclusion}

\begin{conclusion}
Each Wolfe dual normal eigenaxis component\textbf{\ }$\psi_{2i\ast
}\overrightarrow{\mathbf{e}}_{2i\ast}$ on $\mathbf{\mathbf{\psi}}$%
\textbf{\ }encodes an eigen-balanced first and second order statistical moment
about the geometric locus of a correlated extreme data point $\mathbf{x}%
_{2_{i\ast}}$, relative to the eigenloci of all of the eigen-scaled extreme
training points, which determines the eigenlocus of a constrained primal
normal eigenaxis component $\psi_{2_{i\ast}}\mathbf{x}_{2_{i\ast}}$ on
$\mathbf{\tau}_{2}$.
\end{conclusion}

\begin{conclusion}
Any given eigen-balanced pointwise covariance estimate
$\widehat{\operatorname{cov}}_{up_{\updownarrow}}\left(  \mathbf{x}%
_{2_{i_{\ast}}}\right)  $ encodes a distribution of first order coordinates
for an extreme training vector $\mathbf{x}_{2_{i\ast}}$, relative to the
eigen-scaled extreme training points for a given data set. The distribution of
first order coordinates for $\mathbf{x}_{2_{i\ast}}$ describes how the
components of $\mathbf{x}_{2_{i\ast}}$ are distributed within the given
collection of eigen-scaled extreme vectors.
\end{conclusion}

\begin{conclusion}
Returning to Figs $12$ and $23$, the integrated eigenloci of the $\psi
_{2i\ast}\overrightarrow{\mathbf{e}}_{2i\ast}$ Wolfe dual normal eigenaxis
components jointly and implicitly specify the geometric locus of a hyperplane
decision border $H_{-1}$. Likewise, the integrated eigenloci of the
$\psi_{2i\ast}\overrightarrow{\mathbf{e}}_{2i\ast}$ Wolfe dual normal
eigenaxis components jointly and implicitly account for a symmetric
partitioning of a minimal area surface of large covariance that is delineated
by a hyperplane decision boundary $H_{0}$, which is symmetrically located
between a pair of hyperplane decision borders $H_{+1}$ and $H_{-1}$.

\begin{claim}
The square $\left\Vert \psi_{2_{i\ast}}\mathbf{x}_{2_{i\ast}}\right\Vert
_{\min_{c}}^{2}$ of the constrained primal normal eigenaxis
component\textbf{\ }$\psi_{2_{i\ast}}\mathbf{x}_{2_{i\ast}}$\textbf{\ }on
$\mathbf{\mathbf{\tau}}_{2}$\textbf{\ }$\mathbf{\in%
\mathbb{R}
}^{d}$ is the probability of finding the extreme data point $\mathbf{x}%
_{2_{i\ast}}$ in a particular region of $\mathbf{%
\mathbb{R}
}^{d}$, where $\left\Vert \psi_{2_{i\ast}}\mathbf{x}_{2_{i\ast}}\right\Vert
_{\min_{c}}^{2}$ is the total allowed eigenenergy of $\psi_{2_{i\ast}%
}\mathbf{x}_{2_{i\ast}}$.
\end{claim}
\end{conclusion}

The properties exhibited by the total allowed eigenenergy of a Wolf dual
normal eigenlocus are summarized in the next section. Section $16$ will also
outline the fundamental issue that must be resolved for strong dual normal
eigenlocus transforms.

\section{Properties Exhibited by the Total Allowed Eigenenergy of a Wolfe Dual
Normal Eigenlocus $\mathbf{\psi}$}

The eigenloci of the Wolf dual normal eigenaxis components $\left\{
\psi_{i\ast}\overrightarrow{\mathbf{e}}_{i\ast}\right\}  _{i=1}^{l}$ on
$\mathbf{\psi}$ determine the total allowed eigenenergy $\left\Vert
\mathbf{\psi}\right\Vert _{\min_{c}}^{2}$ of $\mathbf{\psi}$%
\begin{align*}
\left\Vert \mathbf{\psi}\right\Vert _{\min_{c}}^{2}  &  =\left(
\sum\nolimits_{i=1}^{l}\psi_{i\ast}\overrightarrow{\mathbf{e}}_{i\ast}%
^{T}\right)  \left(  \sum\nolimits_{i=1}^{l}\psi_{i\ast}%
\overrightarrow{\mathbf{e}}_{i\ast}\right)  \text{,}\\
&  =\sum\nolimits_{i=1}^{l}\psi_{i\ast}^{2}\text{.}%
\end{align*}
Given the uniform geometric and statistical properties which are jointly
exhibited by correlated normal eigenaxis components on $\mathbf{\tau}%
_{1}-\mathbf{\tau}_{2}$ and $\mathbf{\mathbf{\psi}}$, it is concluded that
\emph{a Wolfe dual normal eigenlocus} $\mathbf{\psi}$ \emph{satisfies a
critical minimum eigenenergy constraint which is symmetrically related to the
restriction of the primal normal eigenlocus to the Wolfe dual eigenspace}.

Therefore, consider again Eq.
(\ref{Critical Minimum Eigenenergy of a Wolf Dual Normal Eigenlocus})%
\begin{align*}
\max\mathbf{\psi}^{T}\mathbf{Q\psi}  &  =\lambda_{\max_{\mathbf{\psi}}%
}\left\Vert \mathbf{\psi}\right\Vert _{\min_{c}}^{2}\text{,}\\
&  \cong\left\Vert \mathbf{\tau}\right\Vert _{\min_{c}}^{2}\text{,}%
\end{align*}
and Eq. (\ref{Equilibrium Constraint on Dual Eigen-components})
\[
\sum\nolimits_{i=1}^{l_{1}}\psi_{1_{i\ast}}=\sum\nolimits_{i=1}^{l_{2}}%
\psi_{2_{i\ast}}\text{.}%
\]
Equations (\ref{Equilibrium Constraint on Dual Eigen-components}),
(\ref{Dual Eigen-coordinate Locations Component One}),
(\ref{Unidirectional Scaling Term One1}),
(\ref{Dual Eigen-coordinate Locations Component Two}), and
(\ref{Unidirectional Scaling Term Two1}), which demonstrate how joint
distributions of the normal eigenaxis components of $\mathbf{\psi}$ and
$\mathbf{\tau}$ are symmetrically distributed over the axes of all of the
Wolfe dual normal eigenaxis components $\left\{  \psi_{1_{i\ast}%
}\overrightarrow{\mathbf{e}}_{1i\ast}\right\}  _{i=1}^{l_{1}}$ and $\left\{
\psi_{2_{i\ast}}\overrightarrow{\mathbf{e}}_{2i\ast}\right\}  _{i=1}^{l_{2}}$
and correlated extreme training vectors $\left\{  \mathbf{x}_{1_{i\ast}%
}\right\}  _{i=1}^{l_{1}}$ and $\left\{  \mathbf{x}_{2_{i\ast}}\right\}
_{i=1}^{l_{2}}$, together with Eq.
(\ref{Critical Minimum Eigenenergy of a Wolf Dual Normal Eigenlocus}),
indicate that the critical minimum eigenenergy $\lambda_{\max_{\mathbf{\psi}}%
}\left\Vert \mathbf{\psi}\right\Vert _{\min_{c}}^{2}$ of $\mathbf{\psi}$ is
characterized by joint symmetrical distributions of the eigenenergies of
$\mathbf{\psi}$ and $\mathbf{\tau}$, which are symmetrically distributed over
the Wolfe dual normal eigenaxis components.

It follows that joint distributions of the normal eigenaxis components of
$\mathbf{\psi}$ and $\mathbf{\tau}$ are symmetrically distributed over the
axes of all of the constrained primal normal eigenaxis components $\left\{
\psi_{1_{i\ast}}\mathbf{x}_{1_{i\ast}}\right\}  _{i=1}^{l_{1}}$ and $\left\{
\psi_{2_{i\ast}}\mathbf{x}_{2_{i\ast}}\right\}  _{i=1}^{l_{2}}$, whereby the
critical minimum eigenenergy $\left\Vert \mathbf{\tau}\right\Vert _{\min_{c}%
}^{2}$ of the constrained primal normal eigenlocus $\mathbf{\tau}$ is also
characterized by joint symmetrical distributions of the eigenenergies of
$\mathbf{\psi}$ and $\mathbf{\tau}$.

Section $17$ will demonstrate how the total allowed eigenenergy $\lambda
_{\max_{\mathbf{\psi}}}\left\Vert \mathbf{\psi}\right\Vert _{\min_{c}}^{2}$ of
$\mathbf{\psi}$, which is determined by the eigenloci of the Wolfe dual normal
eigenaxis components, regulates the manner in which joint symmetrical
distributions of the eigenenergies of $\mathbf{\psi}$ and $\mathbf{\tau}$ are
symmetrically distributed over the constrained primal normal eigenlocus
components $\mathbf{\tau}_{1}\ $and$\mathbf{\ \tau}_{2}$ on $\mathbf{\tau}$.

The next section will outline the fundamental issue that must be resolved for
strong dual normal eigenlocus transforms. Finding the state of statistical
equilibrium for the constrained discriminant function of a strong dual normal
eigenlocus $\mathbf{\tau}$ requires finding the right mix of normal eigenaxis
component lengths on $\mathbf{\psi}$ and $\mathbf{\tau}$. The fundamental
problem involves determining the lengths of the Wolfe dual normal eigenaxis
components on $\mathbf{\psi}$, which are the Lagrange multipliers $\left\{
\psi_{i}\right\}  _{i=1}^{N}$ of Eqs (\ref{Wolfe Dual Normal Eigenlocus}) and
(\ref{Vector Form Wolfe Dual}).

\subsection{Finding the Right Component Lengths}

It has been demonstrated that the directions of the constrained primal and the
Wolfe dual normal eigenaxis components are fixed, along with the angles
between all of the extreme vectors. Moreover, the equilibrium constraint of
Eq. (\ref{Equilibrium Constraint on Dual Eigen-components}) on the component
lengths of $\mathbf{\psi}$%
\[
\sum\nolimits_{i=1}^{l_{1}}\psi_{1_{i\ast}}=\sum\nolimits_{i=1}^{l_{2}}%
\psi_{2_{i\ast}}\text{,}%
\]
indicates that the RHS\ of Eq. (\ref{integrated dual loci one1})%
\begin{align*}
\sum\nolimits_{i=1}^{l_{1}}\psi_{1i\ast}  &  =\lambda_{\max_{\mathbf{\psi}}%
}^{-1}\sum\nolimits_{i=1}^{l_{1}}\left\Vert \mathbf{x}_{1_{i\ast}}\right\Vert
\\
&  \times\sum\nolimits_{j=1}^{l_{1}}\psi_{1_{j\ast}}\left\Vert \mathbf{x}%
_{1_{j\ast}}\right\Vert \cos\theta_{\mathbf{x}_{1_{i\ast}}\mathbf{x}%
_{1_{j\ast}}}\\
&  -\lambda_{\max_{\mathbf{\psi}}}^{-1}\sum\nolimits_{i=1}^{l_{1}}\left\Vert
\mathbf{x}_{1_{i\ast}}\right\Vert \\
&  \times\sum\nolimits_{j=1}^{l_{2}}\psi_{2_{j\ast}}\left\Vert \mathbf{x}%
_{2_{j\ast}}\right\Vert \cos\theta_{\mathbf{x}_{1_{i\ast}}\mathbf{x}%
_{2_{j\ast}}}\text{,}%
\end{align*}
must equal the RHS\ of Eq. (\ref{integrated dual loci two1})%
\begin{align*}
\sum\nolimits_{i=1}^{l_{2}}\psi_{2i\ast}  &  =\lambda_{\max_{\mathbf{\psi}}%
}^{-1}\sum\nolimits_{i=1}^{l_{2}}\left\Vert \mathbf{x}_{2_{i\ast}}\right\Vert
\\
&  \times\sum\nolimits_{j=1}^{l_{2}}\psi_{2_{j\ast}}\left\Vert \mathbf{x}%
_{2_{j\ast}}\right\Vert \cos\theta_{\mathbf{x}_{2_{i\ast}}\mathbf{x}%
_{2_{j\ast}}}\\
&  -\lambda_{\max_{\mathbf{\psi}}}^{-1}\sum\nolimits_{i=1}^{l_{2}}\left\Vert
\mathbf{x}_{2_{i\ast}}\right\Vert \\
&  \times\sum\nolimits_{j=1}^{l_{1}}\psi_{1_{j\ast}}\left\Vert \mathbf{x}%
_{1_{j\ast}}\right\Vert \cos\theta_{\mathbf{x}_{2_{i\ast}}\mathbf{x}%
_{1_{j\ast}}}\text{.}%
\end{align*}
Likewise, the RHS of\ Eq. (\ref{integrated dual loci one2})%
\begin{align*}
\sum\nolimits_{i=1}^{l_{1}}\psi_{1i\ast}  &  =\lambda_{\max_{\mathbf{\psi}}%
}^{-1}\sum\nolimits_{i=1}^{l_{1}}\left\Vert \mathbf{x}_{1_{i\ast}}\right\Vert
\\
&  \times\sum\nolimits_{j=1}^{l_{1}}\cos\theta_{\mathbf{x}_{1_{i\ast}%
}\mathbf{x}_{1_{j\ast}}}\left\Vert \mathbf{\tau}_{1}(j)\right\Vert \\
&  -\lambda_{\max_{\mathbf{\psi}}}^{-1}\sum\nolimits_{i=1}^{l_{1}}\left\Vert
\mathbf{x}_{1_{i\ast}}\right\Vert \\
&  \times\sum\nolimits_{j=1}^{l_{2}}\cos\theta_{\mathbf{x}_{1_{i\ast}%
}\mathbf{x}_{2_{j\ast}}}\left\Vert \mathbf{\tau}_{2}(j)\right\Vert \text{,}%
\end{align*}
must equal the RHS\ of Eq. (\ref{integrated dual loci two2})%
\begin{align*}
\sum\nolimits_{i=1}^{l_{2}}\psi_{2i\ast}  &  =\lambda_{\max_{\mathbf{\psi}}%
}^{-1}\sum\nolimits_{i=1}^{l_{2}}\left\Vert \mathbf{x}_{2_{i\ast}}\right\Vert
\\
&  \times\sum\nolimits_{j=1}^{l_{2}}\cos\theta_{\mathbf{x}_{2_{i\ast}%
}\mathbf{x}_{2_{j\ast}}}\left\Vert \mathbf{\tau}_{2}(j)\right\Vert \\
&  -\lambda_{\max_{\mathbf{\psi}}}^{-1}\sum\nolimits_{i=1}^{l_{2}}\left\Vert
\mathbf{x}_{2_{i\ast}}\right\Vert \\
&  \times\sum\nolimits_{j=1}^{l_{1}}\cos\theta_{\mathbf{x}_{2_{i\ast}%
}\mathbf{x}_{1_{j\ast}}}\left\Vert \mathbf{\tau}_{1}(j)\right\Vert \text{.}%
\end{align*}

\subsection{Critical Length Constraints}

The pair of balanced statistical eigenlocus equations outlined directly above
indicate that \emph{all of the magnitudes}%
\[
\left\{  \psi_{1_{i\ast}}|\psi_{1_{i\ast}}>0\right\}  _{i=1}^{l_{1}}\text{,}%
\]
\emph{and}%
\[
\left\{  \psi_{2_{i\ast}}|\psi_{2_{i\ast}}>0\right\}  _{i=1}^{l_{2}}\text{,}%
\]
\emph{of the Wolfe dual normal eigenaxis components on} $\mathbf{\psi}$
\emph{satisfy critical length constraints}, such that the highly
interconnected sets of inner product relationships amongst the Wolfe dual and
the constrained primal normal eigenaxis components in Eqs
(\ref{integrated dual loci one1}) and (\ref{integrated dual loci two1}), or
Eqs (\ref{integrated dual loci one2}) and (\ref{integrated dual loci two2}),
determine proper lengths $\psi_{1i\ast}$ or $\psi_{2i\ast}$ for each Wolfe
dual normal eigenaxis component $\psi_{1i\ast}\overrightarrow{\mathbf{e}%
}_{1i\ast}$ or $\psi_{2i\ast}\overrightarrow{\mathbf{e}}_{2i\ast}$, which
effectively determine the proper length of a correlated, constrained primal
normal eigenaxis component $\psi_{1_{i\ast}}\mathbf{x}_{1_{i\ast}}$ or
$\psi_{2_{i\ast}}\mathbf{x}_{2_{i\ast}}$.

Recall that the real unknowns associated with the inequality constrained
optimization problem in Eq. (\ref{Primal Normal Eigenlocus}) are the
constrained eigen-coordinate locations of a normal eigenaxis $\mathbf{v}$,
which are essentially determined by the magnitudes or lengths of the Wolfe
dual normal eigenaxis components on $\mathbf{\psi}$. It has been shown that
each constrained primal normal eigenaxis component on $\mathbf{\tau}$ is
formed by an eigen-scaled extreme training vector, where the eigen-scale of
each extreme training vector is the magnitude of a correlated Wolfe dual
normal eigenaxis component on $\mathbf{\psi}$. It has also been demonstrated
that each extreme training vector and correlated Wolfe dual normal eigenaxis
component exhibit directional symmetry.

All of the previous analyses and simulation studies indicate that the
magnitudes of each of the Wolfe dual normal eigenaxis components on
$\mathbf{\psi}$ exhibit symmetric proportions which determine properly
proportioned magnitudes for each of the constrained primal normal eigenaxis
components on $\mathbf{\tau}$. It is claimed that the magnitudes of the Wolfe
dual normal eigenaxis components essentially determine the state of
statistical equilibrium which is exhibited by a strong dual normal eigenlocus
$\mathbf{\tau=\tau}_{1}-\mathbf{\tau}_{2}$. In particular, it is claimed that
the state of equilibrium%
\[
\sum\nolimits_{i=1}^{l_{1}}\psi_{1_{i\ast}}=\sum\nolimits_{i=1}^{l_{2}}%
\psi_{2_{i\ast}}\text{,}%
\]
that is satisfied by the component lengths of $\mathbf{\psi}$, which ensures
that joint distributions of the components of $\mathbf{\psi}$ and
$\mathbf{\tau}$ are symmetrically distributed over each of the eigen-scaled
extreme training vectors on $\mathbf{\tau}_{1}$ and $\mathbf{\tau}_{2}$,
determines a point of statistical equilibrium%
\[
\mathbf{\tau=\tau}_{1}-\mathbf{\tau}_{2}\text{,}%
\]
for which a strong dual normal eigenlocus $\mathbf{\tau=\tau}_{1}%
-\mathbf{\tau}_{2}$ exhibits a critical minimum eigenenergy that is
characterized by joint symmetrical distributions of the eigenenergies of
$\mathbf{\psi}$ and $\mathbf{\tau}$.

The next section of the paper will examine the algebraic, geometric, and
statistical nature of the remarkable statistical balancing feat that is
routinely accomplished by solving the inequality constrained optimization
problem in Eq. (\ref{Primal Normal Eigenlocus}). Algebraic expressions will be
developed for the total allowed eigenenergies of $\mathbf{\tau}_{1}$,
$\mathbf{\tau}_{2}$, and $\mathbf{\tau}$. These expressions will be used to
develop statistical expressions for the statistical fulcrum $f_{s}$ and the
symmetric equalizer statistic $\nabla_{eq}$ in Eq.
(\ref{Balancing Factor for Eigenlocus Components}). All of these results will
be used to develop the statistical machinery behind the point of statistical
equilibrium which is determined by the constrained Lagrangian functional
$L_{\Psi\left(  \mathbf{\tau}\right)  }$ of Eq.
(\ref{Lagrangian Normal Eigenlocus}). In the final sections of the analysis,
an algebraic expression called the \emph{strong dual normal eigenlocus
identity} will be developed, in which the total allowed eigenenergies of a
strong dual normal eigenlocus satisfy the law of cosines in a surprisingly
elegant and symmetrical manner.

\section{An Elegant Statistical Balancing Feat}

The strong duality relationships between the constrained primal normal
eigenaxis components on $\mathbf{\tau\in%
\mathbb{R}
}^{d}$ and the Wolfe dual normal eigenaxis components on $\mathbf{\mathbf{\psi
\in%
\mathbb{R}
}}^{N}$ facilitate a remarkable statistical balancing feat, whereby the
constrained primal normal eigenaxis components on $\mathbf{\tau}$
\[
\mathbf{\tau=}\sum\nolimits_{i=1}^{l_{1}}\psi_{1_{i\ast}}\mathbf{x}_{1_{i\ast
}}-\sum\nolimits_{i=1}^{l_{2}}\psi_{2_{i\ast}}\mathbf{x}_{2_{i\ast}}\text{,}%
\]
and $\tau_{0}$%
\[
\tau_{0}=\frac{1}{l}\sum\nolimits_{i=1}^{l}y_{i}\left(  1-\xi_{i}\right)
-\left(  \frac{1}{l}\sum\nolimits_{i=1}^{l}\mathbf{x}_{i\ast}\right)
^{T}\mathbf{\tau}\text{,}%
\]
determine a statistical discriminant function%
\[
D\left(  \mathbf{x}\right)  =\mathbf{\tau}^{T}\mathbf{x}+\tau_{0}\text{,}%
\]
which delineates a bipartite, symmetric partitioning of a region of large
covariance between two overlapping or non-overlapping data distributions in $%
\mathbb{R}
^{d}$. Accordingly, the critical minimum eigenenergies $\left\Vert
\mathbf{\tau}_{1}-\mathbf{\tau}_{2}\right\Vert _{\min_{c}}^{2}$ of the
constrained primal normal eigenlocus components $\mathbf{\tau}_{1}$ and
$\mathbf{\tau}_{2}$ on $\mathbf{\tau}$ satisfy a linear decision boundary and
the bilaterally symmetrical borders which bound it.

It has been demonstrated that the number of constrained primal normal
eigenaxis components used to form $\mathbf{\tau}$ is determined by the number
of extreme data points, which are unscaled (unconstrained) primal normal
eigenaxis components on $\mathbf{\tau}$. It has also been demonstrated that
any given region of large covariance between two data distributions is a
function of extreme training point locations. Previous analyses also indicate
that the statistical contents of the Wolfe dual normal eigenaxis components
play a significant role in determining a bipartite, symmetric partitioning of
a given feature space.

Recall the claim that the total allowed eigenenergy $\left\Vert \mathbf{\tau
}\right\Vert _{\min_{c}}^{2}$ of $\mathbf{\tau}$ is the fundamental geometric
and statistical property of $\mathbf{\tau}$, where $\mathbf{\tau} $ possesses
a critical minimum eigenenergy%
\[
\left\Vert \mathbf{\tau}\right\Vert _{\min_{c}}^{2}=\left\Vert \mathbf{\tau
}_{1}-\mathbf{\tau}_{2}\right\Vert _{\min_{c}}^{2}\text{,}%
\]
for which the total allowed eigenenergies of $\mathbf{\tau}_{1}$ and
$\mathbf{\tau}_{2}$ satisfy a state of statistical equilibrium
\[
\left(  \left\Vert \mathbf{\tau}_{1}\right\Vert _{\min_{c}}^{2}-\left\Vert
\mathbf{\tau}_{1}\right\Vert \left\Vert \mathbf{\tau}_{2}\right\Vert
\cos\theta_{\mathbf{\tau}_{1}\mathbf{\tau}_{2}}\right)  +\nabla_{eq}%
\Leftrightarrow\left(  \left\Vert \mathbf{\tau}_{2}\right\Vert _{\min_{c}}%
^{2}-\left\Vert \mathbf{\tau}_{1}\right\Vert \left\Vert \mathbf{\tau}%
_{2}\right\Vert \cos\theta_{\mathbf{\tau}_{1}\mathbf{\tau}_{2}}\right)
-\nabla_{eq}\text{,}%
\]
in relation to a centrally located statistical fulcrum $f_{s}$, where
$\nabla_{eq}$ is a symmetric equalizer statistic. Recall also that the lengths
of the Wolfe dual normal eigenaxis components are selected to satisfy the
above state of statistical equilibrium, and thereby balance the total allowed
eigenenergies of $\mathbf{\tau}$.

It will now be shown how the state of equilibrium%
\[
\sum\nolimits_{i=1}^{l_{1}}\psi_{1_{i\ast}}=\sum\nolimits_{i=1}^{l_{2}}%
\psi_{2_{i\ast}}\text{,}%
\]
that is satisfied by the component lengths of $\mathbf{\psi}$ effectively
determines a point of statistical equilibrium%
\[
\mathbf{\tau=\tau}_{1}-\mathbf{\tau}_{2}\text{,}%
\]
which exhibits a critical minimum eigenenergy $\left\Vert \mathbf{\tau
}\right\Vert _{\min_{c}}^{2}=\left\Vert \mathbf{\tau}_{1}-\mathbf{\tau}%
_{2}\right\Vert _{\min_{c}}^{2}$ that is characterized by joint symmetrical
distributions of the eigenenergies of $\mathbf{\psi}$ and $\mathbf{\tau}$.

Algebraic expressions will now be developed for the total allowed
eigenenergies of $\mathbf{\tau}_{1}$, $\mathbf{\tau}_{2}$, and $\mathbf{\tau}%
$. These expressions will be used to develop statistical equations for the
statistical fulcrum $f_{s}$ and the symmetric equalizer statistic $\nabla
_{eq}$ in Eq. (\ref{Balancing Factor for Eigenlocus Components}). All of these
results will be used to develop the statistical machinery behind the point of
statistical equilibrium which is determined by the constrained Lagrangian
functional $L_{\Psi\left(  \mathbf{\tau}\right)  }$ of Eq.
(\ref{Lagrangian Normal Eigenlocus}). Figures will be presented that provide
geometric and statistical illustrations of the statistical machinery behind
the balancing feat.

An algebraic expression will be developed, called the strong dual normal
eigenlocus identity, in which the total allowed eigenenergies of a strong dual
normal eigenlocus satisfy the law of cosines in surprisingly symmetrical and
elegant manners. The strong dual normal eigenlocus identity determines the
symmetrical manner in which a strong dual normal eigenlocus $\mathbf{\tau
=\tau_{1}}-\mathbf{\tau}_{2}$ satisfies a linear decision boundary and the
bilaterally symmetrical borders which bound it. All of these results will be
used to demonstrate the critical roles that the Wolfe dual normal eigenaxis
component lengths $\left\{  \left\{  \psi_{1_{i\ast}}\right\}  _{i=1}^{l_{1}%
},\left\{  \psi_{2_{i\ast}}\right\}  _{i=1}^{l_{2}}\right\}  $ and the
equilibrium constraint $\sum\nolimits_{i=1}^{l_{1}}\psi_{1_{i\ast}}%
=\sum\nolimits_{i=1}^{l_{2}}\psi_{2_{i\ast}}$ have in determining a bipartite,
symmetric partitioning of a given feature space. The analysis begins by
returning to the $\tau_{0}$ term.

\subsection*{$\tau_{0}$ Revisited}

Let there be $l$ extreme training points in a collection of training data. Let
there be $l_{1}$ eigen-scaled extreme training points $\psi_{1_{i_{\ast}}%
}\mathbf{x}_{1_{i_{\ast}}}$ that belong to the pattern class $\boldsymbol{X}%
_{1}$ and $l_{2}$ eigen-scaled extreme training points $\psi_{2_{i_{\ast}}%
}\mathbf{x}_{2_{i_{\ast}}}$ that belong to the pattern class $\boldsymbol{X}%
_{2}$. Substituting the expression%
\[
\mathbf{\tau}=\sum\nolimits_{i=1}^{l_{1}}\psi_{1_{i_{\ast}}}\mathbf{x}%
_{1_{i_{\ast}}}-\sum\nolimits_{i=1}^{l_{2}}\psi_{2_{i_{\ast}}}\mathbf{x}%
_{2_{i_{\ast}}}\text{,}%
\]
for $\mathbf{\tau}$ in Eq. (\ref{Pair of Normal Eigenlocus Components}) into
the expression for $\tau_{0}$ in Eq.
(\ref{Normal Eigenlocus Projection Factor}) produces the statistic for the
$\tau_{0}$ term:%
\begin{align}
\tau_{0}  &  =\frac{1}{l}\sum\nolimits_{i=1}^{l}y_{i}\left(  1-\xi_{i}\right)
\label{Normal Eigenlocus Projection Factor Two}\\
&  -\frac{1}{l}\sum\nolimits_{i=1}^{l}\mathbf{x}_{i\ast}^{T}\sum
\nolimits_{j=1}^{l_{1}}\psi_{1_{j_{\ast}}}\mathbf{x}_{1_{j_{\ast}}}\nonumber\\
&  +\frac{1}{l}\sum\nolimits_{i=1}^{l}\mathbf{x}_{i\ast}^{T}\sum
\nolimits_{j=1}^{l_{2}}\psi_{2_{j_{\ast}}}\mathbf{x}_{2_{j_{\ast}}}%
\text{.}\nonumber
\end{align}
It will be shown that $\tau_{0}$ plays a large role in balancing the total
allowed eigenenergies of $\mathbf{\tau}_{1}$ and $\mathbf{\tau}_{2}$. The
significance of $\tau_{0}$ will be clarified shortly.

The next section examines the symmetrical relationships between the total
allowed eigenenergies of $\mathbf{\tau}_{1}$ and $\mathbf{\tau}_{2}$. The
demonstration begins with the law of cosines for strong dual normal eigenlocus
transforms. The law of cosines determines how a constrained primal normal
eigenlocus $\mathbf{\tau=\tau_{1}}-\mathbf{\tau}_{2}$ satisfies a linear
decision boundary and the bilaterally symmetrical borders which bound it.

\subsection{The Law of Cosines for Strong Dual Normal Eigenlocus Transforms}

The critical minimum eigenenergy $\left\Vert \mathbf{\tau}\right\Vert
_{\min_{c}}^{2}$ exhibited by a constrained primal normal eigenlocus
$\mathbf{\tau=\tau_{1}}-\mathbf{\tau}_{2}$ is regulated by the law of cosines.
It will be shown that the law of cosines for strong dual normal eigenlocus
transforms requires that the critical minimum eigenenergy $\left\Vert
\mathbf{\tau}_{1}\right\Vert _{\min_{c}}^{2}$ exhibited by the constrained
primal normal eigenlocus component $\mathbf{\tau_{1}}$ coupled with the inner
product statistic $\left\Vert \mathbf{\tau}_{1}\right\Vert \left\Vert
\mathbf{\tau}_{2}\right\Vert \cos\theta_{\mathbf{\tau}_{1}\mathbf{\tau}_{2}}$%
\[
\left\Vert \mathbf{\tau}_{1}\right\Vert _{\min_{c}}^{2}-\left\Vert
\mathbf{\tau}_{1}\right\Vert \left\Vert \mathbf{\tau}_{2}\right\Vert
\cos\theta_{\mathbf{\tau}_{1}\mathbf{\tau}_{2}}\text{,}%
\]
and the critical minimum eigenenergy $\left\Vert \mathbf{\tau}_{2}\right\Vert
_{\min_{c}}^{2}$ exhibited by the constrained primal normal eigenlocus
component $\mathbf{\tau}_{2}$ coupled with the inner product statistic
$\left\Vert \mathbf{\tau}_{2}\right\Vert \left\Vert \mathbf{\tau}%
_{1}\right\Vert \cos\theta_{\mathbf{\tau}_{2}\mathbf{\tau}_{1}}$%
\[
\left\Vert \mathbf{\tau}_{2}\right\Vert _{\min_{c}}^{2}-\left\Vert
\mathbf{\tau}_{2}\right\Vert \left\Vert \mathbf{\tau}_{1}\right\Vert
\cos\theta_{\mathbf{\tau}_{2}\mathbf{\tau}_{1}}\text{,}%
\]
jointly satisfy the critical minimum eigenenergy $\left\Vert \mathbf{\tau
}\right\Vert _{\min_{c}}^{2}$ exhibited by the constrained primal normal
eigenlocus $\mathbf{\tau}$%
\begin{equation}
\left\{  \left\Vert \mathbf{\tau}_{1}\right\Vert _{\min_{c}}^{2}-\left\Vert
\mathbf{\tau}_{1}\right\Vert \left\Vert \mathbf{\tau}_{2}\right\Vert
\cos\theta_{\mathbf{\tau}_{1}\mathbf{\tau}_{2}}\right\}  +\left\{  \left\Vert
\mathbf{\tau}_{2}\right\Vert _{\min_{c}}^{2}-\left\Vert \mathbf{\tau}%
_{2}\right\Vert \left\Vert \mathbf{\tau}_{1}\right\Vert \cos\theta
_{\mathbf{\tau}_{2}\mathbf{\tau}_{1}}\right\}  =\left\Vert \mathbf{\tau
}\right\Vert _{\min_{c}}^{2}\text{.}
\label{Law of Cosines for Strong Dual Normal Eiigenlocus Transforms}%
\end{equation}
Equation (\ref{Law of Cosines for Strong Dual Normal Eiigenlocus Transforms})
indicates that the critical minimum eigenenergy constraints on $\mathbf{\tau
}_{1}$ and $\mathbf{\tau}_{2}$ include the inner product statistic $\left\Vert
\mathbf{\tau}_{1}\right\Vert \left\Vert \mathbf{\tau}_{2}\right\Vert
\cos\theta_{\mathbf{\tau}_{1}\mathbf{\tau}_{2}}$, which encodes the lengths of
$\mathbf{\tau}_{1}$ and $\mathbf{\tau}_{2}$ and the angle $\theta
_{\mathbf{\tau}_{1}\mathbf{\tau}_{2}}$ between them. The sections that follow
will demonstrate how all of the constrained primal normal eigenaxis components
on a strong dual normal eigenlocus $\mathbf{\tau=\tau_{1}}-\mathbf{\tau}_{2}$
satisfy the law of cosines for strong dual normal eigenlocus transforms.

\subsection{Examining the Total Allowed Eigenenergies of a Strong Dual Normal
Eigenlocus}

A\ strong dual normal eigenlocus $\mathbf{\tau=\tau_{1}}-\mathbf{\tau}_{2}$
satisfies a linear decision boundary and the bilaterally symmetrical borders
which bound it in terms of a critical minimum eigenenergy constraint. The
critical minimum eigenenergy exhibited by a constrained primal normal
eigenlocus is determined by the KKT constraint of Eq. (\ref{KKTE6}).

Let there be $l$ eigen-scaled extreme training points on a constrained primal
normal eigenlocus $\mathbf{\tau=\tau_{1}}-\mathbf{\tau}_{2}$. The KKT
constraint of Eq. (\ref{KKTE6}) requires that the $l$ constrained primal
normal eigenaxis components $\left\{  \psi_{i_{\ast}}\mathbf{x}_{i_{\ast}%
}\right\}  _{i=1}^{l}$ on $\mathbf{\tau}$ satisfy an algebraic system of $l$
strong dual normal eigenlocus equations satisfied as strict equalities:%
\begin{equation}
\psi_{i_{\ast}}\left\{  y_{i}\left(  \mathbf{x}_{i_{\ast}}^{T}\mathbf{\tau
}+\tau_{0}\right)  -1+\xi_{i}\right\}  =0,\ i=1,...,l\text{.}
\label{Minimum Eigenenergy Functional System}%
\end{equation}
Equation (\ref{Minimum Eigenenergy Functional System}) is now used to examine
the critical minimum eigenenergy constraints on $\mathbf{\tau}_{1}$ and
$\mathbf{\tau}_{2}$. The analysis begins with the critical minimum eigenenergy
constraint on $\mathbf{\tau}_{1}$.

\subsubsection{The Total Allowed Eigenenergy of $\mathbf{\tau}_{1}$}

Take any one of the $l_{1}$ eigen-scaled extreme training vectors
$\psi_{1_{i_{\ast}}}\mathbf{x}_{1_{i_{\ast}}}$ that belong to the
$\boldsymbol{X}_{1}$ pattern set: $\left\{  \psi_{1_{i_{\ast}}}\mathbf{x}%
_{1_{i_{\ast}}}\right\}  _{i=1}^{l_{1}}$. Using Eq.
(\ref{Minimum Eigenenergy Functional System}) and letting $y_{i}=+1$, it
follows that the constrained primal normal eigenaxis component $\psi
_{1_{i_{\ast}}}\mathbf{x}_{1_{i_{\ast}}}$ on $\mathbf{\tau}_{1}$ is determined
by a strong dual normal eigenlocus equation satisfied as a strict equality:%
\[
\psi_{1_{i_{\ast}}}\left\{  \left(  \mathbf{x}_{1_{i_{\ast}}}^{T}\mathbf{\tau
}+\tau_{0}\right)  -1+\xi_{i}\right\}  =0\text{,}%
\]
which is part of an algebraic system of $l_{1}$ eigenlocus equations.

Now, take all of the $l_{1}$ eigen-scaled extreme training vectors that belong
to the $\boldsymbol{X}_{1}$ pattern set $\left\{  \psi_{1_{i_{\ast}}%
}\mathbf{x}_{1_{i_{\ast}}}\right\}  _{i=1}^{l_{1}}$. Again using Eq.
(\ref{Minimum Eigenenergy Functional System}) and letting $y_{i}=+1$, it
follows that the complete set of $l_{1}$ constrained primal normal eigenaxis
components $\left\{  \psi_{1_{i_{\ast}}}\mathbf{x}_{1_{i_{\ast}}}\right\}
_{i=1}^{l_{1}}$ on $\mathbf{\tau}_{1}$ is determined by the algebraic system
of $l_{1}$ strong dual normal eigenlocus equations satisfied as strict
equalities:%
\[
\psi_{1_{i_{\ast}}}\mathbf{x}_{1_{i_{\ast}}}^{T}\mathbf{\tau}=\psi
_{1_{i_{\ast}}}\left(  1-\xi_{i}-\tau_{0}\right)  ,\ i=1,...,l_{1}\text{.}%
\]
Using the above expression, it follows that the entire set of $l_{1}\times d$
eigen-transformed extreme vector coordinates of $\left\{  \psi_{1_{i_{\ast}}%
}\mathbf{x}_{1_{i_{\ast}}}\right\}  _{i=1}^{l_{1}}$ satisfies the algebraic
system of $l_{1}$ strong dual normal eigenlocus equations:%
\[
\text{ }(1)\text{ \ }\psi_{1_{1_{\ast}}}\mathbf{x}_{1_{1_{\ast}}}%
^{T}\mathbf{\tau}=\psi_{1_{1_{\ast}}}\left(  1-\xi_{i}-\tau_{0}\right)
\text{,}%
\]%
\[
\text{ }(2)\text{ \ }\psi_{1_{2_{\ast}}}\mathbf{x}_{1_{2_{\ast}}}%
^{T}\mathbf{\tau}=\psi_{1_{2_{\ast}}}\left(  1-\xi_{i}-\tau_{0}\right)
\text{,}%
\]

\[
\vdots
\]%
\[
(l_{1})\text{\ \ }\psi_{1_{l_{\ast}}}\mathbf{x}_{1_{l_{\ast}}}^{T}%
\mathbf{\tau}=\psi_{1_{l_{\ast}}}\left(  1-\xi_{i}-\tau_{0}\right)  \text{,}%
\]
where each constrained primal normal eigenaxis component $\psi_{1_{i_{\ast}}%
}\mathbf{x}_{1_{i_{\ast}}}$ on $\mathbf{\tau}_{1}$ satisfies the statistic:%
\[
\psi_{1_{i\ast}}\mathbf{x}_{1_{i\ast}}^{T}\mathbf{\tau}=\psi_{1_{i_{\ast}}%
}\left(  1-\xi_{i}-\tau_{0}\right)  \text{.}%
\]

An algebraic expression is now developed for the total allowed eigenenergy of
$\mathbf{\tau}_{1}$. Denote the total allowed eigenenergy of $\mathbf{\tau
}_{1}$ by $E_{\mathbf{\tau}_{1}}$ and let $\mathbf{\tau=\tau}_{1}%
-\mathbf{\tau}_{2}$. Summation over the above algebraic system of $l_{1}$
strong dual normal eigenlocus equations produces the following expression for
the total allowed eigenenergy $E_{\mathbf{\tau}_{1}}$ of the constrained
primal normal eigenlocus component $\mathbf{\tau}_{1}$:%
\[
\left(  \sum\nolimits_{i=1}^{l_{1}}\psi_{1_{i_{\ast}}}\mathbf{x}_{1_{i_{\ast}%
}}^{T}\right)  \left(  \mathbf{\tau}_{1}-\mathbf{\tau}_{2}\right)  \cong%
\sum\nolimits_{i=1}^{l_{1}}\psi_{1_{i_{\ast}}}\left(  1-\xi_{i}-\tau
_{0}\right)  \text{,}%
\]
which reduces to%
\[
\mathbf{\tau}_{1}^{T}\mathbf{\tau}_{1}-\mathbf{\tau}_{1}^{T}\mathbf{\tau}%
_{2}\cong\sum\nolimits_{i=1}^{l_{1}}\psi_{1_{i_{\ast}}}\left(  1-\xi_{i}%
-\tau_{0}\right)  \text{,}%
\]
so that the total allowed eigenenergy $E_{\mathbf{\tau}_{1}}$ of the
constrained primal normal eigenlocus component $\mathbf{\tau}_{1}$ satisfies
the equation
\[
\left\Vert \mathbf{\tau}_{1}\right\Vert _{\min_{c}}^{2}-\mathbf{\tau}_{1}%
^{T}\mathbf{\tau}_{2}\cong\sum\nolimits_{i=1}^{l_{1}}\psi_{1_{i_{\ast}}%
}\left(  1-\xi_{i}-\tau_{0}\right)  \text{.}%
\]
It follows that the total allowed eigenenergy $E_{\mathbf{\tau}_{1}}$ of the
constrained primal normal eigenlocus component $\mathbf{\tau}_{1}$ is
determined by the expression%
\begin{equation}
\left\Vert \mathbf{\tau}_{1}\right\Vert _{\min_{c}}^{2}-\left\Vert
\mathbf{\tau}_{1}\right\Vert \left\Vert \mathbf{\tau}_{2}\right\Vert
\cos\theta_{\mathbf{\tau}_{1}\mathbf{\tau}_{2}}\cong\sum\nolimits_{i=1}%
^{l_{1}}\psi_{1_{i_{\ast}}}\left(  1-\xi_{i}-\tau_{0}\right)  \text{,}
\label{Minimal Eigenenergy Normal Eigenlocus One}%
\end{equation}
where the total allowed eigenenergy $E_{\mathbf{\tau}_{1}}$ of $\mathbf{\tau
}_{1}$
\[
E_{\mathbf{\tau}_{1}}\triangleq\left\Vert \mathbf{\tau}_{1}\right\Vert
_{\min_{c}}^{2}-\left\Vert \mathbf{\tau}_{1}\right\Vert \left\Vert
\mathbf{\tau}_{2}\right\Vert \cos\theta_{\mathbf{\tau}_{1}\mathbf{\tau}_{2}%
}\text{,}%
\]
is determined by a statistical equation%
\[
E_{\mathbf{\tau}_{1}}=\sum\nolimits_{i=1}^{l_{1}}\psi_{1_{i_{\ast}}}\left(
1-\xi_{i}-\tau_{0}\right)  \text{,}%
\]
in terms of integrated magnitudes $\sum\nolimits_{i=1}^{l_{1}}\psi
_{1_{i_{\ast}}}$ of Wolfe dual normal eigenaxis components which are
correlated with extreme vectors that belong to the $\boldsymbol{X}_{1}$
pattern set, and the $\tau_{0}$ statistic. The critical minimum eigenenergy
constraint on $\mathbf{\tau}_{2}$ is examined next.

\subsubsection{The Total Allowed Eigenenergy of $\mathbf{\tau}_{2}$}

Take any one of the $l_{2}$ eigen-scaled extreme training vectors
$\psi_{2_{i_{\ast}}}\mathbf{x}_{2_{i_{\ast}}}$ that belong to the
$\boldsymbol{X}_{2}$ pattern set: $\left\{  \psi_{2_{i_{\ast}}}\mathbf{x}%
_{2_{i_{\ast}}}\right\}  _{i=1}^{l_{2}}$. Using Eq.
(\ref{Minimum Eigenenergy Functional System}) and letting $y_{i}=-1$, it
follows that a constrained primal normal eigenaxis component $\psi
_{2_{i_{\ast}}}\mathbf{x}_{2_{i_{\ast}}}$ on $\mathbf{\tau}_{2}$ is determined
by a strong dual normal eigenlocus equation satisfied as a strict equality:%
\[
\psi_{2_{i_{\ast}}}\left\{  \left(  -\mathbf{x}_{2_{i_{\ast}}}^{T}%
\mathbf{\tau}-\tau_{0}\right)  -1+\xi_{i}\right\}  =0\text{,}%
\]
which is part of an algebraic system of $l_{2}$ eigenlocus equations.

Now, take all of the $l_{2}$ eigen-scaled extreme training vectors that belong
to the $\boldsymbol{X}_{2}$ pattern set: $\left\{  \psi_{2_{i_{\ast}}%
}\mathbf{x}_{2_{i_{\ast}}}\right\}  _{i=1}^{l_{2}}$. Again using Eq.
(\ref{Minimum Eigenenergy Functional System}) and letting $y_{i}=-1$, it
follows that the complete set of $l_{2}$ constrained primal normal eigenaxis
components $\left\{  \psi_{2_{i_{\ast}}}\mathbf{x}_{2_{i_{\ast}}}\right\}
_{i=1}^{l_{2}}$ on $\mathbf{\tau}_{2}$ is determined by the algebraic system
of $l_{2}$ strong dual normal eigenlocus equations satisfied as strict
equalities:%
\[
-\psi_{2_{i_{\ast}}}\mathbf{x}_{2_{i_{\ast}}}^{T}\mathbf{\tau}=\psi
_{2_{i_{\ast}}}\left(  1-\xi_{i}+\tau_{0}\right)  ,\text{ \ }i=1,...,l_{2}%
\text{.}%
\]
Using the above expression, it follows that the entire set of $l_{2}\times d$
eigen-transformed extreme vector coordinates of $\left\{  \psi_{2_{i_{\ast}}%
}\mathbf{x}_{2_{i_{\ast}}}\right\}  _{i=1}^{l_{2}}$ satisfies the algebraic
system of $l_{2}$ strong dual normal eigenlocus equations:%
\[
\text{ }(1)\text{ \ }-\psi_{2_{1_{\ast}}}\mathbf{x}_{2_{1_{\ast}}}%
^{T}\mathbf{\tau}=\psi_{2_{1_{\ast}}}\left(  1-\xi_{i}+\tau_{0}\right)
\text{,}%
\]%
\[
(2)\text{ \ }-\psi_{2_{2_{\ast}}}\mathbf{x}_{2_{2_{\ast}}}^{T}\mathbf{\tau
}=\psi_{2_{2_{\ast}}}\left(  1-\xi_{i}+\tau_{0}\right)  \text{,}%
\]%
\[
\vdots
\]%
\[
(l_{2})\ -\psi_{2_{l_{2}\ast}}\mathbf{x}_{2_{_{l_{2}\ast}}}^{T}\mathbf{\tau
}=\psi_{2_{l_{2}\ast}}\left(  1-\xi_{i}+\tau_{0}\right)  \text{,}%
\]
where each constrained primal normal eigenaxis component $\psi_{2_{\ast}%
}\mathbf{x}_{2_{_{\ast}}}$ on $\mathbf{\tau}_{2}$ satisfies the statistic:%
\[
-\psi_{2_{i_{\ast}}}\mathbf{x}_{2_{i_{\ast}}}^{T}\mathbf{\tau}=\psi
_{2_{i_{\ast}}}\left(  1-\xi_{i}+\tau_{0}\right)  \text{.}%
\]

An algebraic expression is now developed for the total allowed eigenenergy of
$\mathbf{\tau}_{2}$. Denote the total allowed eigenenergy of $\mathbf{\tau
}_{2}$ by $E_{\mathbf{\tau}_{2}}$ and let $\mathbf{\tau=\tau}_{1}%
-\mathbf{\tau}_{2}$. Summation over the above algebraic system of $l_{2}$
strong dual normal eigenlocus equations produces the following expression for
the total allowed eigenenergy $E_{\mathbf{\tau}_{2}}$ of the constrained
primal normal eigenlocus component $\mathbf{\tau}_{2}$:%
\[
-\left(  \sum\nolimits_{i=1}^{l_{2}}\psi_{2_{i_{\ast}}}\mathbf{x}_{2_{i_{\ast
}}}^{T}\right)  \left(  \mathbf{\tau}_{1}-\mathbf{\tau}_{2}\right)  \cong%
\sum\nolimits_{i=1}^{l_{2}}\psi_{2_{i_{\ast}}}\left(  1-\xi_{i}+\tau
_{0}\right)  \text{,}%
\]
which reduces to%
\[
\mathbf{\tau}_{2}^{T}\mathbf{\tau}_{2}-\mathbf{\tau}_{2}^{T}\mathbf{\tau}%
_{1}\cong\sum\nolimits_{i=1}^{l_{2}}\psi_{2_{i_{\ast}}}\left(  1-\xi_{i}%
+\tau_{0}\right)  \text{,}%
\]
so that the total allowed eigenenergy $E_{\mathbf{\tau}_{2}}$ of the
constrained primal normal eigenlocus component $\mathbf{\tau}_{2}$ satisfies
the equation%
\[
\left\Vert \mathbf{\tau}_{2}\right\Vert _{\min_{c}}^{2}-\mathbf{\tau}_{2}%
^{T}\mathbf{\tau}_{1}\cong\sum\nolimits_{i=1}^{l_{2}}\psi_{2_{i_{\ast}}%
}\left(  1-\xi_{i}+\tau_{0}\right)  \text{.}%
\]
It follows that the total allowed eigenenergy $E_{\mathbf{\tau}_{2}}$ of the
constrained primal normal eigenlocus component $\mathbf{\tau}_{2}$ is
determined by the expression%
\begin{equation}
\left\Vert \mathbf{\tau}_{2}\right\Vert _{\min_{c}}^{2}-\left\Vert
\mathbf{\tau}_{2}\right\Vert \left\Vert \mathbf{\tau}_{1}\right\Vert
\cos\theta_{\mathbf{\tau}_{2}\mathbf{\tau}_{1}}\cong\sum\nolimits_{i=1}%
^{l_{2}}\psi_{2_{i_{\ast}}}\left(  1-\xi_{i}+\tau_{0}\right)  \text{,}
\label{Minimal Eigenenergy Normal Eigenlocus Two}%
\end{equation}
where the total allowed eigenenergy $E_{\mathbf{\tau}_{2}}$ of $\mathbf{\tau
}_{2}$%
\[
E_{\mathbf{\tau}_{2}}\triangleq\left\Vert \mathbf{\tau}_{2}\right\Vert
_{\min_{c}}^{2}-\left\Vert \mathbf{\tau}_{2}\right\Vert \left\Vert
\mathbf{\tau}_{1}\right\Vert \cos\theta_{\mathbf{\tau}_{2}\mathbf{\tau}_{1}%
}\text{,}%
\]
is determined by a statistical equation%
\[
E_{\mathbf{\tau}_{2}}=\sum\nolimits_{i=1}^{l_{2}}\psi_{2_{i_{\ast}}}\left(
1-\xi_{i}+\tau_{0}\right)  \text{,}%
\]
in terms of integrated magnitudes $\sum\nolimits_{i=1}^{l_{2}}\psi
_{2_{i_{\ast}}}$ of Wolfe dual normal eigenaxis components which are
correlated with extreme vectors that belong to the $\boldsymbol{X}_{2}$
pattern set, and the $\tau_{0}$ statistic. An algebraic expression is now
obtained for the total allowed eigenenergy of a constrained primal normal
eigenlocus $\mathbf{\tau}$. Denote the total allowed eigenenergy $\left\Vert
\mathbf{\tau}\right\Vert _{\min_{c}}^{2}$ of $\mathbf{\tau}$ by
$E_{\mathbf{\tau}}$.

\subsubsection{The Total Allowed Eigenenergy of $\mathbf{\tau}$}

Summation over the complete algebraic system of strong dual normal eigenlocus
equations satisfied by $\mathbf{\tau}_{1}$%
\[
\left(  \sum\nolimits_{i=1}^{l_{1}}\psi_{1_{i_{\ast}}}\mathbf{x}_{1_{i_{\ast}%
}}^{T}\right)  \mathbf{\tau}\cong\sum\nolimits_{i=1}^{l_{1}}\psi_{1_{i_{\ast}%
}}\left(  1-\xi_{i}-\tau_{0}\right)  \text{,}%
\]
and by $\mathbf{\tau}_{2}$%
\[
\left(  -\sum\nolimits_{i=1}^{l_{2}}\psi_{2_{i_{\ast}}}\mathbf{x}_{2_{i\ast}%
}^{T}\right)  \mathbf{\tau}\cong\sum\nolimits_{i=1}^{l_{2}}\psi_{2_{i_{\ast}}%
}\left(  1-\xi_{i}+\tau_{0}\right)  \text{,}%
\]
produces the following expression for the total allowed eigenenergy
$E_{\mathbf{\tau}}$ of $\mathbf{\tau}$
\begin{align*}
&  \left(  \sum\nolimits_{i=1}^{l_{1}}\psi_{1_{i_{\ast}}}\mathbf{x}%
_{1_{i_{\ast}}}^{T}-\sum\nolimits_{i=1}^{l_{2}}\psi_{2_{i_{\ast}}}%
\mathbf{x}_{2_{i\ast}}^{T}\right)  \mathbf{\tau}\\
&  \cong\sum\nolimits_{i=1}^{l_{1}}\psi_{1_{i_{\ast}}}\left(  1-\xi_{i}%
-\tau_{0}\right)  +\sum\nolimits_{i=1}^{l_{2}}\psi_{2_{i_{\ast}}}\left(
1-\xi_{i}+\tau_{0}\right)  \text{,}%
\end{align*}
which reduces to%
\begin{align*}
\left(  \mathbf{\tau}_{1}-\mathbf{\tau}_{2}\right)  ^{T}\mathbf{\tau}  &
\cong\sum\nolimits_{i=1}^{l_{1}}\psi_{1_{j_{\ast}}}\left(  1-\xi_{i}-\tau
_{0}\right) \\
&  +\sum\nolimits_{i=1}^{l_{2}}\psi_{2_{i_{\ast}}}\left(  1-\xi_{i}+\tau
_{0}\right)  \text{,}\\
&  \cong\sum\nolimits_{i=1}^{l}\psi_{i_{\ast}}\left(  1-\xi_{i}\right)
\text{,}%
\end{align*}
so that the total allowed eigenenergy $E_{\mathbf{\tau}}$ of $\mathbf{\tau} $
\[
\left(  \mathbf{\tau}_{1}-\mathbf{\tau}_{2}\right)  ^{T}\mathbf{\tau
=}\left\Vert \mathbf{\tau}\right\Vert _{\min_{c}}^{2}\text{,}%
\]
is determined by a statistical equation
\[
E_{\mathbf{\tau}}=\sum\nolimits_{i=1}^{l}\psi_{i_{\ast}}\left(  1-\xi
_{i}\right)  \text{,}%
\]
solely in terms of the integrated lengths of the Wolfe dual normal eigenaxis
components on $\mathbf{\psi}$. It follows that the total allowed eigenenergy
$E_{\mathbf{\tau}}$ of a constrained primal normal eigenlocus $\mathbf{\tau}$
is determined by the integrated magnitudes $\psi_{i_{\ast}} $ of the Wolfe
dual normal eigenaxis components $\psi_{i\ast}\overrightarrow{\mathbf{e}%
}_{i\ast}$ on $\mathbf{\psi}$%
\begin{align}
\left\Vert \mathbf{\tau}\right\Vert _{\min_{c}}^{2}  &  \cong\sum
\nolimits_{i=1}^{l}\psi_{i_{\ast}}\left(  1-\xi_{i}\right) \label{TAE TAU}\\
&  \cong\sum\nolimits_{i=1}^{l}\psi_{i_{\ast}}-\sum\nolimits_{i=1}^{l}%
\psi_{i_{\ast}}\xi_{i}\text{,}\nonumber
\end{align}
where the regularization parameters $\xi_{i}=\xi\ll1$ are seen to determine
negligible constraints on $E_{\mathbf{\tau}}$.

The next part of the analysis will identify the manner in which the total
allowed eigenenergies of $\mathbf{\tau}_{1}$ and $\mathbf{\tau}_{2}$ are
symmetrically balanced with each other.

\subsection{Balancing the Total Allowed Eigenenergies of $\mathbf{\tau}_{1}$
and $\mathbf{\tau}_{2}$}

Returning to Eq. (\ref{Balancing Factor for Eigenlocus Components}), recall
that the critical minimum eigenenergies of $\mathbf{\tau}_{1}$ and
$\mathbf{\tau}_{2}$ satisfy a state of statistical equilibrium
\[
\left(  \left\Vert \mathbf{\tau}_{1}\right\Vert _{\min_{c}}^{2}-\left\Vert
\mathbf{\tau}_{1}\right\Vert \left\Vert \mathbf{\tau}_{2}\right\Vert
\cos\theta_{\mathbf{\tau}_{1}\mathbf{\tau}_{2}}\right)  +\nabla_{eq}%
\Leftrightarrow\left(  \left\Vert \mathbf{\tau}_{2}\right\Vert _{\min_{c}}%
^{2}-\left\Vert \mathbf{\tau}_{1}\right\Vert \left\Vert \mathbf{\tau}%
_{2}\right\Vert \cos\theta_{\mathbf{\tau}_{1}\mathbf{\tau}_{2}}\right)
-\nabla_{eq}\text{,}%
\]
in relation to a centrally located statistical fulcrum $f_{s}$, where
$\nabla_{eq}$ is a symmetric equalizer statistic. Statistical expressions are
now obtained for the symmetric equalizer statistic $\nabla_{eq}$ and the
statistical fulcrum $f_{s}$.

Using Eq. (\ref{Minimal Eigenenergy Normal Eigenlocus One}) and the KKT
constraint of Eq. (\ref{Equilibrium Constraint on Dual Eigen-components})
\begin{align*}
\left\Vert \mathbf{\tau}_{1}\right\Vert _{\min_{c}}^{2}-\left\Vert
\mathbf{\tau}_{1}\right\Vert \left\Vert \mathbf{\tau}_{2}\right\Vert
\cos\theta_{\mathbf{\tau}_{1}\mathbf{\tau}_{2}}  &  \cong\sum\nolimits_{i=1}%
^{l_{1}}\psi_{1_{i_{\ast}}}\left(  1-\xi_{i}-\tau_{0}\right)  \text{,}\\
&  \cong\frac{1}{2}\sum\nolimits_{i=1}^{l}\psi_{_{i_{\ast}}}\left(  1-\xi
_{i}-\tau_{0}\right)  \text{,}%
\end{align*}
it follows that the total allowed eigenenergy $E_{\mathbf{\tau}_{1}}$ of
$\mathbf{\tau}_{1}$%
\[
E_{\mathbf{\tau}_{1}}\triangleq\left\Vert \mathbf{\tau}_{1}\right\Vert
_{\min_{c}}^{2}-\left\Vert \mathbf{\tau}_{1}\right\Vert \left\Vert
\mathbf{\tau}_{2}\right\Vert \cos\theta_{\mathbf{\tau}_{1}\mathbf{\tau}_{2}%
}\text{,}%
\]
is constrained to satisfy the statistical equation:%
\begin{equation}
E_{\mathbf{\tau}_{1}}=\frac{1}{2}\sum\nolimits_{i=1}^{l}\psi_{_{i_{\ast}}%
}\left(  1-\xi_{i}\right)  -\frac{\tau_{0}}{2}\sum\nolimits_{i=1}^{l}%
\psi_{_{i_{\ast}}}\text{.} \label{TAE TAU COMP1}%
\end{equation}

Using Eq. (\ref{Minimal Eigenenergy Normal Eigenlocus Two}) and the KKT
constraint of Eq. (\ref{Equilibrium Constraint on Dual Eigen-components})
\begin{align*}
\left\Vert \mathbf{\tau}_{2}\right\Vert _{\min_{c}}^{2}-\left\Vert
\mathbf{\tau}_{2}\right\Vert \left\Vert \mathbf{\tau}_{1}\right\Vert
\cos\theta_{\mathbf{\tau}_{2}\mathbf{\tau}_{1}}  &  \cong\sum\nolimits_{i=1}%
^{l_{2}}\psi_{2_{i_{\ast}}}\left(  1-\xi_{i}+\tau_{0}\right)  \text{,}\\
&  \cong\frac{1}{2}\sum\nolimits_{i=1}^{l}\psi_{_{i_{\ast}}}\left(  1-\xi
_{i}+\tau_{0}\right)  \text{,}%
\end{align*}
it follows that the total allowed eigenenergy $E_{\mathbf{\tau}_{2}}$ of
$\mathbf{\tau}_{2}$%
\[
E_{\mathbf{\tau}_{2}}\triangleq\left\Vert \mathbf{\tau}_{2}\right\Vert
_{\min_{c}}^{2}-\left\Vert \mathbf{\tau}_{2}\right\Vert \left\Vert
\mathbf{\tau}_{1}\right\Vert \cos\theta_{\mathbf{\tau}_{2}\mathbf{\tau}_{1}%
}\text{,}%
\]
is constrained to satisfy the statistical equation:%
\begin{equation}
E_{\mathbf{\tau}_{2}}=\frac{1}{2}\sum\nolimits_{i=1}^{l}\psi_{_{i_{\ast}}%
}\left(  1-\xi_{i}\right)  +\frac{\tau_{0}}{2}\sum\nolimits_{i=1}^{l}%
\psi_{_{i_{\ast}}}\text{.} \label{TAE TAU COMP2}%
\end{equation}

Using Eqs (\ref{TAE TAU}), (\ref{TAE TAU COMP1}), and (\ref{TAE TAU COMP2}),
it follows that the total allowed eigenenergies of the constrained primal
normal eigenaxis components on $\mathbf{\tau}_{1}$
\begin{align*}
E_{\mathbf{\tau}_{1}}  &  =\frac{1}{2}\sum\nolimits_{i=1}^{l}\psi_{_{i_{\ast}%
}}\left(  1-\xi_{i}\right)  -\frac{\tau_{0}}{2}\sum\nolimits_{i=1}^{l}%
\psi_{_{i_{\ast}}}\text{,}\\
&  \cong\frac{1}{2}\left\Vert \mathbf{\tau}\right\Vert _{\min_{c}}^{2}\left(
1-\xi_{i}\right)  -\frac{\tau_{0}}{2}\left\Vert \mathbf{\tau}\right\Vert
_{\min_{c}}^{2}\text{,}%
\end{align*}
and the total allowed eigenenergies of the constrained primal normal eigenaxis
components on $\mathbf{\tau}_{2}$
\begin{align*}
E_{\mathbf{\tau}_{2}}  &  =\frac{1}{2}\sum\nolimits_{i=1}^{l}\psi_{_{i_{\ast}%
}}\left(  1-\xi_{i}\right)  +\frac{\tau_{0}}{2}\sum\nolimits_{i=1}^{l}%
\psi_{_{i_{\ast}}}\text{,}\\
&  \cong\frac{1}{2}\left\Vert \mathbf{\tau}\right\Vert _{\min_{c}}^{2}\left(
1-\xi_{i}\right)  +\frac{\tau_{0}}{2}\left\Vert \mathbf{\tau}\right\Vert
_{\min_{c}}^{2}\text{,}%
\end{align*}
are symmetrically balanced with each other by means of the symmetric equalizer
statistic $\nabla_{eq}$:%
\begin{equation}
\nabla_{eq}=\frac{\tau_{0}}{2}\left\Vert \mathbf{\tau}\right\Vert _{\min_{c}%
}^{2}\text{,} \label{Symmetric Difference Factor}%
\end{equation}
in relation to the centrally located statistical fulcrum $f_{s}$:%
\begin{align}
f_{s}  &  =\frac{1}{2}\sum\nolimits_{i=1}^{l}\psi_{_{i_{\ast}}}\left(
1-\xi_{i}\right)  \text{,}\label{Statistical Fulcrum}\\
&  \approx\frac{1}{2}\left\Vert \mathbf{\tau}\right\Vert _{\min_{c}}%
^{2}\text{,}\nonumber
\end{align}
which is half the total allowed eigenenergy $\frac{1}{2}E_{\mathbf{\tau}}$ of
a strong dual normal eigenlocus $\mathbf{\tau}$.

Equations (\ref{Symmetric Difference Factor}) and (\ref{Statistical Fulcrum})
are now used to define the state of statistical equilibrium of a strong dual
normal eigenlocus.

\subsection{The State of Statistical Equilibrium of a Strong Dual Normal
Eigenlocus}

Let $E_{\mathbf{\tau}_{1}}\triangleq\left\Vert \mathbf{\tau}_{1}\right\Vert
_{\min_{c}}^{2}-\left\Vert \mathbf{\tau}_{1}\right\Vert \left\Vert
\mathbf{\tau}_{2}\right\Vert \cos\theta_{\mathbf{\tau}_{1}\mathbf{\tau}_{2}}$
and $E_{\mathbf{\tau}_{2}}\triangleq\left\Vert \mathbf{\tau}_{2}\right\Vert
_{\min_{c}}^{2}-\left\Vert \mathbf{\tau}_{2}\right\Vert \left\Vert
\mathbf{\tau}_{1}\right\Vert \cos\theta_{\mathbf{\tau}_{2}\mathbf{\tau}_{1}}$.
Using Eqs (\ref{Balancing Factor for Eigenlocus Components}),
(\ref{Symmetric Difference Factor}), and (\ref{Statistical Fulcrum}), it
follows that the total allowed eigenenergies of $\mathbf{\tau}_{1}$ and
$\mathbf{\tau}_{2}$ satisfy the state of statistical equilibrium
\begin{equation}
E_{\mathbf{\tau}_{1}}+\frac{\tau_{0}}{2}\left\Vert \mathbf{\tau}\right\Vert
_{\min_{c}}^{2}\Leftrightarrow E_{\mathbf{\tau}_{2}}-\frac{\tau_{0}}%
{2}\left\Vert \mathbf{\tau}\right\Vert _{\min_{c}}^{2}\text{,}
\label{State of Statistical Equilibrium}%
\end{equation}
in relation to the centrally located statistical fulcrum $f_{s}$%
\begin{align*}
f_{s}  &  =\frac{1}{2}\sum\nolimits_{i=1}^{l}\psi_{_{i_{\ast}}}\text{,}\\
&  \cong\frac{1}{2}\left\Vert \mathbf{\tau}\right\Vert _{\min_{c}}^{2}\text{,}%
\end{align*}
which is located at half the critical eigenenergy $\frac{1}{2}E_{\mathbf{\tau
}}$ of a strong dual normal eigenlocus $\mathbf{\tau}$. Clearly, then, the
$\tau_{0}$ term plays a significant role in balancing the total allowed
eigenenergies of $\mathbf{\tau}_{1}$ and $\mathbf{\tau}_{2}$. The geometric
and statistical meaning of Eq. (\ref{State of Statistical Equilibrium}) is
considered next.

\subsection{The Statistical Machinery Behind the Balancing Feat}

It has been shown that correlated normal eigenaxis components on
$\mathbf{\psi}$ and $\mathbf{\tau}$ exhibit directional symmetry. It has also
been shown that joint distributions of normal eigenaxis components on
$\mathbf{\psi}$ and $\mathbf{\tau}$ are symmetrically distributed over the
Wolfe dual normal eigenaxis components $\psi_{1i\ast}%
\overrightarrow{\mathbf{e}}_{1i\ast}$ and $\psi_{2i\ast}%
\overrightarrow{\mathbf{e}}_{2i\ast}$ on $\mathbf{\psi}$ and the constrained
primal normal eigenaxis components $\psi_{1_{i_{\ast}}}\mathbf{x}_{1_{i_{\ast
}}}$ and $\psi_{2_{i\ast}}\mathbf{x}_{2_{i\ast}}$ on $\mathbf{\mathbf{\tau}%
}_{1}$ and $\mathbf{\mathbf{\tau}}_{2}$. In addition, the previous analysis
has demonstrated how strong duality relationships between the component
lengths of $\mathbf{\psi}$ and $\mathbf{\tau}$ ensure that products of joint
component lengths, i.e., the total allowed eigenenergies $\left\Vert
\psi_{1_{i_{\ast}}}\mathbf{x}_{1_{i_{\ast}}}\right\Vert _{\min_{c}}^{2}$ and
$\left\Vert \psi_{2_{i_{\ast}}}\mathbf{x}_{2_{i_{\ast}}}\right\Vert _{\min
_{c}}^{2}$ of the constrained primal normal eigenaxis components
$\psi_{1_{i_{\ast}}}\mathbf{x}_{1_{i_{\ast}}}$ and $\psi_{2_{i\ast}}%
\mathbf{x}_{2_{i\ast}}$, are symmetrically balanced with each other.

A geometric and statistical explanation of Eq.
(\ref{State of Statistical Equilibrium}) is now obtained. The explanation uses
the general machinery of a fulcrum and a lever, where a lever is any rigid
object capable of turning about some fixed point called a fulcrum. If a
fulcrum is placed under directly under a lever's center of gravity, the lever
will remain balanced. If a lever is of uniform dimensions and density, then
the center of gravity is at the geometric center of the lever. Consider for
example, the playground device known as a seesaw or teeter-totter. The center
of gravity is at the geometric center of a teeter-totter, which is where the
fulcrum of a seesaw is located
\citet{Asimov1966}%
.

\subsection{Statistical Machinery of a Statistical Fulcrum and a Statistical
Lever}

Consider an implicit horizontal axis capable of rotating about some fixed
point, where the horizontal axis is a statistical lever, and the fixed point
is a centrally located statistical fulcrum. Let the joint eigenenergies of
$\mathbf{\psi}$ and $\mathbf{\tau}$ be symmetrically distributed over the
statistical lever, and let the statistical fulcrum $f_{s}$ be located directly
under the statistical lever's center of eigenenergy, which is defined to be
half the critical minimum eigenenergy $\frac{1}{2}\left\Vert \mathbf{\tau
}\right\Vert _{\min_{c}}^{2}$ of a strong dual normal eigenlocus
$\mathbf{\tau}$. Using Eq. (\ref{State of Statistical Equilibrium}), let the
statistical lever be subjected to equal and opposite eigenenergies associated
with $\mathbf{\mathbf{\tau}}$, $\mathbf{\mathbf{\tau}}_{1}$, and
$\mathbf{\mathbf{\tau}}_{2}$, in terms of $E_{\mathbf{\tau}_{1}}+\frac
{\tau_{0}}{2}\left\Vert \mathbf{\tau}\right\Vert _{\min_{c}}^{2}$ and
$E_{\mathbf{\tau}_{2}}-\frac{\tau_{0}}{2}\left\Vert \mathbf{\tau}\right\Vert
_{\min_{c}}^{2}$, where the symmetric equalizer statistic $\nabla_{eq}%
=\frac{\tau_{0}}{2}\left\Vert \mathbf{\tau}\right\Vert _{\min_{c}}^{2}$
ensures that the total allowed eigenenergy $E_{\mathbf{\tau}_{1}}$ of
$\mathbf{\mathbf{\tau}}_{1}$ and the total allowed eigenenergy
$E_{\mathbf{\tau}_{2}}$ of $\mathbf{\mathbf{\tau}}_{2}$ are symmetrically
balanced with each other. Given that equal and opposite eigenenergies are
applied to the statistical lever, it follows that the statistical lever
achieves a state of statistical equilibrium.

Figure $29$ illustrates the elegant statistical machinery which ensures that
the total allowed eigenenergies of $\mathbf{\mathbf{\tau}}_{1}$ and
$\mathbf{\mathbf{\tau}}_{2}$ are symmetrically balanced with each other. The
statistical lever is depicted by a gray bar and the statistical fulcrum is
depicted by a purple triangle. The statistical lever, which pivots about the
statistical fulcrum, is subjected to equal and opposite eigenenergies
associated with $\mathbf{\mathbf{\tau}}$, $\mathbf{\mathbf{\tau}}_{1}$, and
$\mathbf{\mathbf{\tau}}_{2}$, and therefore achieves a state of statistical equilibrium.

\begin{center}%
\begin{center}
\includegraphics[
natheight=7.499600in,
natwidth=9.999800in,
height=3.2897in,
width=4.3777in
]%
{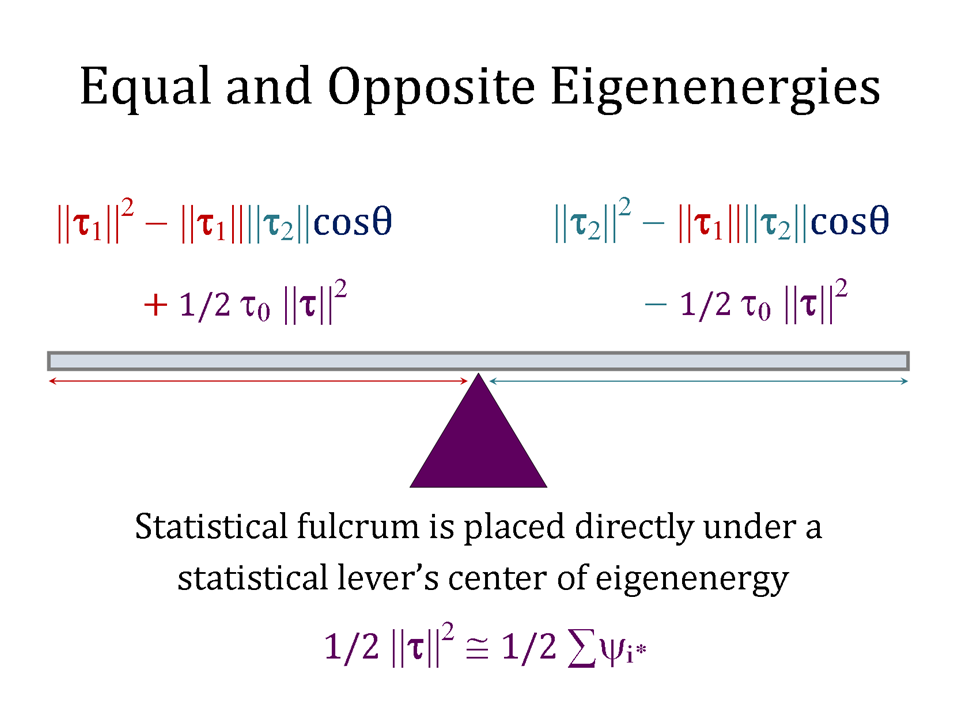}%
\end{center}

\end{center}

\begin{flushleft}
Figure $29$: Illustration of the statistical machinery that is used to
symmetrically balance the eigenenergies of $\mathbf{\mathbf{\tau}}_{1}$ and
$\mathbf{\mathbf{\tau}}_{2}$. The statistical lever, which is subjected to
equal and opposite eigenenergies of $\mathbf{\mathbf{\tau}}$,
$\mathbf{\mathbf{\tau}}_{1}$, and $\mathbf{\mathbf{\tau}}_{2}$, achieves a
state of statistical equilibrium.
\end{flushleft}

\subsection{Characteristics of the State of Statistical Equilibrium}

The state of statistical equilibrium that is achieved by a strong dual normal
eigenlocus results from the total allowed eigenenergy $\left\Vert
\mathbf{\tau}\right\Vert _{\min_{c}}^{2}\cong\left\Vert \mathbf{\tau}%
_{1}-\mathbf{\tau}_{2}\right\Vert _{\min_{c}}^{2}$ of $\mathbf{\tau}$ and the
total allowed eigenenergy $\left\Vert \mathbf{\psi}\right\Vert _{\min_{c}}%
^{2}$ of $\mathbf{\psi}$ being determined by joint symmetrical distributions
of the eigenenergies of $\mathbf{\psi}$ and $\mathbf{\tau}$, whereby the
eigenenergies of the eigen-scaled extreme training points on $\mathbf{\tau
}_{1}$ and $\mathbf{\tau}_{2}$ are distributed in a manner which symmetrically
balances the eigenenergies of $\mathbf{\tau}_{1}$ and $\mathbf{\tau}_{2}$.
Thereby, the total allowed eigenenergy $\left\Vert \mathbf{\tau}\right\Vert
_{\min_{c}}^{2}\cong\left\Vert \mathbf{\tau}_{1}-\mathbf{\tau}_{2}\right\Vert
_{\min_{c}}^{2}$ of $\mathbf{\tau}$ satisfies a linear decision boundary and
the bilaterally symmetrical borders which bound it.

The state of statistical equilibrium is characterized by joint symmetrical
distributions of the normal eigenaxis components on $\mathbf{\psi}$ and
$\mathbf{\tau}$, over all of the eigen-scaled extreme training points on
$\mathbf{\tau}_{1}$ and $\mathbf{\tau}_{2}$, whereby symmetrical distributions
of the critical minimum eigenenergies of $\mathbf{\psi}$ and $\mathbf{\tau}$
jointly satisfy the law of cosines for strong dual normal eigenlocus
transforms in Eq.
(\ref{Law of Cosines for Strong Dual Normal Eiigenlocus Transforms}), in
symmetrical and elegant manners.

Indeed, \emph{all of the eigen-scaled extreme points on} $\mathbf{\tau}_{1}$
\emph{and} $\mathbf{\tau}_{2}$ \emph{possess such eigen-balanced locations},
that the inner product statistic $\left\Vert \mathbf{\tau}_{1}\right\Vert
\left\Vert \mathbf{\tau}_{2}\right\Vert \cos\theta_{\mathbf{\tau}%
_{1}\mathbf{\tau}_{2}}$ coupled with the critical minimum eigenenergy
$\left\Vert \mathbf{\tau}_{1}\right\Vert _{\min_{c}}^{2}$ of the constrained
primal normal eigenlocus component $\mathbf{\tau_{1}}$
\[
\left\Vert \mathbf{\tau}_{1}\right\Vert _{\min_{c}}^{2}-\left\Vert
\mathbf{\tau}_{1}\right\Vert \left\Vert \mathbf{\tau}_{2}\right\Vert
\cos\theta_{\mathbf{\tau}_{1}\mathbf{\tau}_{2}}\text{,}%
\]
and the inner product statistic $\left\Vert \mathbf{\tau}_{2}\right\Vert
\left\Vert \mathbf{\tau}_{1}\right\Vert \cos\theta_{\mathbf{\tau}%
_{2}\mathbf{\tau}_{1}}$ coupled with the critical minimum eigenenergy
$\left\Vert \mathbf{\tau}_{2}\right\Vert _{\min_{c}}^{2}$ of the constrained
primal normal eigenlocus component $\mathbf{\tau}_{2}$
\[
\left\Vert \mathbf{\tau}_{2}\right\Vert _{\min_{c}}^{2}-\left\Vert
\mathbf{\tau}_{2}\right\Vert \left\Vert \mathbf{\tau}_{1}\right\Vert
\cos\theta_{\mathbf{\tau}_{2}\mathbf{\tau}_{1}}\text{,}%
\]
are effectively balanced with the critical minimum eigenenergy $\left\Vert
\mathbf{\tau}\right\Vert _{\min_{c}}^{2}$ of the constrained primal normal
eigenlocus $\mathbf{\tau}$:%
\[
\left\{  \left\Vert \mathbf{\tau}_{1}\right\Vert _{\min_{c}}^{2}-\left\Vert
\mathbf{\tau}_{1}\right\Vert \left\Vert \mathbf{\tau}_{2}\right\Vert
\cos\theta_{\mathbf{\tau}_{1}\mathbf{\tau}_{2}}\right\}  +\left\{  \left\Vert
\mathbf{\tau}_{2}\right\Vert _{\min_{c}}^{2}-\left\Vert \mathbf{\tau}%
_{2}\right\Vert \left\Vert \mathbf{\tau}_{1}\right\Vert \cos\theta
_{\mathbf{\tau}_{1}\mathbf{\tau}_{2}}\right\}  \cong\left\Vert \mathbf{\tau
}\right\Vert _{\min_{c}}^{2}\text{.}%
\]
Correspondingly, \emph{all of the Wolfe dual normal eigenaxis components on}
$\mathbf{\psi}$ \emph{possess such eigen-balanced magnitudes}, that component
lengths of $\mathbf{\psi}$ which regulate the critical minimum eigenenergy of
$\mathbf{\tau}_{1}$%
\[
E_{\mathbf{\tau}_{1}}=\sum\nolimits_{i=1}^{l_{1}}\psi_{1_{i_{\ast}}}-\tau
_{0}\sum\nolimits_{i=1}^{l_{1}}\psi_{1_{i_{\ast}}}\text{,}%
\]
and component lengths of $\mathbf{\psi}$ which regulate the critical minimum
eigenenergy of $\mathbf{\tau}_{2}$%
\[
E_{\mathbf{\tau}_{2}}=\sum\nolimits_{i=1}^{l_{2}}\psi_{2_{i_{\ast}}}+\tau
_{0}\sum\nolimits_{i=1}^{l_{2}}\psi_{2_{i_{\ast}}}\text{,}%
\]
are effectively balanced with component lengths of $\mathbf{\psi}$ which
regulate the critical minimum eigenenergy of $\mathbf{\tau}$:%
\begin{align*}
E_{\mathbf{\tau}}  &  =\sum\nolimits_{i=1}^{l_{1}}\psi_{1_{i_{\ast}}}\left(
1-\tau_{0}\right)  +\sum\nolimits_{i=1}^{l_{2}}\psi_{2_{i_{\ast}}}\left(
1+\tau_{0}\right)  \text{,}\\
&  \cong\sum\nolimits_{i=1}^{l_{1}}\psi_{1_{i_{\ast}}}+\sum\nolimits_{i=1}%
^{l_{2}}\psi_{2_{i_{\ast}}}\text{,}\\
&  \cong\frac{1}{2}\sum\nolimits_{i=1}^{l}\psi_{_{i_{\ast}}}+\frac{1}{2}%
\sum\nolimits_{i=1}^{l}\psi_{_{i_{\ast}}}\text{,}\\
&  \cong\sum\nolimits_{i=1}^{l}\psi_{_{i_{\ast}}}\text{.}%
\end{align*}
Figure $30$ illustrates how the statistical equilibrium point of a strong dual
normal eigenlocus $\mathbf{\tau}$ is characterized by joint symmetrical
distributions of the critical minimum eigenenergies of $\mathbf{\psi}$ and
$\mathbf{\tau}$ which jointly satisfy the law of cosines for strong dual
normal eigenlocus transforms. Thereby, the eigenenergies of the eigen-scaled
extreme training points on $\mathbf{\tau}_{1}$ and $\mathbf{\tau}_{2}$ are
distributed in a symmetric manner which symmetrically balances the
eigenenergies of $\mathbf{\tau}_{1}$ and $\mathbf{\tau}_{2}$.

\begin{center}%
\begin{center}
\includegraphics[
natheight=7.499600in,
natwidth=9.999800in,
height=3.2897in,
width=4.3777in
]%
{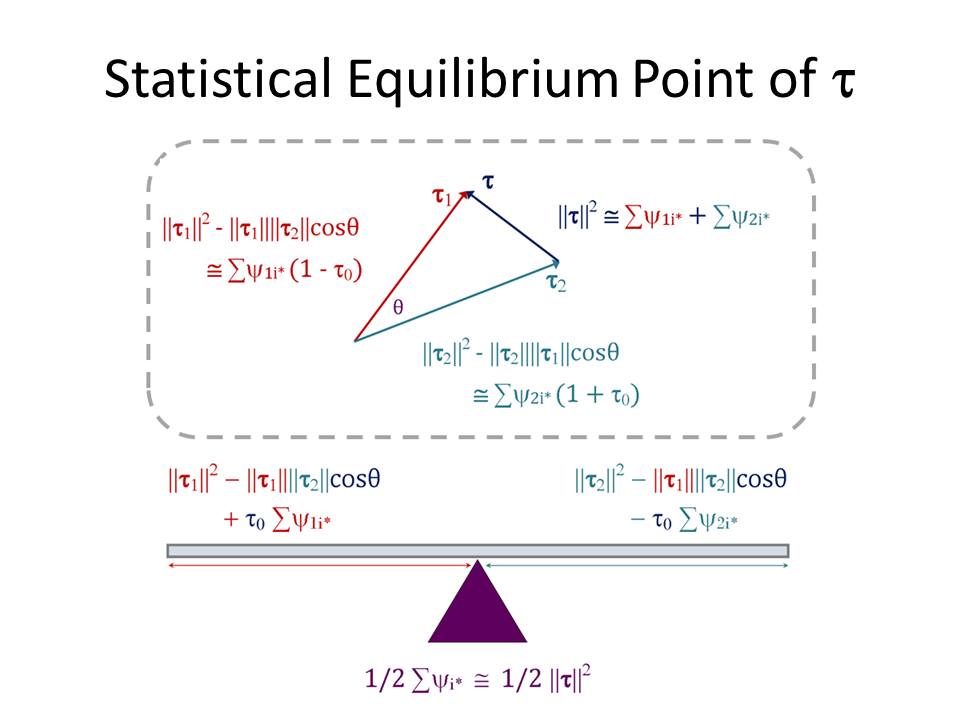}%
\end{center}

\end{center}

\begin{flushleft}
Figure $30$: Illustration of the geometric and topological relationships which
determine the critical minimum eigenenergies and component lengths of a Wolfe
dual and a constrained primal normal eigenlocus. Figure $30$ shows how the
statistical equilibrium point (the dual statistical eigenlocus) of
$\mathbf{\tau}$ is determined by symmetrically balanced eigenenergies of the
constrained pair of primal normal eigenlocus components $\mathbf{\tau}%
_{1}-\mathbf{\tau}_{2}$ on $\mathbf{\tau}$. The critical minimum eigenenergies
of $\mathbf{\psi}$ and $\mathbf{\tau}$ jointly satisfy the law of cosines for
strong dual normal eigenlocus transforms.
\end{flushleft}

The law of cosines which is satisfied by strong dual normal eigenlocus
transforms will be formally referred to as \emph{the strong dual normal
eigenlocus identity}. The components on a Wolfe dual normal eigenlocus
$\mathbf{\psi}$ and a constrained primal normal eigenlocus $\mathbf{\tau=\tau
}_{1}-\mathbf{\tau}_{2}$ satisfy the strong dual normal eigenlocus identity
that is outlined next. Without loss of generality, regularization parameters
$\xi_{i}$ are not included in the identity.

\subsection{The Strong Dual Normal Eigenlocus Identity}

Figure $30$ shows that the total allowed eigenenergies $\left\Vert
\mathbf{\tau}\right\Vert _{\min_{c}}^{2}\cong\left\Vert \mathbf{\tau}%
_{1}-\mathbf{\tau}_{2}\right\Vert _{\min_{c}}^{2}$ of a strong dual normal
eigenlocus $\mathbf{\tau=\tau}_{1}-\mathbf{\tau}_{2}$%
\begin{align*}
\left\Vert \mathbf{\tau}\right\Vert _{\min_{c}}^{2}  &  \cong\left\Vert
\mathbf{\tau}_{1}-\mathbf{\tau}_{2}\right\Vert _{\min_{c}}^{2}\text{,}\\
&  \cong\left\{  \left\Vert \mathbf{\tau}_{1}\right\Vert _{\min_{c}}%
^{2}-\left\Vert \mathbf{\tau}_{1}\right\Vert \left\Vert \mathbf{\tau}%
_{2}\right\Vert \cos\theta_{\mathbf{\tau}_{1}\mathbf{\tau}_{2}}\right\}
+\left\{  \left\Vert \mathbf{\tau}_{2}\right\Vert _{\min_{c}}^{2}-\left\Vert
\mathbf{\tau}_{2}\right\Vert \left\Vert \mathbf{\tau}_{1}\right\Vert
\cos\theta_{\mathbf{\tau}_{1}\mathbf{\tau}_{2}}\right\}  \text{,}%
\end{align*}
and the corresponding total allowed eigenenergy of $\left\Vert \mathbf{\psi
}\right\Vert _{\min_{c}}^{2}$ of a Wolfe dual normal eigenlocus $\mathbf{\psi
}$, \emph{jointly satisfy the law of cosines in a surprisingly elegant and
symmetric manner}%
\begin{align}
\left\Vert \mathbf{\tau}\right\Vert _{\min_{c}}^{2}  &  \cong\left\{
\left\Vert \mathbf{\tau}_{1}\right\Vert _{\min_{c}}^{2}-\left\Vert
\mathbf{\tau}_{1}\right\Vert \left\Vert \mathbf{\tau}_{2}\right\Vert
\cos\theta_{\mathbf{\tau}_{1}\mathbf{\tau}_{2}}\right\}  +\left\{  \left\Vert
\mathbf{\tau}_{2}\right\Vert _{\min_{c}}^{2}-\left\Vert \mathbf{\tau}%
_{2}\right\Vert \left\Vert \mathbf{\tau}_{1}\right\Vert \cos\theta
_{\mathbf{\tau}_{1}\mathbf{\tau}_{2}}\right\}  \text{,}\nonumber\\
&  \cong\sum\nolimits_{i=1}^{l_{1}}\psi_{1_{i_{\ast}}}\left(  1-\tau
_{0}\right)  +\sum\nolimits_{i=1}^{l_{2}}\psi_{2_{i_{\ast}}}\left(  1+\tau
_{0}\right)  \text{,}\nonumber\\
&  \cong\sum\nolimits_{i=1}^{l_{1}}\psi_{1_{i_{\ast}}}+\sum\nolimits_{i=1}%
^{l_{2}}\psi_{2_{i_{\ast}}}\text{,}\nonumber\\
&  \cong\frac{1}{2}\left\Vert \mathbf{\tau}\right\Vert _{\min_{c}}^{2}%
+\frac{1}{2}\left\Vert \mathbf{\tau}\right\Vert _{\min_{c}}^{2}\text{,}%
\nonumber\\
&  \cong\frac{1}{2}\sum\nolimits_{i=1}^{l}\psi_{_{i_{\ast}}}+\frac{1}{2}%
\sum\nolimits_{i=1}^{l}\psi_{_{i_{\ast}}}\text{,}\nonumber\\
&  \cong\left\Vert \mathbf{\tau}_{1}-\mathbf{\tau}_{2}\right\Vert _{\min_{c}%
}^{2}\text{,}\nonumber\\
&  \cong\sum\nolimits_{i=1}^{l}\psi_{_{i_{\ast}}}\text{.}
\label{Strong Dual Normal Eigenlocus Identity}%
\end{align}

Equation (\ref{Strong Dual Normal Eigenlocus Identity}) is given the name of
the strong dual normal eigenlocus identity. Figure $30$ and Eq.
(\ref{Strong Dual Normal Eigenlocus Identity}) illustrate the algebraic and
geometric nature of the symmetrical relationships between the total allowed
eigenenergies of the components of $\mathbf{\tau}$ and the components of
$\mathbf{\psi}$.

Equation (\ref{Strong Dual Normal Eigenlocus Identity}) shows how the total
allowed eigenenergy $E_{\mathbf{\tau}}$ of a strong dual normal eigenlocus
$\mathbf{\tau}$ is regulated by the integrated component lengths
$\sum\nolimits_{i=1}^{l}\psi_{_{i_{\ast}}}$ of a Wolfe dual normal eigenlocus
$\mathbf{\psi}$
\begin{align*}
\left\Vert \mathbf{\tau}\right\Vert _{\min_{c}}^{2}  &  \cong\sum
\nolimits_{i=1}^{l}\psi_{_{i_{\ast}}}\text{,}\\
&  \cong2\sum\nolimits_{i=1}^{l_{1}}\psi_{1_{i_{\ast}}}\text{,}\\
&  \cong2\sum\nolimits_{i=1}^{l_{2}}\psi_{2_{i_{\ast}}}\text{,}%
\end{align*}
where $\sum\nolimits_{i=1}^{l_{1}}\psi_{1_{i_{\ast}}}\equiv\sum\nolimits_{i=1}%
^{l_{2}}\psi_{2_{i_{\ast}}}$, in terms of symmetrically weighted, symmetrical
sets of integrated component lengths%
\begin{align*}
\left\Vert \mathbf{\tau}\right\Vert _{\min_{c}}^{2}  &  \cong\sum
\nolimits_{i=1}^{l_{1}}\psi_{1_{i_{\ast}}}-\tau_{0}\sum\nolimits_{i=1}^{l_{1}%
}\psi_{1_{i_{\ast}}}\\
&  +\sum\nolimits_{i=1}^{l_{2}}\psi_{2_{i_{\ast}}}+\tau_{0}\sum\nolimits_{i=1}%
^{l_{2}}\psi_{2_{i_{\ast}}}\text{,}%
\end{align*}
where the symmetrically weighted, integrated component lengths $\frac{1}%
{2}\left(  1-\tau_{0}\right)  \sum\nolimits_{i=1}^{l}\psi_{_{i_{\ast}}}$ of
$\mathbf{\psi}$ which regulate the total allowed eigenenergy $E_{\mathbf{\tau
}_{1}}$ of $\mathbf{\tau}_{1}$%
\begin{align*}
E_{\mathbf{\tau}_{1}}  &  =\sum\nolimits_{i=1}^{l_{1}}\psi_{1_{i_{\ast}}%
}\left(  1-\tau_{0}\right)  \text{,}\\
&  \cong\frac{1}{2}\left(  1-\tau_{0}\right)  \sum\nolimits_{i=1}^{l}%
\psi_{_{i_{\ast}}}\text{,}\\
&  \cong\frac{1}{2}\left(  1-\tau_{0}\right)  \left\Vert \mathbf{\tau
}\right\Vert _{\min_{c}}^{2}\text{,}%
\end{align*}
and the symmetrically weighted, integrated component lengths $\frac{1}%
{2}\left(  1+\tau_{0}\right)  \sum\nolimits_{i=1}^{l}\psi_{_{i_{\ast}}}$ of
$\mathbf{\psi}$ which regulate the total allowed eigenenergy $E_{\mathbf{\tau
}_{2}}$ of $\mathbf{\tau}_{2}$%
\begin{align*}
E_{\mathbf{\tau}_{2}}  &  =\sum\nolimits_{i=1}^{l_{2}}\psi_{2_{i_{\ast}}%
}\left(  1+\tau_{0}\right)  \text{,}\\
&  \cong\frac{1}{2}\left(  1+\tau_{0}\right)  \sum\nolimits_{i=1}^{l}%
\psi_{_{i_{\ast}}}\text{,}\\
&  \cong\frac{1}{2}\left(  1+\tau_{0}\right)  \left\Vert \mathbf{\tau
}\right\Vert _{\min_{c}}^{2}\text{,}%
\end{align*}
determine symmetrically balanced, critical minimum eigenenergies that sum to
the total allowed eigenenergy $E_{\mathbf{\tau}}$ of $\mathbf{\tau}$%
\begin{align*}
E_{\mathbf{\tau}_{1}}+E_{\mathbf{\tau}_{2}}  &  =\sum\nolimits_{i=1}^{l_{1}%
}\psi_{1_{i_{\ast}}}\left(  1-\tau_{0}\right)  +\sum\nolimits_{i=1}^{l_{2}%
}\psi_{2_{i_{\ast}}}\left(  1+\tau_{0}\right)  \text{,}\\
&  \cong\frac{1}{2}\left(  1-\tau_{0}\right)  \sum\nolimits_{i=1}^{l}%
\psi_{_{i_{\ast}}}+\frac{1}{2}\left(  1+\tau_{0}\right)  \sum\nolimits_{i=1}%
^{l}\psi_{_{i_{\ast}}}\text{,}\\
&  \cong\frac{1}{2}\left(  1-\tau_{0}\right)  \left\Vert \mathbf{\tau
}\right\Vert _{\min_{c}}^{2}+\frac{1}{2}\left(  1+\tau_{0}\right)  \left\Vert
\mathbf{\tau}\right\Vert _{\min_{c}}^{2}\text{,}\\
&  \cong\left\Vert \mathbf{\tau}\right\Vert _{\min_{c}}^{2}\text{,}\\
&  =E_{\mathbf{\tau}}\text{.}%
\end{align*}

It has previously been shown that joint distributions of the normal eigenaxis
components on $\mathbf{\psi}$ and $\mathbf{\tau}$ are symmetrically
distributed over the Wolfe dual normal eigenaxis $\psi_{1i\ast}%
\overrightarrow{\mathbf{e}}_{1i\ast}$ and $\psi_{2i\ast}%
\overrightarrow{\mathbf{e}}_{2i\ast}$ on $\mathbf{\psi}$ and the constrained
primal normal eigenaxis components $\psi_{1_{i_{\ast}}}\mathbf{x}_{1_{i_{\ast
}}}$ and $\psi_{2_{i\ast}}\mathbf{x}_{2_{i\ast}}$ on $\mathbf{\mathbf{\tau}%
}_{1}$ and $\mathbf{\mathbf{\tau}}_{2}$. Figure $30$ and Eq.
(\ref{Strong Dual Normal Eigenlocus Identity}) illustrate how the total
allowed eigenenergies of the constrained primal normal eigenaxis components
$\psi_{1_{i_{\ast}}}\mathbf{x}_{1_{i_{\ast}}}$ and $\psi_{2_{i\ast}}%
\mathbf{x}_{2_{i\ast}}$ on $\mathbf{\mathbf{\tau=\tau}}_{1}-\mathbf{\tau}_{2}$
are regulated by symmetrical relationships between the total allowed
eigenenergies of $\mathbf{\psi}$ and $\mathbf{\tau}$. Indeed, the total
allowed eigenenergies of the constrained primal normal eigenaxis components on
$\mathbf{\tau}_{1}$%
\[
\left\Vert \sum\nolimits_{i=1}^{l_{1}}\psi_{1_{i_{\ast}}}\mathbf{x}%
_{1_{i_{\ast}}}\right\Vert _{\min_{c}}^{2}-\left\Vert \sum\nolimits_{i=1}%
^{l_{1}}\psi_{1_{i_{\ast}}}\mathbf{x}_{1_{i_{\ast}}}\right\Vert \left\Vert
\sum\nolimits_{i=1}^{l_{2}}\psi_{2_{i_{\ast}}}\mathbf{x}_{2_{i_{\ast}}%
}\right\Vert \cos\theta_{\mathbf{\tau}_{1}\mathbf{\tau}_{2}}\text{,}%
\]
are regulated by the statistical equations%
\begin{align*}
E_{\mathbf{\tau}_{1}}  &  =\sum\nolimits_{i=1}^{l_{1}}\psi_{1_{i_{\ast}}%
}\left(  1-\tau_{0}\right)  \text{,}\\
&  \cong\frac{1}{2}\left(  1-\tau_{0}\right)  \sum\nolimits_{i=1}^{l}%
\psi_{_{i_{\ast}}}\text{,}\\
&  \cong\frac{1}{2}\left(  1-\tau_{0}\right)  \left\Vert \mathbf{\tau
}\right\Vert _{\min_{c}}^{2}\text{,}%
\end{align*}
and the total allowed eigenenergies of the constrained primal normal eigenaxis
components on $\mathbf{\tau}_{2}$%
\[
\left\Vert \sum\nolimits_{i=1}^{l_{2}}\psi_{2_{i_{\ast}}}\mathbf{x}%
_{2_{i_{\ast}}}\right\Vert _{\min_{c}}^{2}-\left\Vert \sum\nolimits_{i=1}%
^{l_{1}}\psi_{1_{i_{\ast}}}\mathbf{x}_{1_{i_{\ast}}}\right\Vert \left\Vert
\sum\nolimits_{i=1}^{l_{2}}\psi_{2_{i_{\ast}}}\mathbf{x}_{2_{i_{\ast}}%
}\right\Vert \cos\theta_{\mathbf{\tau}_{1}\mathbf{\tau}_{2}}\text{,}%
\]
are regulated by the statistical equations%
\begin{align*}
E_{\mathbf{\tau}_{2}}  &  =\sum\nolimits_{i=1}^{l_{2}}\psi_{2_{i_{\ast}}%
}\left(  1+\tau_{0}\right)  \text{,}\\
&  \cong\frac{1}{2}\left(  1+\tau_{0}\right)  \sum\nolimits_{i=1}^{l}%
\psi_{_{i_{\ast}}}\text{,}\\
&  \cong\frac{1}{2}\left(  1+\tau_{0}\right)  \left\Vert \mathbf{\tau
}\right\Vert _{\min_{c}}^{2}\text{,}%
\end{align*}
whereby the total allowed eigenenergies of the constrained primal normal
eigenaxis components on $\mathbf{\mathbf{\tau}}_{1}$ and $\mathbf{\mathbf{\tau
}}_{2}$%
\[
\left\Vert \sum\nolimits_{i=1}^{l_{1}}\psi_{1_{i_{\ast}}}\mathbf{x}%
_{1_{i_{\ast}}}-\sum\nolimits_{i=1}^{l_{2}}\psi_{2_{i_{\ast}}}\mathbf{x}%
_{2_{i_{\ast}}}\right\Vert _{\min_{c}}^{2}\text{,}%
\]
are determined by joint symmetrical distributions of the eigenenergies of
$\mathbf{\psi}$ and $\mathbf{\tau}$.

\subsection*{Configurations of Regularized Geometric Architectures}

The learning machine architecture which has been examined in this paper
exhibits a surprising amount of bilateral symmetry for arbitrary data
distributions. Given the previous analyses, it is concluded that robust and
symmetrical linear partitions of arbitrary feature spaces are achieved by
means of symmetrically balanced normal eigenaxis components in dual,
correlated Hilbert spaces.

Returning to Figs. $12$ and $20$, it is concluded that the regularized,
data-driven geometric architectures determined by strong dual normal
eigenlocus transforms are configured by enforcing joint symmetrical
distributions of the eigenenergies of $\mathbf{\psi}$ and $\mathbf{\tau}$ over
the eigen-scaled extreme training vectors on $\mathbf{\tau}_{1}$ and
$\mathbf{\tau}_{2}$, whereby the eigenenergies of the strong dual normal
eigenlocus components $\mathbf{\tau}_{1}$ and $\mathbf{\tau}_{2}$ on
$\mathbf{\tau}$ are symmetrically balanced with each other.

Figure $31$ illustrates how the joint eigenenergies of $\mathbf{\psi}$ and
$\mathbf{\tau}$ are symmetrically distributed over the constrained primal
normal eigenaxis components on the constrained primal normal eigenlocus
$\mathbf{\mathbf{\tau}=\tau}_{1}-\mathbf{\tau}_{2}$.

\begin{center}%
\begin{center}
\includegraphics[
natheight=7.499600in,
natwidth=9.999800in,
height=3.5405in,
width=4.7115in
]%
{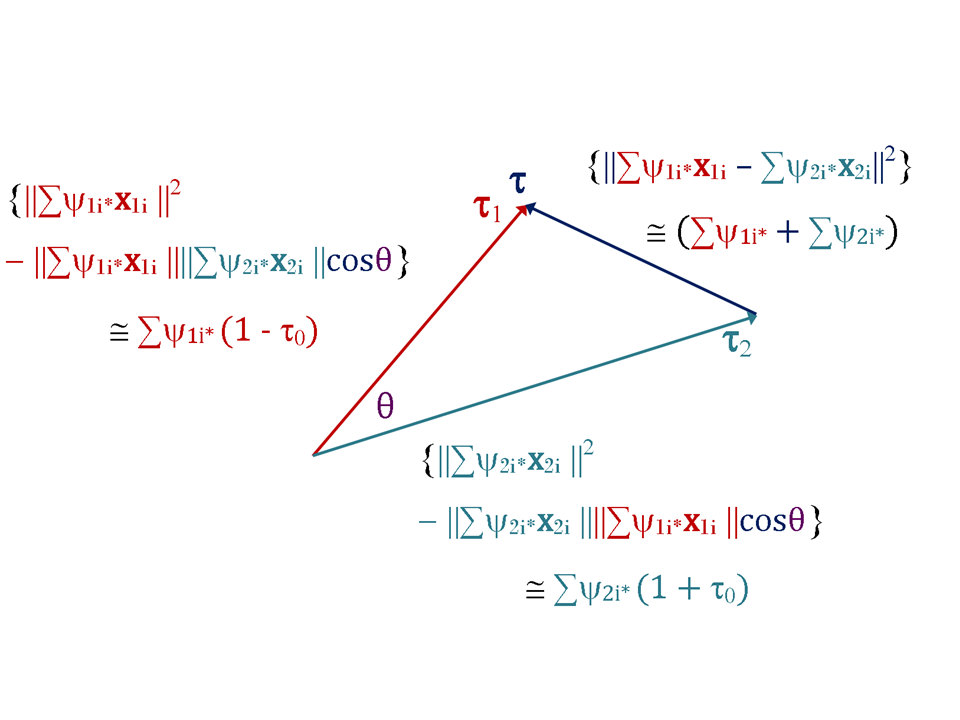}%
\end{center}

\end{center}

\begin{flushleft}
Figure $31$: Illustration of how the total allowed eigenenergies of the
constrained primal normal eigenaxis components on $\mathbf{\mathbf{\tau}}$,
$\mathbf{\mathbf{\tau}}_{1}$, and $\mathbf{\mathbf{\tau}}_{2}$ are determined
by joint symmetrical distributions of the eigenenergies of $\mathbf{\psi}$ and
$\mathbf{\tau}$. The joint eigenenergies of $\mathbf{\psi}$ and $\mathbf{\tau
}$ are symmetrically distributed over the constrained primal normal eigenaxis
components on $\mathbf{\mathbf{\tau}=\tau}_{1}-\mathbf{\tau}_{2}$.
\end{flushleft}

Properties of the symmetric equalizer statistic $\nabla_{eq}$ are examined next.

\subsection*{Geometric and Statistical Properties of the Symmetric Equalizer
Statistic $\nabla_{eq}$}

The statistical expression for the symmetric equalizer statistic $\nabla_{eq}$%
\begin{align*}
\nabla_{eq}  &  =\frac{\tau_{0}}{2}\sum\nolimits_{i=1}^{l}\psi_{i\ast}%
\text{,}\\
&  =\frac{\tau_{0}}{2}\left\Vert \mathbf{\tau}\right\Vert _{\min_{c}}%
^{2}\text{,}\\
&  =\frac{1}{2l}\left\Vert \mathbf{\tau}\right\Vert _{\min_{c}}^{2}%
y_{i}\left(  1-\xi_{i}\right) \\
&  -\frac{1}{2l}\left\Vert \mathbf{\tau}\right\Vert _{\min_{c}}^{2}%
\sum\nolimits_{i=1}^{l}\mathbf{x}_{i\ast}^{T}\sum\nolimits_{j=1}^{l_{1}}%
\psi_{1_{j_{\ast}}}\mathbf{x}_{1_{j_{\ast}}}\\
&  +\frac{1}{2l}\left\Vert \mathbf{\tau}\right\Vert _{\min_{c}}^{2}%
\sum\nolimits_{i=1}^{l}\mathbf{x}_{i\ast}^{T}\sum\nolimits_{j=1}^{l_{2}}%
\psi_{2_{j_{\ast}}}\mathbf{x}_{2_{j_{\ast}}}\text{,}%
\end{align*}
substantiates the significant and joint roles that the KKT constraint of Eq.
(\ref{Equilibrium Constraint on Dual Eigen-components}) and the $\tau_{0}$
term of Eq. (\ref{Normal Eigenlocus Projection Factor Two}) have in achieving
the state of statistical equilibrium that is exhibited by a strong dual normal
eigenlocus $\mathbf{\tau=\tau}_{1}-\mathbf{\tau}_{2}$.

Indeed, half of the total allowed eigenenergy $\frac{1}{2}\left\Vert
\mathbf{\tau}\right\Vert _{\min_{c}}^{2}$ of a strong dual normal eigenlocus
is described by integrated lengths of Wolfe dual normal eigenaxis components
correlated with each pattern category:%
\[
\frac{1}{2}\left\Vert \mathbf{\tau}\right\Vert _{\min_{c}}^{2}=\sum
\nolimits_{i=1}^{l_{1}}\psi_{1_{i_{\ast}}}=\sum\nolimits_{i=1}^{l_{2}}%
\psi_{2_{i_{\ast}}}\text{,}%
\]
where each Wolfe dual normal eigenaxis component exhibits a critical length
which describes the distribution of an extreme training point. Figure $29$
illustrates how the KKT constraint of Eq.
(\ref{Equilibrium Constraint on Dual Eigen-components}) determines a center of
eigenenergy $\frac{1}{2}\sum\nolimits_{i=1}^{l}\psi_{i\ast}$ for a strong dual
normal eigenlocus which is satisfied by half of its total allowed eigenenergy
$\frac{1}{2}\left\Vert \mathbf{\tau}\right\Vert _{\min}^{2}$. Figure $30$
illustrates how the KKT constraint of Eq.
(\ref{Equilibrium Constraint on Dual Eigen-components}) and the $\tau_{0}$
term of Eq. (\ref{Normal Eigenlocus Projection Factor Two}) jointly ensure
that the state of statistical equilibrium exhibited by $\mathbf{\tau=\tau}%
_{1}-\mathbf{\tau}_{2}$ is determined by symmetrically balanced eigenenergies
of $\mathbf{\tau}_{1}$ and $\mathbf{\tau}_{2}$.

Equation (\ref{Normal Eigenlocus Projection Factor Two}) indicates that the
$\tau_{0}$ term describes eigen-balanced correlations between the extreme
training points and the eigen-balanced eigenloci of the constrained primal
normal eigenaxis components on $\mathbf{\tau}_{1}$ and $\mathbf{\tau}_{2}$,
where the lengths of the Wolfe dual normal eigenaxis components and the KKT
constraint of Eq. (\ref{Equilibrium Constraint on Dual Eigen-components}) have
significant and joint roles in balancing highly interconnected sets of inner
product relationships amongst the Wolfe dual and the constrained primal normal
eigenaxis components. Figure $29$ illustrates how the $\tau_{0}$ term ensures
that the state of statistical equilibrium exhibited by $\mathbf{\tau=\tau}%
_{1}-\mathbf{\tau}_{2}$ is determined by equal and opposite eigenenergies of
$\mathbf{\tau}$, $\mathbf{\tau}_{1}$, and $\mathbf{\tau}_{2}$, where the joint
eigenenergies of $\mathbf{\tau}$ and $\mathbf{\psi}$ are symmetrically
distributed over $\mathbf{\tau=\tau}_{1}-\mathbf{\tau}_{2}$.

The expression for the symmetric equalizer statistic $\nabla_{eq}$
substantiates the conclusion that effective and consistent fits of constrained
primal normal eigenaxis components to training data requires satisfying joint
symmetrical distributions of eigenenergies over a constrained primal normal
eigenlocus of eigen-scaled extreme data points.

Figure $32$ illustrates the joint roles that the KKT constraint of Eq.
(\ref{Equilibrium Constraint on Dual Eigen-components}) and the $\tau_{0}$
term of Eq. (\ref{Normal Eigenlocus Projection Factor Two}) have in achieving
the surprisingly elegant statistical balancing feat that is routinely
accomplished by solving the inequality constrained optimization problem of Eq.
(\ref{Primal Normal Eigenlocus}). Given the data-driven symmetrical essence of
this statistical balancing feat, it is recommended that strong dual normal
eigenlocus transforms be applied to equal numbers of training examples from
each pattern class.

\begin{center}%
\begin{center}
\includegraphics[
natheight=7.499600in,
natwidth=9.999800in,
height=3.2897in,
width=4.3777in
]%
{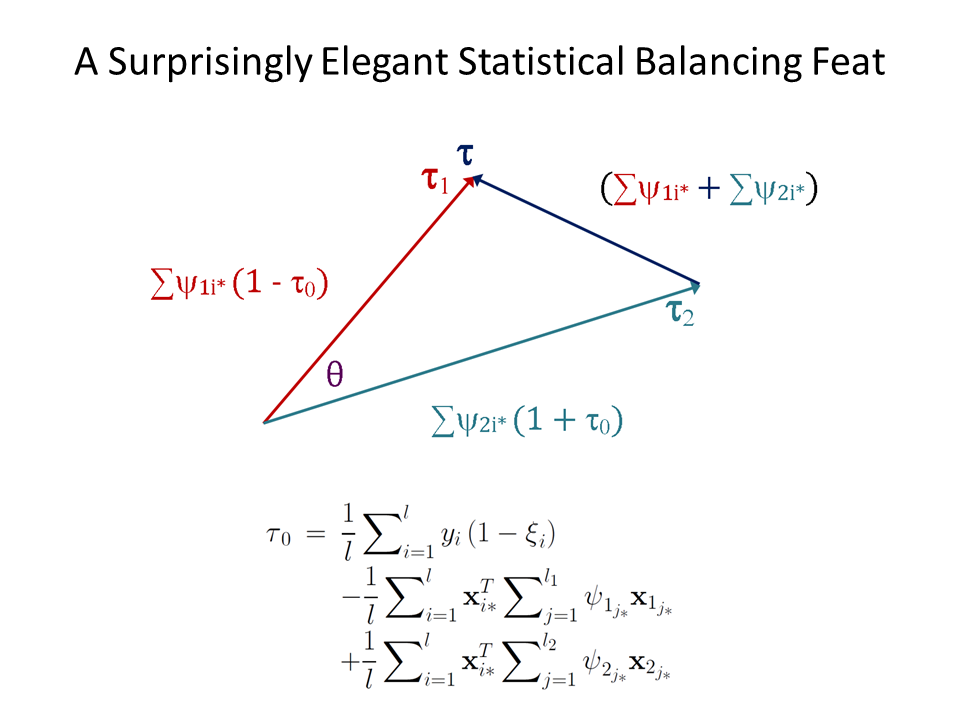}%
\end{center}

\end{center}

\begin{flushleft}
Figure $32$: Illustration of how symmetrical integrated sets $\sum
\nolimits_{i=1}^{l_{1}}\psi_{1_{i_{\ast}}}\equiv\sum\nolimits_{i=1}^{l_{2}%
}\psi_{2_{i_{\ast}}}$ of eigen-balanced magnitudes of Wolfe dual normal
eigenaxis components, both of which are symmetrically balanced by means of the
$\tau_{0}$ statistic, jointly ensure that the total allowed eigenenergies of
$\mathbf{\tau}_{1}$ and $\mathbf{\tau}_{2}$ are symmetrically balanced with
each other.
\end{flushleft}

The next section of the paper will examine probabilistic properties which are
exhibited by strong dual normal eigenlocus discriminant functions $D\left(
\mathbf{x}\right)  =\mathbf{\tau}^{T}\mathbf{x}+\tau_{0}$. An expression will
be obtained for the normal eigenlocus decision rule in Eq.
(\ref{NormalEigenlocusTestStatistic}) which encodes likelihood ratios.
Expressions for the likelihood ratio and the state of statistical equilibrium
will be used to show that constrained normal eigenlocus discriminant functions
describe linear decision boundaries for which class probabilities are
equivalent to each other. The likelihood ratio expression will also be used to
reexamine how width regulation of large covariance decision regions is
accomplished. It will also be shown that strong dual normal eigenlocus
discriminant functions encode Bayes' likelihood ratio for common covariance
data and a robust likelihood ratio for all other data distributions.

\section{Probabilistic Properties Exhibited by the Statistical Equilibrium
Point of a Strong Dual Normal Eigenlocus}

It will now be shown how strong dual discriminant functions $D\left(
\mathbf{x}\right)  =\mathbf{\tau}^{T}\mathbf{x}+\tau_{0}$ encode likelihood
ratios. Recall that each constrained primal normal eigenaxis
component\textbf{\ }$\psi_{1_{i\ast}}\mathbf{x}_{1_{i\ast}}$\textbf{\ }on
$\mathbf{\mathbf{\tau}}_{1}$,\textbf{\ }and each constrained primal normal
eigenaxis component\textbf{\ }$\psi_{2_{i\ast}}\mathbf{x}_{2_{i\ast}}%
$\textbf{\ }on $\mathbf{\mathbf{\tau}}_{2}$, provides a maximum\textit{\ }%
covariance estimate in a principal location, in the form of an eigen-balanced
first and second order statistical moment about the geometric locus of an
extreme data point $\mathbf{x}_{1_{i\ast}}$ or $\mathbf{x}_{2_{i\ast}}$. It
has been demonstrated that any given maximum covariance estimate provides a
measure of how much two groups of eigen-scaled extreme data points and their
common means vary from a given extreme data point. It has also been
demonstrated that any given maximum covariance estimate encodes a distribution
of first order coordinates for an extreme training vector $\mathbf{x}%
_{1_{i\ast}}$ or $\mathbf{x}_{2_{i\ast}}$, relative to the eigen-scaled
extreme vectors for two given data sets. Thereby, any given maximum covariance
estimate describes how the components of an extreme training vector are
distributed within a collection of eigen-scaled extreme training vectors. A
probabilistic explanation for the total allowed eigenenergies of the
constrained primal normal eigenaxis components on $\mathbf{\mathbf{\tau}}$ is
now obtained.

\subsection{A Probabilistic Explanation for the Total Allowed Eigenenergies of
$\mathbf{\mathbf{\tau}}$}

Consider a strong dual normal eigenlocus of constrained primal normal
eigenaxis components $\mathbf{\tau}=\sum\nolimits_{i=1}^{l_{1}}\psi
_{1_{i_{\ast}}}\mathbf{x}_{1_{i_{\ast}}}-\sum\nolimits_{i=1}^{l_{2}}%
\psi_{2_{i_{\ast}}}\mathbf{x}_{2_{i_{\ast}}}$ and take any constrained primal
normal eigenaxis component $\psi_{1_{i\ast}}\mathbf{x}_{1_{i\ast}}$%
\textbf{\ }on $\mathbf{\mathbf{\tau}}_{1}$. Given that $\psi_{1_{i\ast}%
}\mathbf{x}_{1_{i\ast}}$\textbf{\ }provides a maximum\textit{\ }covariance
estimate in a principal location in $\mathbf{%
\mathbb{R}
}^{d}$, it follows that the square $\left\Vert \psi_{1_{i\ast}}\mathbf{x}%
_{1_{i\ast}}\right\Vert _{\min_{c}}^{2}$ of\textbf{\ }$\psi_{1_{i\ast}%
}\mathbf{x}_{1_{i\ast}}$\textbf{\ }is the probability of finding the extreme
data point $\mathbf{x}_{1_{i\ast}}$ in a particular region of $\mathbf{%
\mathbb{R}
}^{d}$, where the integration $\left\Vert \psi_{1_{i\ast}}\mathbf{x}%
_{1_{i\ast}}\right\Vert _{\min_{c}}^{2}$ of the squared vector components of
$\psi_{1_{i\ast}}\mathbf{x}_{1_{i\ast}}$ is the total allowed eigenenergy of
the constrained primal normal eigenaxis component $\psi_{1_{i\ast}}%
\mathbf{x}_{1_{i\ast}}$.

Now take any constrained primal normal eigenaxis component $\psi_{2_{i\ast}%
}\mathbf{x}_{2_{i\ast}}$\textbf{\ }on $\mathbf{\mathbf{\tau}}_{2}$. Given that
$\psi_{2_{i\ast}}\mathbf{x}_{2_{i\ast}}$\textbf{\ }provides a
maximum\textit{\ }covariance estimate in a principal location in $\mathbf{%
\mathbb{R}
}^{d}$, it follows that the square $\left\Vert \psi_{2_{i\ast}}\mathbf{x}%
_{2_{i\ast}}\right\Vert _{\min_{c}}^{2}$ of $\psi_{2_{i\ast}}\mathbf{x}%
_{2_{i\ast}}$ is the probability of finding the extreme data point
$\mathbf{x}_{2_{i\ast}}$ in a particular region of $\mathbf{%
\mathbb{R}
}^{d}$, where the integration $\left\Vert \psi_{2_{i\ast}}\mathbf{x}%
_{2_{i\ast}}\right\Vert _{\min_{c}}^{2}$ of the squared vector components of
$\psi_{2_{i\ast}}\mathbf{x}_{2_{i\ast}}$ is the total allowed eigenenergy of
the constrained primal normal eigenaxis component $\psi_{2_{i\ast}}%
\mathbf{x}_{2_{i\ast}}$.

It follows that the total allowed eigenenergy $\left\Vert \mathbf{\tau
}\right\Vert _{\min_{c}}^{2}$ of $\mathbf{\tau}$%
\begin{align*}
\left\Vert \mathbf{\tau}\right\Vert _{\min_{c}}^{2}  &  \cong\left\Vert
\sum\nolimits_{i=1}^{l_{1}}\psi_{1_{i_{\ast}}}\mathbf{x}_{1_{i_{\ast}}}%
-\sum\nolimits_{i=1}^{l_{2}}\psi_{2_{i_{\ast}}}\mathbf{x}_{2_{i_{\ast}}%
}\right\Vert _{\min_{c}}^{2}\text{,}\\
&  \cong\left\Vert \sum\nolimits_{i=1}^{l_{1}}\psi_{1_{i_{\ast}}}%
\mathbf{x}_{1_{i_{\ast}}}\right\Vert _{\min_{c}}^{2}+\left\Vert \sum
\nolimits_{i=1}^{l_{2}}\psi_{2_{i_{\ast}}}\mathbf{x}_{2_{i_{\ast}}}\right\Vert
_{\min_{c}}^{2}\\
&  -2\left\Vert \sum\nolimits_{i=1}^{l_{1}}\psi_{1_{i_{\ast}}}\mathbf{x}%
_{1_{i_{\ast}}}\right\Vert \left\Vert \sum\nolimits_{i=1}^{l_{2}}%
\psi_{2_{i_{\ast}}}\mathbf{x}_{2_{i_{\ast}}}\right\Vert \cos\theta
_{\mathbf{\tau}_{1}\mathbf{\tau}_{2}}\text{,}%
\end{align*}
describes the probability of finding extreme data points in particular regions
of $\mathbf{%
\mathbb{R}
}^{d}$, which determines the probability of finding data points in regions of
large covariance between either overlapping or non-overlapping data distributions.

It will now be demonstrated that $\mathbf{\tau}$ encodes Bayes' likelihood
ratio for common covariance data distributions and a robust likelihood ratio
for all other data distributions. An expression is first obtained for
$\mathbf{\tau}$ which encodes likelihood ratios.

\subsection{Likelihood Ratios Encoded Within $\mathbf{\tau}$}

Returning to the expression for $\mathbf{\tau}$
\[
\mathbf{\tau}=\sum\nolimits_{i=1}^{l_{1}}\psi_{1_{i_{\ast}}}\mathbf{x}%
_{1_{i_{\ast}}}-\sum\nolimits_{i=1}^{l_{2}}\psi_{2_{i_{\ast}}}\mathbf{x}%
_{2_{i_{\ast}}}\text{,}%
\]
in Eq. (\ref{Pair of Normal Eigenlocus Components}), and substituting the
expressions for $\psi_{1_{i_{\ast}}}$ and $\psi_{2_{i_{\ast}}}$ in Eqs
(\ref{Dual Eigen-coordinate Locations Component One}) and
(\ref{Dual Eigen-coordinate Locations Component Two}) into the above
expression for $\mathbf{\tau}$, provides an expression for $\mathbf{\tau}$
\begin{align}
\mathbf{\tau}  &  =\lambda_{\max_{\psi}}^{-1}\sum\nolimits_{i=1}^{l_{1}}%
\frac{\mathbf{x}_{1_{i_{\ast}}}}{\left\Vert \mathbf{x}_{1_{i_{\ast}}%
}\right\Vert }\left\Vert \mathbf{x}_{1_{i_{\ast}}}\right\Vert ^{2}%
\times\label{Strong Dual Normal Eigenlocus}\\
&  \left[
\begin{array}
[c]{c}%
\sum\nolimits_{j=1}^{l_{1}}\psi_{1_{j\ast}}\left\Vert \mathbf{x}_{1_{j\ast}%
}\right\Vert \cos\theta_{\mathbf{x}_{1_{i\ast}}\mathbf{x}_{1_{j\ast}}}\\
-\sum\nolimits_{j=1}^{l_{2}}\psi_{2_{j\ast}}\left\Vert \mathbf{x}_{2_{j\ast}%
}\right\Vert \cos\theta_{\mathbf{x}_{1_{i\ast}}\mathbf{x}_{2_{j\ast}}}%
\end{array}
\right] \nonumber\\
&  -\lambda_{\max_{\psi}}^{-1}\sum\nolimits_{i=1}^{l_{2}}\frac{\mathbf{x}%
_{2_{i_{\ast}}}}{\left\Vert \mathbf{x}_{2_{i_{\ast}}}\right\Vert }\left\Vert
\mathbf{x}_{2_{i_{\ast}}}\right\Vert ^{2}\times\nonumber\\
&  \left[
\begin{array}
[c]{c}%
\sum\nolimits_{j=1}^{l_{2}}\psi_{2_{j\ast}}\left\Vert \mathbf{x}_{2_{j\ast}%
}\right\Vert \cos\theta_{\mathbf{x}_{2_{i\ast}}\mathbf{x}_{2_{j\ast}}}\\
-\sum\nolimits_{j=1}^{l_{1}}\psi_{1_{j\ast}}\left\Vert \mathbf{x}_{1_{j\ast}%
}\right\Vert \cos\theta_{\mathbf{x}_{2_{i\ast}}\mathbf{x}_{1_{j\ast}}}%
\end{array}
\right]  \text{,}\nonumber
\end{align}
which encodes likelihood ratios, where the terms $\frac{\left\Vert
\mathbf{x}_{1_{i_{\ast}}}\right\Vert }{\left\Vert \mathbf{x}_{1_{i_{\ast}}%
}\right\Vert }$ and $\frac{\left\Vert \mathbf{x}_{2_{i_{\ast}}}\right\Vert
}{\left\Vert \mathbf{x}_{2_{i_{\ast}}}\right\Vert }$ have been introduced and rearranged.

Recall that the eigenloci and corresponding lengths $\psi_{1_{i_{\ast}}}$ or
$\psi_{2_{i_{\ast}}}$ of the Wolfe dual normal eigenaxis components encode
pointwise covariance estimates, i.e., maximum covariance estimates in
principal locations, for the extreme vectors $\mathbf{x}_{1_{i_{\ast}}}$ or
$\mathbf{x}_{2_{i_{\ast}}}$. Given Eqs (\ref{Unidirectional Scaling Term One1}%
) and (\ref{Unidirectional Scaling Term Two1}), it follows that Eq.
(\ref{Strong Dual Normal Eigenlocus}) encodes pointwise covariance estimates
in the form of eigen-balanced, signed magnitudes along the axes of extreme
vectors, where eigen-balanced, signed magnitudes along the axes of the
$\mathbf{x}_{1_{i_{\ast}}}$ extreme vectors are denoted by
\[
\left[
\begin{array}
[c]{c}%
\sum\nolimits_{j=1}^{l_{1}}\psi_{1_{j\ast}}\left\Vert \mathbf{x}_{1_{j\ast}%
}\right\Vert \cos\theta_{\mathbf{x}_{1_{i\ast}}\mathbf{x}_{1_{j\ast}}}\\
-\sum\nolimits_{j=1}^{l_{2}}\psi_{2_{j\ast}}\left\Vert \mathbf{x}_{2_{j\ast}%
}\right\Vert \cos\theta_{\mathbf{x}_{1_{i\ast}}\mathbf{x}_{2_{j\ast}}}%
\end{array}
\right]  \text{,}%
\]
and eigen-balanced, signed magnitudes along the axes of the $\mathbf{x}%
_{2_{i_{\ast}}}$ extreme vectors are denoted by%
\[
\left[
\begin{array}
[c]{c}%
\sum\nolimits_{j=1}^{l_{2}}\psi_{2_{j\ast}}\left\Vert \mathbf{x}_{2_{j\ast}%
}\right\Vert \cos\theta_{\mathbf{x}_{2_{i\ast}}\mathbf{x}_{2_{j\ast}}}\\
-\sum\nolimits_{j=1}^{l_{1}}\psi_{1_{j\ast}}\left\Vert \mathbf{x}_{1_{j\ast}%
}\right\Vert \cos\theta_{\mathbf{x}_{2_{i\ast}}\mathbf{x}_{1_{j\ast}}}%
\end{array}
\right]  \text{.}%
\]
Denote the pointwise covariance estimates for the extreme vectors
$\mathbf{x}_{1_{i_{\ast}}}$ and $\mathbf{x}_{2_{i_{\ast}}}$ in Eq.
(\ref{Strong Dual Normal Eigenlocus}) by $\widehat{\operatorname{cov}%
}_{up_{\updownarrow}}\left(  \mathbf{x}_{1_{i_{\ast}}}\right)  $ and
$\widehat{\operatorname{cov}}_{up_{\updownarrow}}\left(  \mathbf{x}%
_{2_{i_{\ast}}}\right)  $ respectively. Using this notation, the expression
for $\mathbf{\tau}$ in Eq. (\ref{Strong Dual Normal Eigenlocus}) can be
rewritten in the following manner
\begin{align}
\mathbf{\tau}  &  =\lambda_{\max_{\psi}}^{-1}\sum\nolimits_{i=1}^{l_{1}%
}\left\Vert \sqrt{\widehat{\operatorname{cov}}_{up_{\updownarrow}}\left(
\mathbf{x}_{1_{i_{\ast}}}\right)  }\mathbf{x}_{1_{i_{\ast}}}\right\Vert
_{\min_{c}}^{2}\frac{\mathbf{x}_{1_{i_{\ast}}}}{\left\Vert \mathbf{x}%
_{1_{i_{\ast}}}\right\Vert }%
\label{Probabilisitc Expression for Normal Eigenlocus}\\
&  -\lambda_{\max_{\psi}}^{-1}\sum\nolimits_{i=1}^{l_{2}}\left\Vert
\sqrt{\widehat{\operatorname{cov}}_{up_{\updownarrow}}\left(  \mathbf{x}%
_{2_{i_{\ast}}}\right)  }\mathbf{x}_{2_{i_{\ast}}}\right\Vert _{\min_{c}}%
^{2}\frac{\mathbf{x}_{2_{i_{\ast}}}}{\left\Vert \mathbf{x}_{2_{i_{\ast}}%
}\right\Vert }\text{.}\nonumber
\end{align}
It follows that each constrained primal normal eigenaxis component
$\psi_{1i_{\ast}}\mathbf{x}_{1_{i_{\ast}}}$ on $\mathbf{\tau}_{1}$%
\[
\psi_{1i_{\ast}}\mathbf{x}_{1_{i_{\ast}}}\triangleq\lambda_{\max_{\psi}}%
^{-1}\left\Vert \sqrt{\widehat{\operatorname{cov}}_{up_{\updownarrow}}\left(
\mathbf{x}_{1_{i_{\ast}}}\right)  }\mathbf{x}_{1_{i_{\ast}}}\right\Vert
_{\min_{c}}^{2}\frac{\mathbf{x}_{1_{i_{\ast}}}}{\left\Vert \mathbf{x}%
_{1_{i_{\ast}}}\right\Vert }\text{,}%
\]
describes the probability $\lambda_{\max_{\psi}}^{-1}\left\Vert \sqrt
{\widehat{\operatorname{cov}}_{up_{\updownarrow}}\left(  \mathbf{x}%
_{1_{i_{\ast}}}\right)  }\mathbf{x}_{1_{i_{\ast}}}\right\Vert _{\min_{c}}^{2}$
of finding an extreme data point $\mathbf{x}_{1_{i\ast}}$ in a particular
region of $\mathbf{%
\mathbb{R}
}^{d}$. Thereby, the integrated set of constrained primal normal eigenaxis
components on $\mathbf{\tau}_{1}$%
\begin{equation}
\mathbf{\tau}_{1}=\lambda_{\max_{\psi}}^{-1}\sum\nolimits_{i=1}^{l_{1}%
}\left\Vert \sqrt{\widehat{\operatorname{cov}}_{up_{\updownarrow}}\left(
\mathbf{x}_{1_{i_{\ast}}}\right)  }\mathbf{x}_{1_{i_{\ast}}}\right\Vert
_{\min_{c}}^{2}\frac{\mathbf{x}_{1_{i_{\ast}}}}{\left\Vert \mathbf{x}%
_{1_{i_{\ast}}}\right\Vert }\text{,}
\label{Probabilisitc Expression for Normal Eigenlocus Component One}%
\end{equation}
describes the probabilities of finding all of the $\mathbf{x}_{1_{i\ast}}$
extreme data points in particular regions of $\mathbf{%
\mathbb{R}
}^{d}$, where all of the extreme data points $\mathbf{x}_{1_{i\ast}}$ are
located in regions of large covariance between either overlapping or
non-overlapping data distributions.

It also follows that each constrained primal normal eigenaxis component
$\psi_{2i_{\ast}}\mathbf{x}_{2_{i_{\ast}}}$ on $\mathbf{\tau}_{2}$%
\[
\psi_{2_{i_{\ast}}}\mathbf{x}_{2_{i_{\ast}}}\triangleq\lambda_{\max_{\psi}%
}^{-1}\left\Vert \sqrt{\widehat{\operatorname{cov}}_{up_{\updownarrow}}\left(
\mathbf{x}_{2_{i_{\ast}}}\right)  }\mathbf{x}_{2_{i_{\ast}}}\right\Vert
_{\min_{c}}^{2}\frac{\mathbf{x}_{2_{i_{\ast}}}}{\left\Vert \mathbf{x}%
_{2_{i_{\ast}}}\right\Vert }\text{,}%
\]
describes the probability $\lambda_{\max_{\psi}}^{-1}\left\Vert \sqrt
{\widehat{\operatorname{cov}}_{up_{\updownarrow}}\left(  \mathbf{x}%
_{2_{i_{\ast}}}\right)  }\mathbf{x}_{2_{i_{\ast}}}\right\Vert _{\min_{c}}^{2}$
of finding an extreme data point $\mathbf{x}_{2_{i\ast}}$ in a particular
region of $\mathbf{%
\mathbb{R}
}^{d}$. Thereby, the integrated set of constrained primal normal eigenaxis
components on $\mathbf{\tau}_{2}$%
\begin{equation}
\mathbf{\tau}_{2}=\lambda_{\max_{\psi}}^{-1}\sum\nolimits_{i=1}^{l_{2}%
}\left\Vert \sqrt{\widehat{\operatorname{cov}}_{up_{\updownarrow}}\left(
\mathbf{x}_{2_{i_{\ast}}}\right)  }\mathbf{x}_{2_{i_{\ast}}}\right\Vert
_{\min_{c}}^{2}\frac{\mathbf{x}_{2_{i_{\ast}}}}{\left\Vert \mathbf{x}%
_{2_{i_{\ast}}}\right\Vert }\text{,}
\label{Probabilisitc Expression for Normal Eigenlocus Component Two}%
\end{equation}
describes the probabilities of finding all of the $\mathbf{x}_{2_{i\ast}}$
extreme data points in particular regions of $\mathbf{%
\mathbb{R}
}^{d}$, where all of the extreme data points $\mathbf{x}_{2_{i\ast}}$ are
located in regions of large covariance between either overlapping or
non-overlapping data distributions.

It is concluded that the integrated set of constrained primal normal eigenaxis
components on $\mathbf{\tau}_{1}$ and $\mathbf{\tau}_{2}$ in Eq.
(\ref{Probabilisitc Expression for Normal Eigenlocus}) describes the
probabilities of finding each of the extreme data points in particular regions
of $\mathbf{%
\mathbb{R}
}^{d}$, where all of the extreme data points are located in regions of large
covariance between either overlapping or non-overlapping data distributions.
Thereby, it is concluded that Eq.
(\ref{Probabilisitc Expression for Normal Eigenlocus})\emph{ encodes robust
and data-driven likelihood ratios}.

It will now be shown that constrained normal eigenlocus discriminant functions
describe linear decision boundaries for which class probabilities are
equivalent to each other.

\subsubsection{Equivalence of Class Probabilities}

The total allowed eigenenergies $\left\Vert \mathbf{\tau}_{1}-\mathbf{\tau
}_{2}\right\Vert _{\min_{c}}^{2}$ of the constrained primal normal eigenlocus
components $\mathbf{\tau}_{1}$ and $\mathbf{\tau}_{2}$ on $\mathbf{\tau}$
satisfy a linear decision boundary and the bilaterally symmetrical borders
which bound it. Returning to Eq. (\ref{State of Statistical Equilibrium}), it
is concluded that the state of statistical equilibrium which is satisfied by
the total allowed eigenenergies of $\mathbf{\tau}_{1}$ and $\mathbf{\tau}_{2}$%
\[
E_{\mathbf{\tau}_{1}}+\frac{\tau_{0}}{2}\left\Vert \mathbf{\tau}\right\Vert
_{\min_{c}}^{2}\Leftrightarrow E_{\mathbf{\tau}_{2}}-\frac{\tau_{0}}%
{2}\left\Vert \mathbf{\tau}\right\Vert _{\min_{c}}^{2}\text{,}%
\]
in relation to the centrally located statistical fulcrum $f_{s}$%
\[
f_{s}=\frac{1}{2}\left\Vert \mathbf{\tau}\right\Vert _{\min_{c}}^{2}\text{,}%
\]
ensures that \emph{the integrated sum of probabilities encoded with the normal
eigenaxis components on} $\mathbf{\tau}_{1}$:%
\[
\lambda_{\max_{\psi}}^{-1}\sum\nolimits_{i=1}^{l_{1}}\left\Vert \sqrt
{\widehat{\operatorname{cov}}_{up_{\updownarrow}}\left(  \mathbf{x}%
_{1_{i_{\ast}}}\right)  }\mathbf{x}_{1_{i_{\ast}}}\right\Vert _{\min_{c}}%
^{2}\frac{\mathbf{x}_{1_{i_{\ast}}}}{\left\Vert \mathbf{x}_{1_{i_{\ast}}%
}\right\Vert }\text{,}%
\]
\emph{are balanced with the integrated sum of probabilities encoded with the
normal eigenaxis components on} $\mathbf{\tau}_{2}$:%
\[
\lambda_{\max_{\psi}}^{-1}\sum\nolimits_{i=1}^{l_{2}}\left\Vert \sqrt
{\widehat{\operatorname{cov}}_{up_{\updownarrow}}\left(  \mathbf{x}%
_{2_{i_{\ast}}}\right)  }\mathbf{x}_{2_{i_{\ast}}}\right\Vert _{\min_{c}}%
^{2}\frac{\mathbf{x}_{2_{i_{\ast}}}}{\left\Vert \mathbf{x}_{2_{i_{\ast}}%
}\right\Vert }\text{.}%
\]
It follows that the probabilities described by Eq.
(\ref{Probabilisitc Expression for Normal Eigenlocus Component One}) are
equivalent to the probabilities described by Eq.
(\ref{Probabilisitc Expression for Normal Eigenlocus Component Two}). Thereby,
the probabilities $P\left(  \boldsymbol{X}_{1}\right)  $ of the pattern class
$\boldsymbol{X}_{1}$ are equivalent to the probabilities $P\left(
\boldsymbol{X}_{2}\right)  $ of the pattern class $\boldsymbol{X}_{2}$%
\begin{equation}
P\left(  \boldsymbol{X}_{1}\right)  \equiv P\left(  \boldsymbol{X}_{2}\right)
\text{.} \label{Equivalence of Class Probabilities}%
\end{equation}
It is concluded that the statistical equilibrium point of $\mathbf{\tau}$
determines large covariance decision regions, i.e., linear decision boundaries
and regulated linear decision borders, for which class probabilities are
equivalent to each other.

Recall that the geometric loci of a linear decision boundary $D_{0}\left(
\mathbf{x}\right)  $ and its bilaterally symmetrical linear decision borders
$D_{1}\left(  \mathbf{x}\right)  $ and $D_{-1}\left(  \mathbf{x}\right)  $ are
determined by Eqs (\ref{Decision Boundary}), (\ref{Decision Border One}), and
(\ref{Decision Border Two}). Returning to Eqs
(\ref{Width of Linear Decision Region}), (\ref{Large Covariance Region One}),
and (\ref{Large Covariance Region Two}), recall that the eigenloci of the
constrained primal normal eigenlocus components $\mathbf{\tau}_{1}$ and
$\mathbf{\tau}_{2}$ regulate the geometric width, i.e., the breadth, of the
geometric region between the linear decision borders $D_{1}\left(
\mathbf{x}\right)  $ and $D_{-1}\left(  \mathbf{x}\right)  $. The eigenloci of
$\mathbf{\tau}_{1}$ and $\mathbf{\tau}_{2}$ also regulate the span of the
congruent geometric regions between the linear decision boundary $D_{0}\left(
\mathbf{x}\right)  $ and the linear decision borders $D_{+1}\left(
\mathbf{x}\right)  $ and $D_{-1}\left(  \mathbf{x}\right)  $. The next part of
the paper will reconsider how width regulation of large covariance decision
regions is accomplished. The analysis will provide probabilistic explanations
for how Eq. (\ref{Width of Linear Decision Region}) describes geometric
regions of large covariance between non-overlapping data distributions and
overlapping data distributions. The analysis will also provide probabilistic
explanations for how Eqs (\ref{Large Covariance Region One}) and
(\ref{Large Covariance Region Two}) describe disjoint tail regions between
non-overlapping data distributions, and bipartite, joint geometric regions of
large covariance\ between overlapping data distributions.

\subsection{Probabilistic Expressions of Decision Region Widths}

Consider again Eq. (\ref{Width of Linear Decision Region}):%
\[
D_{\left(  D_{1}\left(  \mathbf{x}\right)  -D_{-1}\left(  \mathbf{x}\right)
\right)  }=\frac{2}{\left\Vert \mathbf{\tau}_{1}-\mathbf{\tau}_{2}\right\Vert
}\text{,}%
\]
which describes the width of the geometric region of large covariance between
the linear decision borders $D_{1}\left(  \mathbf{x}\right)  $ and
$D_{-1}\left(  \mathbf{x}\right)  $. Substituting the expressions for
$\mathbf{\tau}_{1}$ in Eq.
(\ref{Probabilisitc Expression for Normal Eigenlocus Component One}) and
$\mathbf{\tau}_{2}$ in Eq.
(\ref{Probabilisitc Expression for Normal Eigenlocus Component Two}) into Eq.
(\ref{Width of Linear Decision Region}) provides a probabilistic expression
for the span of the geometric region of large covariance between the linear
decision borders $D_{1}\left(  \mathbf{x}\right)  $ and $D_{-1}\left(
\mathbf{x}\right)  $:%
\[
D_{\left(  D_{1}\left(  \mathbf{x}\right)  -D_{-1}\left(  \mathbf{x}\right)
\right)  }=2\left\Vert
\begin{array}
[c]{c}%
\lambda_{\max_{\psi}}^{-1}\sum\nolimits_{i=1}^{l_{1}}\left\Vert \sqrt
{\widehat{\operatorname{cov}}_{up_{\updownarrow}}\left(  \mathbf{x}%
_{1_{i_{\ast}}}\right)  }\mathbf{x}_{1_{i_{\ast}}}\right\Vert _{\min_{c}}%
^{2}\frac{\mathbf{x}_{1_{i_{\ast}}}}{\left\Vert \mathbf{x}_{1_{i_{\ast}}%
}\right\Vert }\\
-\lambda_{\max_{\psi}}^{-1}\sum\nolimits_{i=1}^{l_{2}}\left\Vert
\sqrt{\widehat{\operatorname{cov}}_{up_{\updownarrow}}\left(  \mathbf{x}%
_{2_{i_{\ast}}}\right)  }\mathbf{x}_{2_{i_{\ast}}}\right\Vert _{\min_{c}}%
^{2}\frac{\mathbf{x}_{2_{i_{\ast}}}}{\left\Vert \mathbf{x}_{2_{i_{\ast}}%
}\right\Vert }%
\end{array}
\right\Vert ^{-1}\text{.}%
\]
Given the above expression, it is concluded that the width of the geometric
region of large covariance between the linear decision borders $D_{1}\left(
\mathbf{x}\right)  $ and $D_{-1}\left(  \mathbf{x}\right)  $ is regulated by
probabilities that extreme data points are located in particular regions of
$\mathbf{%
\mathbb{R}
}^{d}$.

Now recall that the span of the congruent regions between the linear decision
boundary $D_{0}\left(  \mathbf{x}\right)  $ and the linear decision borders
$D_{+1}\left(  \mathbf{x}\right)  $ and $D_{-1}\left(  \mathbf{x}\right)  $ is
regulated by the expression%
\[
\frac{1}{\left\Vert \mathbf{\tau}_{1}-\mathbf{\tau}_{2}\right\Vert }\text{.}%
\]
Substituting the expressions for $\mathbf{\tau}_{1}$ in Eq.
(\ref{Probabilisitc Expression for Normal Eigenlocus Component One}) and
$\mathbf{\tau}_{2}$ in Eq.
(\ref{Probabilisitc Expression for Normal Eigenlocus Component Two}) into Eqs
(\ref{Large Covariance Region One}) and (\ref{Large Covariance Region Two}),
provides probabilistic expressions for the span of the congruent geometric
regions between the linear decision boundary $D_{0}\left(  \mathbf{x}\right)
$ and the linear decision borders $D_{+1}\left(  \mathbf{x}\right)  $ and
$D_{-1}\left(  \mathbf{x}\right)  $, where the width of the geometric region
of large covariance between the linear decision boundary $D_{0}\left(
\mathbf{x}\right)  $ and the linear decision border $D_{+1}\left(
\mathbf{x}\right)  $ described by the expression:%
\[
D_{\left(  D_{0}\left(  \mathbf{x}\right)  -D_{+1}\left(  \mathbf{x}\right)
\right)  }=\left\Vert
\begin{array}
[c]{c}%
\lambda_{\max_{\psi}}^{-1}\sum\nolimits_{i=1}^{l_{1}}\left\Vert \sqrt
{\widehat{\operatorname{cov}}_{up_{\updownarrow}}\left(  \mathbf{x}%
_{1_{i_{\ast}}}\right)  }\mathbf{x}_{1_{i_{\ast}}}\right\Vert _{\min_{c}}%
^{2}\frac{\mathbf{x}_{1_{i_{\ast}}}}{\left\Vert \mathbf{x}_{1_{i_{\ast}}%
}\right\Vert }\\
-\lambda_{\max_{\psi}}^{-1}\sum\nolimits_{i=1}^{l_{2}}\left\Vert
\sqrt{\widehat{\operatorname{cov}}_{up_{\updownarrow}}\left(  \mathbf{x}%
_{2_{i_{\ast}}}\right)  }\mathbf{x}_{2_{i_{\ast}}}\right\Vert _{\min_{c}}%
^{2}\frac{\mathbf{x}_{2_{i_{\ast}}}}{\left\Vert \mathbf{x}_{2_{i_{\ast}}%
}\right\Vert }%
\end{array}
\right\Vert ^{-1}\text{,}%
\]
and the width of the geometric region of large covariance between the linear
decision boundary $D_{0}\left(  \mathbf{x}\right)  $ and the linear decision
border $D_{-1}\left(  \mathbf{x}\right)  $ described by the equivalent
expression:%
\[
D_{\left(  D_{0}\left(  \mathbf{x}\right)  -D_{-1}\left(  \mathbf{x}\right)
\right)  }=\left\Vert
\begin{array}
[c]{c}%
\lambda_{\max_{\psi}}^{-1}\sum\nolimits_{i=1}^{l_{1}}\left\Vert \sqrt
{\widehat{\operatorname{cov}}_{up_{\updownarrow}}\left(  \mathbf{x}%
_{1_{i_{\ast}}}\right)  }\mathbf{x}_{1_{i_{\ast}}}\right\Vert _{\min_{c}}%
^{2}\frac{\mathbf{x}_{1_{i_{\ast}}}}{\left\Vert \mathbf{x}_{1_{i_{\ast}}%
}\right\Vert }\\
-\lambda_{\max_{\psi}}^{-1}\sum\nolimits_{i=1}^{l_{2}}\left\Vert
\sqrt{\widehat{\operatorname{cov}}_{up_{\updownarrow}}\left(  \mathbf{x}%
_{2_{i_{\ast}}}\right)  }\mathbf{x}_{2_{i_{\ast}}}\right\Vert _{\min_{c}}%
^{2}\frac{\mathbf{x}_{2_{i_{\ast}}}}{\left\Vert \mathbf{x}_{2_{i_{\ast}}%
}\right\Vert }%
\end{array}
\right\Vert ^{-1}\text{,}%
\]
is regulated by probabilities of finding extreme data points within particular
regions of $\mathbf{%
\mathbb{R}
}^{d}$.

Given the above expressions, it is concluded that the geometric loci of a
linear decision boundary and its bilaterally symmetrical linear decision
borders, which are determined by Eqs (\ref{Decision Boundary}),
(\ref{Decision Border One}), and (\ref{Decision Border Two}), are regulated by
probabilities of finding extreme data points within particular regions of $%
\mathbb{R}
^{d}$.

Given all of the above expressions and Eq.
(\ref{Equivalence of Class Probabilities}), it is also concluded that the
geometric loci of a linear decision boundary and its bilaterally symmetrical
linear decision borders, which are determined by Eqs (\ref{Decision Boundary}%
), (\ref{Decision Border One}), and (\ref{Decision Border Two}), are regulated
by class probabilities that are equivalent to each other.

\subsection*{Equivalence Between Bayes' Likelihood Ratio and the Normal
Eigenlocus Likelihood Ratio}

It will now be shown that the normal eigenlocus test statistic in Eq.
(\ref{NormalEigenlocusTestStatistic}) encodes the Bayes' likelihood ratio
expression for similar covariance data distributions. An expression for the
normal eigenlocus likelihood ratio is now obtained.

\subsection{The Normal Eigenlocus Likelihood Ratio}

Consider again the normal eigenlocus test statistic $\Lambda_{\mathbf{\tau}%
}\left(  \mathbf{x}\right)  \overset{H_{1}}{\underset{H_{2}}{\gtrless}}0$ in
Eq. (\ref{NormalEigenlocusTestStatistic})%
\begin{align*}
\Lambda_{\mathbf{\tau}}\left(  \mathbf{x}\right)   &  =\left(  \mathbf{x}%
-\frac{1}{l}\sum\nolimits_{i=1}^{l}\mathbf{x}_{i\ast}\right)  ^{T}%
\mathbf{\tau}_{1}\\
&  -\left(  \mathbf{x}-\frac{1}{l}\sum\nolimits_{i=1}^{l}\mathbf{x}_{i\ast
}\right)  ^{T}\mathbf{\tau}_{2}\\
&  \mathbf{+}\frac{1}{l}\sum\nolimits_{i=1}^{l}y_{i}\left(  1-\xi_{i}\right)
\overset{H_{1}}{\underset{H_{2}}{\gtrless}}0\text{,}%
\end{align*}
where $\mathbf{\tau=\tau}_{1}-\mathbf{\tau}_{2}$. Substituting the expressions
for $\mathbf{\mathbf{\tau}}_{1}$ and $\mathbf{\mathbf{\tau}}_{2}$ in Eqs
(\ref{Probabilisitc Expression for Normal Eigenlocus Component One}) and
(\ref{Probabilisitc Expression for Normal Eigenlocus Component Two}) into the
above expression for $\Lambda_{\mathbf{\tau}}\left(  \mathbf{x}\right)  $
provides an expression for the normal eigenlocus decision rule $\Lambda
_{\mathbf{\tau}_{1}\mathbf{-\tau}_{2}}\left(  \mathbf{x}\right)  $ in terms of
likelihoods%
\begin{align}
\Lambda_{\mathbf{\tau}_{1}\mathbf{-\tau}_{2}}\left(  \mathbf{x}\right)   &
=\left(  \mathbf{x}-\frac{1}{l}\sum\nolimits_{i=1}^{l}\mathbf{x}_{i\ast
}\right)  ^{T}\times\label{Normal Eigenlocus Likelihood Ratio Test}\\
&  \left[  \lambda_{\max_{\psi}}^{-1}\sum\nolimits_{i=1}^{l_{1}}\left\Vert
\sqrt{\operatorname{cov}_{up_{\updownarrow}}\left(  \mathbf{x}_{1_{i_{\ast}}%
}\right)  }\mathbf{x}_{1_{i_{\ast}}}\right\Vert _{\min_{c}}^{2}\frac
{\mathbf{x}_{1_{i_{\ast}}}}{\left\Vert \mathbf{x}_{1_{i_{\ast}}}\right\Vert
}\right] \nonumber\\
&  -\left(  \mathbf{x}-\frac{1}{l}\sum\nolimits_{i=1}^{l}\mathbf{x}_{i\ast
}\right)  ^{T}\times\nonumber\\
&  \left[  \lambda_{\max_{\psi}}^{-1}\sum\nolimits_{i=1}^{l_{2}}\left\Vert
\sqrt{\operatorname{cov}_{up_{\updownarrow}}\left(  \mathbf{x}_{2_{i_{\ast}}%
}\right)  }\mathbf{x}_{2_{i_{\ast}}}\right\Vert _{\min_{c}}^{2}\frac
{\mathbf{x}_{2_{i_{\ast}}}}{\left\Vert \mathbf{x}_{2_{i_{\ast}}}\right\Vert
}\right] \nonumber\\
&  \mathbf{+}\frac{1}{l}\sum\nolimits_{i=1}^{l}y_{i}\left(  1-\xi_{i}\right)
\overset{H_{1}}{\underset{H_{2}}{\gtrless}}0\text{,}\nonumber
\end{align}
where the terms $\lambda_{\max_{\psi}}^{-1}\left\Vert \sqrt{\operatorname{cov}%
_{up_{\updownarrow}}\left(  \mathbf{x}_{1_{i_{\ast}}}\right)  }\mathbf{x}%
_{1_{i_{\ast}}}\right\Vert _{\min_{c}}^{2}$ and $\lambda_{\max_{\psi}}%
^{-1}\left\Vert \sqrt{\operatorname{cov}_{up_{\updownarrow}}\left(
\mathbf{x}_{2_{i_{\ast}}}\right)  }\mathbf{x}_{2_{i_{\ast}}}\right\Vert
_{\min_{c}}^{2}$ describe the likelihood of finding an extreme data point in a
particular region of $%
\mathbb{R}
^{d}$. The normal eigenlocus ratio test in Eq.
(\ref{Normal Eigenlocus Likelihood Ratio Test}) is now compared with Bayes'
decision rule for common covariance data.

\subsection{Comparison of the Normal Eigenlocus Decision Rule with Bayes'
Decision Rule}

Bayes' decision rule and boundary are completely defined within the likelihood
ratio expression $\Lambda\left(  \mathbf{x}\right)  $: \
\begin{align*}
\Lambda\left(  \mathbf{x}\right)   &  =\frac{\left\vert \mathbf{\Sigma}%
_{2}\right\vert ^{1/2}\exp\left\{  -\frac{1}{2}\left(  \mathbf{x}-\mathbf{\mu
}_{1}\right)  ^{T}\mathbf{\Sigma}_{1}^{-1}\left(  \mathbf{x}-\mathbf{\mu}%
_{1}\right)  \right\}  }{\left\vert \mathbf{\Sigma}_{1}\right\vert ^{1/2}%
\exp\left\{  -\frac{1}{2}\left(  \mathbf{x}-\mathbf{\mu}_{2}\right)
^{T}\mathbf{\Sigma}_{2}^{-1}\left(  \mathbf{x}-\mathbf{\mu}_{2}\right)
\right\}  }\\
&  \overset{H_{1}}{\underset{H_{2}}{\gtrless}}\frac{P_{2}\left(  C_{12}%
-C_{22}\right)  }{P_{1}\left(  C_{21}-C_{11}\right)  }\text{,}%
\end{align*}
for Gaussian data, where no costs ($C_{11}=C_{22}=0$) are associated with
correct decisions and $C_{12}=C_{21}=1$ are the costs associated with
incorrect decisions
\citet{Duda2001}%
,
\citet{VanTrees1968}%
. The probabilistic expression of $\Lambda\left(  \mathbf{x}\right)  $ can be
reduced to an algebraic expression by means of a natural logarithm transform
$\ln\left[  \Lambda\left(  \mathbf{x}\right)  \right]  $.

The natural logarithm of the likelihood ratio $\ln\left[  \Lambda\left(
\mathbf{x}\right)  \right]  $:%
\begin{align}
\ln\left[  \Lambda\left(  \mathbf{x}\right)  \right]   &  =\mathbf{x}%
^{T}\left(  \mathbf{\Sigma}_{1}^{-1}\mathbf{\mu}_{1}-\mathbf{\Sigma}_{2}%
^{-1}\mathbf{\mu}_{2}\right) \label{Likelihood Ratio General Gaussian}\\
&  +\frac{1}{2}\mathbf{x}^{T}\left(  \mathbf{\Sigma}_{2}^{-1}\mathbf{x}%
-\mathbf{\Sigma}_{1}^{-1}\mathbf{x}\right) \nonumber\\
&  +\frac{1}{2}\mathbf{\mu}_{2}^{T}\mathbf{\Sigma}_{2}^{-1}\mathbf{\mu}%
_{2}-\frac{1}{2}\mathbf{\mu}_{1}^{T}\mathbf{\Sigma}_{1}^{-1}\mathbf{\mu}%
_{1}\overset{H_{1}}{\underset{H_{2}}{\gtrless}}\eta\text{,}\nonumber
\end{align}%
\[
\eta\triangleq\ln\left(  P_{2}\right)  -\ln\left(  P_{1}\right)  +\frac{1}%
{2}\ln\left(  \left\vert \mathbf{\Sigma}_{1}\right\vert \right)  -\frac{1}%
{2}\ln\left(  \left\vert \mathbf{\Sigma}_{2}\right\vert \right)  \text{,}%
\]
produces an algebraic expression that defines the general form of the
discriminant function for the general Gaussian binary classification problem,
where $\mathbf{\mu}_{1}$ and $\mathbf{\mu}_{2}$ are $d$-component mean
vectors, $\mathbf{\Sigma}_{1}$ and $\mathbf{\Sigma}_{2}$ are $d$-by-$d $
covariance matrices, $\mathbf{\Sigma}^{-1}$ and $\left\vert \mathbf{\Sigma
}\right\vert $ denote the inverse and determinant of a covariance matrix
respectively, $H_{1}$ or $H_{2}$ is the true data category, and the class
probabilities $P_{1}=P_{2}=1/2$.

Bayesian decision theory provides the result that the algebraic expression in
Eq. (\ref{Likelihood Ratio General Gaussian}) describes the geometric loci of
Bayes' decision boundaries for any two classes of data drawn from Gaussian
distributions. Bayes' decision boundaries are defined by regions for which
class probabilities are equivalent to each other, i.e., $P_{1}=P_{2}=1/2$, and
are characterized by the class of hyperquadric decision surfaces which include
hyperplanes, pairs of hyperplanes, hyperspheres, hyperellipsoids,
hyperparaboloids, and hyperhyperboloids
\citet{Duda2001}%
,
\citet{VanTrees1968}%
.

\subsubsection*{Bayes' Decision Rule for Similar Covariance Data}

Letting $\mathbf{\Sigma}_{1}^{-1}=\mathbf{\Sigma}_{2}^{-1}=\mathbf{\Sigma
}^{-1}$ in Eq. (\ref{Likelihood Ratio General Gaussian}) provides an
expression%
\begin{align}
\ln\left[  \Lambda\left(  \mathbf{x}\right)  \right]   &  =\mathbf{x}%
^{T}\mathbf{\Sigma}^{-1}\left(  \mathbf{\mu}_{1}-\mathbf{\mu}_{2}\right)
\label{Bayes' Likelihood Similar Covariance Gaussian Data}\\
&  +\frac{1}{2}\mathbf{\mu}_{2}^{T}\mathbf{\Sigma}^{-1}\mathbf{\mu}_{2}%
-\frac{1}{2}\mathbf{\mu}_{1}^{T}\mathbf{\Sigma}^{-1}\mathbf{\mu}%
_{1}\overset{H_{1}}{\underset{H_{2}}{\gtrless}}\eta\text{,}\nonumber
\end{align}%
\[
\eta\triangleq\ln\left(  P_{2}\right)  -\ln\left(  P_{1}\right)  \text{,}%
\]
which defines the general form of the discriminant function for similar
covariance Gaussian data. Bayesian decision theory provides the result that
Eq. (\ref{Bayes' Likelihood Similar Covariance Gaussian Data}) encodes Bayes'
likelihood ratio for similar covariance Gaussian data. Bayesian decision
theory also provides the result that the algebraic expression in Eq.
(\ref{Bayes' Likelihood Similar Covariance Gaussian Data}) describes the
geometric loci of Bayes' linear decision boundaries
\citet{Duda2001}%
,
\citet{VanTrees1968}%
.

Now, consider again the normal eigenlocus likelihood ratio test in Eq.
(\ref{Normal Eigenlocus Likelihood Ratio Test}). Given that linear kernel SVM
learns Bayes' linear decision boundaries, and given that Eq.
(\ref{Bayes' Likelihood Similar Covariance Gaussian Data}) determines the
locus equation of Bayes' linear decision boundaries, it follows that the
constrained normal eigenlocus discriminant function $\Lambda_{\mathbf{\tau
}_{1}\mathbf{-\tau}_{2}}\left(  \mathbf{x}\right)  $ in Eq.
(\ref{Normal Eigenlocus Likelihood Ratio Test}) determines the locus equation
of linear decision boundaries for similar covariance Gaussian data. Indeed, it
has been shown that the geometric loci of linear decision boundaries
determined by the constrained normal eigenlocus discriminant function
$\Lambda_{\mathbf{\tau}_{1}\mathbf{-\tau}_{2}}\left(  \mathbf{x}\right)  $ in
Eq. (\ref{Normal Eigenlocus Likelihood Ratio Test}) are defined by regions for
which class probabilities are equivalent to each other.

It is concluded that the geometric loci of Bayes' linear decision boundaries
are completely defined within the normal eigenlocus likelihood ratio
expression $\Lambda_{\mathbf{\tau}_{1}\mathbf{-\tau}_{2}}\left(
\mathbf{x}\right)  $ $\overset{H_{1}}{\underset{H_{2}}{\gtrless}}0$ in Eq.
(\ref{Normal Eigenlocus Likelihood Ratio Test}).

Now, \emph{Bayes' decision rule and boundary are completely defined within the
likelihood ratio expression} $\Lambda\left(  \mathbf{x}\right)  $ in Eq.
(\ref{Bayes' Likelihood Similar Covariance Gaussian Data}). Therefore, given
that the constrained normal eigenlocus discriminant function $\Lambda
_{\mathbf{\tau}}\left(  \mathbf{x}\right)  $ describes the geometric loci of
Bayes' linear decision boundaries, it also follows that the likelihood ratio
encoded within the normal eigenlocus test statistic $\Lambda_{\mathbf{\tau
}_{1}\mathbf{-\tau}_{2}}\left(  \mathbf{x}\right)  $ $\overset{H_{1}%
}{\underset{H_{2}}{\gtrless}}0$ in Eq.
(\ref{Normal Eigenlocus Likelihood Ratio Test}) describes Bayes' likelihood
ratio for similar covariance Gaussian data. Numerous simulations studies show
that the normal eigenlocus decision rule in Eq.
(\ref{Normal Eigenlocus Likelihood Ratio Test}) achieves Bayes' error rate for
normally distributed data that have the same covariance matrix, including
homogeneous data distributions.

It is concluded that the likelihood ratio $\Lambda_{\mathbf{\tau}%
_{1}\mathbf{-\tau}_{2}}\left(  \mathbf{x}\right)  $ $\overset{H_{1}%
}{\underset{H_{2}}{\gtrless}}0$ encoded within the discriminant function
$D\left(  \mathbf{x}\right)  =\mathbf{\tau}^{T}\mathbf{x}+\tau_{0}$ determines
Bayes' likelihood ratio for similar covariance Gaussian data.

Clearly, then, the normal eigenlocus test statistic $\Lambda_{\mathbf{\tau
}_{1}\mathbf{-\tau}_{2}}\left(  \mathbf{x}\right)  $ $\overset{H_{1}%
}{\underset{H_{2}}{\gtrless}}0$ in Eq.
(\ref{Normal Eigenlocus Likelihood Ratio Test}) provides a sufficient
statistic for Bayes' decision rule and boundary for similar covariance data.
It is concluded that Bayes' decision rule and boundary are completely defined
within the normal eigenlocus likelihood ratio expression $\Lambda
_{\mathbf{\tau}_{1}\mathbf{-\tau}_{2}}\left(  \mathbf{x}\right)  $
$\overset{H_{1}}{\underset{H_{2}}{\gtrless}}0$ in Eq.
(\ref{Normal Eigenlocus Likelihood Ratio Test}).

Furthermore, given the robust, data-driven likelihood ratio encoded within the
normal eigenlocus decision rule $\Lambda_{\mathbf{\tau}_{1}\mathbf{-\tau}_{2}%
}\left(  \mathbf{x}\right)  $ in Eq.
(\ref{Normal Eigenlocus Likelihood Ratio Test}), it is concluded that the
normal eigenlocus test statistic provides a robust decision statistic for all
other data distributions.

The above analysis substantiates the previous work that is outlined next.

\subsection{Previous Work on Linear Kernel SVMs}

It has been demonstrated that $\Lambda_{\mathbf{\tau}_{1}\mathbf{-\tau}_{2}%
}\left(  \mathbf{x}\right)  \equiv\ln\left[  \Lambda\left(  \mathbf{x}\right)
\right]  $ for the normally distributed training data described next. This
class of Gaussian data is considered to be linearly separable, i.e., a
separating line, plane, or hyperplane exists for all such Gaussian data.

\subsubsection*{Linearly Separable Data}

Linear curves and surfaces provide optimal decision boundaries for normally
distributed data sets that have the same covariance matrix $\mathbf{\Sigma
}_{0}=\mathbf{\Sigma}_{1}=\mathbf{\Sigma}$:%
\begin{align*}
p\left(  \mathbf{x}|H_{0}\right)   &  \sim N\left(  \mathbf{\mu}%
_{0},\mathbf{\Sigma}\right)  \text{,}\\
p\left(  \mathbf{x}|H_{1}\right)   &  \sim N\left(  \mathbf{\mu}%
_{1},\mathbf{\Sigma}\right)  \text{.}%
\end{align*}
This class of problems has been referred to as a linearly separable
classification problem
\citet{Reeves2009}%
.

\subsubsection*{Linearly Separable Classification Problem}

Consider the following binary classification problem:%
\begin{equation}
\mathbf{x}_{i}=\left\{
\begin{array}
[c]{c}%
\mathbf{s}_{0}+\mathbf{n}_{i};\text{ \ }H_{0}\\
\mathbf{s}_{1}+\mathbf{n}_{i};\text{ \ }H_{1}%
\end{array}
\right.  \text{,} \label{Linearly Separable Classification Problem}%
\end{equation}
where $\mathbf{x}_{i}$ is the $i^{th}$ $d\times1$ random vector under
hypotheses $H_{0}$ and $H_{1}$, respectively, $\mathbf{n}_{i}$ is the
corresponding noise vector, and $\mathbf{s}_{0}$ and $\mathbf{s}_{1}$ are
deterministic signal vectors. It is assumed that the $\mathbf{n}_{i}$ are
independent and identically distributed (i.i.d.) zero-mean correlated Gaussian
random vectors with known $d\times d$ noise covariance matrix $R_{N} $.
Because $\mathbf{s}_{0}$ and $\mathbf{s}_{1}$ are deterministic and the noise
is additive and Gaussian, Eq. (\ref{Linearly Separable Classification Problem}%
) defines a linearly separable classification problem. The data points
$\mathbf{x}_{i}$ are optimally partitioned by linear curves and surfaces, and
are said to be linearly separable.

Let $\widetilde{\mathbf{X}}\triangleq\mathbf{D}_{y}\mathbf{X}$, where
$\mathbf{D}_{y}$ is a $N\times N$ diagonal matrix with of training labels
$y_{i}$ and the $N\times d$ data matrix is $\mathbf{X}$ $=%
\begin{pmatrix}
\mathbf{x}_{1}, & \mathbf{x}_{2}, & \ldots, & \mathbf{x}_{N}%
\end{pmatrix}
^{T}$. It has been shown that a strong dual normal eigenlocus decision test
$D\left(  \mathbf{x}\right)  \gtrless_{H_{0}}^{H_{1}}0$ can be written as:%
\begin{align*}
&  \mathbf{m}_{1}^{T}\mathbf{R}_{N}^{-1}\mathbf{x}\overset{H_{1}%
}{\underset{H_{0}}{\gtrless}}\left(  1-p_{0}\right)  \mathbf{m}_{1}%
^{T}\mathbf{R}_{N}^{-1}\mathbf{m}_{1}\\
&  -\left(  1-2p_{0}\right)  \left\{  \frac{1+p_{0}\left(  1-p_{0}\right)
\mathbf{m}_{1}^{T}\mathbf{R}_{N}^{-1}\mathbf{m}_{1}}{2p_{0}\left(
1-p_{0}\right)  \left(  1+\overline{\lambda}\right)  }\right\}  \text{,}%
\end{align*}
where the correlation matrix $\mathbf{R}_{x}\doteq N^{-1}\widetilde{\mathbf{X}%
}^{T}\widetilde{\mathbf{X}}$ can be expressed as $\mathbf{R}_{x}%
=\mathbf{R}_{N}+\left(  1-p_{0}\right)  \mathbf{m}_{1}\mathbf{m}_{1}^{T}$ and
$\mathbf{R}_{x}^{-1}=\left(  \mathbf{R}_{N}\left(  \mathbf{I}+\left(
1-p_{0}\right)  \mathbf{R}_{N}^{-1}\mathbf{m}_{1}\mathbf{m}_{1}^{T}\right)
\right)  ^{-1}$, the quantity $p_{0}\mathbf{m}_{0}+\left(  1-p_{0}\right)
\mathbf{m}_{1}$ $\doteq\sum\limits_{i=1}^{N}\mathbf{x}_{i}/N$, $\mathbf{m}%
_{0}$ and $\mathbf{m}_{1}$ are the means under $H_{0}$ and $H_{1}$
respectively, and $p_{0}$ is the probability that the $H_{0}$ hypothesis is
represented in the training data
\citet{Reeves2009}%
. Without loss of generality, $\mathbf{m}_{0}=0$.

The strong dual normal eigenlocus decision test is now compared with Bayes' test.

\subsubsection*{Comparison with the Bayes' Test}

By way of comparison, recall that the Bayes' Test is:%
\begin{align*}
&  \mathbf{m}_{1}^{T}\mathbf{R}_{N}^{-1}\mathbf{x}\overset{H_{1}%
}{\underset{H_{0}}{\gtrless}}\left(  1-p_{0}\right)  \mathbf{m}_{1}%
^{T}\mathbf{R}_{N}^{-1}\mathbf{m}_{1}\\
&  +\ln\left(  \frac{p_{0}\left(  C_{10}-C_{00}\right)  }{p_{1}\left(
C_{01}-C_{11}\right)  }\right)  \text{.}%
\end{align*}
When $p_{0}=1/2$ and $C_{10}=C_{01}=1$ and $C_{00}=C_{11}=0$ (no costs are
associated with correct decisions), the normal eigenlocus and the Bayes' tests
are equivalent. However, when $p_{0}\neq1/2$, the hypothesis tests are
different. Although the sufficient statistic $\mathbf{m}_{1}^{T}\mathbf{R}%
_{N}^{-1}\mathbf{x}$ is the same, the thresholds are different. This is
because the analogous costs associated with the strong dual normal eigenlocus
test are functions of $p_{0}$, whereas for the Bayes' test, these costs are
fixed. It is concluded that the discriminant function $D\left(  \mathbf{x}%
\right)  $ in Eq. (\ref{Discriminant Function}) encodes an optimal test
statistic that is the minimum probability of error for making a decision for
the training data described by Eq.
(\ref{Linearly Separable Classification Problem}). The above analysis has been
validated with simulation studies
\citet{Reeves2009}%
.

The next section of the paper will consider dual-use of strong dual normal
eigenlocus discriminant functions. Dual-use involves the practical matter of
building robust, scalable, and optimal, probabilistic, multiclass, linear
pattern recognition systems. Dual-use also involves a statistical multimeter
which effectively measures class separability and Bayes' error rate. The
statistical multimeter provides a robust indicator of homogeneous data distributions.

\section{Design of Probabilistic Multiclass Linear Pattern Recognition
Systems}

Given the robust, data-driven likelihood ratio expression encoded within the
strong dual normal eigenlocus discriminant function in Eq.
(\ref{Normal Eigenlocus Likelihood Ratio Test}), it follows that the
constrained discriminant function in Eq. (\ref{Discriminant Function})
describes robust linear decision boundaries for any given data distributions.
Indeed, the data-driven likelihood ratio expression $\Lambda_{\mathbf{\tau
}_{1}\mathbf{-\tau}_{2}}\left(  \mathbf{x}\right)  \overset{H_{1}%
}{\underset{H_{2}}{\gtrless}}0$ encoded within Eq.
(\ref{Normal Eigenlocus Likelihood Ratio Test}) provides a robust test
statistic for both overlapping and non-overlapping data distributions.
Moreover, it has been demonstrated that strong dual normal eigenlocus
transforms produce \emph{regularized and customized statistical decision
systems} for the binary classification task. Figures $17$, $18$, $19$, $20$,
and $22$ illustrate how constrained normal eigenlocus discriminant functions
determine regularized and customized, data-driven geometric architectures that
encode robust decision statistics for the binary classification task.

Given the robust, data-driven likelihood ratio expression encoded within the
probabilistic linear discriminant function $D\left(  \mathbf{x}\right)
=\mathbf{\tau}^{T}\mathbf{x}+\tau_{0}$, it follows that strong dual normal
eigenlocus discriminant functions provide \emph{robust statistical building
blocks for probabilistic, multiclass, linear pattern recognition systems}. A
strong dual statistical decision function $\operatorname{sign}\left(
\Lambda_{\mathbf{\tau}}\left(  \mathbf{x}\right)  \right)  $%
\begin{align*}
\operatorname{sign}\left(  \Lambda_{\mathbf{\tau}}\left(  \mathbf{x}\right)
\right)   &  =\operatorname{sign}\left[  \left(  \mathbf{x}-\frac{1}{l}%
\sum\nolimits_{i=1}^{l}\mathbf{x}_{i\ast}\right)  ^{T}\mathbf{\tau+\cdots
}\right] \\
&  \operatorname{sign}\left[  \cdots+\frac{1}{l}\sum\nolimits_{i=1}^{l}%
y_{i}\left(  1-\xi_{i}\right)  \right]  \text{,}%
\end{align*}
where $\operatorname{sign}\left(  x\right)  \equiv\frac{x}{\left\vert
x\right\vert }$ for $x\neq0$, provides a natural means for discriminating
between multiple classes of data, where robust or optimal decisions can be
made that are based on the largest probabilistic output of decision banks of
strong dual statistical decision functions $\operatorname{sign}\left(
\Lambda_{\mathbf{\tau}}\left(  \mathbf{x}\right)  \right)  $. Figure $33$
illustrates the structure of a scalable statistical building block of a
statistical bank used to build a probabilistic statistical decision engine
which distinguishes between objects in $M$ different pattern classes.

\begin{center}%
\begin{center}
\includegraphics[
natheight=7.499600in,
natwidth=9.999800in,
height=3.2897in,
width=4.3777in
]%
{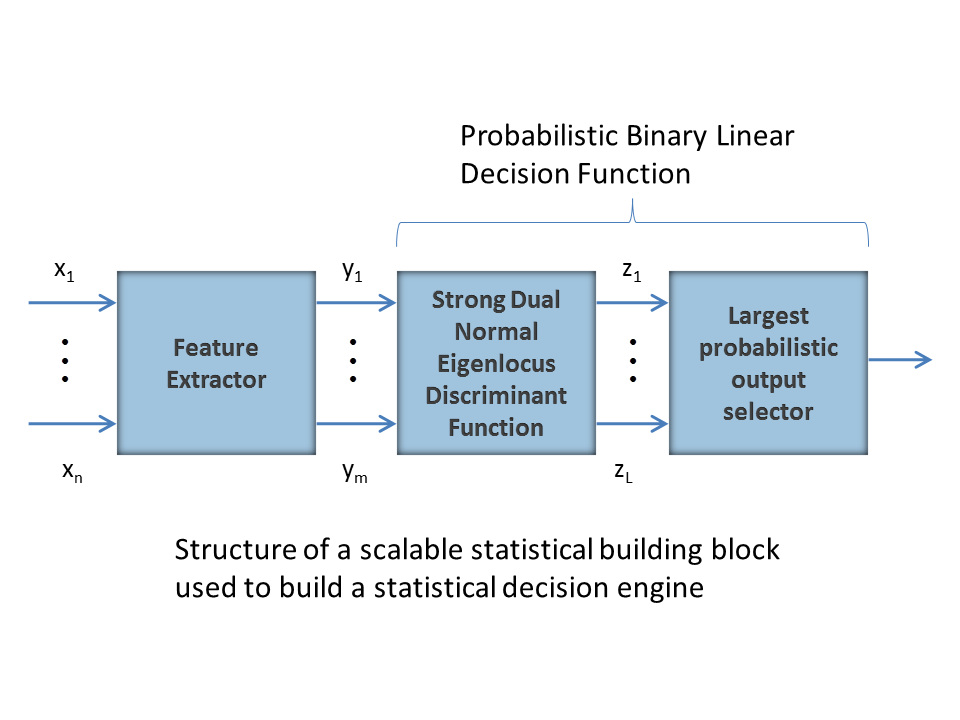}%
\end{center}

\end{center}

\begin{flushleft}
Figure $33$: Illustration of a scalable statistical building block of a
statistical decision bank $D_{dbX_{i}}\left(  \mathbf{x}\right)  $ used to
build a probabilistic statistical decision engine $PD_{e}\left[
\mathbf{x}\right]  $ which distinguishes between objects in $M$ different
pattern classes.
\end{flushleft}

\subsection{Design of Customized Probabilistic Statistical Decision Engines}

Consider the design of a probabilistic, multiclass, linear pattern recognition
system that distinguishes between objects in $M$ different pattern classes.
The powerful statistical machinery encoded within the binary class statistical
decision function $\operatorname{sign}\left(  \Lambda_{\mathbf{\tau}}\left(
\mathbf{x}\right)  \right)  $ enables the design of a customized,
probabilistic statistical decision engine $PD_{e}\left[  \operatorname{sign}%
\left(  \Lambda_{\mathbf{\tau}}\left(  \mathbf{x}\right)  \right)  \right]  $
that recognizes the objects in each of the $M$ pattern classes.

The design of a probabilistic statistical decision engine $PD_{e}\left[
\operatorname{sign}\left(  \Lambda_{\mathbf{\tau}}\left(  \mathbf{x}\right)
\right)  \right]  $ involves designing $M\times(M-1)$ strong dual statistical
decision functions $\operatorname{sign}\left(  \Lambda_{\mathbf{\tau}}\left(
\mathbf{x}\right)  \right)  $, each of which consists of a feature extractor
and a normal eigenlocus discriminant function. Accordingly, a statistical
decision bank $D_{dbX_{i}}\left[  \left\{  \operatorname{sign}\left(
\Lambda_{\mathbf{\tau}_{j}}\left(  \mathbf{x}\right)  \right)  \right\}
_{j=1}^{M-1}\right]  $ can be developed for each given pattern class $X_{i}$
that consists of a bank of $M-1$ statistical decision functions
$\operatorname{sign}\left(  \Lambda_{\mathbf{\tau}}\left(  \mathbf{x}\right)
\right)  $, where the pattern vectors in the given class $X_{i}$ have the
training label $+1$, and the statistical decision bank $D_{dbX_{i}}$ is a
linear combination of customized statistical decision functions%
\[
D_{dbX_{i}}=%
{\textstyle\sum\nolimits_{j=1}^{M-1}}
\operatorname{sign}\left(  \Lambda_{\mathbf{\tau}_{j}}\left(  \mathbf{x}%
\right)  \right)  \text{.}%
\]
The probabilistic statistical decision engine%
\[
PD_{e}\left[  \left\{  D_{dbX_{i}}\left[  \left\{  \operatorname{sign}\left(
\Lambda_{\mathbf{\tau}_{j}}\left(  \mathbf{x}\right)  \right)  \right\}
_{j=1}^{M-1}\right]  \right\}  _{i=1}^{M}\right]  \text{,}%
\]
provides a set of $M\times(M-1)$ decision statistics $\operatorname{sign}%
\left(  \Lambda_{\mathbf{\tau}}\left(  \mathbf{x}\right)  \right)  $, where
the maximum value selector of the statistical decision engine chooses the
pattern class $X_{i}$ for which a statistical decision bank $D_{dbX_{i}%
}\left(  \mathbf{x}\right)  $ has \emph{the maximum probabilistic output}.
Because the statistical decision engine $PD_{e}\left[  \mathbf{x}\right]  $ is
a linear combination of discriminant functions, the overall network complexity
is scale-invariant for the feature space dimension and the number of pattern classes.

Furthermore, if a feature extractor has been developed that generates
non-overlapping feature vectors for all of the $M$ pattern classes, then the
scale-invariance of the statistical decision engine $PD_{e}\left[
\mathbf{x}\right]  $ ensures low estimation variance and optimal
generalization performance for feature vectors that possess optimal
discrimination capacity. All classes of feature vectors drawn from
non-overlapping probability distributions, for which Bayes' error is zero,
naturally exhibit optimal discrimination capacity.

The next section considers the use of normal eigenlocus test statistics to
design effective feature extractors. Strong dual normal eigenlocus decision
functions $\operatorname{sign}\left(  \Lambda_{\mathbf{\tau}}\left(
\mathbf{x}\right)  \right)  $ provide a robust statistical multimeter for
measuring class separability and Bayes' error rate.

\subsection{Practical Dual-Use of Strong Dual Normal Eigenlocus Discriminant
Functions}

Recall that machine learning algorithms for classification systems introduce
four sources of error, i.e., Bayes' error, modeling error, estimation error,
and computational error, into the final classification system. Bayes' error,
i.e., the probability of error, results from overlap between data
distributions. Given the robust, stable, and probabilistic properties of
strong dual discriminant functions, \emph{the fundamental and difficult
problem that remains to be solved is the design of an effective feature
extractor}. Given two or more pattern classes, the design of a feature
extractor involves determining measurements or features which are most
effective for preserving class separability, where class separability is
equivalent to the probability of error due to the Bayes' classifier.
A\ feature extractor produces characteristic signatures, called feature
vectors, that describe the objects in a pattern class, such as fingerprints or
voices. A typical characteristic signature is an ordered sequence of
measurements, whereby each measurement describes a numerical attribute or
feature of an object in a pattern class. Numerical features are random
variables that are characterized by expected values and covariances.

Common examples of characteristic signatures include genetic signatures,
proteomic signatures, geometric shape recognition signatures, chemical
signatures, spectral signatures, biological signatures, radar signatures,
lidar signatures, and multispectral or hyperspectral signatures. This list is
by no means exhaustive. Feature vectors can be extracted from any given
collection of signals or images.

\emph{The probability of error is the key parameter of all statistical pattern
recognition systems.} The amount of overlap between data distributions
determines the Bayes' classification error rate which is the best error rate
that can be achieved by any classifier
\citet{Fukunaga1990}%
.

\subsection{Using Normal Eigenlocus Test Statistics to Design Effective
Feature Extractors}

A\ critical design objective for any statistical pattern recognition system is
to develop a feature extractor that provides distinct statistical signatures
for all of the pattern classes, i.e., negligible or no overlap exists amongst
each pair of data distributions. \emph{Moreover, the criteria to evaluate the
effectiveness of features must be a measure of the overlap or class
separability among data distributions, and not a measure of fit such as the
mean-square error of a statistical model }%
\citet{Fukunaga1990}%
.

In general, Bayes' error rate is difficult to evaluate. Explicit mathematical
expressions are only available for a few special cases. For normal
distributions, calculation of the Bayes' error involves numerical integration,
except for the common covariance case. Alternatively, the Bhattacharyya
distance provides a convenient measure of class separability for two pattern
classes. In addition, the Bhattacharyya distance provides an upper bound of
the Bayes' error, if training data are drawn from Gaussian distributions.
However, the Bhattacharyya distance is difficult to evaluate because the trace
and the determinant of matrices are combined in the criterion
\citet{Fukunaga1990}%
.

\emph{On the other hand, strong dual normal eigenlocus discriminant functions
provide a robust measure of class separability and Bayes' error rate for any
given sets of feature vectors.}

\subsubsection{A Robust Statistical Multimeter}

Strong dual statistical decision functions $\operatorname{sign}\left(
\Lambda_{\mathbf{\tau}}\left(  \mathbf{x}\right)  \right)  $ provide a useful
statistical multimeter for measuring data distribution overlap and Bayes'
error rate. Given the robust, data-driven likelihood ratio test encoded within
strong dual discriminant functions, strong dual statistical decision functions
$\operatorname{sign}\left(  \Lambda_{\mathbf{\tau}}\left(  \mathbf{x}\right)
\right)  $ can be used to estimate data distribution overlap and Bayes' error
rate for any given sets of feature vectors. In addition, strong dual
statistical decision functions $\operatorname{sign}\left(  \Lambda
_{\mathbf{\tau}}\left(  \mathbf{x}\right)  \right)  $ can be used to identify
homogeneous data distributions. Given any homogeneous data distribution, it
has been shown that $\left(  1\right)  $ most, if not all, of the training
data are transformed into constrained primal normal eigenaxis components, and
$\left(  2\right)  $ the error rate of the strong dual discriminant function
is $50\%$.

\subsection{Summary of Practical Dual-Use of Linear Kernel SVMs}

It is concluded that strong dual statistical decision functions
$\operatorname{sign}\left(  \Lambda_{\mathbf{\tau}}\left(  \mathbf{x}\right)
\right)  $ have practical dual-use as $\left(  1\right)  $ statistical
multimeters in the design of effective feature extractors, $\left(  2\right)
$ robust indicators of homogeneous data distributions, and $\left(  3\right)
$ statistical building blocks of statistical decision banks $D_{dbX_{i}%
}\left(  \mathbf{x}\right)  $ used to form probabilistic statistical decision
engines $PD_{e}\left[  \mathbf{x}\right]  $. \emph{Thereby, it is concluded
that linear kernel SVMs are a powerful and robust class of statistical
learning machines which are useful for the design, development, and
implementation of probabilistic, multiclass linear pattern recognition
systems.}

The final sections of the paper will summarize the geometric underpinnings and
statistical machinery of linear kernel SVMs. All of the major findings and
conclusions will be outlined in the next two sections.

\section{Synopsis of Geometric Underpinnings and Statistical Machinery of
Linear Kernel SVMs}

This paper has shown that learning linear decision boundaries from training
data essentially involves learning the locus of a principal eigenaxis, which
has been named a normal eigenaxis. The paper has introduced and developed
locus equations of a normal eigenaxis that describe lines, planes,
hyperplanes, and normal eigenaxes. The paper has shown that the locus of a
normal eigenaxis is an inherent part of any linear curve or surface, where the
geometric locus of a linear curve or surface is encoded within the locus of
its normal eigenaxis.

This paper has demonstrated how the eigen-coordinate locations of a normal
eigenaxis determine the uniform properties exhibited by the points on a linear
curve or surface. The paper has shown that normal eigenaxes of linear loci
provide exclusive and distinctive reference axes. The paper has also
demonstrated that a normal eigenaxis satisfies a linear locus in terms of its
eigenenergy, which is the fundamental property of a normal eigenaxis.

This paper has motivated and developed a dual statistical eigenlocus of normal
eigenaxis components which encodes the eigen-coordinate locations of an
unknown normal eigenaxis of a linear decision boundary. The paper has
introduced and developed the constrained primal and the Wolfe dual eigenlocus
equations of a strong dual normal eigenlocus, which have been shown to provide
a joint statistical estimate of a dual statistical eigenlocus of constrained
primal normal eigenaxis components. The paper has demonstrated that finding a
separating line, plane, or hyperplane requires estimating the strong dual
normal eigenlocus of a linear decision boundary and the bilaterally
symmetrical borders which bound it. The paper has also demonstrated that a
strong dual normal eigenlocus satisfies a linear decision boundary and the
bilaterally symmetrical borders which bound it in terms of a critical minimum eigenenergy.

This paper has demonstrated how the constrained primal and the Wolfe dual
eigenlocus equations of a strong dual normal eigenlocus transform two given
sets of pattern vectors into a dual statistical eigenlocus of constrained
primal normal eigenaxis components, all of which are jointly located in dual
and primal, correlated Hilbert spaces, all of which jointly describe
correlated linear subspaces of\textbf{\ }$%
\mathbb{R}
^{N}$\textbf{\ }and $%
\mathbb{R}
^{d}$,\textbf{\ }each of which encodes the likelihood of finding an extreme
data point in a particular region of $%
\mathbb{R}
^{d}$. Any given dual statistical eigenlocus of constrained primal normal
eigenaxis components delineates and satisfies three, symmetrical linear
partitioning curves or surfaces in $%
\mathbb{R}
^{d}$.

This paper has demonstrated how all of the constrained primal normal eigenaxis
components on a strong dual normal eigenlocus jointly specify a strong dual
statistical decision system that delineates bipartite and symmetrical
geometric regions of large covariance, located between two data distributions
in\ $%
\mathbb{R}
^{d}$, such that these bipartite, congruent geometric regions delineate
$\left(  1\right)  $ bipartite, congruent, non-overlapping geometric regions
of large covariance for any two non-overlapping data distributions, or
$\left(  2\right)  $ bipartite, congruent geometric regions of data
distribution overlap for any two overlapping data distributions. Thereby, this
paper has demonstrated the following:

\begin{itemize}
\item All of the constrained primal normal eigenaxis components on a strong
dual normal eigenlocus jointly specify three, symmetrical hyperplane
partitioning surfaces in $%
\mathbb{R}
^{N}$ that are interconnected with three, symmetrical linear partitioning
curves or surfaces in $%
\mathbb{R}
^{d}$. A strong dual normal eigenlocus delineates and satisfies three,
symmetrical linear partitioning curves or surfaces in $%
\mathbb{R}
^{d}$.

\item The resultant loci of points on all three linear partitioning curves or
surfaces in $%
\mathbb{R}
^{d}$ explicitly and exclusively reference the strong dual normal eigenlocus
of normal eigenaxis components.

\item The geometric loci of a linear decision boundary and its bilaterally
symmetrical linear decision borders are regulated by probabilities of finding
extreme data points within particular regions of $%
\mathbb{R}
^{d}$.

\item Likelihoods encoded within all of the constrained primal normal
eigenaxis components on $\mathbf{\tau}_{1}-\mathbf{\tau}_{2}$ specify the
stochastic behavior of a statistical decision system.

\item All of the constrained primal normal eigenaxis components on a strong
dual normal eigenlocus jointly encode a robust, data-driven likelihood ratio.

\item The fundamental geometric and statistical property exhibited by a
constrained primal normal eigenlocus $\mathbf{\tau=\tau}_{1}-\mathbf{\tau}%
_{2}$ is a critical minimum, i.e., a total allowed, eigenenergy.

\item The total allowed eigenenergy and the statistical equilibrium point of
$\mathbf{\tau}$ are specified by likelihood statistics encoded within
correlated normal eigenaxis components on a Wolfe dual normal eigenlocus
$\mathbf{\psi}$.

\item The constrained primal normal eigenaxis components on $\mathbf{\tau}$
satisfy a point of statistical equilibrium for which the eigenenergies of the
constrained primal normal eigenlocus components $\mathbf{\tau}_{1}%
-\mathbf{\tau}_{2}$ on $\mathbf{\tau}$ are symmetrically balanced with each
other in relation to a centrally located statistical fulcrum, which is half of
the total allowed eigenenergy of a constrained primal normal eigenlocus
$\mathbf{\tau}$.

\item The regularized, data-driven geometric architectures determined by
strong dual normal eigenlocus transforms are configured by enforcing joint
symmetrical distributions of the eigenenergies of $\mathbf{\psi}$ and
$\mathbf{\tau}$ over the eigen-scaled extreme training vectors on
$\mathbf{\tau}_{1}$ and $\mathbf{\tau}_{2}$, whereby the eigenenergies of the
strong dual normal eigenlocus components $\mathbf{\tau}_{1}$ and
$\mathbf{\tau}_{2}$ on $\mathbf{\tau}$ are symmetrically balanced with each other.
\end{itemize}

This paper has shown how the geometric configuration of a Wolfe dual normal
eigenlocus in $%
\mathbb{R}
^{N}$ effectively determines the geometric configuration of a constrained
primal normal eigenlocus in\ $%
\mathbb{R}
^{d}$. Thereby, this paper has demonstrated the following:

\begin{itemize}
\item A Wolfe dual normal eigenlocus delineates and satisfies three,
symmetrical hyperplane partitioning surfaces $H_{0}$, $H_{+1}$, and $H_{-1}$
in $%
\mathbb{R}
^{N}$.

\item Geometric configurations of three hyperplane partitioning surfaces
$H_{0}$, $H_{+1}$, and $H_{-1}$ in $%
\mathbb{R}
^{N}$ regulate the geometric configurations of three linear partitioning
surfaces $D_{0}\left(  \mathbf{x}\right)  $, $D_{+1}\left(  \mathbf{x}\right)
$, and $D_{-1}\left(  \mathbf{x}\right)  $ in $%
\mathbb{R}
^{d}$.

\item Each Wolfe dual normal eigenaxis component on $\mathbf{\psi}$ in $%
\mathbb{R}
^{N}$ encodes a maximum covariance estimate in a principal location
in\textbf{\ }$%
\mathbb{R}
^{d}$, which is determined by an eigen-balanced first and second order
statistical moment about the geometric locus of an extreme data point in\ $%
\mathbb{R}
^{d}$.

\item Each Wolfe dual normal eigenaxis component on $\mathbf{\psi}$
in\textbf{\ }$%
\mathbb{R}
^{N}$ specifies an eigen-scale for an extreme data point in\textbf{\ }$%
\mathbb{R}
^{d}$, whereby each constrained primal normal eigenaxis component on
$\mathbf{\tau}$ in\textbf{\ }$%
\mathbb{R}
^{d}$ encodes the probability of finding an extreme data point in a particular
region of $\mathbf{%
\mathbb{R}
}^{d}$.

\item The direction of each Wolfe dual normal eigenaxis component on
$\mathbf{\psi}$ in $%
\mathbb{R}
^{N}$ is identical to the direction of a correlated, constrained primal normal
eigenaxis component on $\mathbf{\tau}$ in $\mathbf{%
\mathbb{R}
}^{d}$.

\item The lengths of each Wolfe dual normal eigenaxis component on
$\mathbf{\psi}$ in $%
\mathbb{R}
^{N}$ and correlated, constrained primal normal eigenaxis component on
$\mathbf{\tau}$ in $\mathbf{%
\mathbb{R}
}^{d}$ are shaped by identical joint symmetrical distributions of Wolfe dual
and constrained primal normal eigenaxis components.

\item Each Wolfe dual normal eigenaxis component on $\mathbf{\psi}$ in $%
\mathbb{R}
^{N}$ exhibits a length that is shaped by an eigen-balanced pointwise
covariance estimate for a correlated extreme training vector in $%
\mathbb{R}
^{d}$, such that the eigenlocus of each constrained primal normal eigenaxis
component on $\mathbf{\tau}$ in $%
\mathbb{R}
^{d}$\ encodes a maximum\ covariance estimate in a principal location, in the
form of an eigen-balanced first and second order statistical moment about the
geometric locus of the extreme point, which describes the probability of
finding the extreme data point in a particular region of $%
\mathbb{R}
^{d}$.

\item The integrated sum of probabilities encoded within the constrained
primal normal eigenaxis components on $\mathbf{\tau}_{1}$ is balanced with the
integrated sum of probabilities encoded within the constrained primal normal
eigenaxis components on $\mathbf{\tau}_{2}$, so that strong dual normal
eigenlocus transforms determine linear decision boundaries and regulated
linear decision borders for which class probabilities are equivalent to each other.

\item Each constrained primal normal eigenaxis component on $\mathbf{\tau}$ in
$%
\mathbb{R}
^{d}$ is an eigen-scaled extreme vector that encodes an eigenstate of a
statistical decision system which contains a discriminant function that $(1)$
encodes Bayes' likelihood ratio for common covariance training data, and $(2)$
encodes a robust, data-driven likelihood ratio for all other data distributions.
\end{itemize}

In summary, this paper has demonstrated that strong dual normal eigenlocus
transforms generate robust statistical decision systems for a wide variety of
data distributions, including completely overlapping distributions. This paper
has also demonstrated how properly regularized linear kernel SVMs implement
strong dual normal eigenlocus transforms.\hfill

\section{Conclusions}

The problem of learning linear decision boundaries for overlapping sets of
data has been resolved. This long-standing problem was generally deemed
insoluble. The dilemma has been resolved by means of a dual statistical
eigenlocus of normal eigenaxis components, i.e., a strong dual normal
eigenlocus of eigen-scaled extreme data points, all of which encode a robust
likelihood ratio, where all of the eigen-scaled extreme data points sit on and
satisfy the strong dual normal eigenlocus, and all of the points on a
statistical decision system of symmetrical linear partitioning curves or
surfaces explicitly and exclusively reference the strong dual normal eigenlocus.

The discoveries presented in this paper specify effective designs for linear
kernel SVMs. The discoveries also define a statistical model for linear kernel
SVM that represents the relevant aspects of probabilistic, binary, linear
classification systems.

To summarize, an estimation process has been introduced that provides
consistent fits of random data points to unknown normal eigenaxis components
of linear decision boundaries. A computer-implemented method has been
formulated that transforms two sets of pattern vectors, generated by any two
probability distributions whose expected values and covariance structures do
not vary significantly over time, into a dual statistical eigenlocus of normal
eigenaxis components, all of which are jointly and symmetrically located in
correlated, dual and primal Hilbert spaces, all of which jointly describe
correlated, linear subspaces of $%
\mathbb{R}
^{N}$\ and $%
\mathbb{R}
^{d}$,\ all of which encode a robust likelihood ratio, all of which jointly
specify a statistical decision system that delineates a bipartite, symmetric
partitioning of a region of large covariance between two overlapping or
non-overlapping data distributions in $%
\mathbb{R}
^{d}$.

\ The statistical decision system provides a building block for probabilistic,
multiclass linear classifiers, a probabilistic, binary linear classifier for
overlapping and non-overlapping data distributions, an optimal, probabilistic,
binary linear classifier for common covariance data, a statistical gauge for
data distribution overlap and Bayes' error, and a statistical gauge that is a
definitive indicator of homogeneous data distributions.

An upcoming paper will introduce an estimation process that provides
consistent fits of random data points to unknown principal eigenaxis
components of unknown second-order decision boundaries that take the form of
$d$-dimensional circles, ellipses, hyperbolae, and parabolas. The discoveries
presented in the paper are expected to specify effective designs for
polynomial kernel SVMs. The discoveries are also expected to define a
statistical model for polynomial kernel SVM that represents the relevant
aspects of probabilistic, binary, nonlinear classification systems.

\subsection*{Acknowledgment}

The author is indebted to Oscar Gonzalez and Garry Jacyna. The author's
master's thesis
\citet{Reeves1995}
was the primary impetus for this work. The counsel of Oscar Gonzalez motivated
and sustained the trailblazer within the author. Some of the material in this
paper includes portions of the author's PhD dissertation
\citet{Reeves2009}%
. Part of this work would not have occurred without the support of the
author's adviser at the MITRE Corporation. The guidance of Garry Jacyna
enabled the author to successfully navigate the PhD pipeline at George Mason University.

\bibliographystyle{plainnat}
\bibliography{locusedb}

\end{document}